**Phạm Hải Đăng**

# NGHIÊN CỨU, PHÁT TRIỂN PHƯƠNG PHÁP HỌC MÁY HIỆN ĐẠI ĐỂ NÂNG CAO HIỆU NĂNG TÁI ĐỊNH DANH NGƯỜI

## LUẬN ÁN TIẾN SĨ HỆ THỐNG THÔNG TIN

**HÀ NỘI - 2025**

**Phạm Hải Đăng**

# NGHIÊN CỨU, PHÁT TRIỂN PHƯƠNG PHÁP HỌC MÁY HIỆN ĐẠI ĐỂ NÂNG CAO HIỆU NĂNG TÁI ĐỊNH DANH NGƯỜI

Ngành đào tạo: Hệ thống thông tin
Mã số: 9480104

## LUẬN ÁN TIẾN SĨ HỆ THỐNG THÔNG TIN

NGHIÊN CỨU SINH          CÁN BỘ HƯỚNG DẪN

Phạm Hải Đăng     PGS.TS. Nguyễn Ngọc Hóa     PGS.TS. Nguyễn Ngọc Tú

**XÁC NHẬN CỦA ĐƠN VỊ ĐÀO TẠO**

**HÀ NỘI - 2025**

# LỜI CAM ĐOAN

Tôi xin cam đoan các kết quả trình bày trong luận án là công trình nghiên cứu của bản thân Nghiên cứu sinh trong thời gian học tập và nghiên cứu tại Trường Đại học Công nghệ - Đại học Quốc gia Hà Nội dưới sự hướng dẫn của các cán bộ hướng dẫn. Các số liệu, kết quả trình bày trong luận án là hoàn toàn trung thực và chưa từng được công bố trong bất kỳ công trình nào trước đây. Các kết quả sử dụng tham khảo đều đã được trích dẫn đầy đủ và theo đúng quy định.

Hà Nội, ngày 27 tháng 12 năm 2025
Nghiên cứu sinh

Phạm Hải Đăng

# LỜI CẢM ƠN

Luận án này sẽ không thể hoàn thành nếu không có sự giúp đỡ, hỗ trợ, động viên của các thầy, các cô, gia đình, bạn bè, đồng nghiệp và nhà trường.

Đầu tiên, tôi xin bày tỏ lòng biết ơn chân thành và sâu sắc đến PGS.TS. Nguyễn Ngọc Hóa và PGS.TS. Nguyễn Ngọc Tú - tập thể hướng dẫn khoa học đã tận tình chỉ bảo, hướng dẫn, giúp đỡ tôi trong quá trình nghiên cứu và hoàn thiện luận án.

Tôi xin gửi lời cảm ơn đến ban lãnh đạo, cán bộ quản lý, chuyên viên của Trường Đại học Công nghệ - Đại học Quốc gia Hà Nội, cũng như Trường Đại học Khánh Hòa đã tạo mọi điều kiện thuận lợi nhất cho tôi trong quá trình thực hiện luận án.

Tôi xin gửi lời cảm ơn chân thành đến các thầy, các cô trong Bộ môn Các Hệ thống thông tin - Khoa Công nghệ thông tin - Trường Đại học Công nghệ - Đại học Quốc gia Hà Nội đã hỗ trợ, giúp đỡ, động viên tôi trong quá trình làm luận án.

Cuối cùng, tôi xin gửi lời cảm ơn tới tới gia đình, bạn bè, đồng nghiệp đã luôn động viên, giúp đỡ tôi vượt qua khó khăn để đạt được những kết quả nghiên cứu như hôm nay.

Tôi xin chân thành cảm ơn!

Hà Nội, ngày 27 tháng 12 năm 2025
Nghiên cứu sinh

Phạm Hải Đăng

# Mục lục

# Danh mục từ viết tắt

| Viết tắt | Tiếng Anh | Tiếng Việt |
|---|---|---|
| Backbone | Backbone Network | Mạng cốt lõi |
| CMC | Cumulative Match Characteristic | Đặc trưng đối sánh tích lũy |
| CNN | Convolutional Neural Network | Mạng nơ-ron tích chập |
| CV | Computer Vision | Thị giác máy tính |
| DCNN | Deep Convolutional Neural Network | Mạng nơ-ron tích chập sâu |
| DL | Deep Learning | Học sâu |
| DML | Deep Metric Learning | Học độ đo sâu |
| Gallery | Gallery set | Tập tìm kiếm |
| GAP | Global Average Pooling | Phép toán gộp trung bình toàn cục |
| GMP | Global Max Pooling | Phép toán gộp cực đại toàn cục |
| GPU | Graphics Processing Unit | Bộ xử lý đồ họa |
| ID | Identity | Định danh |
| IQA | Image Quality Assessment | Đánh giá chất lượng ảnh |
| mAP | Mean Average Precision | Bình quân độ chính xác trung bình |
| ReID | Re-identification | Tái định danh |
| RR | Re-Ranking | Xếp hạng lại |
| RQ | Research Question | Câu hỏi nghiên cứu |
| SCL | Supervised Contrastive Loss | Hàm mất mát tương phản có giám sát |
| SL | Supervised Learning | Học có giám sát |
| SOTA | State-Of-The-Art | Phương pháp tiên tiến |
| UDA | Unsupervised Domain Adaptation | Thích ứng miền không giám sát |
| USL | Unsupervised Learning | Học không giám sát |
| ViT | Vision Transformer | Mạng Transformer cho thị giác máy tính |

# Danh sách hình vẽ

# Danh sách bảng

# MỞ ĐẦU

## Bối cảnh và động lực nghiên cứu

Trong bối cảnh đô thị hóa và chuyển đổi số ngày càng gia tăng, hệ thống camera giám sát đã trở thành một công cụ thiết yếu trong việc đảm bảo an ninh, quản lý trật tự xã hội và giám sát hành vi trong không gian công cộng. Các camera giám sát thường được lắp đặt tại những vị trí trọng yếu như nhà ga, nút giao thông, trung tâm thương mại, khu dân cư hay các khu vực công cộng đông người, nhằm ghi lại liên tục các hoạt động của con người và phương tiện. Với khả năng theo dõi chuyển động, nhận diện đối tượng và phát hiện hành vi bất thường, các hệ thống giám sát hiện đại cho phép quan sát liên tục 24/7, cảnh báo sớm các tình huống nguy cơ và cung cấp bằng chứng hình ảnh phục vụ công tác điều tra, truy vết. Thực tiễn cho thấy, tại nhiều khu vực áp dụng giám sát chặt chẽ, hệ thống camera đã góp phần đáng kể vào việc giảm thiểu hành vi vi phạm và nâng cao mức độ an toàn công cộng [70, 119].

Một hệ thống camera giám sát điển hình thường bao gồm nhiều thiết bị được bố trí tại các vị trí khác nhau, liên tục thu nhận hình ảnh để tạo ra nguồn dữ liệu đầu vào khổng lồ. Các dữ liệu này sau đó được xử lý thông qua nhiều thành phần chức năng: phát hiện đối tượng để xác định sự xuất hiện của con người trong khung hình; theo dõi đối tượng nhằm ghi lại hành trình di chuyển; trích xuất và biểu diễn đặc trưng để chuyển đổi thông tin hình ảnh sang dạng véc-tơ có thể so sánh; và cuối cùng là lưu trữ, quản lý cũng như phân tích dữ liệu để phục vụ các tác vụ nâng cao như phát hiện hành vi bất thường, phân tích đám đông hay hỗ trợ điều tra hình sự. Tuy nhiên, để đạt hiệu quả toàn diện, hệ thống còn cần có khả năng gắn kết và theo dõi danh tính của một cá nhân không chỉ trong phạm vi một camera mà còn khi họ di chuyển qua nhiều camera khác nhau. Đây chính là vai trò cốt lõi của bài toán tái định danh người. Nhờ tái định danh người, hệ thống có thể truy xuất lịch sử di chuyển, phân tích hành vi, và tích hợp dữ liệu từ nhiều nguồn quan sát thành một bức tranh toàn diện, qua đó trở thành mắt xích then chốt giúp nâng cao giá trị ứng dụng của các hệ thống camera giám sát.

Cụ thể tái định danh người là bài toán xác định và nhận diện lại danh tính cụ thể của một người qua các bức ảnh giám sát khác nhau được thu thập từ một tập ảnh từ các camera có hướng, góc nhìn khác nhau và thường không có thông tin nhận diện rõ ràng như khuôn mặt hoặc dấu vết sinh trắc học. Nói cách khác, tái định danh người đặt ra yêu cầu truy vấn một ảnh người đầu vào (ảnh truy vấn) trong một tập ảnh người đã được thu thập (tập tìm kiếm) nhằm tìm ra những ảnh có khả năng thuộc về cùng một người với ảnh truy vấn. Tái định danh người thu hút sự quan tâm đáng kể từ cộng đồng nghiên cứu trong thập kỷ qua. Việc nhận diện người qua các camera có tọa độ và góc nhìn khác nhau không chỉ là một vấn đề kỹ thuật đầy thách thức mà còn có ý nghĩa thực tiễn sâu sắc trong bối cảnh an ninh đô thị và ứng dụng thông minh. Không giống như nhận dạng khuôn mặt truyền thống vốn bị hạn chế khi khuôn mặt bị che khuất hoặc khi chất lượng ảnh không cao, tái định danh người khai thác đặc trưng toàn thân như màu sắc trang phục, dáng đi, phụ kiện hoặc cấu trúc cơ thể. Tuy nhiên, bài toán này đặc biệt khó do sự biến thiên rất lớn về ngoại hình của cùng một người trong các tình huống khác nhau: thay đổi ánh sáng, tư thế, góc nhìn, che khuất, hoặc thậm chí là thay đổi quần áo.

Trong thời kỳ đầu, tái định danh người được giải quyết theo cách tiếp cận truyền thống sử dụng các đặc trưng thủ công phản ánh các thuộc tính như hình dáng, màu sắc, họa tiết, và kết cấu – chẳng hạn như biểu đồ tần suất màu [49], HOG [24]. Sau khi trích xuất đặc trưng, các nhóm nghiên cứu sử dụng một số độ đo để đo độ tương đồng giữa hai ảnh, ví dụ như độ đo Euclid, Cosin, Mahalanobis, hoặc các thuật toán khoảng cách như KISSME [56] và XQDA [64]. Ngoài ra, một số nghiên cứu còn sử dụng mô hình học máy cổ điển như máy véc-tơ hỗ trợ [22], rừng ngẫu nhiên [9] hoặc tăng cường để phân loại cặp ảnh là cùng người hay khác người. Dù có ưu điểm là dễ cài đặt và không cần lượng dữ liệu huấn luyện lớn, các phương pháp truyền thống bị hạn chế về khả năng tổng quát hóa. Chúng dễ bị ảnh hưởng bởi sự thay đổi ánh sáng, góc nhìn và đặc biệt là sự thay đổi về trang phục của các danh tính giữa các lần xuất hiện. Do đó, trong các môi trường phức tạp, hiệu quả của các phương pháp này thường không đủ để đáp ứng yêu cầu ứng dụng thực tế.

Bắt đầu từ năm 2014, hướng tiếp cận dựa trên học sâu đã nổi lên như một giải pháp hiệu quả cho bài toán tái định danh người [121]. Cốt lõi của hướng tiếp cận này là sử dụng các kiến trúc mạng nơ-ron sâu để tự động trích xuất đặc trưng từ dữ liệu ảnh đầu vào. Khác với phương pháp trích xuất thủ công vốn chỉ tập trung vào một số đặc trưng cố định như màu sắc, biên, hoặc hoa văn, mạng học sâu có khả năng học được các đặc trưng giàu thông tin, có tính phân biệt cao giữa các danh tính, đồng thời mang tính khái quát tốt hơn. Các mạng học sâu hiện đại như ResNet-50 [36], ViT [28] đã được thiết kế lại hoặc điều chỉnh trọng số để phù hợp với bài toán tái định danh bằng cách tích hợp các lớp cụ thể khác nhau nhằm tăng cường khả năng phân biệt và ổn định trong biểu diễn đặc trưng.

Tái định danh người hiện nay có thể được giải quyết thông qua ba hướng tiếp cận chính: học có giám sát, học thích ứng miền không giám sát, và học không giám sát hoàn toàn. Trong hướng tiếp cận học có giám sát, mô hình được huấn luyện trên các tập dữ liệu có nhãn với các hàm mất mát phổ biến như hàm mất mát phân loại [131], hàm mất mát bộ ba [75]. Mục tiêu là tối ưu không gian đặc trưng sao cho đặc trưng của các ảnh cùng danh tính nằm gần nhau, trong khi đặc trưng của các ảnh khác danh tính nằm xa nhau. Nhờ vào tập dữ liệu được gán nhãn đầy đủ, các mô hình học có giám sát đạt được hiệu năng rất cao trên các tập dữ liệu chuẩn.

Tuy nhiên, trong thực tế, mô hình thường được huấn luyện trên một tập dữ liệu nguồn có nhãn, nhưng lại cần hoạt động trên một tập dữ liệu đích không nhãn. Điều này thúc đẩy sự phát triển của các phương pháp thích ứng miền không giám sát. Các phương pháp này tận dụng mô hình đã được huấn luyện trước trên miền nguồn, sau đó điều chỉnh để thích nghi với dữ liệu đích thông qua các kỹ thuật như: tăng cường dữ liệu [133, 21], phân cụm [29, 71], gán nhãn giả, hoặc học biểu diễn bất biến miền [65].

Dẫu vậy, trong nhiều trường hợp thực tế, khi sự khác biệt phân phối giữa miền nguồn và miền đích quá lớn, hoặc việc dựa vào mô hình đã huấn luyện trên miền nguồn trở nên không hiệu quả. Khi đó, các phương pháp học không giám sát hoàn toàn trở thành hướng đi thay thế. Không cần bất kỳ nhãn nào, mô hình USL thường bắt đầu từ một mạng cốt lõi với tham số tiền huấn luyện, sau đó áp dụng các thuật toán phân cụm [29, 71] để gán nhãn giả, từ đó huấn luyện mô hình trên các nhãn này. Qua từng vòng huấn luyện, chất lượng biểu diễn đặc trưng được cải thiện dần theo hướng tăng cường khả năng phân biệt giữa các danh tính.

Với mỗi hướng tiếp cận, vẫn còn tồn tại những vấn đề nghiên cứu chưa được giải quyết triệt để. Do đó, luận án được định hướng nhằm nghiên cứu và phát triển các phương pháp học máy hiện đại nhằm nâng cao hiệu năng tái định danh người trong cả ba thiết lập: học có giám sát, học thích ứng miền không giám sát và học không giám sát hoàn toàn. Luận án không chỉ tập trung khảo sát và phân tích những phương pháp học sâu tiêu biểu trong từng thiết lập, mà còn đề xuất các kỹ thuật mới nhằm khắc phục hạn chế hiện tại, tối ưu hóa không gian đặc trưng và cải thiện độ chính xác tái định danh trong các điều kiện khác nhau. Bên cạnh đó, luận án cũng hướng đến các bài toán có giá trị thực tiễn cao, như giám sát an ninh và xây dựng đô thị thông minh, nơi dữ liệu thường không được gán nhãn đầy đủ và có sự biến thiên đáng kể giữa các miền quan sát.

# Thách thức

Để giải quyết bài toán tái định danh người, một số thách thức cần được khắc phục như sau:

- Sự biến đổi ngoại hình: Cùng một định danh có thể xuất hiện rất khác nhau ở các camera khác nhau do điều kiện ánh sáng thay đổi (ngày và đêm), do thay đổi trang phục, phụ kiện, kiểu tóc hoặc bị che khuất một phần bởi vật thể hoặc người khác. Bên cạnh đó, góc nhìn và sự khác biệt về thông số kỹ thuật giữa các camera cũng khiến hình ảnh của một định danh trông khác nhau.

- Hạn chế về nhãn dữ liệu: Quá trình xây dựng tập dữ liệu tái định danh người đòi hỏi việc thu thập ảnh từ nhiều camera và nhiều vị trí, sau đó gán nhãn thủ công để đảm bảo tính đồng nhất định danh. Đây là công việc tốn kém, mất nhiều thời gian và hầu như bất khả thi khi triển khai ở quy mô rộng. Do đó, rất khó có thể sử dụng phương pháp học có giám sát trong thực tế.

- Khả năng biểu diễn đặc trưng và lựa chọn mạng cốt lõi phù hợp: Làm sao xây dựng được biểu diễn đặc trưng giàu thông tin bằng cách kết hợp đặc trưng toàn cục và cục bộ, đồng thời lựa chọn hoặc thiết kế mạng cốt lõi có khả năng khai thác tối ưu đặc trưng trong các thiết lập học khác nhau.

- Chất lượng nhãn giả và tính ổn định trong huấn luyện không giám sát: Việc gán nhãn giả thường được thực hiện thông qua các phương pháp phân cụm trên không gian đặc trưng và chất lượng nhãn giả phụ thuộc chặt chẽ vào độ chính xác của phân cụm. Nếu nhãn giả bị nhiễu hoặc sai lệch, mô hình không chỉ giảm hiệu năng mà còn dễ rơi vào trạng thái huấn luyện không ổn định, dẫn đến kết quả khó kiểm soát.

- Khả năng mở rộng: Quá trình tìm kiếm và so khớp định danh trong không gian đặc trưng kích thước lớn sẽ tiêu tốn tài nguyên tính toán khổng lồ. Do đó, các thuật tìm kiếm và so khớp cần được thiết kế để vừa đảm bảo độ chính xác cao, vừa tối ưu về chi phí tính toán, nhằm đáp ứng yêu cầu xử lý theo thời gian thực trong quy mô triển khai rộng.

# Mục tiêu và nội dung nghiên cứu

**Mục tiêu của luận án**: Từ bối cảnh, động lực và những thách thức trên, mục tiêu của luận án là nghiên cứu, phát triển các phương pháp mới nhằm nâng cao hiệu năng tái định danh người, tiếp cận theo cả ba hướng học có giám sát, thích ứng miền không giám sát và không giám sát.

**Nội dung nghiên cứu**: Để đạt được mục tiêu trên, các nội dung nghiên cứu chính của luận án được xác lập như sau:

- Nội dung 1: Phân tích, đánh giá tổng quan về bài toán tái định danh người, qua đó tìm hiểu các phương pháp, kỹ thuật hiện có trong bài toán này để từ đó tìm ra khoảng trống nghiên cứu và đề xuất các nội dung nghiên cứu để nâng cao hiệu năng tái định danh người.

- Nội dung 2: Nghiên cứu, phát triển phương pháp nâng cao hiệu năng tái định danh người theo hướng tiếp cận học có giám sát nhằm nâng cao chất lượng biểu diễn đặc trưng từ đó tăng cường khả năng tái định danh người trong điều kiện biến đổi về ngoại hình, ánh sáng, góc nhìn.

- Nội dung 3: Nghiên cứu, phát triển phương pháp nâng cao hiệu năng tái định danh người theo hướng tiếp cận học thích ứng miền không giám sát nhằm khắc phục về hạn chế dữ liệu gán nhãn, giảm sự khác biệt về phân phối giữa hai miền dữ liệu, đồng thời cải thiện khả năng biểu diễn đặc trưng và chất lượng nhãn giả.

- Nội dung 4: Nghiên cứu, phát triển phương pháp nâng cao hiệu năng tái định danh người theo hướng tiếp cận học không giám sát hướng tới giảm sự phụ thuộc hoàn toàn vào dữ liệu gán nhãn, nâng cao khả năng biểu diễn đặc trưng và lựa chọn mạng cốt lõi phù hợp.

# Đối tượng và phạm vi nghiên cứu

Với các mục tiêu và nội dung nghiên cứu đặt ra như trên, đối tượng và phạm vi nghiên cứu của luận án được xác định như sau:

**Đối tượng nghiên cứu**

- Ảnh và tập ảnh người thu được từ các camera giám sát. Cụ thể là tập ảnh từ các bộ dữ liệu chuẩn dùng để đánh giá các mô hình tái định danh người: CUHK03 [61], Market-1501 [127], DukeMTMC-reID [91], MSMT17 [113].

- Hệ thống, mô hình, và phương pháp trích xuất, so sánh đặc trưng hình ảnh nhằm nhận diện chính xác một người xuất hiện trong nhiều camera khác nhau.

**Phạm vi nghiên cứu**

- Nghiên cứu và đưa ra giải pháp học máy hiện đại theo các hướng tiếp cận học có giám sát, học thích ứng miền không giám sát và học không giám sát để nâng cao hiệu năng cho mô hình tái định danh người.

- Nghiên cứu tập trung vào các vấn đề cốt lõi ảnh hưởng trực tiếp đến hiệu năng tái định danh người, bao gồm: biến thể góc nhìn và sự khác biệt giữa các camera; thiếu hoặc không có nhãn trong môi trường thực tế; khả năng biểu diễn đặc trưng và lựa chọn mạng cốt lõi phù hợp; chất lượng nhãn giả và tính ổn định trong huấn luyện không giám sát. Các yếu tố khác như thay đổi điều kiện ánh sáng, thay đổi quần áo và ngoại hình, che khuất, hay khả năng mở rộng trong so khớp danh tính, không nằm trong phạm vi nghiên cứu cụ thể của luận án do thuộc các lớp bài toán chuyên biệt khác.

- Nghiên cứu được thực hiện trên các bộ dữ liệu đánh giá chuẩn là CUHK03 [61], Market-1501 [127], DukeMTMC-reID [91], MSMT17 [113]. Những bộ dữ liệu này cung cấp hình ảnh của nhiều người, được ghi lại trong các điều kiện môi trường và góc nhìn khác nhau, giúp huấn luyện và kiểm tra mô hình tái định danh một cách tổng quát nhất.

- Nghiên cứu được thực hiện trong các bộ dữ liệu có tập đánh giá là tập đóng với giả định tất cả danh tính trong ảnh truy vấn đều tồn tại trong tập tìm kiếm. Điều này khác với tập mở khi dữ liệu cho phép ảnh truy vấn không tồn tại trong tập tìm kiếm, nên mô hình phải biết từ chối hoặc đánh dấu không tìm thấy.

# Phương pháp nghiên cứu

Các phương pháp nghiên cứu được sử dụng trong luận án để nâng cao hiệu năng tái định danh người được trình bày trong các mục dưới đây.

- Phương pháp nghiên cứu lý thuyết: Tìm hiểu các kiến thức nền tảng về trí tuệ nhân tạo, học máy, học sâu và bài toán tái định danh người. Phân tích các hướng tiếp cận như học có giám sát, thích ứng miền không giám sát, không giám sát, cùng các kỹ thuật nâng cao như tăng cường dữ liệu, kết hợp đặc trưng gán nhãn giả, phân cụm và ước lượng độ tin cậy. Nội dung này tạo cơ sở khoa học cho bước đề xuất mô hình và thực nghiệm so sánh kết quả.

- Phương pháp phân tích và tổng kết kinh nghiệm: Thu thập, tổng hợp và đánh giá các công trình nghiên cứu, bài báo khoa học, mã nguồn và kết quả thực nghiệm trước đó. Từ đó giúp nhận diện các xu hướng nghiên cứu hiện tại, đồng thời học hỏi kinh nghiệm từ các mô hình hiệu quả để từ đó làm cơ sở đề xuất hướng cải tiến mới.

- Phương pháp mô hình hóa: Mô hình hóa cụ thể bài toán tái định danh người bằng cách biểu diễn quy trình học đặc trưng, so khớp và đánh giá như một hệ thống gồm các thành phần đầu vào, khối trích xuất đặc trưng, không gian nhúng và hàm đo độ tương đồng. Việc mô hình hóa không chỉ giúp dễ dàng trong quá trình thực nghiệm mà còn tạo điều kiện cho việc triển khai và đánh giá có hệ thống.

- Phương pháp thực nghiệm: Tiến hành thực nghiệm trên các bộ dữ liệu chuẩn: CUHK03, Market-1501, DukeMTMC-reID, và MSMT17 để kiểm tra tính hiệu quả của các mô hình đề xuất. Thực hiện huấn luyện, điều chỉnh các siêu tham số, đánh giá kết quả bằng các tiêu chí như mAP và Rank-n. Ngoài ra, so sánh hiệu năng với các phương pháp tiên tiến hiện có để chứng minh mức độ cải tiến.

# Đóng góp chính

Với mục tiêu nghiên cứu và phạm vi nghiên cứu được xác định rõ ràng, luận án có những đóng góp sau:

- Đóng góp 1: Theo hướng tiếp cận học có giám sát, luận án đề xuất phương pháp "SCM-ReID" sử dụng học tương phản có giám sát kết hợp với bốn hàm mất mát khác là: phân loại, trung tâm, bộ ba và bộ ba trọng tâm giúp mô hình học được đặc trưng có tính phân biệt và khả năng khái quát tốt hơn để nâng cao hiệu năng tái định danh người. Các kết quả nghiên cứu trong đề xuất này được công bố trong công trình [CT2].

- Đóng góp 2: Theo hướng tiếp cận học thích ứng miền không giám sát, luận án đề xuất phương pháp "IQAGA" và "DAPRH" bằng cách kết hợp các chiến lược: thu hẹp khoảng cách phân phối giữa hai miền thông qua GAN và ánh xạ bất biến miền (DIM), đánh giá chất lượng ảnh để điều chỉnh trọng số huấn luyện, tích hợp đặc trưng cục bộ và toàn cục để tăng tính biểu diễn, và tinh chỉnh nhãn giả mềm dựa trên khoảng cách tâm cụm nhằm nâng cao độ ổn định và độ chính xác. Các kết quả nghiên cứu tương ứng với hai đề xuất được công bố trong [CT4] với "IQAGA" và [CT1], [CT5] đối với "DAPRH".

- Đóng góp 3: Theo hướng tiếp cận học không giám sát, luận án đề xuất phương pháp ViTC-UReID tận dụng kiến trúc ViT thay thế CNN để tăng cường biểu diễn đặc trưng toàn cục, đồng thời tích hợp thông tin định danh camera nhằm học đặc trưng phù hợp với từng góc nhìn, từ đó cải thiện hiệu năng tái định danh người không giám sát. Các kết quả nghiên cứu trong đề xuất này được công bố trong công trình [CT3].

# Cấu trúc luận án

Luận án có bốn chương chính, đi kèm với phần mở đầu và kết luận. Các phần có nội dung cụ thể như sau:

- Chương 1 trình bày tổng quan về bài toán tái định danh người, các nền tảng cơ bản liên quan đến học sâu, tổng quan các hướng nghiên cứu trong lĩnh vực tái định danh người, để từ đó xác lập khoảng trống nghiên cứu phục vụ cho xây dựng nội dung nghiên cứu cũng như hướng giải quyết mục tiêu để nâng cao hiệu năng tái định danh người theo các hướng tiếp cận khác nhau.

- Chương 2 trình bày phương pháp đề xuất "SCM-ReID" để nâng cao hiệu năng tái định danh người trong hướng tiếp cận học có giám sát. SCM-ReID sử dụng học tương phản có giám sát kết hợp với bốn hàm mất mát khác là: phân loại, trung tâm, bộ ba và bộ ba trọng tâm. Việc phối hợp các hàm mất mát này giúp tận dụng tốt các đặc điểm bổ trợ lẫn nhau, từ đó nâng cao khả năng phân biệt đặc trưng và cải thiện hiệu năng chung của mô hình.

- Chương 3 trình bày hai phương pháp đề xuất "IQAGA" và "DAPRH" nhằm nâng cao hiệu năng tái định danh người trong hướng tiếp cận học thích ứng miền không giám sát. Cả hai phương pháp đều tập trung cải thiện quá trình huấn luyện với dữ liệu có nhãn trên miền nguồn, trong khi DAPRH còn mở rộng thêm các chiến lược tăng cường hiệu quả huấn luyện trên miền đích không có nhãn, nhằm nâng cao khả năng thích ứng và tổng quát hóa của mô hình trong môi trường thực tế không đồng nhất.

- Chương 4 trình bày phương pháp đề xuất "ViTC-UReID" để nâng cao hiệu năng tái định danh người trong hướng tiếp cận học không giám sát. ViTC-UReID sử dụng kiến trúc mạng cốt lõi ViT thay thế cho các mạng CNN truyền thống, nhằm tăng cường khả năng biểu diễn đặc trưng toàn cục của ảnh. Bên cạnh đó, ViTC-UReID còn tích hợp thêm thông tin về camera để học các đặc trưng có khả năng phân biệt danh tính theo từng góc nhìn, từ đó cải thiện độ chính xác nhận diện trên toàn hệ thống.

- Kết luận và hướng phát triển trình bày tóm tắt các đóng góp chính đã đạt được đối với bài toán tái định danh người, đồng thời đưa ra bàn luận về các kết quả đã đạt được cũng như phân tích những hạn chế còn tồn tại trong các phương pháp đề xuất. Từ đó, luận án đưa ra một số hướng nghiên cứu tiềm năng nhằm mở rộng và nâng cao hiệu năng của các phương pháp tái định danh người.

# Chương 1
# TỔNG QUAN NGHIÊN CỨU
# VÀ KIẾN THỨC NỀN TẢNG

Trong chương này, luận án trình bày tổng quan về bài toán tái định danh người, bao gồm định nghĩa, mục tiêu, cách thức mô hình hóa, hướng tiếp cận giải quyết dựa trên học sâu, các bộ dữ liệu chuẩn được sử dụng phổ biến trong nghiên cứu, cùng với các độ đo đánh giá nhằm xác định mức độ hiệu quả của các mô hình. Sau đó, luận án trình bày các kiến thức nền tảng cơ bản liên quan đến học sâu và bài toán tái định danh người, nhằm cung cấp cơ sở lý thuyết cho các nội dung nghiên cứu tiếp theo. Cuối cùng, luận án trình bày tổng quan các hướng nghiên cứu trong lĩnh vực tái định danh người theo ba hướng tiếp cận, từ đó đưa ra khoảng trống nghiên cứu để làm cơ sở để xác định nội dung nghiên cứu và hướng tiếp cận cụ thể để hướng tới mục tiêu nâng cao hiệu năng tái định danh người.

## 1.1 Giới thiệu bài toán tái định danh người

### 1.1.1 Phát biểu bài toán

Tái định danh người là bài toán nhằm xác định và nhận diện lại danh tính cụ thể của một người qua tập hình ảnh thu được từ nhiều camera. Hình ảnh thu được này thường từ các hệ thống camera giám sát, trong điều kiện có sự thay đổi về góc nhìn, ánh sáng, bối cảnh xung quanh. Các hình ảnh này thường không có độ phân giải cao, từ đó không có các thông tin nhận diện rõ ràng như khuôn mặt, hay các đặc điểm sinh trắc học khác. Chúng ta thường xác định danh tính của các hình ảnh này thông qua khai thác các đặc trưng toàn thân như hình dáng, kích thước, hay các đặc điểm về màu sắc của quần áo, hoặc các đặc trưng cụ thể như giới tính, kiểu tóc hay phụ kiện. Tất cả các thông tin hữu ích tùy trường hợp cụ thể sẽ được sử dụng để tăng hiệu quả tái định danh. Tuy nhiên, bài toán này đặc biệt thách thức do sự biến thiên rất lớn về ngoại hình của cùng một người trong các tình huống khác nhau: thay đổi ánh sáng, tư thế, góc nhìn, che khuất, hoặc thậm chí là thay đổi quần áo.

Hệ thống tái định danh người đầy đủ được minh họa trong Hình 1.1. Trước hết, hình ảnh được các camera giám sát thu thập liên tục từ môi trường thực tế, sau đó được xử lý bằng các mô hình phát hiện người để khoanh vùng đối tượng trong từng

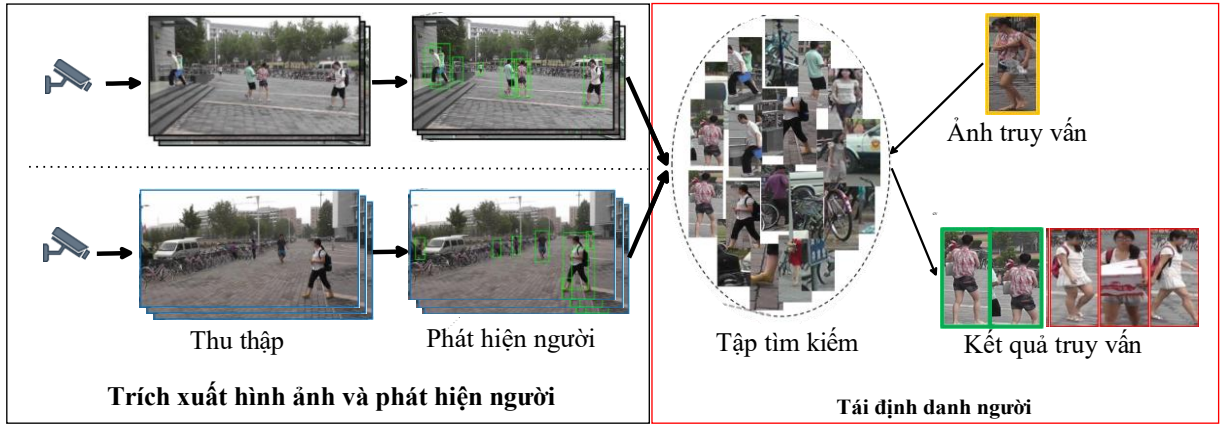

Hình 1.1: Hệ thống tái định danh người đầy đủ.

khung hình. Các vùng ảnh chứa người sau đó sẽ được trích xuất và lưu trữ trong tập tìm kiếm. Khi có một ảnh truy vấn, hệ thống sẽ tiến hành trích xuất đặc trưng và so sánh với các ảnh trong tập tìm kiếm nhằm tìm ra những mẫu có độ tương đồng cao nhất. Kết quả trả về là một danh sách xếp hạng các ứng viên, từ mức độ giống cao đến thấp, qua đó hỗ trợ việc nhận diện lại danh tính hoặc xác định vị trí xuất hiện của đối tượng trong môi trường giám sát.

Tái định danh người có thể được phát biểu dưới dạng bài toán như sau. Cho một tập dữ liệu tìm kiếm $G = \{g_j\}_{j=1}^{N}$, bao gồm $N$ ảnh đại diện cho $M$ danh tính khác nhau. Mỗi ảnh $g_j$ trong tập tìm kiếm tương ứng với một danh tính cụ thể có nhãn định danh đã biết. Đối với một truy vấn $q$, mục tiêu của bài toán là truy xuất ra một danh sách các ảnh trong $G$ có cùng danh tính với $q$, được sắp xếp theo độ tương đồng giảm dần. Hàm mục tiêu của bài toán được biểu diễn trong Công thức 1.1.

$$R(q, G) = \text{Sort}\left(\{g_j\}_{j=1}^{N} \mid \text{sim}(q, g_j)\right) \tag{1.1}$$

trong đó, $\text{sim}(q, g_j)$ là hàm đo độ tương đồng giữa ảnh truy vấn $q$ và ảnh $g_j$ trong tập tìm kiếm. Hàm này có thể là độ đo cosine, khoảng cách Euclid hoặc các hàm đo phổ biến khác trong không gian đặc trưng; $R(q, G)$ là danh sách các ảnh trong tập tìm kiếm được xếp hạng theo mức độ tương đồng với ảnh truy vấn.

Trong nhiều ứng dụng thực tế, chúng ta chỉ quan tâm đến $k$ kết quả có độ tương đồng cao nhất (truy xuất top-$k$). Khi đó, hàm mục tiêu được điều chỉnh như trong Công thức 1.2.

$$R_k(q, G) = \text{Top}_k\left(R(q, G)\right) \tag{1.2}$$

Mục tiêu cuối cùng là đảm bảo rằng các ảnh có cùng danh tính với ảnh truy vấn sẽ xuất hiện ở các vị trí đầu tiên trong danh sách $R_k(q_i, G)$. Kết quả này là cơ sở để đánh giá hiệu quả của hệ thống tái định danh người thông qua các độ đo phổ biến như mAP và Rank-n (chi tiết ở Mục 1.1.7).

## 1.1.2   Phương án so khớp hình ảnh

Để thực hiện so khớp giữa ảnh truy vấn và các ảnh trong tập tìm kiếm, một số phương pháp đơn giản có thể được áp dụng dựa trên so sánh trực tiếp giá trị điểm ảnh hoặc đánh giá mức độ tương đồng cục bộ giữa các vùng ảnh. Ví dụ, phương pháp tương quan chéo chuẩn hóa (NCC) [10] đo mức độ tương quan giữa hai vùng ảnh sau khi chuẩn hóa, với giá trị tương quan càng cao cho thấy mức độ tương đồng càng lớn. Phương pháp tổng bình phương sai khác (SSD) [40] tính tổng bình phương sai khác giữa các điểm ảnh tương ứng, với giá trị càng nhỏ cho thấy hai ảnh càng giống nhau. Trong khi đó, phương pháp chỉ số tương đồng về cấu trúc (SSIM) [112] đánh giá mức độ tương đồng về mặt cấu trúc, độ sáng và độ tương phản, với giá trị càng gần một thể hiện mức độ tương đồng càng cao.

Mặc dù các phương pháp này có ưu điểm là đơn giản và tính toán trực tiếp trên không gian ảnh, chúng lại thiếu tính khái quát và không bền vững khi áp dụng trong điều kiện thực tế, nơi ảnh bị thay đổi bởi ánh sáng, góc nhìn, phông nền hoặc bị che khuất. Do đó, trong các hệ thống tái định danh hiện đại, thay vì so sánh trực tiếp các điểm ảnh, chúng ta ưu tiên so sánh các đặc trưng được trích xuất từ các hình ảnh. Đặc trưng được trích xuất sẽ đại diện cho thông tin nhận dạng có mức trừu tượng cao hơn, giúp mô hình so khớp giữa ảnh truy vấn và ảnh trong tập tìm kiếm một cách ổn định, hiệu quả và có khả năng tổng quát tốt hơn trong các điều kiện môi trường khác nhau.

Vì vậy, trích xuất đặc trưng đóng vai trò trung tâm trong việc xây dựng một không gian véc-tơ đặc trưng có tính phân biệt cao. Hiện nay, thay vì dựa vào các đặc trưng thủ công như HOG [24], hay màu biểu đồ [49] vốn dễ bị ảnh hưởng bởi nhiễu và điều kiện ngoại cảnh, các mạng nơ-ron học sâu như CNN hoặc ViT có khả năng học trực tiếp các đặc trưng nhận dạng từ dữ liệu, nhờ đó trích xuất được các biểu diễn mạnh mẽ và ổn định hơn.

Trong cách tiếp cận học sâu, một mạng cốt lõi sẽ được sử dụng để nhận ảnh người đầu vào và ánh xạ thành một véc-tơ đặc trưng trong không gian nhiều chiều. Các đặc trưng này có khả năng nắm bắt các thông tin nhận dạng cốt lõi như hình dáng cơ thể, kiểu trang phục, màu sắc, hoặc các phụ kiện đặc trưng, trong khi vẫn duy trì tính bất biến trước các thay đổi như góc nhìn, ánh sáng hoặc phông nền. Việc so khớp giữa các ảnh người sau đó được thực hiện bằng cách tính khoảng cách hoặc độ tương đồng giữa các véc-tơ đặc trưng.

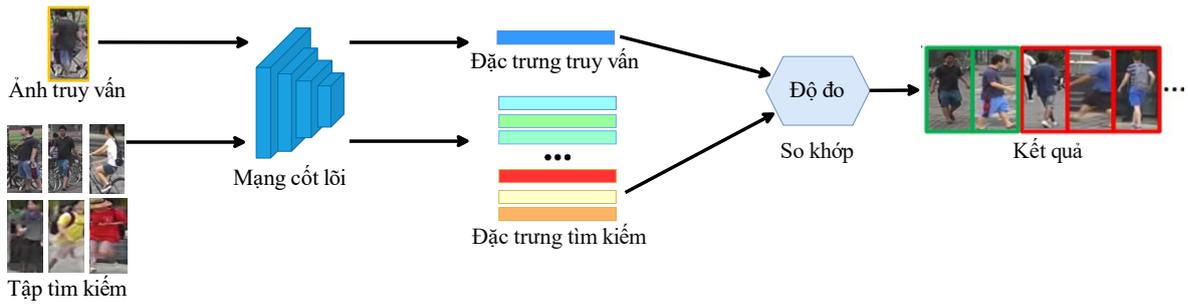

Hình 1.2: Quá trình trích xuất đặc trưng và so khớp.

Hình 1.2 mô hình hóa cụ thể cách thức trích xuất đặc trưng học sâu này. Quá trình bắt đầu với một ảnh truy vấn và một tập tìm kiếm gồm nhiều ảnh người được thu thập từ các camera giám sát khác nhau. Cả ảnh truy vấn và ảnh trong tập tìm kiếm đều được đưa vào mạng cốt lõi để trích xuất các véc-tơ đặc trưng thể hiện thông tin thị giác của từng ảnh. Sau khi thu được đặc trưng truy vấn và đặc trưng của các ảnh trong tập tìm kiếm, hệ thống tiến hành tính toán khoảng cách (độ tương đồng) giữa chúng thông qua một hàm đo như khoảng cách Euclid hoặc độ tương tự cosine. Kết quả của bước so khớp là một danh sách các ảnh trong tập tìm kiếm được sắp xếp theo mức độ tương đồng giảm dần với ảnh truy vấn. Ảnh có độ tương đồng cao – thể hiện khả năng trùng khớp cao – được ưu tiên xếp lên đầu. Để đánh giá xem danh sách này có tốt hay không luận án sử dụng các độ đo mAP và Rank-n. Các độ đo này được trình bày chi tiết trong Mục 1.1.7.

Để mạng cốt lõi có thể trích xuất được đặc trưng phân biệt giữa các định danh khác nhau, chúng ta cần huấn luyện mạng này. Quá trình huấn luyện nhằm điều chỉnh các tham số của mạng, giúp mạng học được cách biểu diễn đặc trưng sao cho hai ảnh thuộc cùng một định danh sẽ có đặc trưng nằm gần nhau, trong khi đặc trưng của hai ảnh thuộc các định danh khác nhau sẽ nằm cách xa nhau trong không gian đặc trưng. Việc huấn luyện này được thực hiện theo ba hướng tiếp cận chính đã được trình bày ở phần trước là học có giám sát, học thích ứng miền không giám sát và học không giám sát.

### 1.1.3   Học có giám sát

Học có giám sát là một trong những hướng tiếp cận truyền thống và hiệu quả nhất trong bài toán tái định danh người. Trong đó, mô hình được huấn luyện dựa trên một tập dữ liệu có nhãn đầy đủ (mỗi ảnh đều gắn với một danh tính cụ thể). Quá trình huấn luyện theo cách tiếp cận học có giám sát được trình bày ở Hình 1.3. Cụ thể, quá trình huấn luyện bắt đầu từ dữ liệu đầu vào là các ảnh từ bộ dữ liệu đã

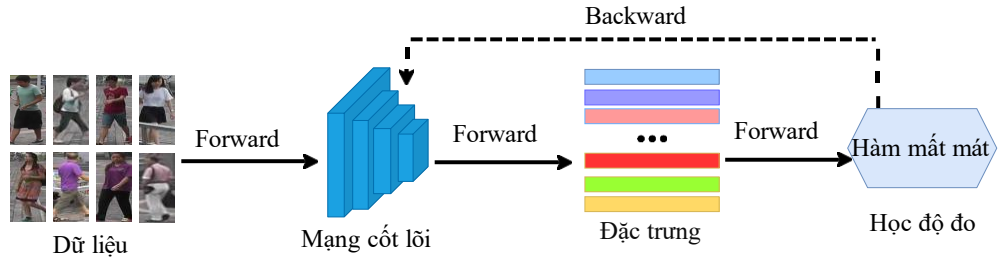

Hình 1.3: Quá trình huấn luyện có giám sát.

được gán nhãn danh tính. Các ảnh này được đưa vào mạng trích xuất đặc trưng, với mục tiêu ánh xạ mỗi ảnh thành một véc-tơ đặc trưng trong không gian biểu diễn. Các véc-tơ này mang thông tin nhận dạng quan trọng, phục vụ cho việc phân biệt giữa các đối tượng. Sau đó, các véc-tơ đặc trưng này được đưa vào mô-đun học độ đo sâu (DML), nơi sử dụng các hàm mất mát như phân loại hoặc bộ ba. Các hàm mất mát này có vai trò đánh giá mức độ sai lệch giữa các đặc trưng được truy xuất và nhãn danh tính tương ứng, hoặc mức độ tương đồng giữa các đặc trưng, từ đó phản ánh khả năng phân biệt của mô hình. Thông qua quá trình lan truyền ngược, các tham số của mạng được cập nhật dựa trên độ dốc của hàm mất mát theo các tham số này, nhằm tối ưu hóa việc học các đặc trưng có tính phân biệt. Nhờ quá trình này, mô hình học sâu dần học được các đặc trưng có tính phân biệt, giúp nhận diện chính xác một danh tính dù xuất hiện ở nhiều camera với điều kiện khác nhau. Đây chính là nền tảng cơ sở để xây dựng một hệ thống tái định danh người hiệu quả và tổng quát hóa tốt.

**Mô hình hóa:** Mục tiêu của hướng tiếp cận này là cải thiện hiệu năng của mô hình tái định danh người $f_\theta$ được huấn luyện và đánh giá trên một tập dữ liệu đã gán nhãn.

Cụ thể, với quá trình huấn luyện cho tập dữ liệu đã gán nhãn thuộc được ký hiệu là D = $\{x_i, y_i\}^N_{i=1}$ trong đó $\{x_i, y_i\}$ là mẫu dữ liệu thứ $i$ gồm ảnh $x_i$ và nhãn danh tính $y_i$, còn N là tổng số mẫu trong tập dữ liệu. Với quá trình truy vấn (đánh giá mô hình), tập dữ liệu dùng để đánh giá là tập tìm kiếm $G$ có cùng nguồn gốc (nghĩa là các ảnh thu được từ cùng tập camera giám sát) với tập dữ liệu huấn luyện. Dựa trên các định nghĩa này, bài toán tái định danh người trong hướng tiếp cận có giám sát được mô hình hóa trong Công thức 1.3.

$$score = F_{eval}\left(F_{SL}(f_\theta, D), G\right) \qquad (1.3)$$

trong đó, $F_{SL}$ và $F_{eval}$ lần lượt biểu thị quá trình huấn luyện có giám sát và quá trình truy vấn. Ngoài ra, *score* là giá trị đại diện cho hiệu năng của mô hình. Trong phạm vi nghiên cứu này, giá trị *score* chính là các chỉ số đánh giá mAP và Rank-n, trong đó giá trị càng cao tương ứng với hiệu quả mô hình càng tốt. Do đó, mục tiêu trong hướng tiếp cận này là tối đa hóa giá trị *score*.

Học có giám sát cho kết quả cao nhưng phương pháp này có một số hạn chế đáng kể. Việc thu thập và gán nhãn dữ liệu định danh người là quá trình rất tốn kém và không khả thi ở quy mô lớn. Bên cạnh đó, mô hình huấn luyện theo cách này thường khó khái quát hóa sang môi trường mới, đặc biệt là khi xuất hiện các người ở trong các camera chưa từng thấy trong quá trình huấn luyện. Điều này hạn chế khả năng ứng dụng thực tế của các mô hình học có giám sát trong hệ thống giám sát nhiều camera ngoài đời thực. Để khắc phục những hạn chế trên, nhiều nghiên cứu đã chuyển sang hướng tiếp cận học thích ứng miền không giám sát trong bài toán tái định danh người. Phương pháp này nhằm giảm sự phụ thuộc vào dữ liệu được gán nhãn, đồng thời tăng khả năng thích ứng với sự thay đổi của môi trường và hạn chế hiện tượng thiên lệch dữ liệu. Hơn nữa, học thích ứng miền không giám sát còn giúp cải thiện khả năng khái quát hóa của hệ thống, đặc biệt trong các kịch bản và miền dữ liệu chưa từng gặp trước đó.

## 1.1.4   Học thích ứng miền không giám sát

Học thích ứng miền không giám sát là một hướng tiếp cận quan trọng trong bài toán tái định danh người, nhằm khắc phục hiện tượng suy giảm hiệu năng khi mô hình được chuyển từ miền nguồn sang miền đích có điều kiện ánh sáng, góc nhìn, môi trường và thiết bị camera khác biệt. Phương pháp này tận dụng mô hình đã được huấn luyện có giám sát trên miền nguồn để áp dụng hiệu quả trong miền đích – nơi dữ liệu không có nhãn. Điều này phản ánh nhu cầu thực tế khi triển khai hệ thống tái định danh trên các mạng lưới camera mới.

Bài toán tái định danh người trong thiết lập thích ứng miền không giám sát được giải quyết thông qua hai giai đoạn như thể hiện trong Hình 1.4. Giai đoạn I, mô hình tái định danh được huấn luyện trước bằng dữ liệu có gán nhãn trên miền nguồn. Dữ liệu này được đưa qua mạng cốt lõi để trích xuất đặc trưng, sau đó các đặc trưng được chuyển đến mô-đun học độ đo sâu, nhằm học không gian đặc trưng có tính phân biệt cao thông qua các hàm mất mát. Sau khi hoàn tất giai đoạn này, trọng số tối ưu của mạng cốt lõi sẽ được dùng để khởi tạo mô hình trong giai đoạn kế tiếp.

Sang giai đoạn II, mô hình đã được khởi tạo sẽ xử lý dữ liệu không nhãn tại miền đích. Tại giai đoạn này, quá trình thường trải qua hai bước là phân cụm gán nhãn giả và huấn luyện. Ở bước đầu tiên (mũi tên màu đỏ), ảnh không nhãn được đưa qua mạng cốt lõi để trích xuất đặc trưng. Những véc-tơ đặc trưng thu được sẽ được gom

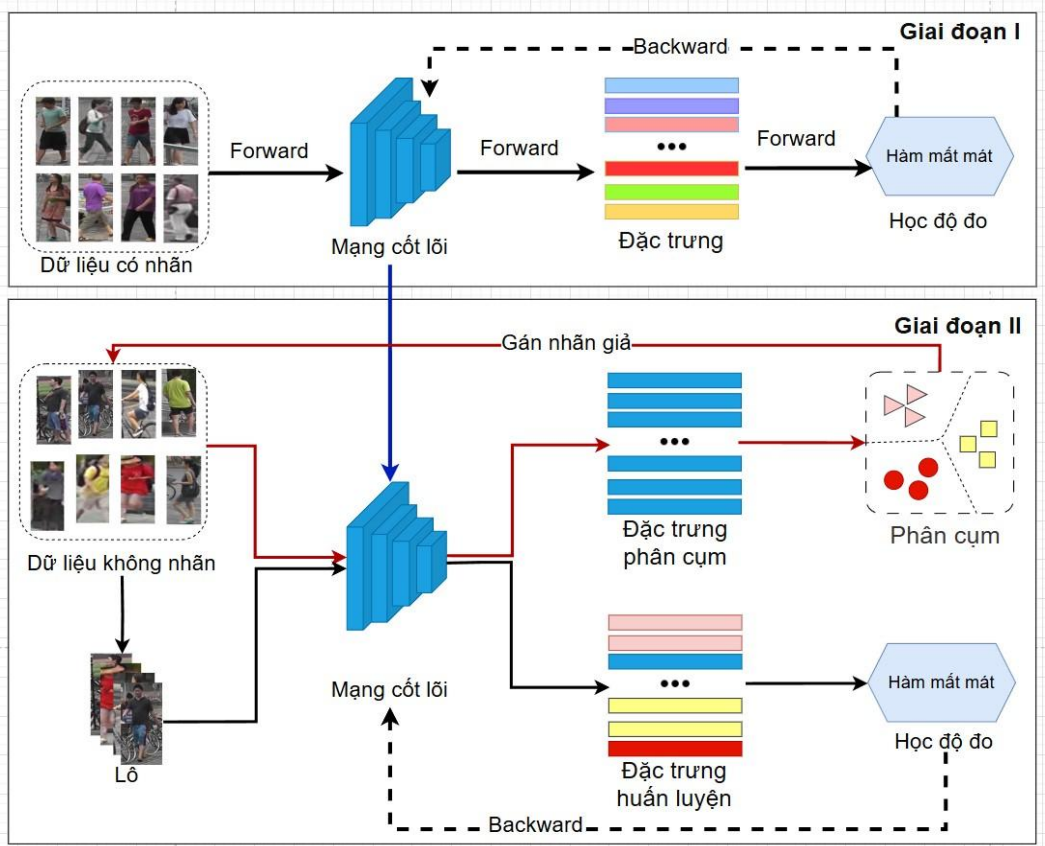

Hình 1.4: Kiến trúc chung hướng tiếp cận học thích ứng miền không giám sát.

nhóm bằng thuật toán phân cụm (K-means [71] hoặc DBSCAN [29]) từ đó tạo ra các nhãn định danh giả định. Các nhãn giả này được gán lại cho tập dữ liệu không nhãn để biến ảnh không nhãn thành ảnh có nhãn. Sau đó, mô hình tiếp tục được huấn luyện theo cách tương tự như trong thiết lập học có giám sát, nhưng sử dụng nhãn giả thay vì nhãn thật (mũi tên màu đen). Quá trình phân cụm và huấn luyện này có thể được lặp lại nhiều lần, nhằm liên tục cải thiện chất lượng của các nhãn giả và tăng độ chính xác của mô hình.

**Mô hình hóa:** Mục tiêu của hướng tiếp cận này là sử dụng mô hình $f_\theta$ được huấn luyện trên một bộ dữ liệu có nhãn (miền nguồn) và học các đặc trưng mới trên một bộ dữ liệu khác không có nhãn (miền đích) để đánh giá kết quả trên tập dữ liệu kiểm tra ở miền đích. Cụ thể, chúng tôi giả định rằng bộ dữ liệu có nhãn là $D_s = \{x_i^s, y_i^s\}_{i=1}^{N_s}$ trong đó $(x_i^s, y_i^s)$ là mẫu dữ liệu có nhãn thứ $i$, $N_s$ là số lượng mẫu trong tập dữ liệu miền nguồn. Tương tự, bộ dữ liệu không nhãn $D_t = \{x_i^t\}_{i=1}^{N_t}$ ở miền đích, trong đó $x_i^t$ và $N_t$ tương ứng với mẫu không có nhãn thứ $i$ và tổng số dữ liệu. Tập dữ liệu dùng để đánh giá là tập tìm kiếm $G_t$ có cùng nguồn gốc (nghĩa là các ảnh thu được từ cùng tập camera giám sát) với tập dữ liệu miền đích. Dựa trên các định nghĩa này, bài toán tái định danh người trong hướng tiếp cận thích ứng miền không giám sát được mô hình hóa trong Công thức 1.4.

$$score = F_{eval}(F_{USL} (F_{SL}(f_\theta, D_s), D_t), G_t) \quad (1.4)$$

trong đó, $F_{SL}$, $F_{USL}$ và $F_{eval}$ lần lượt biểu thị quá trình huấn luyện có giám sát trên miền nguồn, huấn luyện không giám sát trên miền đích và quá trình truy vấn. Ngoài ra, *score* là giá trị đại diện cho hiệu năng của mô hình trên miền đích. Trong phạm vi nghiên cứu này, giá trị *score* chính là các chỉ số đánh giá mAP và Rank-n, trong đó giá trị càng cao tương ứng với hiệu quả mô hình càng tốt. Do đó, mục tiêu của bài toán theo hướng tiếp cận này vẫn là tối đa hóa giá trị *score*.

Mặc dù học thích ứng miền không giám sát giúp giảm đáng kể sự phụ thuộc vào dữ liệu có gán nhãn và cải thiện khả năng thích ứng khi triển khai mô hình sang miền mới, phương pháp này vẫn tồn tại một số hạn chế nhất định. Cụ thể, nó thường yêu cầu một mô hình được huấn luyện trước trên miền nguồn có gán nhãn, và hiệu quả thích ứng miền phụ thuộc vào mức độ tương đồng giữa miền nguồn và miền đích. Trong những trường hợp không có sẵn mô hình tiền huấn luyện phù hợp, hoặc khi hai miền có sự khác biệt lớn về phân phối dữ liệu, phương pháp này sẽ không còn phát huy hiệu quả. Để khắc phục các hạn chế này, nhiều nghiên cứu gần đây đã tập trung vào hướng tiếp cận học không giám sát hoàn toàn trong bài toán tái định danh người. Phương pháp này không yêu cầu bất kỳ dữ liệu có nhãn nào trong cả miền nguồn lẫn miền đích, giúp loại bỏ hoàn toàn chi phí gán nhãn và nâng cao khả năng triển khai thực tế. Học không giám sát hướng đến việc khai thác tối đa cấu trúc nội tại của dữ liệu, từ đó xây dựng mô hình có khả năng phân biệt tốt mà không cần thông tin định danh ban đầu.

## 1.1.5 Học không giám sát

Phương pháp học không giám sát trong tái định danh người không yêu cầu gán nhãn định danh người trong tập dữ liệu, từ đó giúp giảm đáng kể chi phí thu thập, gán nhãn so với phương pháp có giám sát và tăng khả năng ứng dụng trong thực tế.

Quá trình huấn luyện theo cách tiếp cận học không giám sát được trình bày ở Hình 1.5. Cụ thể, quá trình huấn luyện bắt đầu từ với tập dữ liệu ảnh chưa được gán nhãn. Các ảnh này được đưa vào một mạng cốt lõi (CNN hoặc ViT) để trích xuất đặc trưng, biểu diễn dưới dạng các véc-tơ đặc trưng. Các véc-tơ này sau đó được gom nhóm bằng các thuật toán như K-means [71] hoặc DBSCAN [29] để tạo ra các cụm, mỗi cụm đại diện cho một danh tính giả định – gọi là nhãn giả. Các nhãn giả này sẽ được gán ngược lại cho ảnh gốc, cho phép sử dụng chúng như là nhãn tạm thời để huấn luyện mô hình. Tiếp theo, mạng cốt lõi sẽ được huấn luyện

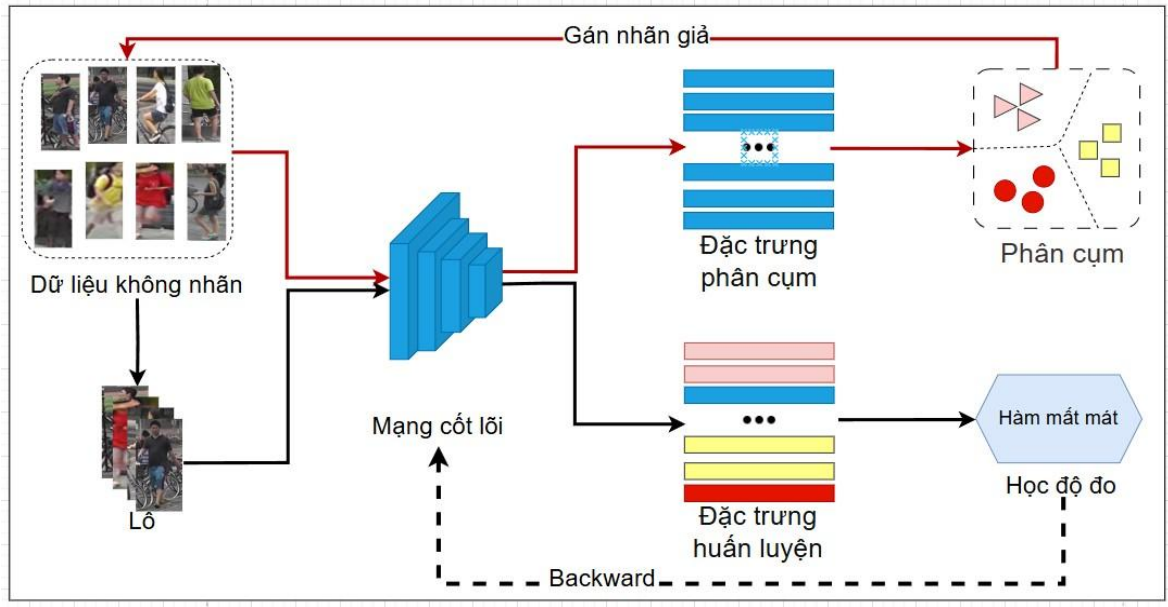

Hình 1.5: Kiến trúc chung hướng tiếp cận học không giám sát.

như trong thiết lập có giám sát với các hàm mất mát như phân loại hoặc InfoNCE [81]. Quy trình này được lặp lại nhiều lần: sau mỗi epoch, đặc trưng được cập nhật, phân cụm được thực hiện, nhãn giả được làm mới và mô hình được huấn luyện lại với thông tin cập nhật. Mặc dù phương pháp này giúp huấn luyện tái định danh mà không cần nhãn thật, nó cũng tiềm ẩn nhiều thách thức như do nhiễu từ nhãn giả, và do đó có thể kết hợp thêm các kỹ thuật như EMA, đo độ tin cậy của nhãn giả, hoặc học có thông tin định danh camera để nâng cao độ chính xác.

**Mô hình hóa:** Mục tiêu của hướng tiếp cận này là sử dụng mô hình $f_\theta$ được huấn luyện trên một bộ dữ liệu không có nhãn. Cụ thể, chúng tôi giả định rằng bộ dữ liệu không nhãn $D = \{x_i\}_{i=1}^{N}$ trong đó $x_i$ là mẫu dữ liệu không có nhãn thứ $i$, $N$ là số lượng mẫu trong tập dữ liệu. Với quá trình truy vấn, tập dữ liệu dùng để đánh giá là tập tìm kiếm $G$ có cùng nguồn gốc với tập dữ liệu huấn luyện. Dựa trên các định nghĩa này, bài toán tái định danh người trong hướng tiếp cận không giám sát được mô hình hóa trong Công thức 1.5.

$$score = \mathrm{F}_{eval}\left(\mathrm{F}_{USL}(f_\theta, \mathrm{D}), G\right) \tag{1.5}$$

trong đó, $\mathrm{F}_{USL}$ và $\mathrm{F}_{eval}$ lần lượt biểu thị quá trình huấn luyện không giám sát và quá trình truy vấn. Ngoài ra, $score$ là giá trị đại diện cho hiệu năng của mô hình. Trong phạm vi nghiên cứu, giá trị $score$ chính là các chỉ số đánh giá mAP và Rank-n, trong đó giá trị càng cao tương ứng với hiệu quả mô hình càng tốt. Do đó, mục tiêu của chúng tôi vẫn là tối đa hóa giá trị $score$.

Mặc dù có ưu thế rõ rệt về chi phí và khả năng triển khai trong thực tế, các phương pháp học không giám sát vẫn đối mặt với nhiều thách thức. Cụ thể, việc gán nhãn giả thường thiếu độ chính xác, quá trình phân cụm không cố định, và tốc độ hội tụ thường chậm hơn so với các mô hình học có giám sát. Ngoài ra, mô hình cũng nhạy cảm với nhiễu trong dữ liệu và khó duy trì hiệu năng ổn định trong các tình huống phức tạp, đặc biệt là khi có sự thay đổi lớn về ánh sáng, góc nhìn hoặc bối cảnh giữa các camera.

## 1.1.6 Dữ liệu huấn luyện và đánh giá

Trong nghiên cứu về tái định danh người, việc huấn luyện và đánh giá hiệu quả mô hình thường được thực hiện trên các tập dữ liệu chuẩn đã được công bố rộng rãi trong cộng đồng nghiên cứu. Các tập dữ liệu này khác nhau về quy mô, điều kiện môi trường, số lượng danh tính, và số lượng camera. Bốn bộ dữ liệu phổ biến nhất được sử dụng bao gồm CUHK03 [61], Market-1501 [127], DukeMTMC-ReID [91] và MSMT17 [113].

- **CUHK03**: Bộ dữ liệu CUHK03 được công bố năm 2014, bao gồm tổng cộng 13164 ảnh của 1467 danh tính, được chụp bởi 6 camera giám sát khác nhau tại Đại học Trung văn Hồng Kông (CUHK). Trong đó, 767 danh tính đầu tiên với 7368 ảnh được sử dụng cho giai đoạn huấn luyện. 700 danh tính còn lại được sử dụng cho tập kiểm thử với 1400 ảnh truy vấn và 5328 ảnh trong tập tìm kiếm. Ngoài ra, bộ dữ liệu này còn cung cấp hai tập có nhãn được gán thủ công (CUHK03-Labeled) và nhãn được tạo ra bởi công cụ tự động (CUHK03-Detected).

- **Market-1501**: Bộ dữ liệu Market-1501 bao gồm tổng cộng 32668 ảnh của 1501 danh tính được thu thập trước một siêu thị trong khuôn viên trường đại học. Trong đó, 750 danh tính đầu tiên với 12936 ảnh được sử dụng cho giai đoạn huấn luyện, còn 751 danh tính còn lại với 19732 ảnh được sử dụng cho kiểm thử. Tập kiểm thử này tiếp tục được chia thành 3368 ảnh truy vấn và 15913 ảnh trong tập tìm kiếm.

- **DukeMTMC-reID**: Bộ dữ liệu DukeMTMC-reID bao gồm tổng cộng 36411 ảnh của 1812 danh tính, được thu thập từ 8 camera giám sát đặt tại khuôn viên trường Đại học Duke. Trong đó, 702 danh tính đầu tiên với 16522 ảnh được sử dụng cho giai đoạn huấn luyện, còn 702 danh tính khác với tổng cộng 19889 ảnh được sử dụng cho kiểm thử. Tập kiểm thử này tiếp tục được chia thành 2228 ảnh truy vấn và 17661 ảnh trong tập tìm kiếm.

- **MSMT17**: Bộ dữ liệu MSMT17 bao gồm tổng cộng 126441 ảnh của 4101 danh tính, được thu thập từ hệ thống 15 camera giám sát bao gồm cả camera cố định và di động, trong các điều kiện ánh sáng và môi trường thực tế khác nhau. Trong đó, 1041 danh tính đầu tiên với 32621 ảnh được sử dụng cho huấn luyện. 3060 danh tính còn lại với 93820 ảnh được sử dụng cho kiểm thử, chia thành 11659 ảnh truy vấn và 82161 ảnh trong tập tìm kiếm.

Cụ thể, các thông số của từng bộ dữ liệu được thể hiện trong Bảng 1.1.

Bảng 1.1: Thông số các bộ dữ liệu phổ biến trong bài toán tái định danh người

| Bộ dữ liệu | Công bố | Số lượng ảnh | Số lượng danh tính | Số lượng camera | Tập huấn luyện (ảnh/danh tính) | Tập đánh giá (tìm kiếm/truy vấn) |
|---|---|---|---|---|---|---|
| CUHK03 | 2014 | 13164 | 1467 | 6 | 7368/767 | 5328/1400 |
| Market-1501 | 2015 | 32668 | 1501 | 6 | 12936/750 | 15913/3368 |
| DukeMTMC-reID | 2016 | 36411 | 1812 | 8 | 16522/702 | 17661/2268 |
| MSMT17 | 2018 | 126441 | 4101 | 15 | 32621/1041 | 82161/11659 |

Qua các số liệu thống kê trong Bảng 1.1, có thể nhận thấy rằng CUHK03 là bộ dữ liệu được công bố sớm nhất, với quy mô tương đối nhỏ và gọn, thích hợp cho các nghiên cứu cơ bản hoặc thử nghiệm ban đầu trong bài toán tái định danh người. Market-1501 và DukeMTMC-reID, được công bố trong các năm tiếp theo (2015 và 2016), là những bộ dữ liệu ngoài trời với ảnh được phát hiện tự động từ nhiều camera, phản ánh rõ các yếu tố gây nhiễu như thay đổi góc nhìn, điều kiện ánh sáng và tình trạng che khuất – những đặc điểm thường gặp trong môi trường giám sát thực tế. MSMT17, với quy mô lớn nhất và độ đa dạng cao về điều kiện thu thập, được đánh giá là bộ dữ liệu có độ thách thức cao nhất hiện nay, phù hợp để kiểm chứng khả năng tổng quát hóa và độ ổn định của các mô hình tái định người trong môi trường triển khai thực tế.

Việc sử dụng đồng thời nhiều bộ dữ liệu trong quá trình huấn luyện và kiểm thử không chỉ giúp kiểm tra hiệu suất mô hình ở các cấp độ quy mô khác nhau, mà còn cho phép đánh giá toàn diện về khả năng tổng quát hóa, tính ổn định và khả năng thích ứng của mô hình trong các kịch bản thực tế. Ngoài ra, do đây đều là những bộ dữ liệu chuẩn được cộng đồng nghiên cứu sử dụng rộng rãi, việc thực nghiệm trên chúng còn đảm bảo tính khách quan và khả năng so sánh trực tiếp với các phương pháp hiện có.

Trong luận án này, cách viết giản lược bộ dữ liệu được sử dụng nhằm đảm bảo tính ngắn gọn và nhất quán trong trình bày. Cụ thể, "CUHK" được dùng để chỉ "CUHK03", "Market" để chỉ "Market-1501", "Duke" để chỉ "DukeMTMC-reID", và "MSMT" để chỉ "MSMT17". Đối với các thiết lập học thích ứng miền không giám sát, ký hiệu dạng "Market-to-Duke" được sử dụng để biểu thị quá trình huấn luyện có giám sát trên tập dữ liệu nguồn "Market", huấn luyện không giám sát và đánh giá trên tập dữ liệu đích "Duke". Cách viết tương tự cũng được áp dụng cho các thiết lập như "Duke-to-Market", "Market-to-MSMT" và "Duke-to-MSMT".

Tất cả các bộ dữ liệu sử dụng trong nghiên cứu đều được gán sẵn nhãn danh tính. Tuy nhiên, việc sử dụng các nhãn này phụ thuộc vào phương pháp học cụ thể được áp dụng. Trong thiết lập học có giám sát, nhãn danh tính được sử dụng đầy đủ trong cả quá trình huấn luyện và đánh giá. Đối với học không giám sát, toàn bộ nhãn danh tính đều bị loại bỏ khỏi quá trình huấn luyện. Trong khi đó, với thiết lập học thích ứng miền không giám sát, chỉ sử dụng nhãn danh tính từ miền nguồn, còn nhãn từ miền đích không được khai thác nhằm đảm bảo tính đúng đắn và khách quan trong bài toán thích ứng giữa hai miền.

## 1.1.7 Độ đo đánh giá

Trong bài toán tái định danh người, hiệu năng của mô hình thường được đánh giá dựa trên hai chỉ số chính: Đặc trưng tích lũy theo hạng (CMC) và bình quân độ chính xác trung bình (mAP). Các chỉ số này đánh giá khả năng mô hình truy xuất đúng danh tính từ tập tìm kiếm dựa trên một ảnh truy vấn, xét cả vị trí xếp hạng và mức độ đầy đủ trong việc truy xuất.

### 1.1.7.1 Đặc trưng tích lũy theo hạng

**Định nghĩa:** CMC đo xác suất danh tính của ảnh truy vấn xuất hiện trong $k$ ảnh đầu tiên được xếp hạng từ tập tìm kiếm. Nó phản ánh khả năng của mô hình trong việc nhận diện chính xác danh tính trong số các kết quả đứng đầu.

**Công thức:** Gọi $R_q$ là vị trí của kết quả đúng đầu tiên tương ứng với truy vấn $q$. Khi đó, CMC tại hạng $n$ được tính theo Công thức 1.6.

$$\text{Rank-n} = \frac{1}{N} \sum_{q=1}^{N} \mathbf{1}(R_q \leq n) \tag{1.6}$$

trong đó, $N$ là tổng số ảnh truy vấn và $\mathbf{1}(\cdot)$ là hàm chỉ thị có giá trị bằng 1 nếu $(R_q \leq n)$, bằng 0 trong trường hợp ngược lại.

Độ chính xác tại Rank-1 là chỉ số phổ biến nhất, cho biết ảnh đầu tiên trong tập kết quả đúng với định danh của ảnh truy vấn.

#### 1.1.7.2 Bình quân độ chính xác trung bình

**Định nghĩa:** mAP đánh giá chất lượng xếp hạng bằng cách xét đến vị trí của tất cả các ảnh so khớp đúng trong tập kết quả của một truy vấn. Đây là chỉ số đặc biệt hữu ích trong các tình huống có nhiều ảnh đúng tương ứng với một truy vấn.

**Công thức:** Với một truy vấn $q$, độ chính xác trung bình (AP) được tính theo Công thức 1.7.

$$\text{AP}(q) = \frac{1}{m_q} \sum_{k=1}^{N} P(k) \cdot \text{rel}(k) \tag{1.7}$$

trong đó, $m_q$ là số lượng ảnh đúng với truy vấn $q$, $N$ là số lượng ảnh trong danh sách kết quả truy vấn, $P(k)$ là độ chính xác tại hạng $k$ (tức là số lượng ảnh đúng trong top-k chia cho k) và rel(k) là chỉ số nhị phân có giá trị bằng 1 nếu ảnh tại vị trí $k$ là đúng, ngược lại có giá trị bằng 0.

Bình quân độ chính xác trung bình trên tất cả các truy vấn được tính theo Công thức 1.8.

$$\text{mAP} = \frac{1}{M} \sum_{q=1}^{M} \text{AP}(q) \tag{1.8}$$

trong đó, $M$ là tổng số lượng ảnh truy vấn.

Chỉ số mAP xem xét cả sự hiện diện và vị trí xếp hạng của tất cả các ảnh định danh đúng, giúp đánh giá hiệu năng mô hình một cách toàn diện, đặc biệt trong các bài toán truy xuất nhiều ảnh.

## 1.2 Kiến thức nền tảng

### 1.2.1 Mạng cốt lõi trích xuất đặc trưng

Đây là phần đầu tiên và quan trọng nhất trong kiến trúc mạng nơ-ron học sâu. Nhiệm vụ của mạng cốt lõi này là ánh xạ dữ liệu thô thành một đặc trưng đại diện có ý nghĩa để dùng cho các tác vụ như phân loại, phát hiện, nhận diện. Luận án sẽ giới thiệu một số mạng cốt lõi nổi tiếng hay được sử dụng.

#### 1.2.1.1 Mạng cốt lõi CNN

**LeNet-5**: được giới thiệu bởi Yann LeCun và cộng sự vào năm 1998 [58], Lenet-5 là một trong những mạng CNN đầu tiên và có ảnh hưởng sâu rộng. LeNet-5 là một mạng tương đối nhỏ, nhưng tích hợp đầy đủ các thành phần cốt lõi của kiến trúc học sâu: lớp tích chập, lớp gộp và lớp kết nối đầy đủ. Nhờ khả năng trích xuất và học được các đặc trưng phân cấp từ dữ liệu hình ảnh, LeNet-5 hoạt động rất hiệu quả trong việc nhận dạng ký tự viết tay. Mô hình này đã đặt nền móng cho nhiều hệ thống thị giác máy tính hiện đại ngày nay.

**AlexNet**: AlexNet là một trong những mạng CNN tiên phong, được phát triển bởi Alex Krizhevsky và cộng sự tại Đại học Toronto [57]. Mạng đã tạo nên tiếng vang lớn khi giành chiến thắng trong cuộc thi nhận dạng hình ảnh quy mô lớn (ILSVRC) năm 2012, với độ chính xác vượt trội so với các mô hình trước đó. AlexNet được xem là bước ngoặt quan trọng đưa học sâu trở thành trung tâm trong lĩnh vực thị giác máy tính. Với kiến trúc gồm 8 lớp (năm lớp tích chập và ba lớp kết nối đầy đủ), 62 triệu tham số, và các cải tiến như hàm kích hoạt ReLU, Dropout để giảm quá khớp, cũng như huấn luyện song song trên GPU, AlexNet đã chứng minh khả năng xử lý và học hiệu quả từ dữ liệu hình ảnh quy mô lớn.

**VGGNet**: VGGNet là một kiến trúc mạng CNN nổi tiếng do nhóm hình học thị giác tại Đại học Oxford phát triển năm 2014 [93]. Mạng này nổi bật với thiết kế đơn giản, đồng nhất khi chỉ sử dụng các lớp tích chập 3x3 và gộp cực đại 2x2, xếp chồng để tăng độ sâu. Các biến thể phổ biến như VGG16 và VGG19 lần lượt có 16 và 19 lớp có trọng số. VGGNet đạt hiệu suất cao trong nhận dạng ảnh và từng đạt vị trí thứ hai trong cuộc thi ImageNet 2014. Tuy nhiên, mô hình có số lượng tham số rất lớn (khoảng 138 triệu đối với VGG16), đòi hỏi tài nguyên tính toán cao và dễ bị quá khớp nếu không xử lý tốt. Dù vậy, VGGNet vẫn được sử dụng rộng rãi trong thị giác máy tính và học chuyển giao.

**ResNet**: ResNet là một kiến trúc mạng CNN được giới thiệu năm 2015 nhằm giải quyết vấn đề suy giảm hiệu suất khi tăng số lớp mạng [36]. Điểm nổi bật của ResNet là sử dụng các khối dư với kết nối tắt, cho phép mô hình học phần dư thay vì toàn bộ hàm cần tối ưu. Nhờ đó, ResNet có thể huấn luyện các mạng rất sâu như ResNet-50, ResNet-101 hay ResNet-152 mà vẫn duy trì hiệu suất cao. Kiến trúc này đã giành giải nhất trong cuộc thi ImageNet 2015 và trở thành nền tảng cho nhiều mô hình thị giác máy tính hiện đại.

**Inception**: Inception là một kiến trúc mạng CNN do Google phát triển [97], nổi bật với thiết kế các Inception module – kết hợp nhiều loại tích chập và gộp trong cùng một khối để trích xuất đặc trưng ở nhiều mức độ. Mô hình sử dụng tích chập 1x1 để giảm số chiều và tối ưu tính toán, giúp đạt hiệu quả cao với ít tham số hơn. Inception đã giành chiến thắng trong cuộc thi ImageNet 2014 dưới tên GoogLeNet và tiếp tục được cải tiến qua các phiên bản như Inception v3 và Inception-ResNet.

**MobileNet**: MobileNet là một kiến trúc mạng nơ-ron nhẹ do Google phát triển nhằm tối ưu cho thiết bị di động và nhúng [43]. Mô hình sử dụng kỹ thuật tích chập phân tách theo chiều sâu để giảm mạnh số tham số và chi phí tính toán mà vẫn giữ hiệu suất tốt. Các phiên bản từ MobileNetV1 đến MobileNetV3 liên tục cải tiến về độ chính xác và hiệu quả, khiến MobileNet trở thành lựa chọn phổ biến cho các ứng dụng thị giác máy tính trên thiết bị có tài nguyên hạn chế.

**EfficientNet**: EfficientNet là một kiến trúc mạng nơ-ron hiệu quả cao do Google phát triển [98], nổi bật với kỹ thuật chia tỉ lệ phức hợp giúp mở rộng đồng thời độ sâu, độ rộng và độ phân giải của mạng. Mô hình sử dụng MBConv kết hợp với khối SE để tối ưu hiệu suất, cho phép đạt độ chính xác cao với số tham số và chi phí tính toán thấp. EfficientNet đã nhanh chóng trở thành lựa chọn phổ biến trong nhiều ứng dụng thị giác máy tính hiện đại.

**ResNeSt**: ResNeSt là một biến thể nâng cao của ResNet [122], kết hợp cơ chế tập trung, giúp mạng tập trung tốt hơn vào các đặc trưng quan trọng trong ảnh. Nhờ cấu trúc attention phân nhánh, ResNeSt đạt hiệu suất cao trong nhiều nhiệm vụ thị giác máy tính mà không làm tăng đáng kể độ phức tạp mô hình, trở thành lựa chọn mạnh mẽ thay thế cho ResNet trong nhiều hệ thống hiện đại.

**CBNetV2**: CBNetV2 là một kiến trúc mạng cốt lõi tổng hợp được thiết kế để nâng cao khả năng trích xuất đặc trưng cho các nhiệm vụ như phát hiện và phân đoạn ảnh [63]. Mô hình kết hợp nhiều mạng cốt lõi giống nhau theo cấu trúc kết hợp, trong đó các mạng cốt lõi phụ hỗ trợ mạng cốt lõi chính thông qua các module tổng hợp. CBNetV2 cải thiện hiệu suất đáng kể mà vẫn duy trì tính linh hoạt và khả năng tích hợp với các kiến trúc mạng cốt lõi phổ biến hiện nay như ResNet hoặc Swin Transformer.

Trong phạm vi nghiên cứu của luận án, Resnet-50 là mạng cốt lõi đầu tiên với khoảng 25,6 triệu tham số được lựa chọn cho bài toán tái định danh người vì nó đã chứng minh được hiệu quả cao trong các nghiên cứu trước đây, đặc biệt với các mô hình như BoT [68], SpCL [32], SCL [54]. Resnet-50 có kiến trúc sâu vừa phải, dễ dàng tinh chỉnh, có khả năng trích xuất đặc trưng mạnh mẽ và dễ tích hợp vào các phương pháp cải tiến khác như tập trung. Đây cũng là mạng cốt lõi chuẩn trong đa số các nghiên cứu về tái định danh người hiện nay, giúp thuận tiện và khách quan trong việc so sánh và đánh giá kết quả.

### 1.2.1.2 Mạng cốt lõi ViT

Các phương pháp tiếp cận giải quyết bài toán tái định danh phụ thuộc rất nhiều vào mạng CNN do khả năng mạnh mẽ của chúng trong việc trích xuất các đặc điểm trực quan cục bộ và phân cấp không gian. Tuy nhiên, CNN thường gặp khó khăn trong việc nắm bắt các đặc điểm có mối liên kết sâu và thông tin ngữ cảnh toàn cục,

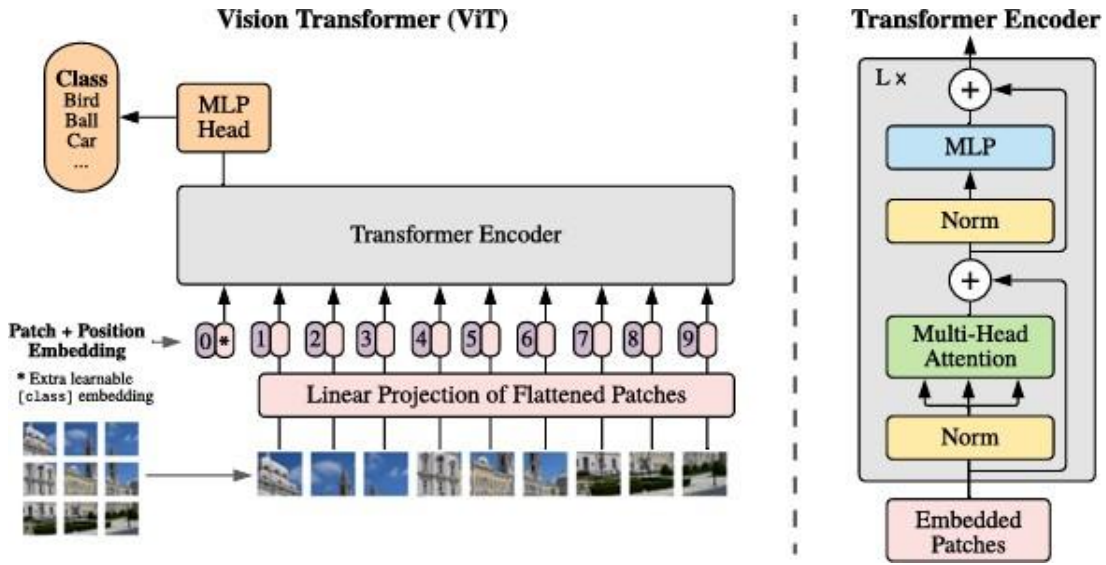

Hình 1.6: Kiến trúc mạng cốt lõi ViT [28].

vốn rất quan trọng trong việc xử lý các thách thức về hình dáng, tư thế, độ sáng và độ che khuất. Nhằm khắc phục những hạn chế trên, Vision Transformer (ViT) đã được Dosovitskiy và các cộng sự [28] đề xuất như một hướng tiếp cận mới, tận dụng sức mạnh của cơ chế tự chú ý trong kiến trúc Transformer – vốn rất thành công trong lĩnh vực xử lý ngôn ngữ tự nhiên. ViT cho phép mô hình hóa quan hệ toàn cục giữa các vùng ảnh một cách trực tiếp, đồng thời bỏ qua hoàn toàn việc sử dụng các tầng tích chập. Nhờ đó, ViT có khả năng học các đặc trưng mạnh mẽ hơn, đặc biệt khi được huấn luyện trên các tập dữ liệu lớn. Các mô hình ViT đã đạt được kết quả vượt trội trên nhiều đánh giá như ImageNet, và cũng được ứng dụng thành công trong các bài toán nhận dạng khuôn mặt, phân đoạn ảnh và tái định danh người.

Mô hình ViT cụ thể được trình bày trong Hình 1.6. Cụ thể, mô hình sử dụng kiến trúc Transformer tiêu chuẩn để xử lý ảnh đầu vào bằng cách chia ảnh thành các ô nhỏ nhỏ có kích thước cố định. Ảnh $x \in R^{H \times W \times C}$ được chia thành $N = \frac{HW}{P^2}$ ô, mỗi ô sau đó được làm phẳng và ánh xạ sang không gian véc-tơ có kích thước cố định $D$ thông qua một lớp tuyến tính học được. Đây gọi là quá trình xây dựng đặc trưng.

Để mô hình có thể học được thông tin toàn cục của ảnh, một véc-tơ nhúng đặc biệt (tương tự [class] trong BERT) được thêm vào đầu chuỗi patch. Véc-tơ này sẽ tích lũy thông tin từ toàn bộ ảnh trong quá trình truyền qua các lớp Transformer, và sau cùng sẽ được sử dụng làm biểu diễn toàn ảnh cho các tác vụ như phân loại.

Chuỗi các véc-tơ này sẽ được đưa vào khối mã hóa Transformer Encoder – thành phần cốt lõi của mô hình, bao gồm nhiều lớp chú ý đa đầu ra, các lớp tuyến tính, chuẩn hóa, và cơ chế phần dư. Đặc biệt, một token đặc biệt gọi là [CLS] được thêm vào đầu chuỗi để đóng vai trò tóm tắt thông tin của toàn bộ ảnh. Sau khi đi qua các lớp mã hóa Transformer, đầu ra của [CLS] sẽ được dùng làm biểu diễn đặc trưng toàn cục của ảnh và đưa vào lớp phân loại để xác định nhãn đầu ra.

ViT mang lại nhiều ưu điểm nổi bật so với CNN truyền thống, đặc biệt là khả năng học các mối quan hệ toàn cục giữa các vùng ảnh mà không bị giới hạn bởi nhân tích chập cục bộ. Khi được huấn luyện trên các tập dữ liệu lớn như ImageNet-21k hoặc JFT-300M, ViT cho thấy hiệu năng vượt trội so với các mạng CNN mạnh như ResNet. Tuy nhiên, do thiếu các đặc trưng như tính cục bộ và bất biến vốn có trong CNN, ViT thường yêu cầu khối lượng dữ liệu huấn luyện lớn và kỹ thuật tiền huấn luyện mạnh để đạt hiệu quả tối ưu. Dù vậy, với sự phát triển của các phiên bản cải tiến như DeiT [101] hay Swin Transformer [67], mô hình này ngày càng trở nên phổ biến và được ứng dụng rộng rãi trong nhiều bài toán thị giác như phân loại ảnh, nhận diện đối tượng, phân đoạn ảnh, và đặc biệt là trong tái định danh người.

## 1.2.2 Học độ đo sâu

Trong quá trình huấn luyện tái định danh người theo hướng có giám sát, việc học dựa theo các độ đo học sâu DML đóng vai trò then chốt trong việc hướng dẫn mô hình học các biểu diễn đặc trưng có tính phân biệt cao giữa các danh tính. Mục tiêu chính của DML là hướng dẫn mô hình điều chỉnh các tham số sao cho trong không gian đặc trưng, các ảnh thuộc cùng một danh tính được ánh xạ đến các véc-tơ gần nhau, trong khi các ảnh của những người khác danh tính nằm ở xa nhau. Cách tiếp cận này giúp tăng cường khả năng phân biệt và nhận dạng của hệ thống. Việc học độ đo thường được thực hiện thông qua các hàm mất mát đặc trưng, như phân loại, bộ ba, tương phản nhằm định hướng rõ ràng cho mô hình trong quá trình tối ưu. Nhờ đó, mô hình có thể học được không gian biểu diễn ổn định, giúp tăng độ chính xác trong việc nhận diện đối tượng dưới nhiều góc nhìn và điều kiện khác nhau.

### 1.2.2.1 Hàm mất mát phân loại

Hàm mất mát phân loại [131] là một trong những lựa chọn phổ biến để huấn luyện các mô hình học sâu trong bài toán tái định danh người. Mục tiêu chính của hàm mất mát này là thu hẹp khoảng cách giữa các đặc trưng thuộc cùng một danh tính, đồng thời tăng cường sự phân tách giữa các đặc trưng thuộc các danh tính khác nhau. Thông qua việc tối ưu hóa hàm mất mát phân loại, mô hình học được cách gán nhãn chính xác cho mỗi ảnh đầu vào, từ đó hình thành không gian đặc trưng có khả năng phân biệt cao. Trong quá trình huấn luyện, mỗi danh tính trong tập dữ liệu được xem như một lớp riêng biệt, và mô hình được tối ưu để tăng xác suất dự đoán đúng nhãn lớp tương ứng với từng mẫu ảnh. Điều này giúp mạng học sâu trích xuất các đặc trưng đại diện một cách hiệu quả và nhất quán cho mỗi người.

Hàm mất mát phân loại thường được định nghĩa dưới dạng hàm mất mát softmax kết hợp với cross-entropy, cụ thể được biểu diễn dưới dạng Công thức 1.9.

$$L_{ce} = -\log p\,(y_i \mid x_i) \tag{1.9}$$

trong đó, $x_i$ là mẫu tại chỉ số $i$; $y_i$ và $p(y_i \mid x_i)$ lần lượt là nhãn nhận dạng đúng và xác suất dự đoán của nhãn đúng cho mẫu $i$.

### 1.2.2.2 Hàm mất mát bộ ba

Hàm mất mát bộ ba [27] là một hàm mất mát phổ biến và hiệu quả, thường được áp dụng trong các mô hình học sâu cho bài toán tái định danh người. Hàm mất mát này hoạt động dựa trên các bộ ba mẫu dữ liệu bao gồm:

- Anchor (A): một ảnh đại diện cho một danh tính (ví dụ: ảnh của người X),

- Positive (P): một ảnh khác của cùng danh tính với anchor (người X), và

- Negative (N): một ảnh của một danh tính khác (người Y).

Mục tiêu của hàm bộ ba là học một không gian nhúng sao cho khoảng cách giữa Anchor và Positive nhỏ hơn khoảng cách giữa Anchor và Negative ít nhất là một giá trị biên độ định trước. Điều này giúp tăng tính phân biệt của đặc trưng và cải thiện hiệu quả nhận dạng. Hàm mất mát bộ ba được biểu diễn theo Công thức 1.10.

$$L_{tri} = \left[ \|f_i - f_p\|_2^2 - \|f_i - f_n\|_2^2 + \alpha \right]_+ \tag{1.10}$$

trong đó, $f_i, f_p, f_n$ lần lượt là các véc-tơ đặc trưng của anchor, positive và negative của mẫu $i$; $\alpha$ là tham số biên độ quy định khoảng cách tối thiểu mà mô hình cần đảm bảo giữa khoảng cách giữa các cặp cùng định danh và cặp khác định danh; $[\cdot]_+$ đại diện cho hàm hinge [19], có giá trị bằng 0 nếu biểu thức trong hàm âm.

### 1.2.2.3 Hàm mất mát trung tâm

Để tăng cường khả năng phân biệt của các đặc trưng học được trong bài toán tái định danh người, hàm mất mát trung tâm [68] được sử dụng để đảm bảo rằng các đặc trưng của các mẫu cùng danh tính nằm gần nhau trong không gian đặc trưng của trung tâm lớp của danh tính đó. Nó thường được sử dụng cùng với các hàm mất mát khác trước đó để cải thiện quá trình huấn luyện các mô hình học sâu trong các bài toán tái định danh. Mục tiêu của hàm mất mát trung tâm là tối thiểu hóa khoảng cách giữa các đặc trưng của cùng một lớp với trung tâm lớp tương ứng của chúng. Cụ thể, nó giúp giảm sự biến thiên nội lớp bằng cách kéo các đặc trưng của cùng một danh tính lại gần trung tâm lớp, nghĩa là các đặc trưng của cùng một danh tính sẽ xích lại gần nhau. Hàm mất mát trung tâm được tính theo Công thức 1.11.

$$L_{ct} = \frac{1}{2} \|f_i - c_{y_i}\|_2^2 \tag{1.11}$$

trong đó, $f_i$ và $y_i$ lần lượt là véc-tơ đặc trưng và nhãn của mẫu $i$; $c_y$ là trung tâm lớp của các đặc trưng sâu tương ứng với lớp $y_i$.

### 1.2.2.4 Hàm mất mát bộ ba trọng tâm

Hàm mất mát bộ ba trọng tâm [116] là một biến thể của hàm bộ ba truyền thống, nhằm cải thiện thêm khả năng phân biệt của các véc-tơ đặc trưng trong bài toán tái định danh. Hàm mất mát này giới thiệu một phương pháp dựa trên tâm cụm, làm cho không gian đặc trưng trở nên có cấu trúc và ổn định hơn so với việc sử dụng các mẫu định danh làm tham chiếu trong việc tạo bộ ba. Cụ thể, hàm này sửa đổi hàm mất mát triplet bằng cách thay thế các mẫu Anchor, Positive và Negative của từng danh tính bằng các tâm cụm (hoặc đại diện trung bình) cho mỗi lớp. Cụ thể, thay vì đánh giá khoảng cách giữa Anchor-Positive và Anchor-Negative của hình ảnh A, hàm bộ ba trọng tâm đo lường khoảng cách giữa hình ảnh A với tâm cụm của lớp cùng danh tính $c_p$ và tâm cụm của lớp khác danh tính $c_n$. Mục tiêu là đảm bảo rằng tâm cụm của mỗi lớp gần các mẫu cùng định danh của nó hơn so với các mẫu khác định danh từ các lớp khác. Hàm này được biểu diễn trong Công thức 1.12.

$$L_{ctl} = \left[ \|f_i - c_p\|_2^2 - \|f_i - c_n\|_2^2 + \alpha \right]_+ \tag{1.12}$$

trong đó, $f_i$ là véc-tơ đặc trưng của mẫu $i$; $c_p$ và $c_n$ lần lượt là tâm cụm của lớp cùng định danh và lớp khác định danh với anchor $i$; $\alpha$ là tham số biên độ quy định khoảng cách tối thiểu mà mô hình cần đảm bảo giữa khoảng cách của mẫu tới tâm cụm cùng định danh với tâm cụm khác định danh; và [·]$_+$ đại diện cho hàm hinge [19], có giá trị bằng 0 nếu biểu thức trong hàm âm.

### 1.2.2.5 Hàm mất mát InfoNCE

InfoNCE [81] là một hàm mất mát phổ biến trong học tương phản, đặc biệt trong các mô hình học tự giám sát. Mục tiêu chính của InfoNCE là học được biểu diễn đặc trưng sao cho các cặp mẫu cùng danh tính – thường là các phiên bản tăng cường dữ liệu của cùng một đối tượng – có đặc trưng gần nhau, trong khi các mẫu khác danh tính – là các đối tượng khác nhau – bị đẩy ra xa trong không gian đặc trưng. Hàm mất mát này hoạt động như một bài toán phân loại, trong đó mô hình cố gắng nhận diện đúng mẫu cùng danh tính trong tập gồm nhiều mẫu khác danh tính, bằng cách tối đa hóa xác suất của mẫu cùng danh tính thông qua hàm softmax. InfoNCE thường sử dụng độ tương đồng cosine giữa các véc-tơ đặc trưng, kết hợp với một tham số điều chỉnh để điều chỉnh mức độ của phân phối xác suất. Nhờ khả năng học biểu diễn mạnh mẽ mà không cần nhãn, InfoNCE đã trở thành nền tảng trong nhiều phương pháp như SimCLR [16], MoCo [35] và được ứng dụng rộng rãi trong các bài toán như phân cụm, truy xuất hình ảnh, và tái định danh người không giám sát. Hàm mất mát InfoNCE được biểu diễn theo Công thức 1.13.

$$L_{\text{InfoNCE}} = -\log \frac{\exp(\text{sim}(x, x^+)/\tau)}{\exp(\text{sim}(x, x^+)/\tau) + \sum_{i=1}^{N} \exp(\text{sim}(x, x_i^-)/\tau)} \quad (1.13)$$

trong đó, $x$ là một mẫu anchor, $x^+$ là mẫu positive và $x^-$ là mẫu negative, $N$ là số lượng mẫu negative, $sim(,)$ là độ tương đồng giữa hai mẫu, $\tau$ là tham số điều chỉnh.

### 1.2.2.6 Hàm mất mát tương phản tự giám sát

Trong học tương phản tự giám sát, các mô hình học cách phân biệt giữa các đối tượng khác nhau mà không dựa vào nhãn rõ ràng. Nó sử dụng các phương pháp tăng cường dữ liệu để tạo ra các góc nhìn khác nhau (ví dụ: các phiên bản cắt hoặc xoay) của cùng một hình ảnh và coi mỗi cặp tăng cường là các mẫu cùng định danh. Các hình ảnh khác nhau được coi là các mẫu khác định danh. Mục tiêu là đưa các véc-tơ đặc trưng của các cặp cùng định danh lại gần nhau trong khi đẩy các véc-tơ đặc trưng của các mẫu khác định danh ra xa nhau. Phương pháp này đã thành công trong việc học các đại diện chất lượng cao cho nhiều nhiệm vụ thị giác máy tính mà không cần dữ liệu có nhãn. Hàm mất mát tương phản tự giám sát này được biểu diễn trong Công thức 1.14.

$$L_{ssl} = -\log \frac{\exp\left(\text{sim}\left(\mathbf{z}_i, \mathbf{z}_i^{'}\right)/\tau\right)}{\sum_{a \in N(j)} \exp\left(\text{sim}\left(\mathbf{z}_i, \mathbf{z}_a\right)/\tau\right)} \qquad (1.14)$$

ở đây, $z_i = g(f(Aug(x_i)))$, $z_i^{'} = g(f(Aug(x_i)))$, $z_j = g(f(Aug(x_j)))$ với $x_i$ và $x_j$ là mẫu hình ảnh tại chỉ số $i$ và $j$; $Aug(\cdot)$ là hàm tăng cường dữ liệu; $f(\cdot)$ là mạng cốt lõi; $g(\cdot)$ là một đầu ra mạng nơ-ron nhỏ giúp ánh xạ các biểu diễn vào không gian của hàm mất mát tương phản; $\tau$ là tham số điều chỉnh; $\text{sim}(u, v) = u^\top v/(|u||v|)$ đại diện cho tích vô hướng giữa hai véc-tơ $u$ và $v$ đã được chuẩn hóa $l_2$; $N(j)$ là tập các mẫu khác định danh.

### 1.2.2.7 Hàm mất mát tương phản có giám sát

Trong bối cảnh học có giám sát, hàm mất mát tương phản tự giám sát trong Công thức 1.14 không thể giải quyết tình huống mà nhiều mẫu được nhận diện là thuộc cùng một danh tính. Do vậy, Khosla và cộng sự [54] đã đề xuất hàm mất mát tương phản có giám sát để tổng quát hóa cho bất kỳ số lượng mẫu cùng danh tính. Hàm này được biểu diễn trong Công thức 1.15.

$$L_{\text{sup}} = \frac{-1}{|P(i)|} \sum_{p \in P(i)} \log \frac{\exp\left(\text{sim}(\mathbf{z}_i, \mathbf{z}_p)/\tau\right)}{\sum_{a \in N(j)} \exp\left(\text{sim}\left(\mathbf{z}_i, \mathbf{z}_a\right)/\tau\right)} \qquad (1.15)$$

trong đó, ký hiệu trong biểu thức này giống như trong Công thức 1.14. Thêm vào đó, $P(i)$ là tập hợp các ảnh có cùng định danh với $x_i$.

SCL sử dụng hoàn toàn thông tin có nhãn bằng cách xem xét tất cả các cặp cùng danh tính và khác danh tính trong một lô. Khác với các hàm mất mát tương phản truyền thống hoặc hàm bộ ba, vốn chỉ tập trung vào một cặp cùng và khác định danh cho mỗi ảnh mốc, SCL so sánh đồng thời ảnh mốc với nhiều cặp cùng và khác định danh. Tính chất này rất quan trọng đối với các bài toán như tái định danh người, khi cùng một định danh có thể xuất hiện dưới các góc nhìn và điều kiện ánh sáng khác nhau. Thêm vào đó, phương pháp này có thể xử lý sự biến thiên nội lớp và sự tương đồng giữa các lớp đáng kể, điều này phổ biến trong các trường hợp tái định danh người, giúp cải thiện khả năng của mô hình trong việc nhận dạng chính xác các định danh qua các camera và điều kiện khác nhau. Kết quả là, quá trình huấn luyện trở nên hiệu quả hơn.

## 1.2.3 Kỹ thuật xếp hạng lại

Trong bài toán tái định danh người, sau khi mô hình trích xuất đặc trưng cho ảnh truy vấn và ảnh trong tập tìm kiếm, quá trình truy xuất danh tính thường được thực hiện dựa trên khoảng cách Euclid hoặc cosine giữa các véc-tơ đặc trưng. Tuy nhiên, do các yếu tố như sự thay đổi góc nhìn, ánh sáng, hoặc nhiễu trong đặc trưng, danh sách xếp hạng ban đầu có thể không phản ánh chính xác mức độ tương đồng về danh tính giữa các mẫu. Nhằm cải thiện hiệu năng tái định danh, kỹ thuật xếp hạng lại (RR) được sử dụng như một bước hậu xử lý, giúp điều chỉnh lại thứ tự các kết quả truy xuất dựa trên ngữ cảnh và mối quan hệ giữa các mẫu ảnh.

Một trong những phương pháp xếp hạng lại hiệu quả và nổi bật được đề xuất bởi Zhong và cộng sự [132] là kỹ thuật mã hóa k-reciprocal. Ý tưởng chính của phương pháp này là thay vì chỉ dựa vào khoảng cách trực tiếp giữa truy vấn và các mẫu, RR sẽ xem xét mối quan hệ qua lại giữa các mẫu thông qua hàng xóm gần nhất. Cụ thể, với một ảnh truy vấn $q$, tập hàng xóm gần nhất $N_k(q)$ được xác định bằng cách chọn ra $k$ ảnh có khoảng cách nhỏ nhất. Một mẫu $g$ được coi là k-reciprocal của $q$ nếu $g \in N_k(q)$ và đồng thời $q \in N_k(g)$. Điều này thể hiện rằng hai ảnh đều coi nhau là hàng xóm gần nhất, phản ánh mối quan hệ hai chiều và do đó được đánh giá là đáng tin cậy hơn. Dựa trên tập các hàng xóm k-reciprocal, một đặc trưng mở rộng được tạo ra để đại diện cho ảnh truy vấn.

Sau đó, phương pháp tính khoảng cách được mở rộng bằng cách kết hợp giữa khoảng cách Euclid ban đầu và khoảng cách Jaccard giữa các tập hàng xóm. Cụ thể, khoảng cách Jaccard giữa hai mẫu được tính dựa trên tỉ lệ giữa phần giao và phần hợp của các tập k-reciprocal tương ứng trong Công thức 1.16.

$$D_{\text{Jaccard}}(q, g) = 1 - \frac{|R(q) \cap R(g)|}{|R(q) \cup R(g)|} \tag{1.16}$$

trong đó, $R(q)$ và $R(g)$ lần lượt là tập các hàng xóm k-reciprocal của truy vấn $q$ và ảnh $g$. Cuối cùng, khoảng cách tổng hợp được tính bằng Công thức 1.17.

$$D_{\text{final}}(q, g) = (1 - \lambda) \cdot D_{\text{Jaccard}}(q, g) + \lambda \cdot D_{\text{Euclid}}(q, g) \tag{1.17}$$

với $\lambda \in [0, 1]$ là hệ số điều chỉnh cân bằng giữa thông tin ngữ cảnh và khoảng cách trực tiếp.

Phương pháp xếp hạng lại k-reciprocal cho thấy hiệu quả rõ rệt trong việc cải thiện hiệu năng tái định danh người, đặc biệt là độ chính xác mAP và Rank-1. Một ưu điểm lớn của phương pháp này là không yêu cầu thay đổi hay huấn luyện lại mô hình học đặc trưng ban đầu, đồng thời có thể dễ dàng tích hợp vào luồng hệ thống hiện có. Tuy nhiên, chi phí tính toán cao và sự phụ thuộc vào các tham số như $k$ và $\lambda$ cũng là những yếu tố cần cân nhắc khi áp dụng trong thực tế.

## 1.2.4 Thuật toán phân cụm

Trong các phương pháp tái định danh người không giám sát, chiến lược phân cụm và gán nhãn giả là một hướng tiếp cận phổ biến và mang lại hiệu quả cao. Quy trình này bắt đầu bằng việc trích xuất đặc trưng từ ảnh thông qua một mạng cốt lõi được khởi tạo từ mô hình tiền huấn luyện. Các đặc trưng thu được sau đó được đưa vào các thuật toán phân cụm như K-Means [71] hoặc DBSCAN [29] nhằm nhóm các mẫu thành các cụm, trong đó mỗi cụm tương ứng với một nhãn giả và các mẫu trong cùng cụm sẽ được gán cùng nhãn này. Nhờ ưu thế không yêu cầu xác định trước số lượng cụm như K-Means cũng như khả năng phát hiện các cụm có hình dạng bất kỳ, DBSCAN trở thành một thuật toán phân cụm được ưa chuộng trong các phương pháp tái định danh người không giám sát.

Cụ thể, DBSCAN là một thuật toán phân cụm dựa trên mật độ, được thiết kế để nhận dạng các nhóm điểm hình thành những vùng có mật độ cao trong không gian đặc trưng, đồng thời tách biệt chúng khỏi các điểm nhiễu hoặc ngoại lai. Thuật toán dựa vào hai tham số cốt lõi: bán kính lân cận $\varepsilon$ và số điểm tối thiểu minPts cần có để tạo thành một vùng mật độ đủ lớn. Với mỗi điểm $x_i$, tập lân cận $N_\varepsilon(x_i)$ của điểm này được định nghĩa trong Công thức 1.18.

$$N_\varepsilon(x_i) = \{x_j \mid \text{dist}(x_i, x_j) \leq \varepsilon\} \tag{1.18}$$

trong đó, $\text{dist}(x_i, x_j)$ là hàm tính khoảng cách giữa hai điểm $x_i$ và $x_j$.

Một điểm được xem là điểm lõi nếu $|N_\varepsilon(x_i)| \geq$ minPts; nếu không phải điểm lõi nhưng nằm trong lân cận của một điểm lõi, nó được xem là điểm biên; các điểm còn lại được coi là nhiễu. Dựa trên các định nghĩa này, DBSCAN phân biệt được các điểm thuộc cụm và các điểm nằm ngoài cụm. Những mẫu không được phân cụm thường bị loại bỏ trước giai đoạn huấn luyện để duy trì độ tin cậy của nhãn giả, trong khi các mẫu thuộc cụm sẽ được gán nhãn giả và sử dụng trong việc tính toán hàm mất mát cũng như tối ưu hóa mô hình. Cách tiếp cận này cho phép mô hình học được các biểu diễn đặc trưng có ý nghĩa mà không cần đến nhãn thật, qua đó đặc biệt phù hợp với các bối cảnh ứng dụng thực tế nơi dữ liệu không có nhãn. Tuy nhiên, do bản chất không có nhãn thật, quá trình phân cụm có thể dẫn đến việc gán nhãn giả sai. Vì vậy, các phương pháp hiện đại thường kết hợp thêm các chiến lược như nhãn mềm [104], Teacher-Student [99] hoặc học tự điều chỉnh [32] nhằm cải thiện tính ổn định và độ chính xác của mô hình.

## 1.2.5 Tăng cường dữ liệu

Trong bài toán tái định danh người, dữ liệu huấn luyện thường hạn chế về số lượng và chưa phản ánh đầy đủ sự đa dạng của môi trường thực tế. Điều này dễ dẫn đến hiện tượng học quá khớp, khi mô hình ghi nhớ chi tiết huấn luyện thay vì nắm bắt các đặc trưng tổng quát. Trên thực tế, đối tượng có thể xuất hiện dưới nhiều góc nhìn, tư thế, điều kiện ánh sáng khác nhau hoặc bị che khuất, khiến việc thu thập dữ liệu bao quát tất cả tình huống trở nên khó khăn. Do đó, việc áp dụng các kỹ thuật tăng cường dữ liệu là cần thiết nhằm tạo ra nhiều biến thể nhân tạo từ ảnh gốc, giúp mở rộng và đa dạng hóa tập huấn luyện. Nhờ vậy, mô hình có khả năng học được những đặc trưng ổn định và quan trọng, đồng thời giảm phụ thuộc vào chi tiết không đặc thù. Trong các nghiên cứu về tái định danh người, tăng cường dữ liệu đã trở thành quy trình chuẩn nhằm mở rộng và đa dạng hóa tập huấn luyện. Các phép biến đổi được áp dụng bao gồm: thay đổi kích thước ảnh về $256 \times 128$; lật ngang ngẫu nhiên với xác suất 50%; cắt ngẫu nhiên về kích thước $256 \times 128$ sau khi thêm viền 10 pixel; và xóa một vùng chữ nhật ngẫu nhiên với xác suất 50%. Những thiết lập này không chỉ giúp làm phong phú dữ liệu mà còn góp phần hạn chế hiện tượng quá khớp, đồng thời nâng cao khả năng khái quát hóa của mô hình. Ý nghĩa cụ thể của các phép biến đổi này được trình bày theo dưới đây.

- Thay đổi kích thước ảnh: Đây là bước tiền xử lý cơ bản trong hầu hết hệ thống thị giác máy tính, nhằm đưa toàn bộ ảnh đầu vào về cùng kích thước cố định. Việc chuẩn hóa kích thước giúp thống nhất dữ liệu, giảm sự biến thiên độ phân giải và độ phức tạp tính toán, đồng thời tạo điều kiện để mô hình tập trung vào nội dung ảnh. Tuy nhiên, quá trình này có thể làm biến dạng hình học hoặc mất chi tiết, do đó thường được kết hợp với các kỹ thuật tăng cường dữ liệu khác để cân bằng hiệu quả.

- Lật ngang ngẫu nhiên: Ảnh được lật đối xứng trái–phải với một xác suất định trước (thường là 0.5). Phép biến đổi này giúp mô hình học được tính bất biến trước sự thay đổi hướng nhìn của đối tượng, đồng thời nhân rộng số lượng mẫu mà không cần thu thập thêm dữ liệu. Trong bài toán tái định danh người, kỹ thuật này đặc biệt hữu ích khi người quan sát có thể xuất hiện từ nhiều góc độ khác nhau.

- Cắt ngẫu nhiên kết hợp thêm viền: Ảnh đầu vào được mở rộng bằng cách thêm viền ở bốn phía, sau đó cắt ngẫu nhiên một vùng theo đúng kích thước yêu cầu. Cách làm này khiến mô hình trong mỗi lần huấn luyện sẽ nhìn thấy những bố

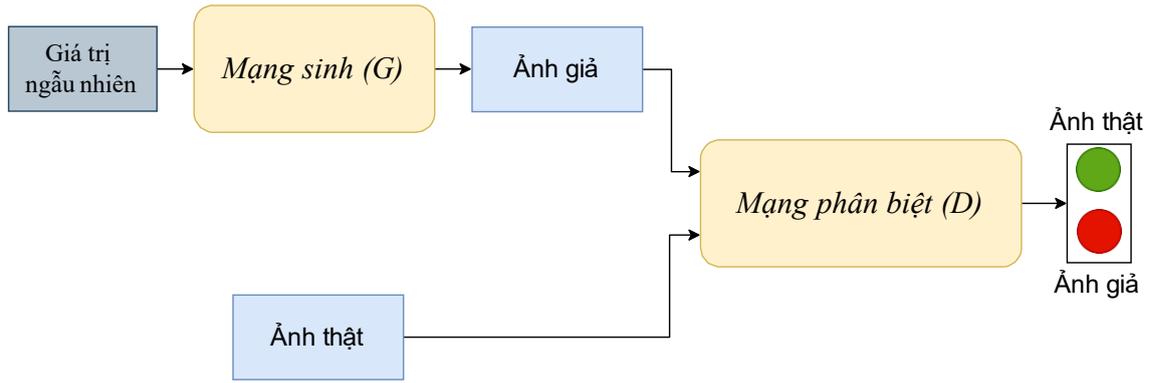

Hình 1.7: Kiến trúc mạng đối kháng tạo sinh.

cục khác nhau, từ đó tăng khả năng học tính bất biến về vị trí đối tượng trong khung hình. Kỹ thuật này góp phần giảm hiện tượng quá khớp với hình ảnh gốc, đồng thời làm dữ liệu huấn luyện đa dạng và gần hơn với thực tế.

- Xóa một vùng ngẫu nhiên: Chọn một vùng chữ nhật ngẫu nhiên và thay thế bằng giá trị điểm ảnh cố định. Cách tiếp cận này giúp mô hình học được tính ổn định khi đối tượng bị che khuất một phần, giảm nguy cơ học thuộc chi tiết không quan trọng và nâng cao khả năng khái quát hóa. Với ưu điểm đơn giản, dễ triển khai và không làm tăng chi phí tính toán, phương pháp này hiện được ứng dụng rộng rãi trong bài toán tái định danh người, đặc biệt là các trường hợp người bị che khuất.

## 1.2.6 Mạng đối kháng tạo sinh

Mạng đối kháng tạo sinh (GAN) là một mô hình sinh dữ liệu dựa trên học sâu, được Goodfellow và cộng sự đề xuất vào năm 2014 [33]. GAN học phân bố dữ liệu thực thông qua cơ chế đối kháng giữa hai mạng nơ-ron, gồm mạng sinh ($G$) và mạng phân biệt ($D$). Trong đó, $G$ có nhiệm vụ sinh ra các mẫu dữ liệu giả sao cho có phân bố gần giống dữ liệu thật, trong khi $D$ được huấn luyện để phân biệt dữ liệu thật và dữ liệu giả do $G$ sinh ra như thể hiện trong Hình 1.7. Cụ thể, $G$ nhận đầu vào là nhiễu ngẫu nhiên $z \sim p_z(z)$ và ánh xạ thành mẫu giả $x_{fake} = G(z)$. Mục tiêu của $G$ là tạo ra dữ liệu đủ giống thật để đánh lừa $D$ và được cực tiểu hóa giá trị theo Công thức 1.19.

$$\min_G V(G) = \mathsf{E}_{z \sim p_z(z)}[\log(1 - D(G(z)))] \qquad (1.19)$$

trong đó, $G$ cố gắng tối đa hóa xác suất để $D$ dự đoán dữ liệu giả là thật.

Ngược lại, $D$ được thiết kế nhằm phân biệt chính xác dữ liệu thật $x_{real} \sim p_{data}$ và dữ liệu giả do $G$ sinh ra. Mục tiêu của $D$ là cực đại hóa giá trị theo Công thức 1.20.

$$\max_{D} V(D) = \mathsf{E}_{x \sim p_{data}(x)}[\log D(x)] + \mathsf{E}_{z \sim p_z(z)}[\log(1 - D(G(z)))] \qquad (1.20)$$

Toàn bộ quá trình GAN được huấn luyện theo bài toán đối kháng dạng cực tiểu - cực đại được biểu diễn trong Công thức 1.21.

$$\min_{G} \max_{D} V(D, G) = \mathsf{E}_{x \sim p_{data}(x)}[\log D(x)] + \mathsf{E}_{z \sim p_z(z)}[\log(1 - D(G(z)))] \qquad (1.21)$$

GAN mang lại nhiều ưu điểm nổi bật, đặc biệt là khả năng sinh dữ liệu chất lượng tương đối cao và đa dạng, tốc độ sinh mẫu nhanh. Dựa trên kiến trúc GAN gốc, nhiều biến thể đã được phát triển nhằm cải thiện tính ổn định, chất lượng mẫu sinh và mở rộng ứng dụng. DCGAN [88] khai thác cấu trúc CNN cho cả $G$ và $D$, giúp huấn luyện ổn định hơn trên dữ liệu ảnh. WGAN [7] và WGAN-GP [34] thay thế phân kỳ Jensen-Shannon bằng khoảng cách Wasserstein, giúp cải thiện độ dốc và giảm hiện tượng suy giảm đa dạng. Các mô hình như CycleGAN [135] và pix2pix [48] tập trung vào dịch chuyển ảnh giữa hai miền khác nhau, đặc biệt hiệu quả trong các bài toán chuyển đổi ảnh. Ngoài ra, các mô hình như StyleGAN [53] mang lại chất lượng sinh ảnh vượt trội nhờ cơ chế điều khiển phong cách ở nhiều mức độ trong không gian đặc trưng.

Trên cơ sở các hướng phát triển này, StarGAN [21] được đề xuất nhằm mở rộng khả năng dịch chuyển ảnh đa miền bằng một mô hình duy nhất. Khác với CycleGAN [135] vốn chỉ hỗ trợ dịch giữa hai miền cố định, StarGAN tích hợp véc-tơ miền mục tiêu $c$ vào quá trình sinh ảnh, cho phép $G$ tạo ra ảnh được chuyển đổi theo nhiều phong cách hoặc nhiều miền khác nhau. Đồng thời, $D$ trong StarGAN không chỉ đánh giá tính thật-giả mà còn dự đoán miền của ảnh đầu vào, giúp mô hình học hiệu quả hơn trong bối cảnh đa miền. Nhờ đó, StarGAN trở thành công cụ linh hoạt cho các ứng dụng như thay đổi thuộc tính khuôn mặt, điều chỉnh phong cách camera và thích ứng miền trong các bài toán thị giác máy tính.

# 1.3 Tổng quan nghiên cứu liên quan

## 1.3.1 Tái định danh người dựa trên học có giám sát

Có thể thấy, quá trình huấn luyện có giám sát trong bài toán tái định danh người thường bao gồm hai thành phần chính: trích xuất đặc trưng từ mạng cốt lõi và học độ đo. Trong đó, đặc trưng hình ảnh có thể được trích xuất dưới dạng đặc trưng toàn cục hoặc đặc trưng cục bộ. Đặc trưng toàn cục là một trong những hướng tiếp cận sớm khi học sâu được áp dụng vào bài toán tái định danh [105, 85, 68]. Phương pháp này có ưu điểm đơn giản và hiệu quả, tuy nhiên lại gặp hạn chế trong việc biểu diễn các chi tiết quan trọng của người trong ảnh, đặc biệt là trong những tình huống có sự thay đổi về tư thế, góc nhìn hoặc vật che khuất. Chính vì vậy, việc tập trung khai thác các đặc trưng cục bộ giàu thông tin và mang tính phân biệt cao đang trở thành một hướng nghiên cứu quan trọng nhằm cải thiện độ chính xác của hệ thống tái định danh.

Các nghiên cứu như [111, 76] sử dụng cách chia ảnh thành một số lượng vùng cố định nhất định để trích xuất đặc trưng cục bộ. Tuy nhiên, điểm chung của các phương pháp này là chưa xem xét hoặc mô hình hóa mối liên kết giữa các vùng cục bộ, dẫn đến hạn chế trong việc khai thác quan hệ không gian giữa các bộ phận cơ thể người. Thêm vào đó, phương pháp PCB [95] là một bước tiến quan trọng trong tái định danh khi đề xuất phương pháp chia đặc trưng thành các dải ngang cố định tương ứng với từng phần cơ thể người. Cách tiếp cận này giúp mô hình học được các đặc trưng cục bộ mà không cần ước lượng tư thế. Một cải tiến khác là RPP còn giúp tăng tính nhất quán của đặc trưng trong từng phần. Tiếp nối hướng tiếp cận này, mô hình MGN [106] sử dụng kiến trúc nhiều nhánh để khai thác đặc trưng ở ba mức độ khác nhau: toàn cục, chia hai, chia ba. Việc kết hợp đặc trưng đa mức độ giúp cải thiện rõ rệt hiệu năng nhận dạng. Trong khi đó, Wang và cộng sự [109] đề xuất phương pháp ProNet giữ lại bộ phân loại ở bước suy luận như phép chiếu sang không gian nguyên mẫu, giúp tăng tính phân biệt và ProNet++ bổ sung thiết kế đa mức sử dụng đặc trưng cục bộ để nâng cao kết quả. Gần đây, AGW [120] đóng vai trò như một chuẩn mạnh trong lĩnh vực tái định danh, kết hợp kỹ thuật gộp trung bình tổng quát (GeM), chú ý phi cục bộ và hàm mất mát bộ ba có trọng số. AGW tập trung cân bằng giữa học đặc trưng toàn cục và cục bộ, đồng thời giữ được sự đơn giản trong tính toán. Tổng thể, các mô hình này đã chứng minh rằng việc kết hợp đặc trưng cục bộ và toàn cục – đặc biệt qua các kiến trúc đa nhánh – là chìa khóa để giải quyết các thách thức như che khuất, thay đổi tư thế và sự tương đồng thị giác trong các hệ thống nhận dạng người.

Về phần DML, đây kỹ thuật nhằm học một không gian nhúng trên không gian đặc trưng sao cho đặc trưng của các ảnh của cùng định danh được đưa lại gần nhau, trong khi đặc trưng của các ảnh khác định danh nằm xa nhau [5, 121, 120]. Một trong những hướng tiếp cận sớm là mô hình IDE của Zheng và cộng sự [131], xem quá trình huấn luyện tái định danh người như một bài toán phân loại đa lớp đối với ảnh người. Cách tiếp cận này dễ huấn luyện, tận dụng tốt các kiến trúc phân loại và đạt độ chính xác cao ở thời điểm công bố, nhưng chỉ hoạt động hiệu quả với dữ liệu gán nhãn đầy đủ và khó khái quát sang định danh mới. Tiếp theo, Ding và cộng sự [27] đề xuất sử dụng các bộ ba ảnh từ cùng một định danh và khác định danh để xây dựng mô hình tái định danh. Trong khi Hermans và cộng sự [39] nhấn mạnh hiệu quả của hàm mất mát bộ ba trong việc học các đặc trưng phân biệt mà không cần bước học độ đo riêng biệt. Phương pháp này giúp mô hình học đặc trưng phân biệt và thích hợp khi xuất hiện định danh mới, song số lượng bộ ba tăng nhanh khiến quá trình huấn luyện phức tạp, đòi hỏi chiến lược lựa chọn mẫu hiệu quả. Tiếp đến, Luo và cộng sự [68] đã đề xuất một phương pháp nền tảng mạnh mẽ cho tái định danh bằng cách tích hợp ba loại mất mát: phân loại, trung tâm, bộ ba. Việc thêm hàm mất mát trung tâm giúp tăng tính chặt chẽ nội cụm, đưa đặc trưng cùng định danh về gần tâm lớp, từ đó cải thiện độ ổn định và tốc độ hội tụ. Tuy vậy, phương pháp này yêu cầu lưu trữ và cập nhật véc-tơ tâm, làm tăng chi phí bộ nhớ khi số định danh lớn. Gần đây, Wieczorek và cộng sự [116] đề xuất sử dụng hàm mất mát bộ ba trọng tâm thay thế các mẫu riêng lẻ bằng tâm cụm nhằm giảm nhiễu và bớt phụ thuộc vào chiến lược chọn bộ ba. Phương pháp này giúp mô hình hội tụ nhanh và ổn định hơn, nhưng hiệu quả vẫn phụ thuộc vào chất lượng tâm cụm và dễ bị ảnh hưởng khi dữ liệu mất cân bằng. Tổng thể, mỗi phương pháp DML đều có ưu và nhược điểm riêng, trong đó, các tiếp cận kết hợp nhiều hàm mất mát cho phép tận dụng các ưu thế bổ sung, nâng cao khả năng phân biệt, tính khái quát và tốc độ hội tụ, nhưng đồng thời cũng làm tăng chi phí tính toán và độ phức tạp trong huấn luyện.

Để tóm tắt nghiên cứu có liên quan, Bảng 1.2 được trình bày để so sánh các phương pháp giải quyết bài toán tái định danh theo hướng tiếp cận học có giám sát dựa trên các thông tin: Phương pháp, Nơi công bố, Cách tiếp cận, Bộ dữ liệu và Kết quả (mAP(%)). Các tên viết tắt Market (M), Duke (D), MSMT (MT), CUHK03(L) (CL), CUHK(D) (CD) được sử dụng để biểu thị các tập dữ liệu tương ứng trong quá trình huấn luyện và đánh giá.

Bảng 1.2: Tóm tắt kết quả một số nghiên cứu tái định danh người có giám sát

| Phương pháp | Nơi công bố | Cách tiếp cận | Bộ dữ liệu | mAP (%) |
|---|---|---|---|---|
| PCB [95] | CVPR 2018 | Chia đặc trưng thành các dải ngang cố định tương ứng với từng phần cơ thể người sau đó sử dụng RPP để tăng tính nhất quán của đặc trưng trong từng phần. | M | 91.9 |
| BoT [68] | CVPR 2019 | Đề xuất một phương pháp cơ sở đơn giản và hiệu quả bằng cách sử dụng sáu phương án nâng cao hiệu quả mô hình. | M, D | 94.2, 89.1 |
| CTL [116] | ICONIP 2021 | Đề xuất hàm mất mát bộ ba trọng tâm nhằm tăng cường khả năng phân biệt và cải thiện hiệu năng truy xuất hình ảnh. | M, D | 98.3, 96.1 |
| MCTL [6] | JoBD 2023 | Giới thiệu một biến thể của hàm mất mát bộ ba trọng tâm nhằm tăng cường trọng tâm vào khoảng cách giữa các lớp. | M, D | 98.63, 97.3 |
| CTupletL [11] | 3PGCIC 2023 | Sử dụng đại diện trọng tâm trung bình được chọn ngẫu nhiên của các lớp trong mỗi lô. | M, CL, CD | 98.57, 94.38, 93.39 |
| PRONET ++ [109] | arXiv 2023 | Đề xuất một quan điểm mới trong việc diễn giải bộ phân loại như một ánh xạ từ đặc trưng ảnh đến nguyên mẫu lớp. | M, CL, CD, MT | 95.3, 91.9, 89.2, 80.0 |
| CA-Jaccard [18] | CVPR 2024 | Trình bày khoảng cách Jaccard có tính đến thông tin camera để cải thiện độ chính xác. | M, MT | 94.5, 45.3 |

## 1.3.2 Tái định danh người dựa trên học thích ứng miền không giám sát

Vấn đề cần giải quyết đầu tiên trong bài toán UDA là sự khác biệt giữa hai phân phối dữ liệu trong miền nguồn và miền đích. Để giải quyết sự khác biệt này, các nghiên cứu đề xuất phương án giảm thiểu sự khác biệt ở cấp hình ảnh hoặc ở cấp đặc trưng. Ở cấp hình ảnh các mô hình tạo sinh (GAN) thường được sử dụng để chuyển phong cách từ miền nguồn sang miền đích trong khi vẫn giữ nguyên danh tính của ảnh chuyển đổi nhằm giúp giảm sự chênh lệch miền và tăng hiệu quả học đặc trưng. Theo nghiên cứu của [15], các phương pháp tăng cường dữ liệu dựa trên GAN thường được áp dụng có điều kiện, tập trung vào các yếu tố như: tư thế, ánh sáng, phong cách camera, hậu cảnh và cấu trúc tổng quát. Về tư thế, FD-GAN [31] và PN-GAN [86] sinh ảnh với tư thế mới dựa trên thông tin 2D, sau này còn được khai thác cho bài toán thích ứng miền. Về ánh sáng, [47] xây dựng khung nhận dạng bất biến với điều kiện chiếu sáng nhằm loại bỏ ảnh hưởng của biến thiên ánh sáng đến đặc trưng nhận dạng. Với phong cách camera, CamStyle [133] và CMFC [102] lần lượt sử dụng CycleGAN và StarGAN để chuyển đổi phong cách giữa các camera, thu hẹp chênh lệch miền. Đối với hậu cảnh, SBSGAN [46] được thiết kế để thay thế nền ảnh, giảm tác động của bối cảnh đến quá trình tái định danh không giám sát. Cuối cùng, DG-Net [129] và DG-Net++ [138] tái tô màu ảnh người dựa trên phân phối màu của ảnh khác, qua đó học biểu diễn danh tính rời rạc ít phụ thuộc vào cấu trúc hình ảnh.

Ở cấp độ đặc trưng, mô hình được huấn luyện sao cho phân phối đặc trưng của miền nguồn và miền đích trở nên gần nhau. Một cách tiếp cận phổ biến là sử dụng độ lệch trung bình tối đa (MMD), được Mekhazni và cộng sự đề xuất [73], nhằm căn chỉnh phân phối khoảng cách giữa các cặp mẫu nguồn và đích. Tuy nhiên, phương pháp này phụ thuộc vào chất lượng hình ảnh của dữ liệu và có thể bỏ qua thông tin ngữ cảnh trong đặc trưng gốc. Trong khi đó, Liu và cộng sự [65] đưa ra phương pháp DIM+GLO, kết hợp DIM để học ánh xạ bất biến miền với GLO nhằm tối ưu đồng thời khoảng cách toàn cục và cục bộ trong không gian đặc trưng. Cách tiếp cận này không chỉ thu hẹp sự khác biệt miền mà còn duy trì khả năng phân biệt danh tính, nhờ đó đạt hiệu quả cao trong bối cảnh thích ứng miền không giám sát. Tuy nhiên, nhược điểm của DIM+GLO là làm tăng độ phức tạp mô hình và đòi hỏi nhiều tài nguyên tính toán hơn trong quá trình huấn luyện.

Tiếp theo, trong UDA, việc chỉ sử dụng đặc trưng toàn cục thường làm mất thông tin cục bộ quan trọng, dẫn đến suy giảm khả năng tổng quát hóa trên miền đích. Do đó, các nghiên cứu gần đây kết hợp đặc trưng cục bộ hoặc biểu diễn đa mức nhằm nâng cao tính phân biệt. Đồng thời, nhãn giả sinh ra từ phân cụm trong môi trường không nhãn thường không ổn định, đòi hỏi các kỹ thuật bổ trợ như học tương phản, học đồng thuận đa nhánh hoặc gán trọng số theo độ tin cậy để tinh chỉnh nhãn và cải thiện hiệu quả huấn luyện. Những hướng tiếp cận này cho thấy hiệu quả rõ rệt trong việc cải thiện hiệu năng của mô hình tái định danh người trong hướng tiếp cận thích ứng miền không giám sát.

Cụ thể, Ge và cộng sự đã đóng góp hai phương pháp đáng chú ý là MMT [30] và SpCL [32]. MMT sử dụng hai mô hình Teacher–Student cập nhật qua EMA với nhãn mềm và kết hợp hai hàm mất mát nhằm giảm nhiễu và ổn định huấn luyện. Phương pháp này giúp tăng độ tin cậy của nhãn giả và cải thiện sự hội tụ, tuy nhiên cấu trúc phức tạp do yêu cầu hai cặp Teacher–Student song song và vẫn phụ thuộc vào chất lượng nhãn giả ban đầu. Trong khi đó, SpCL đề xuất khung học tương phản tự điều chỉnh với bộ nhớ kết hợp, tích hợp các cấp độ giám sát khác nhau và tinh chỉnh nhãn cụm một cách linh hoạt để tăng hiệu năng. Cách tiếp cận này cải thiện rõ rệt khả năng phân biệt và khái quát hóa của mô hình, song vẫn dễ bị ảnh hưởng bởi sai lệch trong quá trình phân cụm giai đoạn đầu. Zheng và cộng sự cũng đề xuất hai hướng tiếp cận khác nhau. UNRN [125] tận dụng độ bất định giữa mô hình Teacher và Student để điều chỉnh trọng số của mẫu trong các hàm mất mát giúp giảm tác động của nhãn sai, nhưng đòi hỏi thiết kế cơ chế đo bất định phức tạp. GLT [126] gán và tinh chỉnh nhãn giả theo nhóm ngay trong quá trình huấn luyện, giúp cải thiện hiệu quả học trên dữ liệu không gán nhãn, nhưng hiệu năng vẫn phụ thuộc vào chất lượng phân cụm và có khả năng gộp nhầm nhóm. Bertocco và cộng sự [8] sử dụng hàm mất mát đơn giản chỉ với một siêu tham số và chiến lược tạo bộ ba dựa trên đa dạng camera để chọn mẫu tin cậy, đồng thời áp dụng cơ chế tự tập hợp trọng số từ nhiều vòng huấn luyện nhằm tăng độ khái quát và tránh quá khớp trên miền mục tiêu. Ưu điểm của phương pháp này là thiết kế gọn nhẹ, dễ huấn luyện và tổng quát tốt, nhưng nhược điểm là kém linh hoạt khi dữ liệu có ít thông tin camera. Wang và cộng sự [110] đề xuất mô-đun AWB tích hợp vào mô hình học qua lại lẫn nhau để tăng tính bổ sung giữa hai mạng và giảm nhiễu trong nhãn giả. Phương pháp này giúp nâng cao tính ổn định, nhưng thêm mô-đun khiến chi phí huấn luyện tăng. Yuanpeng Tu [102] đề xuất phương pháp CMFC kết hợp chuyển ảnh giữa hai miền bằng StarGAN và phân cụm đa đặc trưng (toàn cục và bộ phận), cùng chiến lược gán nhãn giả phối hợp để giảm thiên lệch và cải thiện biểu diễn đặc trưng. Phương pháp có ưu điểm là giảm thiên lệch và khai thác được thông tin toàn cục – cục bộ; tuy nhiên có nhược điểm là phụ thuộc vào GAN và chi phí tính toán cao. Bên cạnh đó, Tian và Sun [100] giới thiệu CDCL, sử dụng ảnh xám một phần để giảm nhiễu màu, kiến trúc hai nhánh để tách biệt thông tin màu sắc và chiến lược tinh chỉnh cụm nhằm nâng cao độ tin cậy của nhãn giả. Phương pháp này giúp tăng độ tin cậy của nhãn giả, nhưng kiến trúc phức tạp và nhạy cảm với tham số. Ding và Zhou [26] đề xuất $LF^2$ kết hợp đặc trưng toàn cục và cục bộ thông qua chiến lược Teacher-Student và module hợp nhất có thể huấn luyện được, giúp tạo nhãn giả đa dạng và cải thiện hiệu năng tái định danh người thích ứng miền không giám sát. Ưu điểm là tạo nhãn giả đa dạng và cải thiện hiệu năng, nhưng điều này cũng gây ra sự phức tạp cho mô hình và phụ thuộc vào sự cân bằng giữa đặc trưng toàn cục và cục bộ. Wu và cộng sự [117] đề xuất phương pháp MCRN gồm ba thành phần chính: MCM lưu nhiều tâm đại diện cho mỗi cụm để giảm nhiễu nhãn; DSCL chỉ so sánh mẫu trong cùng miền để học đặc trưng tốt hơn; và SONI tạo thêm mẫu âm có thông tin để tăng độ phân biệt. Cách này tăng tính phân biệt và giảm nhiễu, nhưng chi phí bộ nhớ và tính toán cao. Gần đây, Mamedov và cộng sự [72] đề xuất ReMix, huấn luyện chung trên dữ liệu đa camera có nhãn và dữ liệu đơn camera không nhãn. Phương pháp này cải thiện tính tổng quát và tận dụng dữ liệu dễ thu thập, nhưng phụ thuộc vào chiến lược lấy mẫu và thiết kế hàm mất mát phức tạp. Để tóm tắt nghiên cứu có liên quan trong phần này, Bảng 1.3 được trình bày để so sánh các phương pháp giải quyết bài toán tái định danh theo hướng tiếp cận thích ứng miền không giám sát với các thông tin: Phương pháp, Nơi công bố, Cách tiếp cận, Bộ dữ liệu và kết quả (mAP(%)). Các thiết lập học thích ứng miền không giám sát được ký hiệu rút gọn "M2D" cho "Market-to-Duke", "D2M" cho "Duke-to-Market", "M2MT" cho "Market-to-MSMT", "D2MT" cho "Duke-to-MSMT".

Bảng 1.3: Tóm tắt kết quả một số nghiên cứu tái định danh người thích ứng miền không giám sát

| Phương pháp | Nơi công bố | Cách tiếp cận | Bộ dữ liệu | mAP (%) |
|---|---|---|---|---|
| DIMGLO [65] | ICM 2020 | Kết hợp DIM và GLO nhằm giải quyết đồng thời sự khác biệt dữ liệu giữa hai miền và thiếu nhãn. | D2M, M2D, D2MT, M2MT | 65.1, 58.3, 24.4, 20.7 |
| MMT [30] | ICLR 2020 | Sử dụng hai mô hình Teacher–Student cập nhật qua EMA, sử dụng nhãn mềm để giảm nhiễu và kết hợp hai loại mất mát để ổn định quá trình huấn luyện. | D2M, M2D, D2MT, M2MT | 71.2, 65.1, 23.3, 22.9 |

| Phương pháp | Nơi công bố | Cách tiếp cận | Bộ dữ liệu | mAP (%) |
|---|---|---|---|---|
| SpCL [32] | NIPS 2020 | Đề xuất khung học tương phản tự điều chỉnh SpCL với bộ nhớ kết hợp, kết hợp nhiều cấp độ giám sát và tinh chỉnh dần nhãn cụm. | D2M, M2D, D2MT, M2MT | 76.7, 68.8, 26.5, 26.8 |
| UNRN [125] | AAAI 2021 | Tối ưu theo độ bất định bằng cách đo độ nhất quán giữa Teacher và Student để điều chỉnh trọng số mẫu, giảm ảnh hưởng của nhãn giả sai. | D2M, M2D, D2MT, M2MT | 78.1, 69.1, 26.2, 25.3 |
| GLT [126] | CVPR 2021 | Gán và tinh chỉnh nhãn giả theo nhóm ngay trong lúc huấn luyện. | D2M, M2D, D2MT, M2MT | 79.5, 69.2, 27.7, 26.5 |
| CCTSE [8] | TIFS 2021 | Sử dụng hàm mất mát đơn giản chỉ với một siêu tham số và chiến lược tạo bộ ba dựa trên đa dạng camera để chọn mẫu tin cậy, đồng thời áp dụng cơ chế tự tập hợp trọng số từ nhiều vòng huấn luyện nhằm tăng độ khái quát và tránh quá khớp trên miền mục tiêu. | D2M, M2D, D2MT, M2MT | 78.4, 72.6, 34.5, 33.2 |
| AWB [110] | TIP 2022 | Đề xuất mô-đun AWB tích hợp vào mô hình học qua lại lẫn nhau để tăng tính bổ sung giữa hai mạng và giảm nhiễu trong nhãn giả. | D2M, M2D, D2MT, M2MT | 81.0, 70.9, 29.0, 29.0 |
| CMFC [102] | HCMA 2022 | Kết hợp chuyển ảnh giữa hai miền bằng StarGAN và phân cụm đa đặc trưng, cùng chiến lược gán nhãn giả phối hợp để giảm thiên lệch và cải thiện biểu diễn đặc trưng. | D2M, M2D | 81.0, 71.2 |

| Phương pháp | Nơi công bố | Cách tiếp cận | Bộ dữ liệu | mAP (%) |
|---|---|---|---|---|
| CDCL [100] | KBS 2023 | Sử dụng ảnh xám một phần để giảm nhiễu màu, kiến trúc hai nhánh để tách biệt thông tin màu sắc và chiến lược tinh chỉnh cụm nhằm nâng cao độ tin cậy của nhãn giả. | D2M, M2D, D2MT, M2MT | 81.5, 70.2, 30.3, 29.5 |
| L$F_2$ [26] | ICPR 2022 | Kết hợp đặc trưng toàn cục và cục bộ thông qua chiến lược Teacher-Student và module hợp nhất có thể huấn luyện được, giúp tạo nhãn giả đa dạng và cải thiện hiệu năng. | D2M, M2D | 83.2, 73.5 |
| MCRN [117] | AAAI 2022 | Giới thiệu cơ chế học nhiều tâm cụm nhằm cải thiện biểu diễn đặc trưng và giảm ảnh hưởng của nhiễu trong quá trình gán nhãn giả. | D2M, M2D, D2MT, M2MT | 83.8, 71.5, 35.7, 32.8 |

Tổng thể, các phương pháp UDA trong tái định danh người được trình bày ở trên đã cho thấy sự tiến hóa rõ rệt về mặt chiến lược và kỹ thuật. Từ những tiếp cận đơn giản giảm sự khác biệt giữa hai miền dữ liệu, trích xuất đặc trưng toàn cục hoặc phân cụm cứng, đến các phương pháp phức tạp tích hợp nhiều mô hình, nhiều nhánh và chiến lược huấn luyện như học tương phản, nhãn mềm, đồng thuận đa mô hình, cơ chế trọng số tin cậy và tự điều chỉnh. Mỗi phương pháp đều mang lại những đóng góp cụ thể, như giảm sự khác biệt giữa dữ liệu giữa miền nguồn và nguồn đích, tăng cường khả năng khái quát hóa, giảm ảnh hưởng của nhiễu nhãn, và nâng cao hiệu quả thích ứng trong các môi trường thực tế không đồng nhất. Xu hướng chung cho thấy việc kết hợp nhiều kỹ thuật và điều chỉnh mô hình theo thời gian là hướng tiếp cận hứa hẹn giúp UDA đạt được hiệu năng ngày càng gần với mô hình có giám sát.

## 1.3.3 Tái định danh người dựa trên học không giám sát

Nhiều nghiên cứu đã được đề xuất nhằm cải thiện hiệu năng của mô hình trong bài toán tái định danh người không giám sát. Các hướng tiếp cận này tập trung vào việc giảm nhiễu trong quá trình gán nhãn giả, tăng cường tính nhất quán trong không gian đặc trưng, và tận dụng thông tin bổ sung như định danh camera để cải thiện khả năng học của mô hình.

Cụ thể, Ge và cộng sự giới thiệu SpCL [32] có chiến lược học tự điều chỉnh và bộ nhớ kết hợp nhằm lựa chọn mẫu tin cậy trong huấn luyện, cải thiện độ ổn định và độ chính xác với ưu điểm nổi bật là giảm nhiễu trong nhãn giả nhưng vẫn dễ bị ảnh hưởng bởi sai lệch trong quá trình phân cụm giai đoạn đầu. Wang và cộng sự phát triển CAP [108], sử dụng đại diện theo từng camera để xử lý sự sai lệch đặc trưng đa góc, qua đó cải thiện độ chính xác gán nhãn. Phương pháp này đặc biệt hiệu quả khi có thông tin camera đầy đủ nhưng khó mở rộng trong môi trường thiếu thông tin. Chen và cộng sự đề xuất ICE [14], tận dụng tương đồng mềm và thứ hạng để sinh nhãn cố định ổn định, kết hợp học tương phản nhằm giảm phương sai nội lớp. Mặc dù tăng cường tính ổn định, ICE có độ phức tạp cao và dễ suy giảm hiệu quả khi các cụm có sự chồng lấn mạnh. Dai và cộng sự [23] giới thiệu học tương phản theo cụm, tối ưu đồng thời nhất quán nội cụm và khác biệt liên cụm, với kho bộ nhớ lưu tâm cụm hỗ trợ huấn luyện ổn định. Phương pháp này có ưu điểm là cân bằng được tính phân biệt trong và giữa cụm, nhưng tốn thêm bộ nhớ và nhạy cảm với tham số phân cụm. Cho và cộng sự phát triển PPLR [20] sử dụng tinh chỉnh nhãn giả dựa trên độ tương hợp giữa đặc trưng toàn cục và bộ phận thông qua chỉ số đồng ý chéo. Phương pháp này khai thác hiệu quả thông tin đa mức nhưng dễ bị ảnh hưởng bởi chất lượng tách đặc trưng bộ phận, đặc biệt khi ảnh bị che khuất. Zhang và cộng sự đưa ra ISE [123], sử dụng nội suy giữa các mẫu trong cùng cụm để tạo điểm hỗ trợ ẩn, giúp tăng mật độ đặc trưng nhưng có nguy cơ sinh ra mẫu ảo thiếu tính đại diện. Zhao và cộng sự đề xuất DHCL [124] , một khung học tương phản lai động, phân biệt giữa các mẫu đã gán nhãn và chưa phân cụm, đồng thời kết hợp học tương phản ở cả mức cá thể và cụm nhằm tăng cường khả năng phân biệt và tổng quát hóa của mô hình. Tuy nhiên, mô hình này có cấu trúc phức tạp và chi phí tính toán cao. Tương tự như CAP và ICE, Liu và cộng sự đã giới thiệu CER [66], cũng tận dụng thông tin về camera để xử lý vấn đề về nhiễu trong nhãn giả nhưng vẫn bị hạn chế ở môi trường thiếu thông tin camera. Một hướng phát triển khác là sử dụng mạng tích chập đồ thị (GCN) để nâng cao khả năng biểu diễn đặc trưng tổng quát. CGMAL được phát triển bởi Ran và cộng sự [90] đề xuất áp dụng GCN, đồng thời giới thiệu một cơ chế huấn luyện đối kháng nhằm chuyển giao tri thức mà GCN đã học được sang bộ trích xuất đặc trưng nhưng đòi hỏi chi phí cao trong xây dựng và tối ưu đồ thị. Tương tự, Zhu và cộng sự [134] cũng dựa trên các mạng GCN và kết hợp thêm kỹ thuật tăng độ phân giải cho ảnh để cải thiện độ chính xác của quá trình học, song việc xử lý ảnh độ phân giải cao lại tiêu tốn tài nguyên và phụ thuộc mạnh vào thiết kế GCN.

Các nghiên cứu trên chủ yếu sử dụng mạng nơ-ron tích chập làm kiến trúc cốt lõi, tuy nhiên trong thực tế, mô hình gặp một số hạn chế như khó nắm bắt mối quan hệ không gian dài hạn, thiếu khả năng biểu diễn thông tin ngữ nghĩa cấp cao, và dễ suy giảm hiệu năng khi gặp với các điều kiện bất lợi như che khuất, ánh sáng yếu hay thay đổi góc nhìn. Điều này thúc đẩy việc nghiên cứu các kiến trúc hiện đại hơn như Transformer [103] để cải thiện khả năng biểu diễn ngữ cảnh và tính tổng quát của mô hình. Một số nghiên cứu đã ứng dụng ViT vào làm mạng trích xuất đặc trưng chính và đạt được những kết quả đầy hứa hẹn.

Cụ thể, Luo và cộng sự [69] đề xuất sử dụng học tự giám sát trên tập LUPerson, đồng thời giới thiệu chỉ số CFS để chọn dữ liệu tiền huấn luyện gần miền tái định danh. Phương pháp còn kết hợp với mô-đun ICS nhằm giảm khoảng cách miền và cải thiện hiệu năng trong cả thiết lập có giám sát và không giám sát. Phương pháp này giúp nâng cao tính tổng quát và hiệu năng trong cả thiết lập tự giám sát, song có hạn chế là cần tính toán CFS trên tập lớn gây tốn kém tài nguyên. Tiếp nối hướng tiếp cận này, Zhu và cộng sự đề xuất phương pháp PASS [136] sử dụng token [PART] để học và so khớp thông tin riêng biệt của từng vùng ảnh thay vì ghép toàn bộ đặc trưng cục bộ với đặc trưng toàn cục như các phương pháp thông thường để tăng khả năng nhận dạng chi tiết, tuy nhiên, số lượng token nhiều cũng khiến chi phí tính toán tăng và dễ phát sinh nhiễu khi vùng ảnh kém chất lượng. Trong khi đó, Li và cộng sự [60] đề xuất kiến trúc hai nhánh để trích xuất đặc trưng đa mức, và kết hợp với phương pháp O2CAP, một chiến lược học tương phản kết hợp giữa thông tin trực tuyến và ngoại tuyến – nhằm tăng hiệu quả học trong thiết lập không giám sát, nhưng độ phức tạp mô hình lại là thách thức khi mở rộng. Tập trung vào việc khai thác mẫu khó, Ji và cộng sự [50] đề xuất khung học đặc trưng đối kháng ACFL giúp tăng khả năng phân biệt bằng cách tạo ra mẫu khó từ các quan hệ có độ tin cậy cao giữa các ảnh chưa gán nhãn, sau đó huấn luyện mô hình học đặc trưng phân biệt tốt hơn, song phương pháp này lại có nguy cơ tạo ra mẫu giả không thực sự đại diện. Nhằm giải quyết đồng thời vấn đề nhiễu từ các vùng trong ViT và sự không nhất quán đặc trưng trong học tương phản, Zhu và cộng sự [137] đề xuất phương pháp TCMM gồm hai thành phần: ViT Token Constraint để giảm nhiễu từ các vùng trong ViT và kho bộ nhớ đa tỉ lệ nhằm duy trì tính nhất quán đặc trưng, đồng thời khai thác hiệu quả các mẫu ngoại lai thường bị bỏ qua trong các phương pháp trước, nhưng cũng làm tăng đáng kể chi phí bộ nhớ và độ phức tạp huấn luyện. Cuối cùng, Li và Gong [59] đề xuất phương pháp tinh chỉnh trực tiếp bộ mã hóa ảnh của CLIP bằng hàm mất mát tương phản nguyên mẫu cho bài toán tái định danh, thay thế cho học truy vấn vốn không được sử dụng trong bối cảnh hình ảnh không có nhãn ngữ nghĩa. Phương pháp đạt hiệu quả cạnh tranh trong cả thiết lập có giám sát và không giám sát, vượt qua CLIP-ReID trên nhiều tập dữ liệu.

Để tóm tắt nghiên cứu có liên quan trong phần này, Bảng 1.4 được trình bày để so sánh các phương pháp giải quyết bài toán tái định danh theo hướng tiếp cận học không giám sát với các thông tin: Phương pháp, Nơi công bố, Cách tiếp cận, Bộ dữ liệu và kết quả (mAP(%)). Các tên viết tắt Market-1501 (M), Duke (D), MSMT17 (MT) được sử dụng để biểu thị các tập dữ liệu tương ứng trong quá trình huấn luyện và đánh giá.

Bảng 1.4: Tóm tắt kết quả một số nghiên cứu tái định danh người không giám sát

| Phương pháp | Nơi công bố | Cách tiếp cận | Bộ dữ liệu | mAP (%) |
|---|---|---|---|---|
| SpCL [32] | NIPS 2020 | Sử dụng chiến lược học tự điều chỉnh và bộ nhớ kết hợp nhằm lựa chọn mẫu tin cậy trong huấn luyện. | M, MT, | 73.1, 36.9 |
| CAP [108] | AAAI 2021 | Sử dụng đại diện theo từng camera để xử lý sự sai lệch đặc trưng đa góc quay, qua đó cải thiện độ chính xác gán nhãn. | M, D, MT, | 79.2, 67.3, 36.9 |
| ICE [14] | ICCV 2021 | Tận dụng tương đồng mềm và thứ hạng để sinh nhãn cố định, kết hợp học tương phản nhằm giảm phương sai nội lớp. | M, D, MT, | 82.3, 69.9, 38.9 |
| CC [23] | ACCV 2022 | Giới thiệu học tương phản theo cụm, tối ưu đồng thời nhất quán nội cụm và khác biệt liên cụm, với bộ nhớ động lưu tâm cụm hỗ trợ huấn luyện ổn định. | M, MT | 83.0. 33.0 |
| PPLR [20] | CVPR 2022 | Tinh chỉnh nhãn giả dựa trên độ tương hợp giữa đặc trưng toàn cục và bộ phận thông qua chỉ số cross-agreement. | M, MT | 84.4, 42.2 |
| ISE [123] | CVPR 2022 | Sử dụng nội suy giữa các mẫu trong cùng cụm để tạo điểm hỗ trợ ẩn, giúp tăng mật độ đặc trưng. | M, MT | 85.3, 37.0 |

| Phương pháp | Nơi công bố | Cách tiếp cận | Bộ dữ liệu | mAP (%) |
|---|---|---|---|---|
| Trans-ReID-SSL [69] | Arxiv 2021 | Sử dụng học tự giám sát trên tập LUPerson, kết hợp với CFS và ICS nhằm giảm khoảng cách miền và cải thiện hiệu năng. | M, MT | 89.6, 50.6 |
| PASS [136] | ECCV 2022 | Sử dụng token [PART] để học và so khớp thông tin riêng biệt của từng vùng ảnh. | M, MT | 88.5, 41.0 |
| PCL-CLIP [59] | Arxiv 2023 | Tinh chỉnh trực tiếp bộ mã hóa ảnh của CLIP bằng hàm mất mát tương phản nguyên mẫu. | M, MT | 88.4, 65.5 |
| hline TMGF [60] | WACV 2023 | Sử dụng kiến trúc hai nhánh để trích xuất đặc trưng đa mức, và kết hợp với phương pháp O2CAP, một chiến lược học tương phản kết hợp giữa thông tin trực tuyến và ngoại tuyến. | M, D, MT | 89.5, 76.8, 58.2 |
| ACFL-ViT [50] | PR 2024 | Sử dụng khung học đặc trưng đối kháng ACFL giúp tăng khả năng phân biệt bằng cách tạo ra mẫu khó từ các quan hệ có độ tin cậy cao giữa các ảnh chưa gán nhãn. | M, MT | 89.1, 45.7 |
| TCMM [137] | Arxiv 2025 | Sử dụng ViT Token Constraint để giảm nhiễu từ các vùng trong ViT và kho bộ nhớ đa tỉ lệ nhằm duy trì tính nhất quán đặc trưng. | M, MT | 90.5, 52.0 |

Nhìn chung, các phương pháp trên đã góp phần nâng cao hiệu quả mô hình tái định danh người không giám sát thông qua các hướng tiếp cận như học tương phản, tinh chỉnh nhãn giả, tăng cường dữ liệu và khai thác đặc trưng đa mức, nhằm giải quyết các thách thức như nhiễu nhãn, sai lệch giữa camera và khác biệt miền. Đặc biệt, xu hướng gần đây tập trung vào việc ứng dụng mạng cốt lõi ViT cùng với các thiết kế mô hình và chiến lược huấn luyện chuyên biệt như trích đặc trưng cục bộ, mô hình đa nhánh, sinh mẫu khó và tận dụng mô hình tiền huấn luyện lớn, giúp thu hẹp khoảng cách với các phương pháp có giám sát.

# 1.4 Các nghiên cứu trong nước

Tại Việt Nam, hướng nghiên cứu về tái định danh người mới xuất hiện hơn một thập kỷ qua và đã phát triển qua nhiều giai đoạn, từ các phương pháp truyền thống đến học sâu hiện đại, ngày càng gắn kết hơn với ứng dụng hệ thống giám sát thông minh. Một trong những phương pháp đầu tiên được đề xuất của tác giả Trần Thanh Toàn và cộng sự [4] năm 2015 hướng tới việc loại bỏ vùng nền trong khung ảnh đối tượng đã được phát hiện và củng cố đặc trưng tách từ các mảnh ghép cục bộ dùng để phân biệt hai đối tượng khác nhau. Việc kết hợp này nhằm nâng cao mức độ tin cậy của việc truy vấn cùng một đối tượng ở hai camera khác nhau với góc nhìn không chồng lấp. Tiếp theo, Bao và cộng sự đã công bố phương pháp xếp hạng lại dựa trên thông tin đồng xuất hiện [78]. Những năm tiếp theo, nhóm nghiên cứu của Lan và cộng sự tại trường Đại học Bách khoa Hà Nội đã có những công bố liên quan tái định danh người. Một số nghiên cứu đã tập trung vào việc cải tiến đặc trưng thủ công như KDES [83] hay GOG [80] kết hợp với các kỹ thuật xử lý nền và loại bỏ bóng, qua đó nâng cao độ ổn định và khả năng phân biệt trong môi trường giám sát. Bên cạnh đó, nhiều công trình đã khai thác hướng ảnh–video thông qua cơ chế tổng hợp đặc trưng tuần tự hoặc các chiến lược kết hợp nhằm tận dụng thông tin từ chuỗi khung hình, giúp cải thiện hiệu năng so với cách tiếp cận ảnh đơn [41, 79].

Một số nghiên cứu khác hướng tới xây dựng một khung thống nhất bao gồm phát hiện, trích xuất đặc trưng và so khớp, cho thấy tiềm năng ứng dụng thực tế cao hơn, song hiệu năng vẫn phụ thuộc mạnh vào độ chính xác của khâu phát hiện người [42].

Kết quả từ hướng nghiên cứu cụ thể đó, năm 2021, Nguyễn Thúy Bình đã bảo vệ thành công luận án với chủ đề "Person Re-identification in a Surveillance Camera Network" [1]. Luận án nghiên cứu về tái định danh người trong hệ thống camera giám sát đưa ra hai đóng góp chính. Thứ nhất, đề xuất phương pháp tái định danh dựa trên nhiều thể hiện với bốn bước cơ bản: lựa chọn khung hình đại diện, trích chọn đặc trưng, tổng hợp và đối sánh. Thay vì sử dụng toàn bộ khung hình, chỉ một số khung hình đại diện (theo chu kỳ hoặc một nhóm khung hình đặc trưng) được chọn để giảm chi phí tính toán, dung lượng lưu trữ nhưng vẫn duy trì độ chính xác. Thứ hai, luận án khai thác hiệu quả riêng của từng loại đặc trưng bằng cách kết hợp chúng theo nhiều chiến lược, bao gồm cả gán trọng số bằng nhau và trọng số thích nghi với người cần truy vấn, từ đó nâng cao hiệu quả biểu diễn và tái định danh trong các ngữ cảnh khác nhau.

Một luận án khác của Nguyễn Hồng Quân với chủ đề "Nghiên cứu và phát triển phương pháp trích chọn đặc trưng dựa trên học sâu trong tái định danh người" [3] đã tập trung vào hai hướng cải tiến chính cho bài toán tái định danh người từ dữ liệu camera giám sát. Hướng thứ nhất là cải tiến mạng ResNet-50 nhằm khai thác thông tin cục bộ, đồng thời đề xuất phương pháp đối sánh dựa trên độ đo EMD để tăng khả năng phân biệt trong điều kiện dữ liệu biến thiên. Bên cạnh đó, một phương pháp nén mạng học sâu cũng được phát triển và thử nghiệm trên mạng VGG16 và VGG16-SSD, qua đó cho thấy tính khả thi khi triển khai trên phần cứng FPGA (vi mạch dùng cấu trúc mảng phần tử lô-gic có thể lập trình được). Hướng thứ hai là cải tiến mô hình tái định danh dựa trên chuỗi hình ảnh, trong đó các biến thể RNN được áp dụng để tổng hợp đặc trưng theo trình tự khung hình, giúp tăng độ ổn định và cải thiện chất lượng nhận dạng trong dữ liệu video.

Giai đoạn gần đây, các công trình nghiên cứu đã bắt nhịp với xu hướng quốc tế, chú trọng nhiều hơn đến khía cạnh ứng dụng thực tiễn trong hệ thống giám sát thông minh. Tại Trường Đại học Công nghệ Thông tin - ĐHQG TP.HCM, nhóm nghiên cứu của Nghĩa và cộng sự đã xây dựng hệ thống tìm kiếm và phân tích thông tin con người từ camera giám sát, bước đầu tích hợp khả năng tái định danh vào bài toán giám sát tổng thể [2]. Tại đại học Bách Khoa Hà Nội và Viện CMC, hướng kết hợp đặc trưng khuôn mặt và thân thể được triển khai để nâng cao độ chính xác trong điều kiện đối tượng mặc trang phục tương tự hoặc bị che khuất [77]. Tiếp theo, Huy và cộng sự [12] công bố phương pháp kết hợp đặc trưng bất định (UFFM) với kết hợp độ đô thông minh (WDA) nhằm nâng cao tính ổn định và hiệu quả trên các bộ dữ liệu đánh giá tiêu chuẩn. Năm 2025, nhóm này tiếp tục đề xuất KWF để cải tiến bước xếp hạng lại [13].

Nhìn chung, nghiên cứu về tái định danh người tại Việt Nam đã có những bước tiến rõ rệt, từ việc khai thác các đặc trưng thủ công truyền thống đến việc ứng dụng các phương pháp học sâu hiện đại, từ bài toán ảnh đơn trên các bộ dữ liệu quy mô nhỏ như VIPeR, ETHZ đến các phương pháp tiên tiến áp dụng trên những bộ dữ liệu tiêu chuẩn lớn hơn như CUHK, Market, DukeC và MSMT. Đồng thời, nhiều hướng tiếp cận trong nước đã bắt nhịp với xu thế quốc tế, chẳng hạn như khai thác Transformer, mô hình bất định, kỹ thuật xếp hạng lại thông minh hay kết hợp đa phương pháp. Tiếp nối định hướng đó, trong luận án này, chúng tôi đề xuất một số phương pháp mới dựa trên các kiến trúc học sâu hiện đại, với mục tiêu chính là nâng cao hiệu năng tổng thể tái định danh người, để từ đó có thể mở rộng khả năng ứng dụng thực tiễn của bài toán trong hệ thống giám sát thông minh.

# 1.5    Khoảng trống và hướng tiếp cận nghiên cứu

Trên cơ sở các đánh giá và phân tích tổng quan về bài toán tái định danh người cũng như những phương pháp và kỹ thuật đã được các nhóm nghiên cứu triển khai, ba khoảng trống và hướng tiếp cận nghiên cứu được luận án xác định như sau.

- *Khoảng trống nghiên cứu thứ nhất*: Trong hướng tiếp cận học có giám sát, các mô hình học sâu kết hợp với DML đã trở thành cách tiếp cận chính trong bài toán tái định danh người nhờ khả năng học biểu diễn đặc trưng có tính phân biệt cao thông qua việc tối ưu các hàm mất mát. Tuy nhiên, các hàm mất mát phổ biến hiện nay vẫn tồn tại nhiều hạn chế: hàm mất mát phân loại khó tổng quát hóa các danh tính chưa thấy; hàm mất mát trung tâm thiếu khả năng phân biệt liên lớp; hàm mất mát bộ ba phụ thuộc vào chọn mẫu khó và thiếu ổn định; hàm mất mát bộ ba trọng tâm tăng độ ổn định nhưng khó duy trì tâm cụm chính xác với tập lớn. Bên cạnh đó, mặc dù hàm mất mát tương phản có giám sát đã chứng minh hiệu quả vượt trội trong việc khai thác thông tin nhãn giám sát, song chúng vẫn chưa được áp dụng một cách hệ thống cho bài toán tái định danh người và chưa khai thác triệt để ưu thế của thông tin nhãn khi không kết hợp cùng các hàm mất mát khác.

  Từ khoảng trống nghiên cứu thứ nhất, luận án xác định *hướng tiếp cận thứ nhất* là nghiên cứu, phân tích ưu và nhược điểm của các hàm mất mát phổ biến được sử dụng trong bài toán tái định danh người. Qua đó, luận án đề xuất chiến lược kết hợp hàm tương phản có giám sát với bốn hàm mất mát khác: phân loại, trung tâm, bộ ba và bộ ba trong tâm nhằm tận dụng toàn diện các đặc tính bổ sung lẫn nhau giữa chúng. Hướng tiếp cận này sẽ góp phần tăng cường khả năng phân biệt đặc trưng và nâng cao hiệu năng tổng thể của hệ thống tái định danh người trong thiết lập học có giám sát.

- *Khoảng trống nghiên cứu thứ hai*: Trong hướng tiếp cận học thích ứng miền không giám sát, thách thức lớn nhất là sự khác biệt về phân phối dữ liệu giữa miền nguồn và miền đích, khiến mô hình huấn luyện trên miền nguồn thường giảm mạnh hiệu năng khi áp dụng sang miền đích. Để khắc phục, cần có các phương pháp thu hẹp khoảng cách giữa hai miền như chuyển đổi phong cách ảnh hoặc học đặc trưng bất biến miền. Bên cạnh đó, việc chỉ khai thác đặc trưng toàn cục giúp tăng tính phân biệt nhưng bỏ qua đặc trưng cục bộ giàu ngữ cảnh, làm hạn chế khả năng biểu diễn của mô hình. Do đó, cần kết hợp cả hai loại đặc trưng để nâng cao hiệu quả huấn luyện. Cuối cùng, độ tin cậy của nhãn giả vẫn là vấn đề lớn, do phân cụm sai có thể làm nhiễu quá trình huấn luyện.

Từ khoảng trống nghiên cứu thứ hai, luận án xác định *hướng tiếp cận thứ hai* là nghiên cứu, phát triển phương pháp nâng cao hiệu năng của bài toán tái định danh người trong thiết lập học thích ứng miền không giám sát thông qua các cách tiếp cận chính: giảm sự sai lệch giữa phân phối đặc trưng của miền nguồn và miền đích, đánh giá chất lượng ảnh, tăng cường khả năng biểu diễn đặc trưng bằng cách kết hợp thông tin từ cả đặc trưng toàn cục và đặc trưng cục bộ, và tinh chỉnh nhãn giả nhằm nâng cao độ chính xác và độ tin cậy của nhãn huấn luyện trong miền mục tiêu.

- *Khoảng trống nghiên cứu thứ ba*: Trong hướng tiếp cận học không giám sát, mạng cốt lõi ResNet-50 được sử dụng rộng rãi nhờ tính ổn định, nhưng còn hạn chế trong việc mô hình hóa quan hệ không gian giữa các vùng cơ thể. Mặt khác, mạng cốt lõi ViT có thể khắc phục nhược điểm này nhờ khả năng tự chú ý, nhưng phần lớn nghiên cứu hiện nay chỉ khai thác đặc trưng toàn cục, trong khi đặc trưng cục bộ chưa được tận dụng hiệu quả. Bên cạnh đó, thông tin định danh camera, nếu được tích hợp một cách phù hợp vào quá trình huấn luyện, có thể giúp điều chỉnh không gian đặc trưng theo ngữ cảnh quan sát, từ đó nâng cao hiệu quả phân biệt liên miền trong các thiết lập không giám sát.

Từ khoảng trống nghiên cứu thứ ba, luận án xác định *hướng tiếp cận thứ ba* là nghiên cứu, phân tích ưu và nhược điểm của kiến trúc CNN và ViT để làm cơ sở cho đề xuất sử dụng ViT nhằm khai thác hiệu quả thông tin ngữ cảnh toàn cục và chi tiết cục bộ thông qua cơ chế tự chú ý. Ngoài ra, việc tích hợp thông tin định danh camera vào quá trình huấn luyện giúp giảm thiểu độ lệch miền do sự khác biệt về góc nhìn, điều kiện ánh sáng và độ phân giải, qua đó tăng cường tính nhất quán và khả năng thích ứng giữa các nguồn camera.

Với ba khoảng trống nghiên cứu đã được xác định, ba hướng tiếp cận mà luận án đề xuất mang tính bao quát và phản ánh toàn diện các thiết lập phổ biến trong thực tiễn của bài toán tái định danh người.

Thứ nhất, học có giám sát được khai thác trong bối cảnh mô hình được huấn luyện trên tập dữ liệu có đầy đủ nhãn danh tính. Hướng tiếp cận này cho phép tận dụng triệt để thông tin nhãn để xây dựng không gian biểu diễn đặc trưng có tính phân biệt cao, qua đó hình thành nền tảng mô hình mạnh mẽ. Đây cũng là thiết lập chuẩn trong hầu hết các nghiên cứu hiện nay, đồng thời là cơ sở quan trọng để so sánh, đánh giá hiệu quả của các cách tiếp cận khác trong điều kiện thiếu hoặc không có nhãn.

Thứ hai, trong các ứng dụng thực tế nơi mô hình cần được triển khai sang miền đích không có nhãn, luận án tiếp cận theo hướng học thích ứng miền không giám sát. Trong thiết lập này, mô hình được huấn luyện trên miền nguồn có nhãn và sau đó thích ứng với miền đích thông qua một loạt kỹ thuật như điều chỉnh đặc trưng để giảm độ chênh lệch phân phối giữa hai miền, phân cụm dữ liệu, gán nhãn giả, tinh chỉnh nhãn giả nhằm tăng độ tin cậy, và kết hợp đặc trưng để nâng cao tính ổn định của mô hình. Cách tiếp cận này phản ánh thực tế triển khai của nhiều hệ thống giám sát quy mô lớn, nơi dữ liệu liên tục thay đổi và việc gán nhãn thủ công là không khả thi.

Thứ ba, trong kịch bản thách thức hơn, khi cả miền nguồn và miền đích đều không có nhãn, luận án phát tiếp cận theo hướng học không giám sát hoàn toàn. Phương pháp này tận dụng các kỹ thuật như phân cụm, tự huấn luyện, học đặc trưng tự giám sát hoặc kết hợp với các nguồn thông tin phụ trợ khác như định danh camera hoặc thông tin theo ngữ cảnh để củng cố cấu trúc cụm. Mục tiêu là tối ưu hiệu năng mô hình mà không cần bất kỳ nhãn danh tính nào, từ đó mở rộng khả năng ứng dụng cho các tình huống thực tế đòi hỏi tính linh hoạt cao, nơi việc thu thập hoặc gán nhãn dữ liệu gần như không thể thực hiện.

Tổng thể, ba hướng tiếp cận này bổ sung cho nhau, bao quát các mức độ sẵn có của nhãn từ đầy đủ, thiếu nhãn đến hoàn toàn không có nhãn. Điều này giúp luận án không chỉ giải quyết ba khoảng trống nghiên cứu mà còn cung cấp hướng tiếp cận phù hợp với yêu cầu triển khai của hệ thống tái định danh người trong thực tế.

## 1.6    Kết luận chương

Chương 1 đã giới thiệu cụ thể về bài toán tái định danh người, trình bày các kiến thức nền tảng cũng như đưa ra tổng quan các nghiên cứu liên quan. Trên cơ sở đó, luận án đã xác định được những khoảng trống nghiên cứu, từ đó đề xuất các hướng tiếp cận phù hợp nhằm nâng cao hiệu năng tái định danh người theo ba hướng chính là học có giám sát, học thích ứng miền không giám sát và học không giám sát.

Trong các chương tiếp theo, luận án sẽ trình bày chi tiết từng phương pháp đề xuất tương ứng với ba hướng tiếp cận trên, nhằm giải quyết bài toán tái định danh người một cách toàn diện, có hệ thống và hiệu quả.

# Chương 2
# NÂNG CAO HIỆU NĂNG
# TÁI ĐỊNH DANH NGƯỜI CÓ GIÁM SÁT
# DỰA TRÊN KẾT HỢP CÁC HÀM MẤT MÁT

Trong chương này, luận án trình bày phương pháp đề xuất "SCM-ReID" để nâng cao hiệu năng tái định danh người theo hướng tiếp cận có giám sát bằng cách sử dụng hàm mất mát tương phản có giám sát kết hợp với bốn hàm mất mát phổ biến gồm: phân loại, trung tâm, bộ ba và bộ ba trọng tâm. Trong đó, hàm mất mát tương phản có giám sát đóng vai trò thúc đẩy mô hình học được các biểu diễn đặc trưng có tính phân biệt cao, thông qua việc khai thác đồng thời mối quan hệ giữa các mẫu cùng và khác định danh trong cùng một lô huấn luyện. Bốn hàm mất mát còn lại hỗ trợ tối ưu hóa ranh giới phân lớp, đảm bảo sự ổn định của không gian đặc trưng và tăng cường độ chính xác trong việc phân biệt các danh tính. Việc tích hợp các hàm mất mát này cho phép phát huy đồng thời những ưu điểm riêng biệt của từng hàm, từ đó nâng cao chất lượng biểu diễn đặc trưng và hiệu quả tổng thể của mô hình. Các kết quả nghiên cứu trong chương này được công bố tại công trình [CT2].

## 2.1 Đặt vấn đề

Trong bối cảnh của bài toán tái định danh người, nhiều hàm mất mát khác nhau đã được đề xuất để tối ưu hiệu năng của mô hình. Mỗi hàm mất mát có những ưu và nhược điểm riêng, cũng như phải đối mặt với những hạn chế cụ thể [96]. Trong số đó, hàm mất mát phân loại được sử dụng rộng rãi trong bài toán tái định danh người vì tính đơn giản và sự ổn định trong quá trình huấn luyện [131]. Nó giúp mô hình học các đặc điểm cụ thể của danh tính bằng cách phân loại trực tiếp từng người trong tập huấn luyện. Tuy nhiên, nó không tối ưu hóa không gian đặc trưng cho các tác vụ truy xuất và gặp khó khăn khi phải nhận diện các danh tính không có trong tập huấn luyện. Để cải thiện tính chặt chẽ trong các lớp, hàm mất mát trung tâm được sử dụng để giảm thiểu khoảng cách giữa các đặc trưng và tâm cụm của chúng [68], nhưng nó thiếu sự tách biệt giữa các lớp và làm tăng thêm độ phức tạp do thuật toán cần cập nhật tâm cụm một cách linh hoạt. Đối với không gian đặc trưng phân biệt hơn, hàm mất mát bộ ba được sử dụng để tối ưu hóa trực tiếp bằng cách áp dụng biên độ giữa các cặp ảnh của danh tính giống nhau và khác nhau [27]. Nó sẽ giúp cho mô hình biểu diễn đặc trưng của các mẫu có cùng danh tính ở gần nhau trong khi các danh tính khác nhau nằm xa nhau. Tuy nhiên, hàm bộ ba đòi hỏi phải lựa chọn cẩn thận các bộ ba hình ảnh để có thể kết hợp hiệu quả. Quá trình này có thể tốn kém về mặt tính toán và không ổn định trong quá trình huấn luyện. Việc lựa chọn bộ ba không chất lượng có thể dẫn đến quá trình huấn luyện hội tụ chậm hoặc hiệu năng không tối ưu. Dựa trên hàm mất mát bộ ba, hàm mất mát bộ ba trọng tâm sử dụng trọng tâm của các lớp thay vì các mẫu riêng lẻ [116]. Phương pháp này làm tăng tính ổn định của quá trình huấn luyện, giảm sự phụ thuộc vào khai thác mẫu cứng và thúc đẩy cả tính chặt chẽ trong lớp và khả năng tách biệt giữa các lớp. Tuy nhiên, việc duy trì các trọng tâm chính xác, đặc biệt là trong các tập dữ liệu quy mô lớn, có thể tăng thêm chi phí tính toán.

Bên cạnh đó, các nghiên cứu trong lĩnh vực nhận dạng gần đây đã đưa ra phương pháp học tương phản tự giám sát để tận dụng dữ liệu không có nhãn quy mô lớn mà không cần phải gán nhãn thủ công [16]. Mặc dù phương pháp này đạt được kết quả khả quan, nhưng nó vẫn chưa tận dụng được lợi thế nhãn của dữ liệu. Để giải quyết hạn chế này, Khosla và cộng sự [54] đề xuất mở rộng khái niệm học tương phản tự giám sát thành học tương phản có giám sát. Cụ thể, học tương phản có giám sát đưa ra hàm mất mát tương phản có giám sát (SCL). Đây là phiên bản tổng quát hơn của hàm mất mát bộ ba [114] và N-loss [94] bằng cách so sánh từng mẫu định danh với cùng lúc nhiều mẫu cùng định danh và khác định danh khác trong một lô. Việc sử dụng nhiều mẫu tương đồng và khác nhau cho mỗi mẫu định danh giúp mô hình hoạt động tốt hơn, đặc biệt là trong các tác vụ như tái định danh người, trong đó cùng một người có thể xuất hiện dưới nhiều góc nhìn, điều kiện ánh sáng và bối cảnh khác nhau.

Hiện nay, nhiều phương pháp tái định danh người dựa vào các hàm mất mát riêng lẻ hoặc kết hợp tuyến tính nhiều hàm để cải thiện hiệu năng. Tuy nhiên, chúng vẫn chưa giải quyết được hoàn toàn những thách thức cốt lõi để đạt được kết quả vượt trội. Điều này làm nổi bật một hướng tiếp cận quan trọng: làm thế nào để thiết kế một chiến lược huấn luyện thống nhất, có khả năng kết hợp các điểm mạnh của các hàm mất mát khác nhau nhằm tăng tính đồng nhất trong từng lớp, nâng cao khả năng phân biệt giữa các lớp, và cải thiện độ bền vững tổng thể của mô hình trong các môi trường đa dạng và quy mô lớn.

Xuất phát từ nhận định đó, luận án hướng đến việc đề xuất một phương pháp toàn diện hơn, không chỉ tận dụng cấu trúc quan hệ được SCL mang lại, mà còn củng cố các thông tin hữu ích qua các hàm mất mát truyền thống. Điều này dẫn tới câu hỏi nghiên cứu: Có thể phát triển một phương pháp thống nhất kết hợp hiệu quả SCL với bốn hàm mất mát phổ biến khác - phân loại, trung tâm, bộ ba và bộ ba trọng tâm - để nâng cao hiệu năng tái định danh người không?

Để trả lời câu hỏi này, luận án đề xuất phương pháp kết hợp học tương phản có giám sát với học độ đo sâu, có tên là "SCM-ReID", nhằm nâng cao hiệu năng tái định danh người. Cụ thể, chúng tôi tích hợp SCL [54] cùng với bốn hàm mất mát phổ biến: phân loại [128], trung tâm [115], bộ ba [75], và bộ ba trọng tâm [116] từ mô hình cơ sở [116]. Mục tiêu trong phương pháp đề xuất này là khai thác sức mạnh của hàm mất mát tương phản có giám sát kết hợp với sự tương hỗ giữa bốn loại hàm mất mát phổ biến nhằm nâng cao khả năng phân biệt đặc trưng, tăng cường độ bền vững của mô hình trước nhiễu, đồng thời đạt được hiệu năng tiên tiến trên các bộ dữ liệu chuẩn trong bài toán tái định danh người. Để đánh giá hiệu quả của phương pháp SCM-ReID, luận án đã thực hiện các thực nghiệm toàn diện trên ba bộ dữ liệu chuẩn trong bài toán tái định danh người: Market-1501 [127], CUHK03-Detected [61] và CUHK03-Labeled [61]. Kết quả cho thấy phương pháp đề xuất SCM-ReID có hiệu năng vượt trội hơn các phương pháp SOTA hiện có về cả mAP và Rank-n.

Khái quát lại, các đóng góp nổi bật của luận án trong hướng tiếp cận này bao gồm:

- Đề xuất sử dụng học tương phản có giám sát cho bài toán tái định danh người. Cụ thể, phương pháp sử dụng SCL để tận dụng đồng thời nhiều mẫu tương đồng và khác nhau cho mỗi mẫu định danh, giúp tối ưu hóa độ chặt chẽ trong lớp và sự tách biệt giữa các lớp.

- Đề xuất phương pháp SCM-ReID nhằm tích hợp SCL với bốn hàm mất mát thông dụng: phân loại, trung tâm, bộ ba, bộ ba trọng tâm. Phương pháp này tối ưu hóa hiệu năng của mô hình tái định danh người bằng cách tận dụng các lợi thế của SCL và các hàm mất mát khác.

- Thực nghiệm toàn diện trên ba bộ dữ liệu tái định danh người là Market-1501, CUHK03-Detected và CUHK03-Labeled nhằm đánh giá hiệu quả của phương pháp được đề xuất. Kết quả cho thấy SCM-ReID đạt hiệu năng cao so với các phương pháp SOTA hiện tại, khẳng định tính ưu việt và khả năng áp dụng rộng rãi của phương pháp.

Trong các phần tiếp theo, luận án sẽ trình bày chi tiết về phương pháp đề xuất SCM-ReID cùng với các thực nghiệm và đánh giá toàn diện.

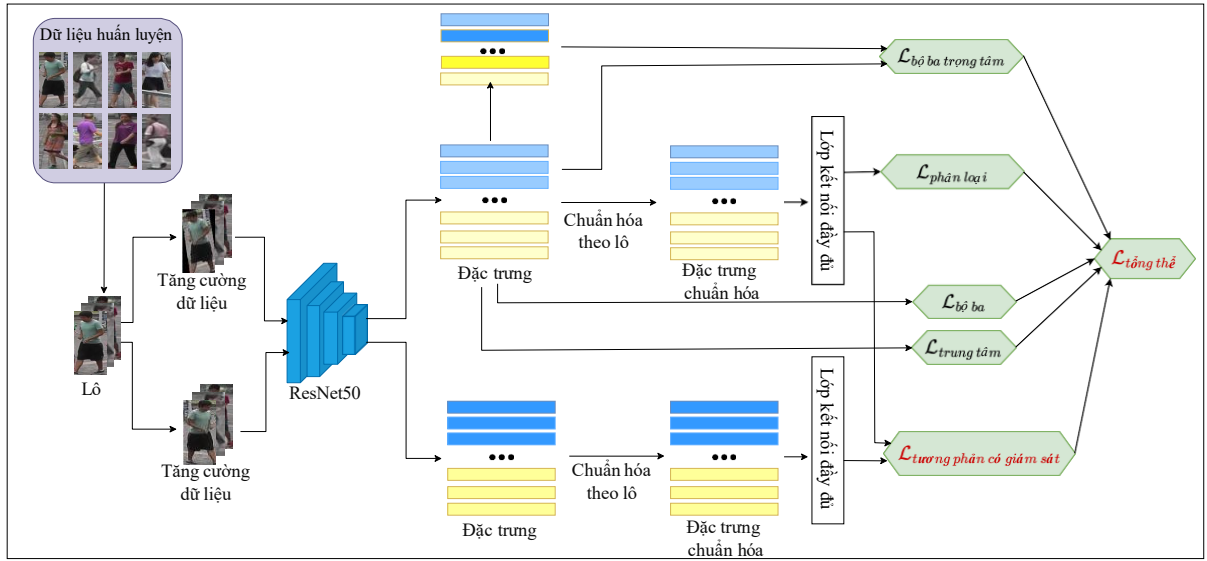

Hình 2.1: Mô hình kiến trúc của phương pháp đề xuất - SCM-ReID.

## 2.2 Phương pháp đề xuất

### 2.2.1 Hướng tiếp cận

Để giải quyết những thách thức đã được trình bày, luận án đề xuất phương pháp nhằm nâng cao hiệu năng tái định danh người trong hướng tiếp cận học có giám sát bằng cách kết hợp học tương phản có giám sát kết hợp với học độ đo sâu có tên là **SCM-ReID**. Cụ thể, chúng tôi tích hợp SCL [54] cùng với bốn hàm mất mát nổi tiếng từ mô hình cơ sở [116]. Phương pháp đề xuất được minh họa trong Hình 2.1.

Trong SCM-ReID, đầu tiên chúng tôi chia dữ liệu huấn luyện thành các lô và tạo ra hai bản sao của mỗi lô bằng cách thực hiện hai lần tăng cường dữ liệu với các kỹ thuật như trong Mục 1.2.5. Cả hai bản sao sau đó được đưa qua mạng cốt lõi để thu được véc-tơ đặc trưng. Trong phương pháp này, chúng tôi sử dụng ResNet-50 làm mạng cốt lõi, tạo ra một véc-tơ đặc trưng với kích thước 2048 chiều. Véc-tơ đặc trưng đầu tiên được sử dụng trực tiếp để tính toán các hàm mất mát trung tâm, bộ ba, và bộ ba trọng tâm. Đồng thời, nó được truyền qua lớp chuẩn hóa lô (BN) và lớp kết nối đầy đủ (FC) để tạo ra một véc-tơ đầu ra cho việc tính toán hàm mất mát phân loại. Tương tự, véc-tơ đặc trưng thứ hai từ lô bản sao thứ hai cũng được truyền qua BN và FC để thu được một véc-tơ đầu ra khác. Từ đó, SCL sẽ được tính toán sử dụng hai véc-tơ đầu ra cuối cùng từ các lô bản sao đầu tiên và thứ hai. Tổng giá trị mất mát từ năm hàm mất mát sẽ hướng dẫn quá trình huấn luyện mô hình. Sau khi huấn luyện, mô hình tối ưu sẽ được sử dụng để đánh giá kết quả. Phần tiếp theo luận án sẽ trình chi tiết về mô hình cơ sở và cách tích hợp SCL vào mô hình.

## 2.2.2   Mô hình cơ sở

Bốn hàm mất mát được sử dụng trong mô hình cơ sở, kế thừa từ nghiên cứu của Wieczorek và cộng sự [116] bao gồm là: phân loại, trung tâm, bộ ba, bộ ba trọng tâm. Công thức của bốn hàm mất mát này được tóm tắt dưới đây.

**Hàm mất mát phân loại**: Hàm mất mát phân loại [131] là lựa chọn phổ biến trong huấn luyện mô hình học sâu cho bài toán tái định danh người. Mỗi danh tính được xem như một lớp, và mô hình được tối ưu để tăng xác suất dự đoán đúng nhãn tương ứng. Việc tối ưu hóa hàm mất mát này giúp thu hẹp đặc trưng cùng danh tính, đồng thời tăng cường phân biệt giữa các danh tính khác nhau, từ đó hình thành không gian đặc trưng ổn định và có khả năng phân biệt cao. Hàm mất mát này được biểu diễn theo Công thức 2.1.

$$L_{ce} = -\log p\,(y_i \mid x_i) \tag{2.1}$$

trong đó, $x_i$ là mẫu tại chỉ số $i$; $y_i$ và $p(y_i \mid x_i)$ lần lượt là nhãn nhận dạng đúng và xác suất dự đoán của nhãn đúng cho mẫu $i$.

**Hàm mất mát trung tâm**: Hàm mất mát trung tâm [68] được sử dụng để tăng cường khả năng phân biệt bằng cách giảm biến thiên nội lớp trong không gian đặc trưng. Cụ thể, hàm này kéo các đặc trưng của cùng một danh tính tiến gần đến trung tâm lớp tương ứng, giúp các mẫu cùng danh tính có biểu diễn nhất quán hơn. Thông thường, hàm mất mát trung tâm thường được kết hợp với các hàm mất mát khác nhằm cải thiện hiệu quả huấn luyện trong bài toán tái định danh người. Hàm mất mát này được biểu diễn theo Công thức 2.2.

$$L_{ct} = \frac{1}{2}\|f_i - c_{y_i}\|_2^2 \tag{2.2}$$

trong đó, $f_i$ và $y_i$ lần lượt là véc-tơ đặc trưng và nhãn của mẫu $i$; $c_y$ là trung tâm lớp của các đặc trưng sâu tương ứng với lớp $y_i$.

**Hàm mất mát bộ ba**: Hàm mất mát bộ ba [27] là một phương pháp hiệu quả trong tái định danh người, nhằm học không gian nhúng sao cho ảnh cùng danh tính gần nhau hơn ảnh khác danh tính ít nhất một ngưỡng biên độ giá trị. Hàm mất mát này được biểu diễn theo Công thức 2.3.

$$L_{tri} = \|f_i - f_p\|_2^2 - \|f_i - f_n\|_2^2 + \alpha_{\,+} \tag{2.3}$$

trong đó, $f_i, f_p, f_n$ lần lượt là các véc-tơ đặc trưng của mẫu gốc, mẫu cùng định danh và mẫu khác định danh của mẫu $i$; $\alpha$ là tham số biên độ quy định khoảng cách tối thiểu mà mô hình cần đảm bảo giữa khoảng cách giữa các cặp cùng định danh và cặp khác định danh; $[\cdot]_+$ đại diện cho hàm hinge có giá trị bằng 0 nếu biểu thức trong hàm âm.

**Hàm mất mát bộ ba trọng tâm**: Hàm mất mát bộ ba trọng tâm [116] cải tiến từ hàm mất mát bộ ba bằng cách thay thế các mẫu cụ thể bằng trọng tâm cụm của lớp, giúp tăng tính phân biệt và ổn định của không gian đặc trưng. Nó tối ưu khoảng cách giữa mẫu và trọng tâm cụm cùng lớp so với tâm cụm khác lớp. Hàm mất mát này được biểu diễn theo Công thức 2.4.

$$L_{ctl} = \left[ \|f_i - c_p\|_2^2 - \|f_i - c_n\|_2^2 + \alpha \right]_+ \tag{2.4}$$

trong đó, $f_i$ là véc-tơ đặc trưng của mẫu $i$; $c_p$ và $c_n$ lần lượt là tâm cụm của lớp cùng định danh và lớp khác định danh với anchor $i$; $\alpha$ là tham số biên độ quy định khoảng cách tối thiểu mà mô hình cần đảm bảo giữa khoảng cách của mẫu tới tâm cụm cùng định danh với tâm cụm khác định danh; và $[\cdot]_+$ đại diện cho hàm hinge có giá trị bằng 0 nếu biểu thức trong hàm âm.

Dựa trên các hàm mất mát này, hàm mất mát tổng quát của mô hình cơ sở được trình bày trong Công thức 2.5.

$$L_{bl} = L_{ce} + \lambda_{tri} L_{tri} + \lambda_{ct} L_{ct} + \lambda_{ctl} L_{ctl} \tag{2.5}$$

trong đó, $\lambda_{tri}, \lambda_{ct}, \lambda_{ctl}$ là các siêu tham số điều chỉnh ảnh hưởng của từng hàm mất mát với hàm mất mát tổng.

Trong phạm vi luận án, các giá trị siêu tham số do tác giả [116] đề xuất ($\lambda_{tri}$ bằng 1.0, $\lambda_{ct}$ bằng 0.0005 và $\lambda_{ctl}$ bằng 1.0) được giữ nguyên, thay vì được tối ưu hoá lại trên toàn bộ không gian siêu tham số. Các giá trị này đã được kiểm chứng thực nghiệm trên nhiều bộ dữ liệu chuẩn trong bài toán tái định danh người, và việc tái tối ưu hoá sẽ làm phát sinh chi phí tính toán đáng kể trong khi không nằm trong trọng tâm đóng góp của luận án. Việc kế thừa trực tiếp các siêu tham số tối ưu giúp đảm bảo tính so sánh công bằng giữa phương pháp đề xuất và phương pháp gốc.

## 2.2.3 Hàm mất mát tương phản có giám sát

Hàm mất mát tương phản có giám sát [54] tận dụng toàn bộ thông tin có nhãn trong một lô bằng cách so sánh mỗi mẫu với tất cả các mẫu cùng và khác danh tính, thay vì chỉ một cặp cùng và khác danh tính như hàm mất mát bộ ba. Nhờ đó, mô hình có thể xử lý tốt hơn sự biến thiên nội lớp và độ tương đồng giữa các lớp, giúp nâng cao hiệu quả nhận dạng trong các điều kiện và góc nhìn khác nhau. Hàm này được biểu diễn lại như trong Công thức 2.6.

$$L_{sup} = \frac{-1}{|P(i)|} \sum_{p \in P(i)} \log \frac{\exp(sim(z_i, z_p)/\tau)}{\sum_{a \in N(j)} \exp(sim(z_i, z_a)/\tau)} \qquad (2.6)$$

ở đây, $z_i = g(f(Aug(x_i)))$, với $x_i$ là mẫu hình ảnh tại chỉ số $i$; $Aug(\cdot)$ là hàm tăng cường dữ liệu; $f(\cdot)$ là mạng cốt lõi; $g(\cdot)$ là một mạng nơ-ron nhỏ giúp ánh xạ các biểu diễn vào không gian của hàm mất mát tương phản; $\tau$ là tham số điều chỉnh; $sim(u, v) = u^\top v / (\|u\|\|v\|)$ đại diện cho tích vô hướng giữa hai véc-tơ $u$ và $v$ đã được chuẩn hóa $l_2$; $N(j)$ đại diện cho tập khác danh tính; $P(i)$ đại diện cho tập cùng danh tính.

## 2.2.4   Hàm mất mát tổng thể

Kết hợp SCL từ Công thức 2.6 với các hàm mất mát trong Công thức 2.5, hàm mục tiêu tối ưu của phương pháp đề xuất được biểu diễn trong Công thức 2.7.

$$L_{overall} = L_{bl} + \lambda_{sup} L_{sup} \qquad (2.7)$$

trong đó, $\lambda_{sup}$ là tham số điều chỉnh ảnh hưởng của SCL. Giá trị này được phân tích cụ thể trong Mục 2.3.2.1.

Bên cạnh đó, quy trình huấn luyện chi tiết của phương pháp SCM–ReID được mô tả trong Thuật toán 2.1. Cụ thể, đầu vào bao gồm mô hình $F(\cdot, \theta)$ được khởi tạo từ trọng số mô hình tiền huấn luyện ImageNet, tập dữ liệu huấn luyện $\{x_i, y_i\}_{i=1}^{N}$, các siêu tham số $\lambda_{tri}, \lambda_{ct}, \lambda_{ctl}, \lambda_{sup}$, cùng số vòng lặp huấn luyện max_epoch. Trước hết, dữ liệu được phân chia thành các lô với kích thước $B$. Mỗi ảnh $x_i$ trong lô sau đó được tăng cường độc lập hai lần nhằm thu được hai phiên bản khác nhau của cùng một danh tính, ký hiệu $x_{i1}$ và $x_{i2}$. Tiếp theo, mạng cốt lõi $F(\cdot, \theta)$ trích xuất các đặc trưng $f_{i1}$ và $f_{i2}$ tương ứng, đồng thời các đặc trưng này được ánh xạ sang không gian biểu diễn của hàm mất mát tương phản có giám sát thông qua hàm $g(\cdot)$ để thu được các véc-tơ $z_{i1}$ và $z_{i2}$. Trên cơ sở các đặc trưng thu được, các hàm mất mát được tính toán theo từng công thức tương ứng, sau đó được tổng hợp lại trong hàm mất mát tổng thể $L_{overall}$ theo Công thức 2.7 với các giá trị siêu tham số tương ứng. Cuối cùng, lan truyền ngược được sử dụng để cập nhật tham số $\theta$ của mô hình. Toàn bộ quy trình được lặp lại qua các vòng huấn luyện nhằm thu được mô hình $F(\cdot, \theta)$ với trọng số tối ưu.

---

**Thuật toán 2.1** SCM-ReID

---

**Đầu vào:** Mô hình $F(\cdot, \theta)$ với trọng số của mô hình tiền huấn luyện ImageNet, dữ liệu huấn luyện $\{x_i, y_i\}_{i=1}^N$ ; tham số: $\lambda_{tri}$, $\lambda_{ct}$, $\lambda_{ctl}$ và $\lambda_{sup}$; max_epoch.

1: **for each** epoch $\in [1,$ max_epoch$]$ **do**
2:  Chia lô $\{(x_i, y_i)\}_{i=1}^B$
3:  **for each** *mini-batch* **do**
4:    Tăng cường dữ liệu: $x_{i1} = Aug(x_i)$, $x_{i2} = Aug(x_i)$
5:    Tính $f_{i1} = F(x_{i1}, \theta)$, $f_{i2} = F(x_{i2}, \theta)$;
6:    Tính $z_{i1} = g(f_{i1})$, $z_{i2} = g(f_{i2})$;
7:    Tính L$_{ce}$ sử dụng Công thức 2.1;
8:    Tính L$_{tri}$ sử dụng Công thức 2.3;
9:    Tính L$_{ct}$ sử dụng Công thức 2.2;
10:    Tính L$_{ctl}$ sử dụng Công thức 2.4;
11:    Tính L$_{bl}$ sử dụng Công thức 2.5;
12:    Tính L$_{sup}$ sử dụng Công thức 2.6;
13:    Tính hàm mất mát tổng thể L$_{overall}$ sử dụng Công thức 2.7;
14:    Lan truyền ngược và cập nhật trọng số $\theta$ trong mô hình $F$;
15:   **end for**
16: **end for**

**Đầu ra:** Mô hình $F(\cdot, \theta)$ với trọng số tối ưu $\theta$.

---

## 2.3   Thực nghiệm và đánh giá kết quả

Để đánh giá hiệu quả của phương pháp đề xuất SCM-ReID, các thực nghiệm toàn diện được thực hiện trên các bộ dữ liệu Market-1501, CUHK03(L) và CUHK(D) với một máy tính trang bị một GPU NVIDIA A100 80GB VRAM nhằm trả lời các câu hỏi nghiên cứu.

- RQ1: Việc tích hợp SCL có giúp cải thiện hiệu năng tái định danh người hay không? Giá trị tối ưu của tham số $\lambda_{sup}$ là bao nhiêu để đạt được hiệu năng tổng thể tốt nhất cho SCM-ReID?

- RQ2: Kích thước ảnh đầu vào ảnh hưởng như thế nào đến hiệu năng của SCM-ReID?

- RQ3: Kích thước lô ảnh hưởng ra sao đến hiệu quả hoạt động của SCM-ReID?

- RQ4: Hiệu suất tính toán của SCM-ReID so với phương pháp cơ sở như thế nào?

- RQ5: Trong những trường hợp nào SCM-ReID không nhận diện đúng danh tính?

- RQ6: Phương pháp SCM-ReID có hiệu năng vượt trội hơn các phương pháp tiên tiến trong bài toán tái định danh người theo hướng tiếp cận học có giám sát không?

Phần tiếp theo sẽ trình bày chi tiết về quá trình huấn luyện mô hình, kết quả thực nghiệm, cũng như phân tích để trả lời các câu hỏi nghiên cứu đã được đặt ra.

## 2.3.1 Chi tiết cài đặt

**Huấn luyện**: Mô hình được triển khai huấn luyện với các tham số dựa trên thiết lập của các nghiên cứu cơ sở [54, 116]. Cụ thể, mạng cốt lõi sử dụng ResNet-50 với trọng số khởi tạo từ mô hình tiền huấn luyện trên ImageNet. Trong giai đoạn huấn luyện, mỗi lô bao gồm 16 định danh, mỗi định danh có 4 hình ảnh, tạo ra kích thước lô là 64. Hơn nữa, mỗi ảnh trước khi huấn luyện được tăng cường theo các kỹ thuật trong Mục 1.2.5. Hai bộ tối ưu hóa, Adam và SGD được sử dụng để tối ưu các tham số của mô hình. Bộ tối ưu hóa Adam được sử dụng để huấn luyện mô hình, với tốc độ học là $3.5 \times 10^{-4}$, giảm 10 lần ở epoch thứ 40 và 70. Giống như trong các nghiên cứu [68, 116], bộ tối ưu hóa SGD được sử dụng cho hàm mất mát trung tâm với tốc độ học là 0.5. Mô hình được huấn luyện ba lần, mỗi lần huấn luyện trong 120 epoch. Đóng góp này tập trung vào tác động của SCL, vì vậy luận án chỉ điều chỉnh tham số $\lambda_{sup}$, trong khi các tham số còn lại được kế thừa từ nghiên cứu trước [116]: $\lambda_{tri}$ là 1.0, $\lambda_{ct}$ là 0.0005, và $\lambda_{ctl}$ là 1.0.

## 2.3.2 Phân tích và đánh giá kết quả

### 2.3.2.1 Ảnh hưởng của tham số $\lambda_{sup}$

Đối với câu hỏi nghiên cứu RQ1, chúng tôi phân tích ảnh hưởng của trọng số $\lambda_{sup}$ trong Công thức 2.7 bằng cách điều chỉnh giá trị từ 0.0 đến 1.0. Kết quả thực nghiệm được trình bày trong Hình 2.2. Đối với bộ dữ liệu Market, kết quả mAP đạt được 98.30% tại $\lambda_{sup}$ là 0.0 (không sử dụng SCL) và đạt giá trị cao nhất là 98.84% khi $\lambda_{sup}$ là 0.2. Sau đó, hiệu năng của mô hình giảm dần khi giá trị của $\lambda_{sup}$ giảm. Đối với các bộ dữ liệu CUHK(L) và CUHK(D), cũng có kết quả theo xu hướng tương

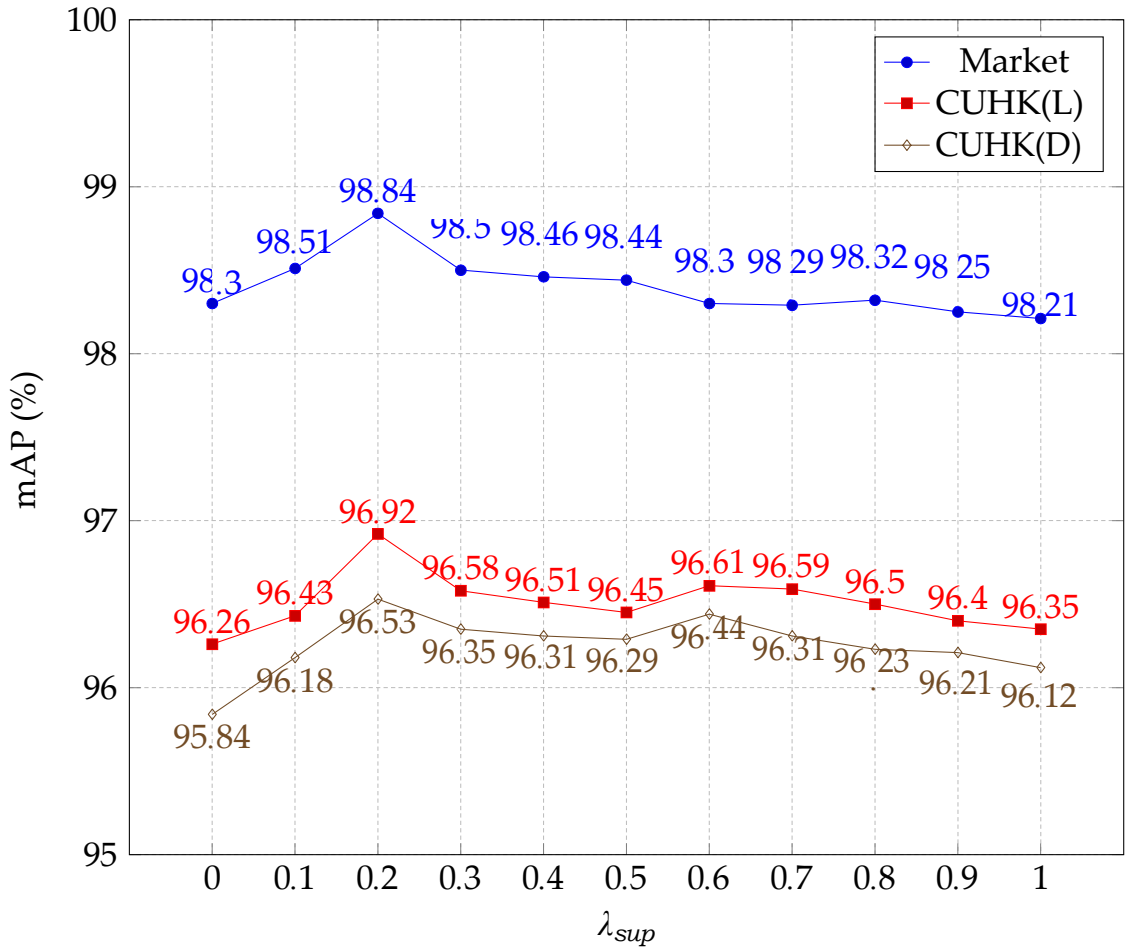

Hình 2.2: Ảnh hưởng của tham số $\lambda_{sup}$.

tự, với hiệu năng tăng từ 96.26% và 95.84% khi $\lambda_{sup}$ là 0.0 lên đến giá trị cao nhất là 96.92% và 96.53%, tương ứng, tại $\lambda_{sup}$ là 0.2. Sau đó, kết quả giảm dần, theo xu hướng giống như trong bộ dữ liệu Market. Vì vậy, SCL thực sự cải thiện hiệu năng tái định danh người, và giá trị tối ưu của $\lambda_{sup}$ cho hiệu năng tổng thể của SCM-ReID là 0.2 cho cả ba bộ dữ liệu.

### 2.3.2.2 Ảnh hưởng của kích thước hình ảnh

Chúng tôi tiến hành việc đánh giá ảnh hưởng của kích thước hình ảnh đến hiệu năng của mô hình để trả lời câu hỏi RQ2. Trong bài toán tái định danh người, việc lựa chọn kích thước ảnh phù hợp là rất quan trọng nhằm đảm bảo đủ thông tin chi tiết về đối tượng, đồng thời duy trì độ phức tạp tính toán ở mức hợp lý cho các mô hình học sâu. Bốn kích thước ảnh phổ biến được thực nghiệm bao gồm: 256×128, 384×128, 384×192 và 224×224 pixel. Các lựa chọn này giúp cho cho sự cân bằng giữa độ phân giải và hiệu năng xử lý của mạng CNN.

Bảng 2.1: Hiệu năng SCM-ReID với kích thước đầu vào hình ảnh khác nhau

| Kích thước ảnh | Market | | | | CUHK03(L) | | | | CUHK03(D) | | | |
|---|---|---|---|---|---|---|---|---|---|---|---|---|
| (pixels x pixels) | mAP | R1 | R5 | R10 | mAP | R1 | R5 | R10 | mAP | R1 | R5 | R10 |
| 256x128 | **98.84** | **98.66** | **99.14** | 99.5 | **96.92** | **95.57** | 98.57 | 99.14 | **96.53** | **95.14** | 98.35 | 99 |
| 224x224 | 98.42 | 98.07 | 98.46 | 99.52 | 96.46 | 94.99 | 98.5 | 99 | 95.89 | 94.28 | 98.21 | 98.78 |
| 384x128 | 98.34 | 97.75 | 98.81 | 99.55 | 96.43 | 94.92 | 98.43 | **99.28** | 96.27 | 94.64 | 98.28 | **99.11** |
| 384x192 | 98.75 | 98.49 | **99.14** | **99.7** | 96.68 | 95.06 | **98.64** | 99.14 | 96.45 | 94.92 | **98.71** | 98.93 |

Như thể hiện trong Bảng 2.1, kích thước ảnh 256×128 đạt hiệu năng cao nhất. Kích thước này duy trì tỷ lệ khung hình khoảng 2:1, phù hợp với dáng người trong ảnh giám sát, giúp bảo toàn tỷ lệ hình thể và hỗ trợ trích xuất hiệu quả các đặc trưng như hoa văn trang phục và dáng vóc.

Mặc dù kích thước 384×192 cũng giữ tỷ lệ 2:1 nhưng ở mức phân giải cao, giúp mô hình có thể tiếp cận được với nhiều chi tiết hình ảnh hơn, song kết quả thực nghiệm cho thấy nó không mang lại cải thiện đáng kể về hiệu năng. Lý do có thể là do độ phân giải cao làm tăng độ phức tạp tính toán, dễ gây hiện tượng quá khớp trên tập huấn luyện, đặc biệt trong các tập dữ liệu có kích thước trung bình như Market hoặc CUHK.

Trong khi đó, kích thước ảnh 224×224 sử dụng tỷ lệ khung hình gần vuông (1:1) sẽ không phản ánh đúng cấu trúc hình thể người trong các ảnh giám sát. Do đó, mô hình học sâu có thể gặp khó khăn trong việc học được các đặc trưng hình học quan trọng. Hệ quả là hiệu năng của mô hình ở kích thước này thấp hơn rõ rệt so với các kích thước giữ tỷ lệ 2:1.

### 2.3.2.3    Ảnh hưởng của kích thước lô

Để trả lời câu hỏi RQ3, chúng tôi đánh giá ảnh hưởng của kích thước lô đối với hiệu năng trong SCM-ReID. Cụ thể, chúng tôi thay đổi kích thước lô $B = P \times K$ hình ảnh, trong đó $P$ đại diện cho số lượng danh tính khác nhau và $K$ là số lượng hình ảnh khác nhau của mỗi danh tính. Chúng tôi thay đổi giá trị $B$ từ 32 đến 256 (32 là kích thước lô nhỏ nhất và 256 là kích thước lô lớn nhất để có thể chạy được với GPU có 80GB VRAM). Kết quả thực nghiệm được trình bày trong Bảng 2.2.

Đối với cả ba bộ dữ liệu, các kích thước lô nhỏ hơn, chẳng hạn như 4x8, 8x8 và 16x4, có xu hướng mang lại kết quả tốt nhất trên các chỉ số mAP, R1, R5 và R10. Trong số này, kích thước lô 16x4 nổi bật là có hiệu năng ổn định cao nhất trên tất cả các bộ dữ liệu, với giá trị mAP đạt 98.84% trên Market, 96.92% trên CUHK03(L), và 96.53% trên CUHK03(D). Ngược lại, các kích thước lô lớn hơn, đặc biệt là 64x4,

Bảng 2.2: Hiệu năng SCM-ReID với kích thước lô khác nhau

| Kích thước lô | Market | | | | CUHK03(L) | | | | CUHK03(D) | | | |
|---|---|---|---|---|---|---|---|---|---|---|---|---|
| (P x K) | mAP | R1 | R5 | R10 | mAP | R1 | R5 | R10 | mAP | R1 | R5 | R10 |
| 4x8 | 98.62 | 98.37 | 99.08 | 99.64 | 95.22 | 93.56 | 97.21 | 98.50 | 94.90 | 92.78 | 97.53 | 98.71 |
| 4x16 | 98.19 | 97.86 | 98.69 | 96.44 | 92.45 | 90.06 | 95.49 | 97.00 | 91.93 | 89.20 | 95.14 | 96.57 |
| 4x32 | 98.48 | 98.22 | 98.99 | 99.55 | 90.93 | 88.48 | 93.92 | 95.21 | 91.73 | 89.48 | 94.28 | 95.49 |
| 4x64 | 98.44 | 98.25 | 98.99 | 99.58 | 90.81 | 88.41 | 93.78 | 95.21 | 90.65 | 88.34 | 93.78 | 94.92 |
| 8x4 | 98.51 | 98.05 | 99.02 | <u>99.67</u> | <u>96.53</u> | <u>95.14</u> | <u>98.21</u> | **99.14** | 96.21 | 94.64 | <u>98.35</u> | <u>99.07</u> |
| 8x8 | <u>98.76</u> | <u>98.49</u> | **99.20** | **99.73** | 96.22 | 94.78 | 98.14 | 98.71 | <u>96.26</u> | <u>94.92</u> | **98.36** | 98.86 |
| 8x16 | 98.47 | 98.28 | 98.87 | 99.61 | 93.12 | 90.70 | 95.85 | 97.50 | 93.03 | 90.49 | 96.35 | 97.21 |
| 8x32 | 98.49 | 98.25 | 98.99 | 99.44 | 92.29 | 89.84 | 95.28 | 96.35 | 90.27 | 94.99 | 96.85 |
| 16x4 | **98.84** | **98.66** | <u>99.14</u> | 99.50 | **96.92** | **95.57** | **98.57** | **99.14** | **96.53** | **95.14** | <u>98.35</u> | 99.00 |
| 16x8 | 98.45 | 98.16 | 98.90 | 99.44 | 96.20 | 94.42 | 98.21 | 98.86 | 95.47 | 94.71 | 98.05 | **99.14** |
| 16x16 | 98.27 | 97.92 | 98.90 | 99.50 | 94.37 | 92.35 | 96.85 | 97.85 | 94.07 | 92.13 | 96.49 | 97.35 |
| 32x4 | 98.15 | 97.56 | 98.75 | 99.61 | 96.38 | 94.92 | 98.20 | <u>99.07</u> | 95.86 | 94.28 | 98.07 | 99.00 |
| 32x8 | 98.14 | 97.80 | 98.57 | 99.40 | 96.23 | 94.56 | 98.18 | 98.93 | 95.47 | 93.49 | 97.78 | 99.00 |
| 64x4 | 97.81 | 97.39 | 98.49 | 99.32 | 95.52 | 93.56 | 98.14 | 98.86 | 95.29 | 93.42 | 97.64 | 98.64 |

thể hiện sự suy giảm rõ rệt về hiệu năng với giá trị mAP thấp nhất là 97.81% trên Market, 95.52% trên CUHK03(L), và 95.29% trên CUHK03(D). Từ đó có thể thấy rằng mô hình có thể gặp khó khăn trong việc tổng quát hiệu quả với các lô lớn, cụ thể là do số lượng danh tính trong mỗi lô tăng lên, gây ra sự biến thiên và nhiễu lớn hơn.

Cụ thể, đối với bộ dữ liệu Market, khi $K$ tăng (ví dụ: 4×8, 4×16, 4×32, 4×64), mô hình khi huấn luyện sẽ dùng nhiều hình ảnh hơn cho mỗi danh tính trong quá trình huấn luyện, nhưng mAP và độ chính xác Rank-n thay đổi rất ít. Tuy nhiên, khi $P$ tăng (ví dụ: 8x4, 16x4, 32x4, 64x4), có nghĩa là nhiều danh tính hơn được bao gồm trong mỗi lô, và số lượng hình ảnh cho mỗi danh tính không thay đổi, hiệu năng của mô hình đạt đỉnh tại $P$ bằng 16 và sau đó giảm dần khi $P$ đạt 64. Đối với cả hai bộ dữ liệu CUHK, một xu hướng tương tự được quan sát khi giữ $K$ cố định và thay đổi $P$. Tuy nhiên, khi $K$ tăng, cả mAP và độ chính xác Rank-n đều giảm dần. Kích thước $K$ lớn cho phép mô hình học được nhiều biến thể hơn cho mỗi định danh, nhưng nó cũng tạo ra thêm nhiễu cho mô hình, cuối cùng làm suy giảm hiệu năng trên bộ dữ liệu CUHK.

### 2.3.2.4 Hiệu suất tính toán

Để phân tích thêm tác động thực tiễn của phương pháp đề xuất và trả lời câu hỏi RQ4, chúng tôi đánh giá hiệu suất tính toán của SCM-ReID so với mô hình cơ sở trên bộ dữ liệu Market với kích thước lô là 16x4 như một ví dụ đại diện. Trong

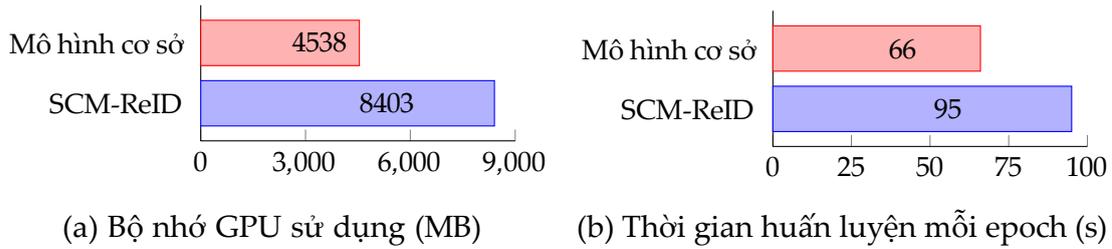

(a) Bộ nhớ GPU sử dụng (MB)     (b) Thời gian huấn luyện mỗi epoch (s)

Hình 2.3: Hiệu suất tính toán của Mô hình cơ sở và SCM-ReID trên bộ dữ liệu Market với kích thước lô là 16x4.

khi việc thêm SCL cải thiện đáng kể khả năng phân biệt đặc trưng và hiệu năng tổng thể của mô hình tái định danh người, nó cũng gây ra thêm chi phí tính toán trong quá trình huấn luyện. Chúng tôi sử dụng công cụ phân tích tài nguyên của PyTorch để đo lường mức tiêu thụ bộ nhớ GPU tối đa và thời gian huấn luyện mỗi epoch. Như được minh họa trong Hình 2.3, SCM-ReID tiêu thụ tối đa 8403 MB bộ nhớ GPU, trong khi mô hình cơ sở chỉ tiêu thụ 4538 MB. Sự tăng này chủ yếu do các so sánh theo cặp và sự giám sát mở rộng do SCL mang lại, làm tăng mức sử dụng bộ nhớ trong quá trình huấn luyện. Về tốc độ huấn luyện, SCM-ReID tốn 95 giây cho mỗi epoch, trong khi mô hình cơ sở chỉ mất 66 giây, tăng 44%. Chi phí thêm này đến từ việc thêm SCL và làm cho quá trình tối ưu hóa trở nên phức tạp hơn. Đáng chú ý, SCL không thêm bất kỳ tham số huấn luyện mới nào, làm cho chi phí suy luận tương đương với các mô hình cơ sở. Mặc dù có thêm chi phí huấn luyện, nhưng những cải tiến trong mAP và Rank-n cho thấy hiệu quả tương xứng với chi phí huấn luyện bổ sung. Thông qua việc áp dụng SCL, mô hình SCM-ReID học được một không gian đặc trưng có tính phân biệt cao và ổn định hơn, từ đó nâng cao khả năng tổng quát hóa trên các tập dữ liệu tái định danh người.

### 2.3.2.5    Trực quan hóa

Để đánh giá thêm hiệu quả của mô hình SCM-ReID, chúng tôi trình bày kết quả trực quan trên bộ dữ liệu Market-1501 như một trường hợp đại diện. Như minh họa trong Hình 2.4, mười kết quả truy xuất đầu tiên được hiển thị cho một số ảnh truy vấn tiêu biểu, sử dụng cả mô hình cơ sở và phương pháp đề xuất SCM-ReID. Các ảnh đúng được đánh dấu bằng khung viền màu xanh, trong khi các ảnh sai được đánh dấu bằng khung viền màu đỏ.

So sánh trực quan cho thấy SCM-ReID truy xuất kết quả chính xác và ổn định hơn, đặc biệt trong các tình huống khó như che khuất, thay đổi góc nhìn, và sự biến đổi ngoại hình. Không giống như mô hình cơ sở — vốn dễ nhầm lẫn giữa các cá nhân có trang phục hoặc tư thế tương tự — SCM-ReID thể hiện khả năng trích xuất đặc

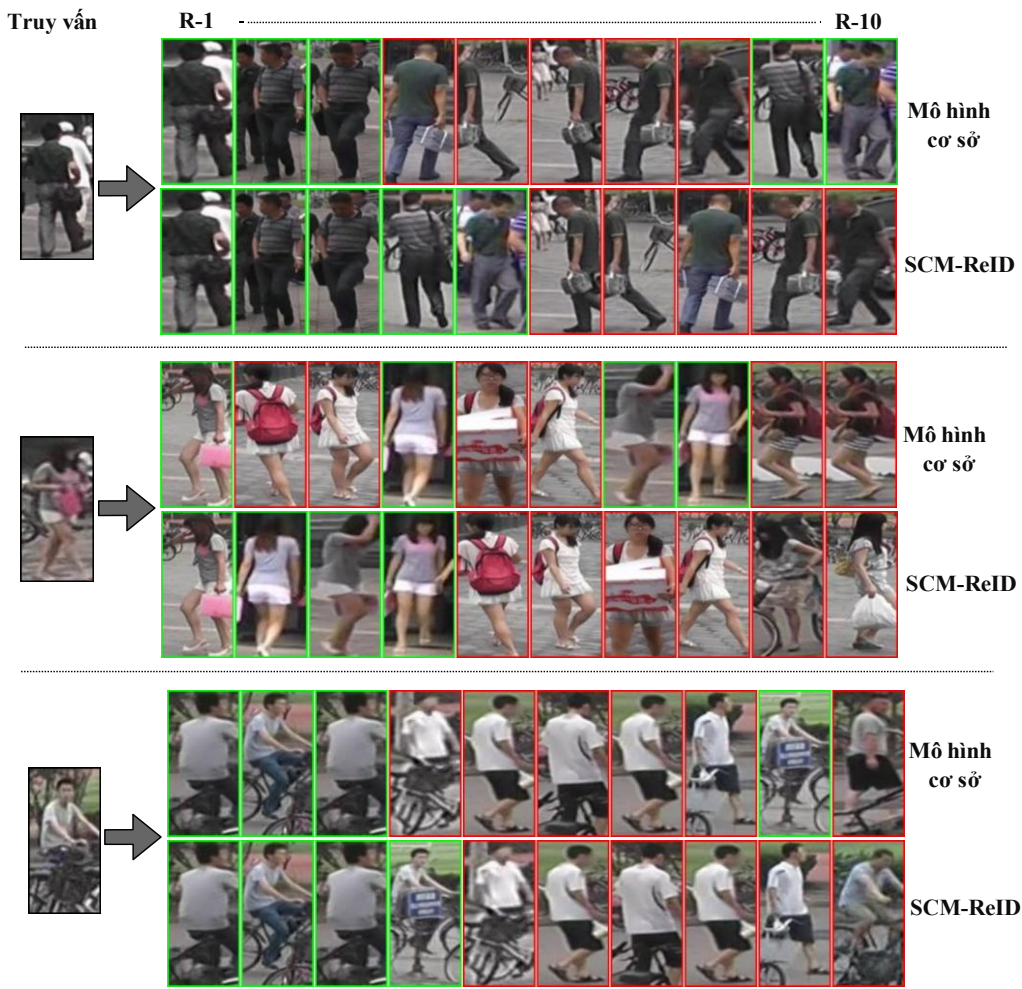

Hình 2.4: Kết quả truy vấn của mô hình cơ sở và SCM-ReID trên bộ dữ liệu Market. Kết quả đúng được đánh dấu bằng viền màu xanh, kết quả sai được đánh dấu bằng viền màu đỏ.

trưng phân biệt tốt hơn, cũng như duy trì tính nhất quán cao trong cùng một danh tính. Điều này được thúc đẩy bởi cơ chế học tương phản có giám sát cùng thiết kế hàm mất mát tích hợp hiệu quả.

Những kết quả này chứng minh khả năng tổng quát vượt trội của SCM-ReID trong các điều kiện phức tạp. Tuy nhiên, như được phân tích trong Mục 2.3.2.6, một số trường hợp mà mô hình không thể định danh đúng trong Rank-10, điều này cho thấy tiềm năng cải tiến và mở ra hướng nghiên cứu tiếp theo trong tương lai.

### 2.3.2.6   Trường hợp khó định danh

Mặc dù SCM-ReID đã đạt được hiệu năng cao nhờ tận dụng các ưu điểm của SCL và các hàm mất mát khác, phương pháp này vẫn chưa thể hoàn toàn vượt qua

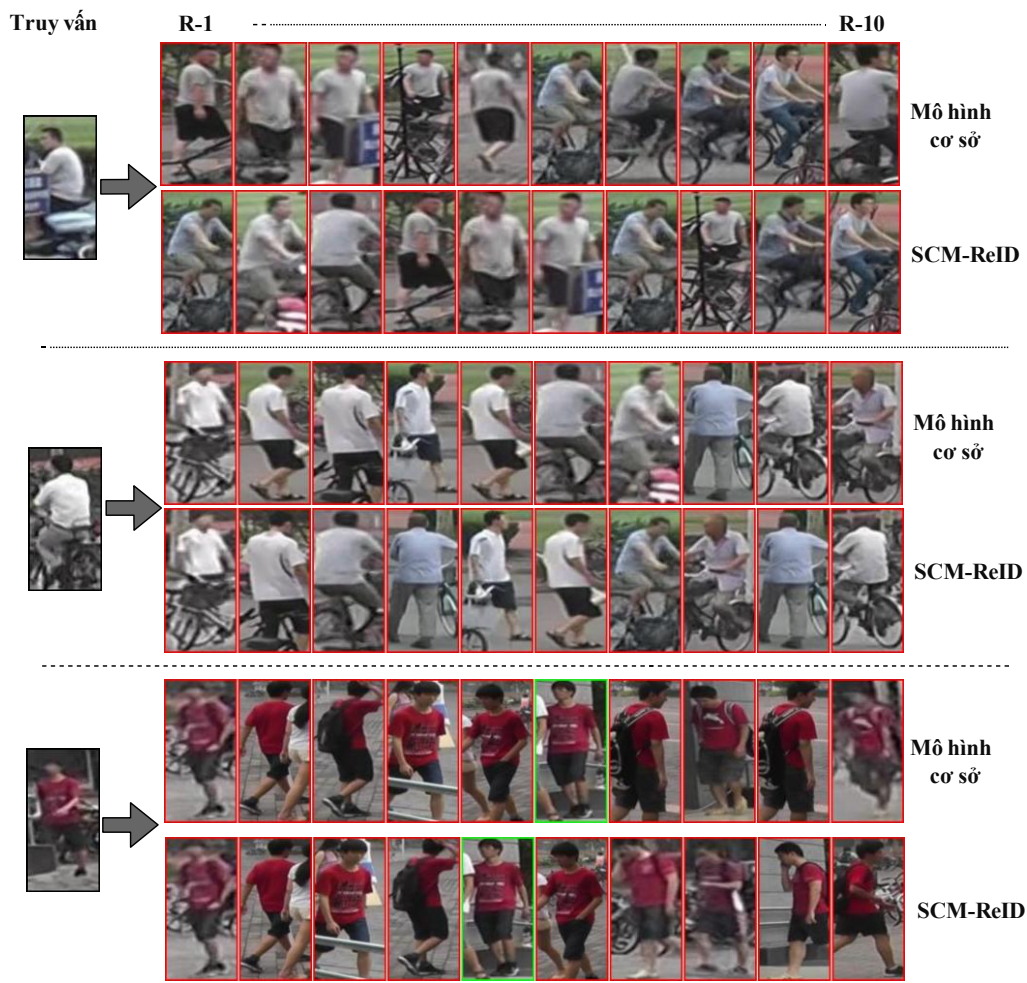

Hình 2.5: Trường hợp khó định danh trên bộ dữ liệu Market. Kết quả đúng đánh dấu bằng viền màu xanh, kết quả sai đánh dấu bằng viền màu đỏ.

những giới hạn hiện tại của bài toán tái định danh người, cả những vấn đề vốn có của bài toán và những vấn đề quan sát được trong mô hình cơ sở. Như được chỉ ra trong Hình 2.5, vẫn có một số trường hợp mô hình không thể định danh đúng người trong Rank-10, điều này trả lời cho câu hỏi RQ5.

Những hạn chế này chủ yếu được quan sát trong các tình huống thách thức cao, bao gồm che khuất lớn, thay đổi góc nhìn, và sự tương đồng cao về ngoại hình giữa các danh tính khác nhau (phong cách hoặc màu sắc trang phục tương tự). Cụ thể, khi phần lớn cơ thể của người bị che khuất hoặc khi hai cá nhân mặc trang phục gần như giống hệt nhau, ngay cả SCM-ReID cũng gặp khó khăn trong việc phân biệt chúng, thường xuyên đưa ra các kết quả sai tương tự như mô hình cơ sở.

Một yếu tố góp phần đáng kể khác đến từ chất lượng hình ảnh trong các bộ dữ liệu chuẩn. Nhiều mẫu ảnh có độ phân giải thấp, điều kiện ánh sáng không đảm bảo,

Bảng 2.3: So sánh hiệu năng của SCM-ReID với các phương pháp tiên tiến

| Phương pháp | Market | | | | CUHK03(L) | | | | CUHK03(D) | | | |
|---|---|---|---|---|---|---|---|---|---|---|---|---|
| | mAP | R1 | R5 | R10 | mAP | R1 | R5 | R10 | mAP | R1 | R5 | R10 |
| BoT (CVPR 19) | 94.2 | 95.4 | - | - | - | - | - | - | - | - | - | - |
| CTL (ICONIP 21) | 98.3 | 98 | 98.6 | <u>99.5</u> | - | - | - | - | - | - | - | - |
| MCTL (JoBD 23) | <u>98.63</u> | <u>98.4</u> | <u>99.1</u> | 99.6 | - | - | - | - | - | - | - | - |
| CTupletL (3PGCIC 23) | 98.57 | 98.3 | - | - | <u>94.38</u> | <u>92.3</u> | - | - | <u>93.39</u> | <u>91.10</u> | - | - |
| PRONET++ (arXiv 23) | 95.3 | 96.4 | - | - | 91.9 | 90.6 | - | - | 89.2 | 87.8 | - | - |
| CA-Jaccard (CVPR 24) | 94.5 | 96.2 | - | - | - | - | - | - | - | - | - | - |
| **SCM-ReID (Ours)** | **98.84** | **98.66** | **99.14** | <u>99.5</u> | **96.92** | **95.57** | **98.57** | **99.14** | **96.53** | **95.14** | **98.35** | **99.00** |

hoặc nền ảnh bị nhiễu, khiến các đặc trưng phân biệt bị che khuất và làm suy giảm khả năng học biểu diễn đặc trưng của mô hình. Những hạn chế này cho thấy rằng, mặc dù SCM-ReID đã mang lại những cải thiện đáng kể, vẫn cần tiếp tục phát triển các kỹ thuật nâng cao hơn để xử lý triệt để các trường hợp khó định danh. Hướng tiếp cận tiềm năng bao gồm tích hợp cơ chế chú ý theo bộ phận hoặc trích xuất đặc trưng ở mức độ tinh vi và ngữ cảnh sâu hơn.

## 2.3.3 So sánh với các phương pháp tiên tiến

Trong phần này, luận án tiến hành so sánh phương pháp đề xuất SCM-ReID với các phương pháp tiên tiến trong những năm gần đây. Như được trình bày trong Bảng 2.3, SCM-ReID đạt mAP là 98.84% và Rank-1 là 98.66% trên Market, mAP là 96.92% và Rank-1 là 95.57% trên CUHK(L), và 96.53% và 95.14% trên CUHK(D), đây đều là các giá trị cao nhất trên từng bộ dữ liệu. Điều này cho thấy tính hiệu quả và độ tin cậy của mô hình trong việc nhận dạng chính xác các danh tính qua các bộ dữ liệu khác nhau.

Khi so sánh với các phương pháp khác, SCM-ReID đạt mức cải thiện đáng kể trên bộ dữ liệu Market khoảng 4.64% về mAP và 3.26% về Rank-1 so với phương pháp nền tảng BoT [68] vốn chỉ dùng ba hàm mất mát là phân loại, bộ ba và trung tâm. Ngoài ra, so với các mô hình nâng cao hơn như CTL [116] sử dụng bốn hàm mất mát và MCTL [6] điều chỉnh hàm mất mát bộ ba trọng tâm, SCM-ReID vẫn cho thấy những cải thiện ổn định, với mức tăng 0.54% và 0.21% về mAP, cùng 0.66% và 0.26% về Rank-1 trên tập dữ liệu Market. Mặc dù các cải thiện này tương đối khiêm tốn, chúng cho thấy tính hiệu quả bổ sung của chiến lược tích hợp nhiều hàm mất mát trong SCM-ReID.

Bên cạnh đó, so với phương pháp dùng hàm mất mát đối sánh nhiều tâm cụm gần đây [11], SCM-ReID đạt được hiệu năng tốt hơn trên tất cả các tập dữ liệu. Cụ thể, mô hình ghi nhận mức cải thiện về mAP dao động từ 0.27% đến 3.14%, và về

Rank-1 từ 0.36% đến 4.04%, phản ánh sự tiến bộ rõ rệt trong khả năng nhận diện chính xác.

Tổng hợp các kết quả trên, có thể thấy rằng SCM-ReID duy trì sự hiệu quả và ổn định trên cả ba bộ dữ liệu. Việc kết hợp hàm mất mát tương phản có giám sát với bốn hàm mất mát khác giúp mô hình học được không gian đặc trưng phân tách tốt hơn, giảm hiện tượng lệch miền và tăng khả năng tổng quát hóa đối với các tập dữ liệu. Những phân tích này cung cấp câu trả lời cho câu hỏi nghiên cứu RQ6.

## 2.4    Kết luận chương

Chương này tập trung vào việc giải quyết thách thức hiện có để cải thiện hiệu năng mô hình tái định danh người theo hướng tiếp cận có giám sát. Bằng cách tận dụng mạng cốt lõi Resnet-50 sử dụng học tương phản có giám sát và học độ đo sâu, chúng tôi đề xuất phương pháp SCM-ReID, kết hợp hàm mất mát tương phản có giám sát với bốn hàm mất mát khác: phân loại, trung tâm, bộ ba, bộ ba trọng tâm để đạt được kết quả vượt trội. Các thực nghiệm thực hiện trên các bộ dữ liệu tiêu chuẩn như Market, CUHK03(L) và CUHK03(D) cho thấy SCM-ReID vượt trội so với các phương pháp tiên tiến trong học có giám sát cho tái định danh người. Những kết quả này xác nhận rằng các mục tiêu nghiên cứu, thiết kế một phương pháp huấn luyện thống nhất nhằm nâng cao hiệu năng tái định danh người, đã đạt được. Hơn nữa, những kết quả này nhấn mạnh triển vọng của SCM-ReID trong học có giám sát, giúp nâng cao độ chính xác trong các ứng dụng giám sát và nhiệm vụ tổng quát hóa miền.

Mặc dù SCM-ReID đạt được kết quả thực nghiệm tốt, phương pháp vẫn tồn tại một số hạn chế cần được khắc phục. Việc tích hợp hàm mất mát tương phản có giám sát làm tăng độ nhạy cảm với kích thước của lô, đồng thời đòi hỏi cao hơn về bộ nhớ và chi phí tính toán so với mô hình cơ sở. Ngoài ra, SCM-ReID vẫn chưa thể giải quyết triệt để các tình huống đặc biệt khó như che khuất lớn, thay đổi góc nhìn, hoặc các trường hợp có mức độ tương đồng cao về ngoại hình giữa các danh tính. Một điểm hạn chế khác là sự phụ thuộc vào mạng cốt lõi ResNet-50 và cách kết hợp các hàm mất mát được xác định trước. Trong các nghiên cứu tiếp theo, chúng tôi sẽ mở rộng đánh giá trên các kiến trúc mạng cốt lõi và bộ dữ liệu phong phú hơn để kiểm chứng độ bền vững của mô hình. Đồng thời, chúng tôi sẽ tiếp tục nghiên cứu kết hợp học tương phản có giám sát với các kỹ thuật học độ đo sâu khác nhằm nâng cao khả năng kết hợp các hàm mất mát, hướng tới mục tiêu cải thiện hiệu năng tổng thể của bài toán tái định danh người.

# Chương 3

# NÂNG CAO HIỆU NĂNG TÁI ĐỊNH DANH NGƯỜI THÍCH ỨNG MIỀN KHÔNG GIÁM SÁT DỰA TRÊN TĂNG CƯỜNG DỮ LIỆU, KẾT HỢP ĐẶC TRƯNG VÀ TINH CHỈNH NHÃN GIẢ

Trong chương này, luận án trình bày các phương pháp đề xuất nhằm nâng cao hiệu năng tái định danh người theo hướng tiếp cận học thích ứng miền không giám sát. Cụ thể, đề xuất đầu tiên "IAQGA" tập trung vào việc thu hẹp khoảng cách phân phối giữa miền nguồn và miền đích ở mức hình ảnh, thông qua việc sử dụng GAN để tăng cường dữ liệu tạo ra các ảnh mang phong cách đặc trưng của miền đích. Đồng thời, phương pháp còn tích hợp thuật toán đánh giá chất lượng ảnh để giảm thiểu tác động tiêu cực từ các ảnh sinh ra có chất lượng thấp trong quá trình huấn luyện.

Trong đề xuất thứ hai mang tên "DAPRH", bên cạnh việc sử dụng GAN để giảm thiểu sự sai lệch phân phối ở mức hình ảnh, phương pháp còn tích hợp DIM nhằm thu hẹp sự khác biệt giữa các miền ở mức đặc trưng. Ngoài ra, phương pháp này còn tập trung vào việc học không giám sát ở miền đích thông qua việc kết hợp đặc trưng toàn thể, bao gồm cả đặc trưng toàn cục và đặc trưng cục bộ, cùng với kỹ thuật tinh chỉnh nhãn giả, nhằm cải thiện chất lượng biểu diễn và tăng cường khả năng phân biệt của mô hình.

Các kết quả nghiên cứu trong chương này được công bố tại công trình [CT1], [CT4] và [CT5].

## 3.1  Đặt vấn đề

Các kết quả trình bày ở chương trước đã cho thấy rằng mô hình tái định danh người có thể đạt hiệu năng vượt trội khi được huấn luyện và đánh giá trên cùng một bộ dữ liệu. Tuy nhiên, các nghiên cứu thực nghiệm đã chỉ ra rằng hiệu năng của mô hình thường suy giảm đáng kể khi được áp dụng trên các bộ dữ liệu khác nhau, chủ yếu do sự khác biệt lớn giữa các bộ dữ liệu [52]. Như minh họa trong Hình 3.1, các bộ dữ liệu thường được thu thập từ nhiều camera ở các vị trí khác nhau, dẫn đến sự khác biệt rõ rệt về dáng người, điều kiện chiếu sáng và phong cách camera. Do đó,

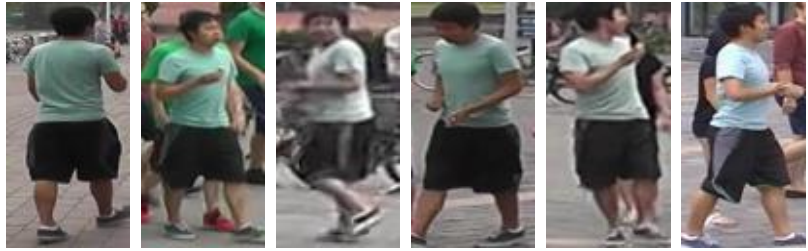

(a) Hình ảnh người từ các camera c1:c6 trong bộ dữ liệu Market.

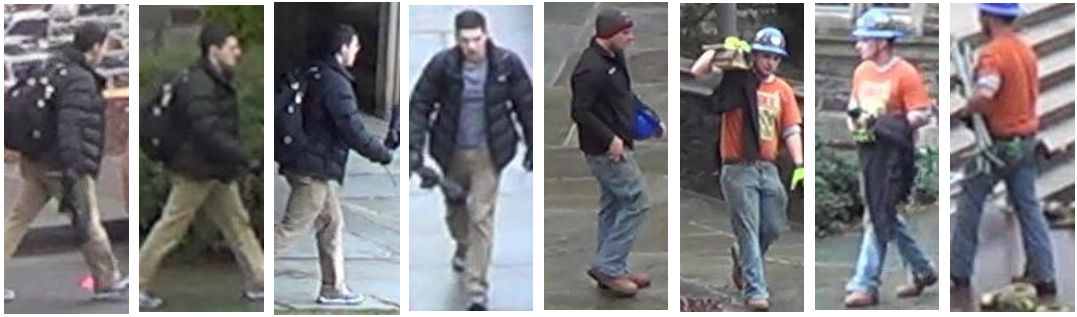

(b) Hình ảnh người từ các camera c1:c8 trong bộ dữ liệu Duke.

Hình 3.1: Các ảnh mẫu trong tập dữ liệu Market và Duke. Do sự khác biệt về ánh sáng, phong cách và bối cảnh giữa hai miền dữ liệu, việc duy trì hiệu năng của mô hình khi được huấn luyện trên một miền và đánh giá trên một miền khác đặt ra một thách thức đáng kể.

các mô hình tái định danh người được huấn luyện trên một miền cụ thể thường cho thấy hiệu năng suy giảm đáng kể khi được áp dụng sang một miền khác. Như kết quả thể hiện trong Bảng 3.1, khi mô hình được huấn luyện và đánh giá trên cùng một tập dữ liệu với chỉ hai hàm mất mát là phân loại và bộ ba như thể hiện trong Công thức 3.1.

$$L_{BL} = L_{ce} + L_{tr} \tag{3.1}$$

Đối với mô hình được huấn luyện trên bộ dữ liệu Market, kết quả đánh giá trên tập đánh giá của bộ dữ liệu này cho thấy hiệu năng rất cao, với mAP đạt 80.1% và Rank-1 đạt 92.8%. Tuy nhiên, khi sử dụng mô hình này để đánh giá trên tập đánh giá của bộ dữ liệu Duke thì hiệu năng suy giảm đáng kể, với mAP chỉ còn 25.8% và Rank-1 giảm xuống 43.7%. Điều này phản ánh rõ rệt tác động của sự khác biệt phân phối giữa các miền đến khả năng tổng quát hóa của mô hình.

Để khắc phục hiện tượng này, học thích ứng miền không giám sát (UDA) đã được đề xuất trong bài toán tái định danh người như một hướng tiếp cận tiềm năng. Trong thiết lập UDA, mô hình được huấn luyện có giám sát trên một miền nguồn – nơi có nhãn danh tính đầy đủ, và được đánh giá truy vấn trên một miền đích – nơi không có nhãn định danh. Mục tiêu đặt ra là làm sao để mô hình thích ứng hiệu

Bảng 3.1: Sự suy giảm hiệu năng khi đánh giá trên các tập dữ liệu khác nhau

| Tập | Tập đánh giá | | | |
|---|---|---|---|---|
| huấn | Market | | Duke | |
| luyện | mAP | R1 | mAP | R1 |
| Market | 80.1 | 92.8 | 25.8 | 43.7 |
| Duke | 26.2 | 55.3 | 70.2 | 84.1 |

quả với miền đích, để giảm sự chênh lệch về phân phối dữ liệu giữa hai miền.

Trong quá trình nghiên cứu, chúng tôi tiếp cận bài toán tái định danh người UDA theo hướng tách thành hai vấn đề cốt lõi: (i) cải thiện chất lượng và tính đại diện của dữ liệu miền nguồn cho học có giám sát để xây dựng mô hình tổng quát tốt hơn, và (ii) tận dụng triệt để dữ liệu miền đích thông qua học không giám sát để nâng cao hiệu năng tái định danh người.

Ở bước đầu tiên, chúng tôi phát triển phương pháp IQAGA (**I**mage **Q**uality **A**ssessment-Driven Learning with **G**AN-Based Dataset **A**ugmentation), sử dụng thích ứng miền mức hình ảnh để tạo ảnh chuyển đổi phong cách từ miền nguồn sang miền đích bằng StarGAN [21]. Tuy nhiên, một số lượng lớn ảnh sinh ra bởi StarGAN có chất lượng kém, thậm chí còn mất hẳn đặc trưng nhận diện danh tính. Nhằm giảm thiểu ảnh hưởng xấu của các ảnh GAN kém chất lượng này, chúng tôi đã thêm mô-đun đánh giá chất lượng ảnh thông qua chuẩn hóa đặc trưng [55] và từ đó điều chỉnh trọng số ảnh hưởng của từng ảnh trong quá trình huấn luyện. Nhờ đó, mô hình được huấn luyện với dữ liệu phong phú nhưng vẫn đảm bảo ổn định và chính xác. Kiến trúc tổng thể của IQAGA được biểu diễn trong Hình 3.2.

Dẫu vậy, IQAGA mới chỉ cải thiện hiệu quả ở giai đoạn huấn luyện có giám sát trên miền nguồn và chưa khai thác triệt để thông tin từ miền đích. Do đó, ở giai đoạn tiếp theo, chúng tôi phát triển phương pháp DAPRH (GAN-based **D**ata **A**ugmentation and **P**seudo-label **R**efinement with **H**olistic Features), nhằm khắc phục những hạn chế này. DAPRH tiếp tục sử dụng GAN để tăng cường dữ liệu cho tập dữ liệu huấn luyện, nhưng không áp dụng cơ chế đánh giá chất lượng ảnh như IQAGA. Thực tế cho thấy, việc sử dụng toàn bộ ảnh sinh bởi GAN trong IQAGA đưa vào mô hình huấn luyện nhiều ảnh nhiễu, làm suy giảm hiệu quả huấn luyện. Khác với IQAGA, DAPRH chỉ chọn lọc một tỷ lệ ảnh nhất định được sinh ra từ GAN để tăng cường dữ liệu, nhằm giảm thiểu tác động tiêu cực của ảnh xấu. Tỷ lệ này được điều chỉnh linh hoạt trong quá trình huấn luyện để tìm ra giá trị tối ưu, từ đó không chỉ giảm nhiễu mà còn tiết kiệm chi phí tính toán liên quan đến chuẩn hóa véc-tơ đặc trưng sử dụng trong phương pháp đánh giá chất lượng ảnh.

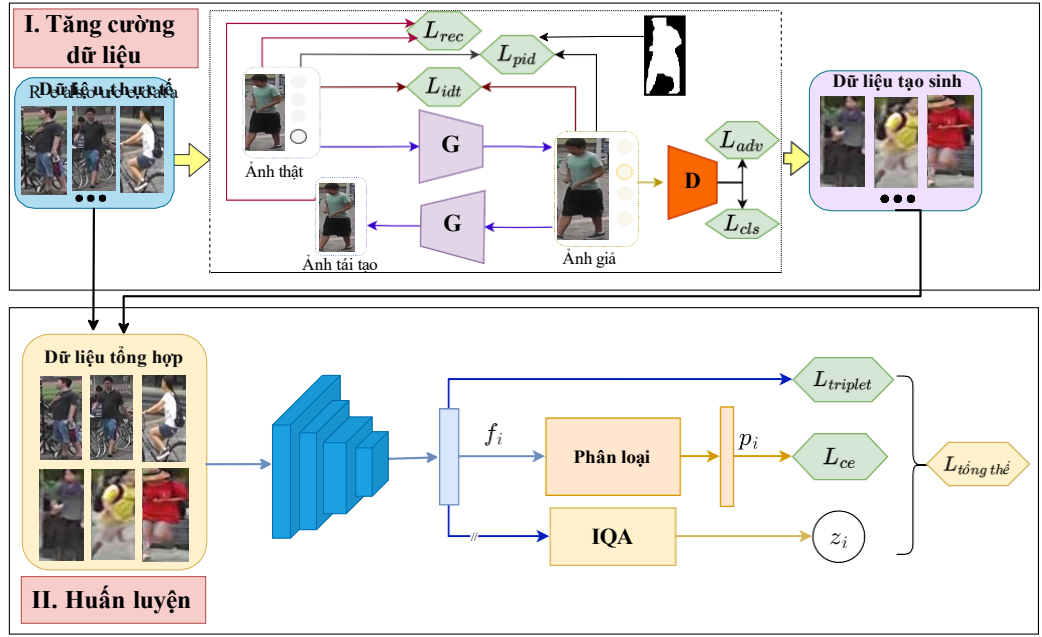

Hình 3.2: Kiến trúc phương pháp đề xuất - IQAGA.

Thêm vào đó, DAPRH còn tích hợp cơ chế ánh xạ bất biến miền (DIM) nhằm thu hẹp khoảng cách đặc trưng giữa hai miền ở mức đặc trưng ngay trong quá trình huấn luyện có giám sát. Quan trọng hơn, DAPRH tập trung vào giai đoạn học không giám sát trên miền đích thông qua tinh chỉnh nhãn giả dựa trên khoảng cách cụm, kết hợp biểu diễn toàn cục và cục bộ, và áp dụng kiến trúc Teacher-Student để duy trì tính ổn định trong huấn luyện. Phương pháp đề xuất DAPRH được thể hiện trong Hình 3.3 và có các đóng góp chính sau.

- Đề xuất sử dụng phương pháp GAN thích ứng miền ở mức hình ảnh dựa trên StarGAN [21] để sinh ảnh chuyển đổi phong cách theo miền đích nhằm tăng cường dữ liệu cho tập huấn luyện, đồng thời áp dụng kỹ thuật DIM [65] ở mức đặc trưng để giảm thiểu khoảng cách phân phối giữa miền nguồn và miền đích.

- Đề xuất khai thác đồng thời thông tin từ đặc trưng cục bộ và đặc trưng toàn cục như một đặc trưng tổng thể nhằm đảm bảo mô hình tiếp nhận đầy đủ thông tin cần thiết trong quá trình huấn luyện.

- Đề xuất cơ chế tinh chỉnh nhãn giả đơn giản bằng cách sử dụng nhãn mục tiêu mềm thay vì nhãn cứng truyền thống, qua đó cải thiện hiệu quả của quá trình học không giám sát dựa trên kiến trúc Teacher-Student.

- Tiến hành các thực nghiệm toàn diện trên ba tập dữ liệu phổ biến gồm Market, Duke và MSMT để đánh giá hiệu quả của phương pháp đề xuất.

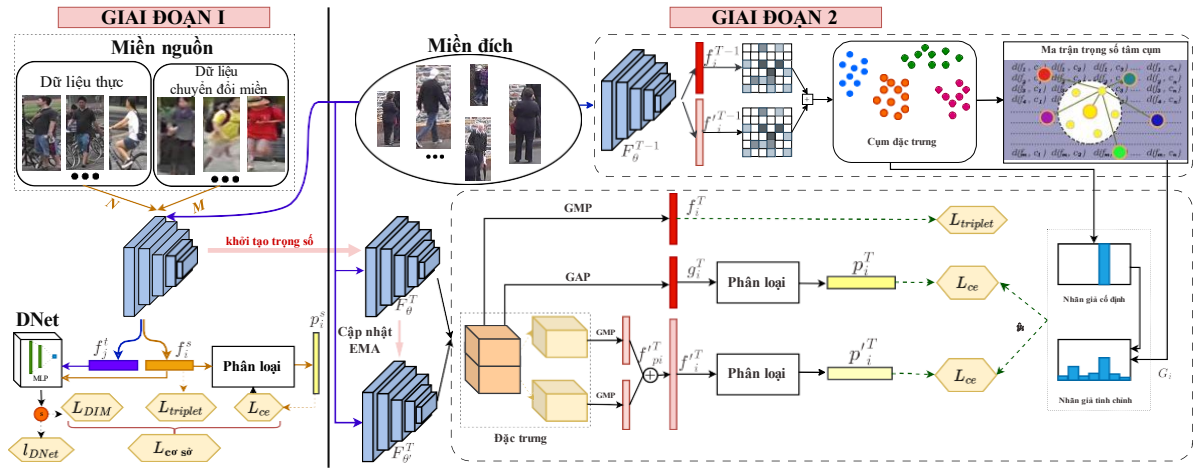

Hình 3.3: Kiến trúc phương pháp đề xuất - DAPRH.

Trong phần tiếp theo, luận án sẽ trình bày cụ thể các chi tiết, kỹ thuật được sử dụng trong IQAGA và DAPRH để nâng cao hiệu năng mô hình tái định danh người theo hướng tiếp cận học thích ứng miền không giám sát.

## 3.2 Phương pháp nâng cao học có giám sát trên miền nguồn

Đây là hướng tiếp cận đầu tiên nhằm cải thiện hiệu năng của mô hình trong giai đoạn học có giám sát trên miền nguồn, từ đó tạo nền tảng vững chắc để nâng cao khả năng tổng quát của mô hình khi đánh giá trên miền đích.

### 3.2.1 Thích ứng miền mức hình ảnh

Như đã đề cập trong Mục 3.1, một trong những yếu tố chính có thể ảnh hưởng đến hiệu năng của mô hình trên miền chưa thấy là sự khác biệt lớn về đặc trưng giữa các miền. Để khắc phục vấn đề này, chúng tôi đề xuất sử dụng StarGAN [21] để thực hiện chuyển đổi hình ảnh theo phong cách của miền đích. Cách tiếp cận này được xem như một hình thức thích ứng miền ở mức hình ảnh cho phép thu hẹp khoảng cách biểu diễn giữa hai miền ngay từ dữ liệu đầu vào. Nhờ đó, mô hình có thể học được các đặc trưng phù hợp hơn với miền đích, từ đó cải thiện đáng kể hiệu năng khi đánh giá trên dữ liệu trên miền đích.

Khác với các phương pháp GAN khác vốn chỉ có khả năng chuyển đổi hình ảnh giữa hai miền cụ thể [135, 133], StarGAN có ưu điểm là có thể học và chuyển đổi giữa nhiều miền cùng lúc. Trong nghiên cứu này, chúng tôi xem mỗi camera như một miền riêng biệt, do đó miền đích sẽ bao gồm tổng cộng $C_t$ miền, trong đó $C_t$ là số lượng camera trong miền đích. Mục tiêu là chuyển đổi mỗi ảnh $(x_i^s, y_i^s)$ từ $D_s$ sang $C_t$ phong cách camera khác nhau, đồng thời giữ nguyên thông tin nhãn định danh trong quá trình chuyển đổi. Quá trình này được biểu diễn theo Công thức 3.2.

$$G((x_i^s, y_i^s), c) \rightarrow (x_i^t, y_i^s) \tag{3.2}$$

Bên cạnh ba hàm mất mát cơ bản của StarGAN, bao gồm: hàm mất mát đối kháng - $L_{adv}$, hàm mất mát phân loại miền - $L_{cls}$, hàm mất mát tái cấu trúc - $L_{rec}$. Trong phương pháp IQAGA, chúng tôi bổ sung thêm hai hàm mất mát để duy trì nội dung liên quan đến danh tính người trong ảnh đầu vào, theo nghiên cứu của Yuanpeng Tu [102] với sự thay đổi là sử dụng YOLOv8 [51] để thực hiện phân vùng ảnh.

- Hàm mất mát ánh xạ danh tính - ($L_{idt}$): giúp bảo toàn thành phần màu sắc giữa miền nguồn và ảnh được chuyển đổi theo phong cách miền đích.

- Hàm mất mát bảo toàn danh tính người - ($L_{pid}$): được tính bằng cách đo sự thay đổi ở vùng tiền cảnh người trước và sau khi chuyển đổi.

Mục tiêu tổng thể để tối ưu hóa mạng sinh $G$ và mạng phân biệt $D$ được biểu diễn trong Công thức 3.3.

$$
\begin{aligned}
L_D &= -L_{adv} + \lambda_{cls} L_{cls}^r \\
L_G &= L_{adv} + \lambda_{cls} L_{cls}^f + \lambda_{rec} L_{rec} + \lambda_{idt} L_{idt} + \lambda_{pid} L_{pid}
\end{aligned}
\tag{3.3}
$$

trong đó, các hệ số $\lambda_{cls}, \lambda_{rec}, \lambda_{idt}$ và $\lambda_{pid}$ là các siêu tham số điều chỉnh mức độ ảnh hưởng tương đối của các thành phần mất mát phân loại miền, tái tạo, ánh xạ định danh và bảo toàn định danh so với mất mát đối kháng. Trong các thực nghiệm của chúng tôi, các hệ số này được thiết lập kế thừa từ nghiên cứu của Yuanpeng Tu [102] với các giá trị lần lượt là: $\lambda_{cls} = 1$, $\lambda_{rec} = 10$, $\lambda_{idt} = 1$, và $\lambda_{pid} = 10$.

Sau giai đoạn này, bộ dữ liệu tăng cường thu được được ký hiệu là $D_{syn}$. Với phương pháp IQAGA, toàn bộ ảnh sinh ra được kết hợp với tập dữ liệu thực $D_s$ để tạo thành bộ dữ liệu huấn luyện tổng hợp $D_{cb}$. Việc khai thác toàn bộ ảnh tăng cường giúp tăng tính đa dạng trong huấn luyện, từ đó sẽ nâng cao khả năng khái quát của mô hình. Tuy nhiên, khi xem xét ảnh sinh ra, chúng tôi nhận thấy có rất nhiều ảnh chất lượng kém (như minh họa trong Hình 3.4). Có những ảnh chất lượng kém tới mức không giữ được đặc trưng nhận diện người, gây ảnh hưởng tiêu cực đến hiệu quả huấn luyện. Vậy nên trong phương pháp IQAGA chúng tôi đề xuất thêm

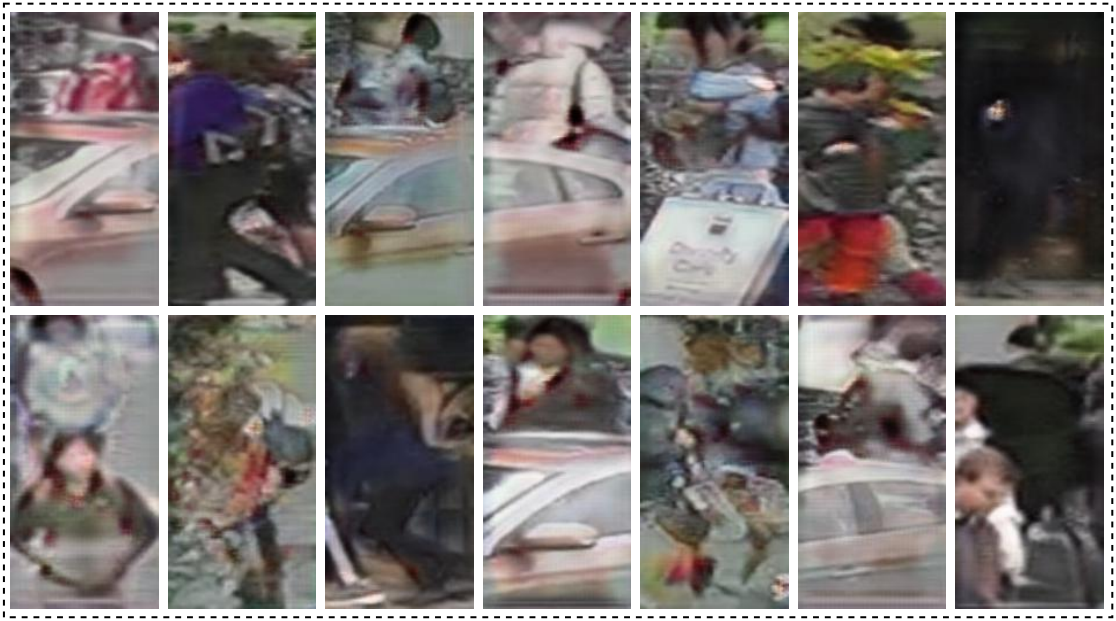

Hình 3.4: Ảnh chất lượng xấu sinh ra bởi GAN.

một kỹ thuật đánh giá chất lượng ảnh dựa trên chuẩn hóa đặc trưng (được trình bày tại Mục 3.2.3) để làm giảm sự tác động của các ảnh chất lượng xấu lên mô hình. Khác với IQAGA, phương pháp DAPRH chỉ sử dụng một phần dữ liệu tăng cường để tăng thêm dữ liệu cho quá trình học có giám sát trên miền nguồn. Nguyên nhân là do khi sử dụng tất cả các ảnh sinh ra từ GAN sẽ làm gia tăng tỉ lệ ảnh chất lượng xấu gây tiêu cực đến hiệu quả của mô hình. Ngay cả khi dùng phương pháp đánh giá chất lượng ảnh để giảm ảnh hưởng đến mô hình, một số lượng nhất định ảnh xấu vẫn gây ảnh hưởng không tốt đến mô hình. Do vậy, thay vì sử dụng tất cả các ảnh sinh ra, phương pháp chỉ lấy một phần theo một tỉ lệ nhất định là $N : M$ trong đó $N$ là tỉ lệ ảnh thực, $M$ là tỉ lệ ảnh tăng cường. Tỷ lệ này được xác định thông qua thực nghiệm để tìm ra giá trị tối ưu trong Mục 3.4.3.3. Cách tiếp cận này không chỉ giúp hạn chế ảnh hưởng tiêu cực từ ảnh chất lượng xấu mà còn giảm chi phí tính toán liên quan đến bước chuẩn hóa véc-tơ đặc trưng trong quá trình đánh giá chất lượng ảnh. Lưu ý rằng, sau bước này, trong phương pháp DAPRH, khi chúng tôi đề cập đến tập dữ liệu ảnh nguồn, đây chính là tập dữ liệu đã được kết hợp giữa ảnh thực và ảnh tăng cường theo tỉ lệ tối ưu. Tập dữ liệu này sẽ được sử dụng làm đầu vào cho các giai đoạn tiếp theo trong quá trình huấn luyện mô hình.

## 3.2.2 Thích ứng miền mức đặc trưng

Ngoài việc sử dụng StarGAN để tăng cường dữ liệu, chúng tôi còn sử dụng phương pháp ánh xạ bất biến miền DIM [65] để giảm thiểu sự khác biệt phân phối đặc trưng

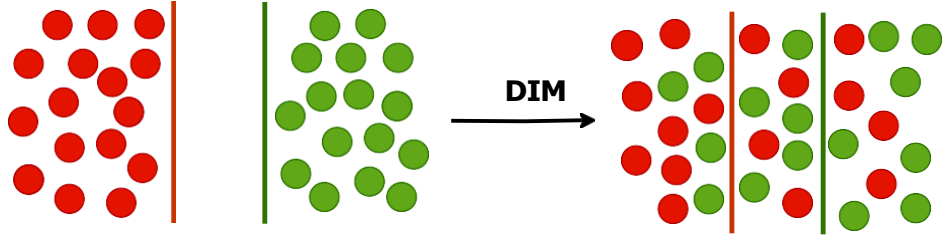

Hình 3.5: DIM: Giảm sự khác biệt phân phối ở mức đặc trưng.● thể hiện đặc trưng ở miền nguồn.● thể hiện đặc trưng ở miền đích.

giữa miền nguồn và miền đích ở mức đặc trưng. DIM bao gồm một bộ trích xuất đặc trưng và một bộ phân loại miền gọi là DNet. Bộ trích xuất đặc trưng học cách biểu diễn các đặc trưng bất biến miền, trong khi bộ phân loại miền đảm bảo sinh ra các đặc trưng phân biệt cho tái định danh người. Điều này được minh họa trong Hình 3.5.

DNet sử dụng cả thông tin từ miền nguồn và miền đích làm đầu vào, và tạo ra các điểm số nhận dạng miền trong phạm vi [0,1]. Giá trị mục tiêu của các điểm số nhận dạng miền là 1 đối với $f_i^s$ từ miền nguồn $D_s^*$ và 0 đối với $f_j^t$ từ miền đích $D_t$. Bằng cách sử dụng sai số bình phương trung bình, DNet sẽ được tối ưu hóa bởi hàm mất mát được biểu diễn trong Công thức 3.4.

$$l_{DNet} = \frac{1}{N_s}\sum_{i=0}^{N_s} DNet\left(f_i^s\right) - \left(\frac{1}{N_t}\sum_{j=0}^{N_t}(DNet(f_j^t))^2\right)$$  (3.4)

Sau đó, sử dụng hàm mất mát DIM để trích xuất các đặc trưng với điểm số nhận dạng miền 0.5, chúng tôi sử dụng DNet để giám sát các mô hình DCNN $f_\theta$ với hàm mất mát mục tiêu DIM được biểu diễn trong Công thức 3.5.

$$L_{DIM}(\Omega) = \frac{1}{N_s}\sum_{i=0}^{N_s}(DNet(f_i^s) - 0.5)^2 + \frac{1}{N_t}\sum_{j=0}^{N_t}(DNet(f_j^t) - 0.5)^2$$  (3.5)

Theo nghiên cứu [65], bằng cách thiết lập điểm số nhận dạng miền là 0.5, các đặc trưng từ hai miền được làm cho không thể phân biệt, từ đó hình thành một không gian đặc trưng chung. Giá trị 0.5 được xem là ngưỡng tại đó DNet không còn khả năng xác định liệu một đặc trưng cụ thể xuất phát từ miền nguồn hay miền đích.

Kết hợp với hai hàm mất mát ở Công thức 3.1, hàm mất mát tổng quát trong giai đoạn học có giám sát trên miền nguồn này được biểu diễn theo Công thức 3.6.

$$L_{SL} = L_{ce} + L_{tr} + \lambda_{DIM} * L_{DIM}$$  (3.6)

trong đó, $\lambda_{DIM}$ là trọng số hàm mất mát DIM.

Chú ý rằng, trong mỗi vòng lặp, DNet sẽ được huấn luyện trước với Công thức 3.4. Sau đó, mô hình tái định danh $f_\theta$ sẽ được huấn luyện với Công thức 3.6.

Tổng thể, hai kỹ thuật chính là (i) thích ứng miền mức hình ảnh bằng StarGAN và (ii) thích ứng miền mức đặc trưng bằng DIM đã được đề xuất sử dụng trong quá trình huấn luyện có giám sát trên miền nguồn của "DAPRH". Hai thành phần này đóng vai trò then chốt trong việc nâng cao hiệu quả tổng thể của mô hình và được minh họa cụ thể trong Hình 3.3.

## 3.2.3 Đánh giá chất lượng ảnh và điều chỉnh trọng số huấn luyện

Do tính không hoàn hảo của tập dữ liệu tổng hợp $D_{cb}$, đặc biệt là sự xuất hiện của nhiều mẫu sinh ra từ GAN có chất lượng thấp (như minh họa trong Hình 3.4), việc sử dụng toàn bộ dữ liệu trong quá trình huấn luyện có thể gây nhiễu và làm suy giảm hiệu quả học biểu diễn đặc trưng. Để khắc phục vấn đề này, chúng tôi áp dụng một phương pháp đánh giá nhằm xác định và giảm sự ảnh hưởng của chúng tới mô hình huấn luyện. Cụ thể, chúng tôi đề xuất sử dụng một thuật toán đánh giá chất lượng ảnh dựa trên chuẩn hóa véc-tơ đặc trưng, trong đó giá trị chuẩn hóa của đặc trưng được sử dụng làm chỉ báo cho chất lượng ảnh. Cách tiếp cận này được xây dựng dựa trên nhận định rằng chuẩn hóa đặc trưng có tương quan với chất lượng hình ảnh, như đã được chỉ ra trong [55]. Cụ thể, chuẩn hóa đặc trưng được sử dụng để ước lượng giá trị chất lượng $z_i$ của ảnh $x_i$ được biểu diễn trong Công thức 3.7.

$$z_i = \left\lfloor \frac{\|f_i\| - \mu_f}{\sigma_f / h} \right\rfloor_{-1}^{1} \in \ [-1, 1] \tag{3.7}$$

trong đó, $\mu_f$ và $\sigma_f$ lần lượt là giá trị trung bình và độ lệch chuẩn của tất cả các véc-tơ chuẩn hóa đặc trưng $\|f_i\|$ trong một lô đầu vào. Hàm toán học $\lfloor \cdot \rfloor$ được sử dụng để giới hạn giá trị tính toán trong khoảng từ $-1$ đến $1$. Ngoài ra, tham số $h$ được dùng để điều chỉnh mức độ tập trung của giá trị đầu ra. Theo thiết lập từ nghiên cứu [55], giá trị $h$ được cài đặt là 0.33, nhằm mục đích đưa giá trị $z_i$ nằm trong vùng $[-1, 1]$ nhiều nhất có thể.

Dựa trên chỉ số chất lượng này, hàm mất mát tổng thể được biểu diễn trong Công thức 3.8.

$$L(x_i, y_i) = (z_i * \lambda_z + 1) * (L_{ce}(x_i, y_i) + L_{tri}(x_i)) \tag{3.8}$$

ở đây, $\lambda_z$ là một siêu tham số có giá trị trong khoảng $[0.0, 1.0]$, được sử dụng để điều chỉnh mức độ ảnh hưởng của giá trị $z_i$. Lưu ý rằng theo Công thức 3.7, giá trị $z_i$ nằm trong khoảng $[-1, 1]$. Do đó, tích $z_i * \lambda_z$ sẽ nằm trong khoảng $[-\lambda_z, \lambda_z] \subseteq [-1, 1]$.

Để đảm bảo giá trị này không âm, một hằng số 1 được cộng thêm, tức là biểu thức cuối cùng trở thành $z_i * \lambda_z + 1$, đảm bảo đầu ra luôn nằm trong khoảng $[1 - \lambda_z, 1 + \lambda_z] \subseteq [0, 2]$.

Cơ chế điều chỉnh trọng số này cho phép sự khác biệt về chất lượng giữa các mẫu dữ liệu được mô hình nhận biết. Cụ thể, trọng số ảnh hưởng sẽ có giá trị cao khi ảnh có chất lượng cao và có giá trị thấp khi ảnh có chất lượng thấp. Dựa trên giá trị của hàm đánh giá chất lượng ảnh, hàm mất mát tổng thể của mô hình sẽ được điều chỉnh tương ứng. Trong quá trình huấn luyện, trọng số của mô hình sẽ được cập nhật dựa trên hàm mất mát cuối cùng nhằm tối ưu hiệu năng, và kết quả tốt nhất sẽ được lựa chọn để sử dụng cho giai đoạn tiếp theo – giai đoạn đánh giá.

Tổng thể, hai kỹ thuật chính là (i) thích ứng miền mức hình ảnh bằng StarGAN và (ii) đánh giá chất lượng ảnh để điều chỉnh trọng số huấn luyện được đề xuất sử dụng trong quá trình huấn luyện có giám sát trên miền nguồn được trình bày trong đề xuất "IQAGA". Hai thành phần này đóng vai trò then chốt trong việc nâng cao hiệu quả tổng thể của mô hình và được minh họa cụ thể trong Hình 3.2.

# 3.3 Phương pháp nâng cao học thích ứng miền không giám sát trên miền đích

Trong giai đoạn huấn luyện không giám sát trên miền đích, hai chiến lược chính được sử dụng là: (i) sử dụng đặc trưng toàn cục và đặc trưng cục bộ từ các phần của hình ảnh để kết hợp như một đặc trưng tổng thể để thu thập tất cả thông tin cần thiết trong suốt quá trình học, (ii) phát triển một phương pháp tinh chỉnh nhãn giả đơn giản sử dụng nhãn mềm thay vì nhãn cứng để giảm thiểu tác động của các nhãn nhiễu và học dựa trên kiến trúc Teacher-Student để nâng cao hiệu năng trong quá trình học không giám sát.

## 3.3.1 Đặc trưng tổng thể dựa trên đặc trưng toàn cục và cục bộ

Từ những thành tựu của các công trình gần đây [20, 38, 117, 102], chúng tôi không chỉ quan tâm đến thông tin từ các đặc trưng toàn cục mà còn sử dụng thông tin từ các phần cục bộ của hình ảnh để cải thiện sự tương đồng của các hình ảnh mục tiêu và cải thiện nhãn giả. Như được minh họa trong giai đoạn II của Hình 3.3, mô hình học sâu trong giai đoạn này bao gồm hai nhánh cùng sử dụng một mạng cốt lõi.

- **Nhánh đặc trưng toàn cục:** Để lấy đặc trưng toàn cục của một hình ảnh $x_i$ tại epoch $T$, trước tiên, chúng tôi cần trích xuất bản đồ đặc trưng tương ứng $F_\theta^T(x_i) \in R^{C \times H \times W}$, trong đó $C$, $H$ và $W$ lần lượt là thông tin của kênh, chiều cao và chiều rộng của hình ảnh. Với biểu diễn này, khác với một số công trình trước đây, để lấy đặc trưng toàn cục $f_i^T$, chúng tôi thêm một lớp GMP [89] song song với lớp GAP trên bản đồ đặc trưng. Véc-tơ GMP sẽ được sử dụng để phân cụm và học qua hàm mất mát bộ ba, trong khi đặc trưng GAP sẽ được sử dụng hàm mất mát phân loại. Sau đó, đặc trưng toàn cục bổ sung $g_i^T$ được xuất ra từ nhánh đó sẽ được truyền vào bộ phân loại và sử dụng để tính toán hàm mất mát nhận dạng như trình bày ở phần dưới cùng của giai đoạn II trong Hình 3.3. Vì có nhiều thông tin phân biệt hơn về ảnh định danh trong các vùng có giá trị phản hồi cao hơn, phương pháp này cho phép mô hình khai thác thông tin nổi bật được thu thập bởi lớp gộp cực đại.

- **Nhánh đặc trưng cục bộ:** Bên cạnh thông tin toàn cục, chúng tôi cũng xem xét các ưu điểm đến từ thông tin cục bộ được trích xuất từ các vùng của hình ảnh. Tương tự như các nghiên cứu trước [20, 26, 102], các đặc trưng cục bộ có thể thu được bằng cách chia ngang bản đồ đặc trưng thành $K$ vùng có kích thước bằng nhau, và sau đó áp dụng lớp GMP trên từng vùng. Kết quả là một tập hợp gồm $K$ đặc trưng $\{f_{p0}^i, f_{p1}^i, ..., f_{pk}^i | f_{pi}^i \in R^{C \times \frac{H}{K} \times W}\}$ được tạo ra. Khác với một số nghiên cứu trước đây, chúng tôi không áp dụng trực tiếp các hàm mất mát thường dùng trên các đặc trưng thành phần này mà nối $K$ đặc trưng phần lại thành một véc-tơ cục bộ toàn diện $f'_i$. Đặc trưng này sau đó chỉ được sử dụng bởi hàm mất mát nhận dạng và được gán nhãn tương ứng với nhánh toàn cục.

Trong phần này, giá trị $K$ được lựa chọn là 2 nhằm đảm bảo khả năng triển khai trên hệ thống huấn luyện của chúng tôi sử dụng GPU với bộ nhớ giới hạn là 16GB VRAM. Bên cạnh đó, việc huấn luyện đồng thời cả hai nhánh - cục bộ và toàn cục - được thực hiện bằng cách sử dụng hàm mất mát nhận dạng theo Công thức 3.9.

$$L_{id}(p_i, p'_i, y_i) = L_{ce}(p_i, y_i) + L_{ce}(p'_i, y_i) \tag{3.9}$$

## 3.3.2  Tinh chỉnh nhãn giả dựa trên tâm cụm

Một trong những vấn đề không thể tránh khỏi khi sử dụng thuật toán phân cụm để tạo ra nhãn giả là sự xuất hiện của nhiễu. Ví dụ như khi các cá nhân mặc trang phục có màu sắc giống nhau, thuật toán phân cụm, cụ thể là DBSCAN, có thể xảy

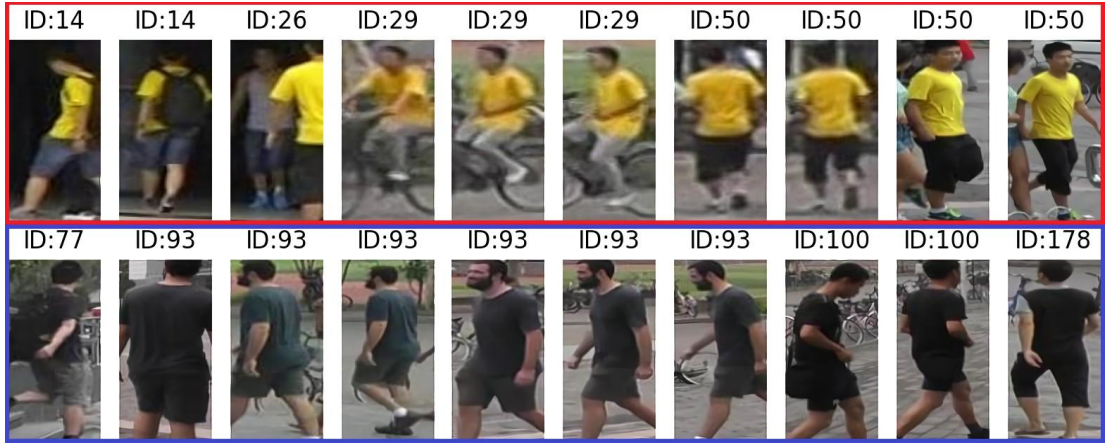

Hình 3.6: Ví dụ về vấn đề nhiễu nhãn.

ra nhầm lẫn, dẫn đến việc hình thành các cụm không nhất quán. Ví dụ minh họa trong Hình 3.6 cho thấy cụm được khoanh đỏ gồm bốn người khác nhau nhưng đều mặc áo vàng, trong khi cụm màu xanh cũng có bốn người với trang phục gần như giống hệt. Kết quả là các mẫu bị gán nhãn sai có thể ảnh hưởng tiêu cực đến quá trình huấn luyện của mô hình, làm giảm hiệu quả học đặc trưng và khả năng phân biệt giữa các danh tính.

Do đó, để giảm thiểu tác động của nhãn nhiễu, nhãn giả cố định $y_i$ sẽ được cập nhật theo Công thức 3.10.

$$\hat{\mathbf{y}}_i = (1 - \alpha)\mathbf{y}_i + \alpha\mathbf{G}_i \quad (3.10)$$

với $\hat{\mathbf{y}} \in \mathsf{R}^K$ được định nghĩa là nhãn giả đã được tinh chỉnh, đây là nhãn mềm thay vì nhãn cứng cố định ban đầu. $\alpha$ là một siêu tham số được sử dụng để điều chỉnh sự sửa đổi. Hơn nữa, $\mathbf{G}_i$ là một véc-tơ trọng số biểu diễn mối quan hệ khoảng cách giữa véc-tơ đặc trưng $i$ và tất cả các tâm cụm trong không gian đặc trưng. Mục tiêu của việc gán nhãn này là khuyến khích các mẫu tiếp cận không chỉ tâm cụm ban đầu mà còn các tâm cụm lân cận, nơi thông tin nhận dạng của chúng có thể được ánh xạ.

Để tính toán $\mathbf{G}_i$, khoảng cách tổng thể của các tâm cụm ban đầu và các véc-tơ đặc trưng được xây dựng. Giả sử ma trận $\mathbf{G}$ đại diện cho khoảng cách đó được tính theo Công thức 3.11.

$$G_{i,j} = \frac{exp(d_E(f_i, m_j)/\tau)}{\mathsf{L}_{j=1}^{|C|} exp(d_E(f_i, m_j)/\tau)} \quad (3.11)$$

là giá trị khoảng cách Euclid giữa véc-tơ đặc trưng $f_i$ tương ứng với mẫu thứ $i$-th từ $\mathsf{D}_t$ và véc-tơ tâm cụm $m_j$ tương ứng với cụm thứ $j$-th từ bộ tâm cụm ban đầu $\mathsf{C}$, được tính bằng Công thức 3.12.

$$m_j = \frac{1}{|\mathsf{C}_j|} \sum_{f_t \in \mathsf{C}_j} f_t \tag{3.12}$$

trong đó, $\mathsf{C}_j$ đại diện cho tập hợp các mẫu trong cụm thứ *j*. |.| đại diện cho phép toán lấy số lượng phần tử trong tập hợp và $\tau$ là siêu tham số điều chỉnh.

Chiến lược tinh chỉnh này giúp cải thiện sự đa dạng của nhãn giả đã tinh chỉnh $\hat{y}$. Đối với một mẫu ở biên của cụm, chiến lược của chúng tôi sẽ giảm độ tin cậy của việc học từ nhãn cứng vì nó có khả năng bị gán nhãn sai cao. Hơn nữa, mô hình sẽ học cách làm đa dạng các đặc trưng bằng cách tăng sự khác biệt giữa mẫu này và những mẫu khác thuộc về cụm của nó, gần với tâm cụm.

Ngoài ra, do sự tồn tại các mẫu nhiễu trong cụm, như đã đề cập, các véc-tơ tâm cụm tính trực tiếp theo Công thức 3.12 có thể không đáng tin cậy. Do đó, Miao và cộng sự [74] đề xuất sử dụng giá trị Silhouette [92] để đánh giá mối tương quan giữa các mẫu và các tâm cụm theo Công thức 3.13.

$$s_i = \frac{b_i - a_i}{\max(a_i, b_i)} \in [-1, 1] \tag{3.13}$$

trong đó, $a_i$ đại diện cho khoảng cách trung bình giữa véc-tơ đặc trưng *thứ i* và các mẫu khác trong cùng một cụm được khởi tạo bởi DBSCAN và $b_i$ là khoảng cách của véc-tơ đó đến các mẫu trong cụm lân cận gần nhất trong không gian đặc trưng.

Một mẫu sẽ được coi là gần các điểm trong cùng một cụm và xa các điểm trong các cụm khác nếu giá trị Silhouette cao. Nói cách khác, nhãn của nó có độ tin cậy cao, và Công thức 3.12 có thể được tính toán lại theo Công thức 3.14.

$$m_j = \frac{1}{|S_j|} \sum_{f_i \in S_j} f_i \text{ với } S_j = \{f_t \in \mathsf{C}_j \mid s_t > \sigma\} \tag{3.14}$$

ở đây, một tập con với độ tin cậy $S_j$ được chọn từ cụm ban đầu $\mathsf{C}_j$ dựa theo ngưỡng chọn lọc $\sigma$. Quá trình lọc này sẽ làm giảm tác động của các hình ảnh chất lượng thấp trong quá trình huấn luyện khiến mô hình có thể xác định danh tính sai.

Độ đo Silhouette được sử dụng để đánh giá mức độ phù hợp của mỗi mẫu với cụm mà nó được gán, bằng cách so sánh độ tương đồng giữa mẫu đó với các mẫu trong cùng cụm và các cụm lân cận. Giá trị Silhouette nằm trong khoảng từ -1 đến 1; trong đó, giá trị càng cao cho thấy mẫu càng phù hợp với cụm hiện tại, trong khi giá trị âm thường chỉ ra rằng mẫu có thể đã bị gán sai cụm, do cụm khác có độ tương đồng cao hơn [92]. Dựa trên đặc điểm này, chúng tôi đặt ngưỡng $\sigma$ bằng 0 để loại bỏ các mẫu có độ tin cậy thấp, tức là những mẫu ít có khả năng thực sự thuộc về tâm cụm được gán nhãn giả.

### 3.3.3 Huấn luyện dựa trên kiến trúc Teacher-Student

Trong giai đoạn II, dữ liệu ở miền đích do không có nhãn nên phần nào tính chất động và khó dự đoán. Số lượng hình ảnh mang nhãn giả do DBSCAN tạo ra thay đổi theo từng epoch, trong khi số lượng mẫu trong mỗi lớp cũng không cố định và rất khó để kiểm soát. Tình huống này đặt ra những thách thức đặc thù, nhưng cũng đồng thời mở ra cơ hội cho mô hình học được sự tổng quát từ sự biến động và tính linh hoạt của dữ liệu:

- Sự biến động về số lượng mẫu mang nhãn giả qua từng epoch đồng nghĩa với việc mô hình được huấn luyện trên một tập dữ liệu liên tục thay đổi và đa dạng. Điều này giúp mô hình học được những đặc trưng tổng quát hơn, tăng khả năng khái quát hoá và khả năng thích nghi với dữ liệu chưa từng thấy.

- Sự không đồng đều về số lượng danh tính trong mỗi cụm là một thách thức phổ biến khi áp dụng các thuật toán phân cụm không giám sát như DBSCAN. Mất cân bằng cụm có thể dẫn đến thiên lệch trong quá trình huấn luyện, khi các cụm lớn ảnh hưởng quá mức đến việc cập nhật tham số mô hình.

- Việc không thể kiểm soát trước số lượng danh tính trong mỗi cụm đòi hỏi mô hình phải có tính linh hoạt cao và khả năng thích ứng với phân phối lớp không đồng đều. Mô hình cần học một cách hiệu quả bất kể sự biến động trong cấu trúc dữ liệu đầu vào.

Như được minh họa trong Hình 3.3, trong giai đoạn II, chúng tôi đề xuất sử dụng kiến trúc Teacher-Student để xử lý vấn đề sự biến động và không thể đoán trước trong giai đoạn học không giám sát. Đầu tiên, chúng tôi cũng sử dụng mô hình Teacher để sinh ra các đặc trưng của mẫu, sau đó áp dụng DBSCAN để tạo ra nhãn giả như được minh họa trong Hình 3.3. Sau đó, chúng tôi sử dụng phương pháp tinh chỉnh nhãn giả dựa trên tâm cụm được trình bày trong Mục 3.3.2 để làm cho nhãn giả chính xác hơn là chỉ tin vào các dự đoán của Teacher.

Tiếp theo, mô hình Student được huấn luyện dựa trên cả dữ liệu đầu vào và các dự đoán từ mô hình Teacher trong mỗi epoch. Tương tự như các phương pháp trong [130] và [99], chúng tôi huấn luyện song song hai mô hình, trong đó mỗi mô hình đồng thời đảm nhiệm vai trò của Teacher $F_{\theta'}$ và Student $F_\theta$. Với mỗi lô hình ảnh đầu vào, mô hình Student liên tục học các đặc trưng mới, trong khi mô hình Teacher được cập nhật ít thường xuyên hơn, thông qua cơ chế lấy trung bình trọng số từ tham số của Student. Mô hình Teacher đóng vai trò định hướng, giúp Student học

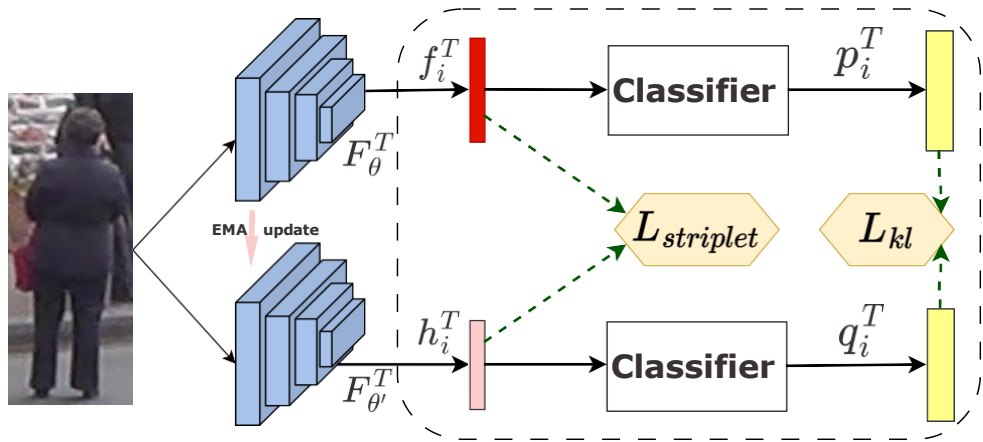

Hình 3.7: Tổng quát về kiến trúc Teacher-Student.

một cách ổn định hơn, đặc biệt trong các tình huống dữ liệu đầu vào không đồng nhất hoặc chứa nhiễu. Thiết kế này có một số ưu điểm:

- Khả năng chống nhiễu: Kiến trúc Teacher-Student nổi tiếng là có khả năng chống nhiễu dữ liệu. Trong trường hợp này, sự thay đổi trong số lượng hình ảnh có nhãn giả ở mỗi epoch và số lượng nhãn trong mỗi lớp có thể là nguồn gốc của nhiễu.

- Xử lý dữ liệu không cân bằng: Dữ liệu không cân bằng có thể được xử lý tốt bởi kiến trúc Teacher-Student. Sự mất cân bằng lớp có thể phát sinh từ sự thay đổi lớn trong số lượng thành viên trong mỗi cụm, nhưng kiến trúc này có thể xử lý được điều đó.

- Độ nhất quán dự đoán: Mặc dù dữ liệu đầu vào có thể bị lệch hoặc không cân bằng, kiến trúc Teacher-Student thúc đẩy độ nhất quán trong dự đoán. Kết quả đầu ra của mô hình Teacher được sao chép trong mô hình Student thông qua quá trình huấn luyện.

- Bộ nhớ về các trạng thái trước đó: Mô hình Teacher duy trì một "bộ nhớ" các trạng thái trước đó của nó, giúp ổn định các dự đoán theo thời gian. Điều này đặc biệt hữu ích khi xử lý dữ liệu biến thiên, vì nó cho phép mô hình đưa ra các dự đoán nhất quán mặc dù có sự thay đổi trong dữ liệu đầu vào.

Để thiết lập, chúng tôi sử dụng hai hàm mất mát để cho phép mô hình Student học từ mô hình Teacher: hàm mất mát Kullback–Leibler [17] $L_{kl}(p_i, q_i)$ và hàm mất mát Soft-Triplet [84] $L_{stri}(f_i, h_i)$. Trong các hàm mất mát này, $p_i$, $q_i$, $f_i$, và $h_i$ đại diện cho đầu ra của lớp phân loại và các véc-tơ đặc trưng được trích xuất từ mô hình Student và mô hình Teacher như được thể hiện trong Hình 3.7. Do đó, các hàm này cho phép đánh giá sự khác biệt trong phân phối các dự đoán giữa mô hình Teacher và mô hình Student. Sau đó, mô hình Teacher $\theta'$ được học bởi mạng Student $\theta$ trong bước huấn luyện $t$ theo chiến lược trung bình cập nhật hàm mũ (EMA) theo Công thức 3.15.

$$\theta'_t = w * \theta'_{t-1} + (1 - w) * \theta_t \tag{3.15}$$

Điều này đảm bảo rằng mô hình Teacher vẫn ổn định ngay cả trong quá trình huấn luyện. Cần lưu ý rằng trọng số điều khiển $w$ thường có giá trị tối ưu là 0.99 [99].

### 3.3.4 Hàm mất mát tổng thể

Cuối cùng, để tối ưu hóa mô hình tái định danh người trong giai đoạn II, luận án sử dụng hàm mất mát nhận dạng từ Công thức 3.9, hàm mất mát bộ ba từ Công thức 1.10, và hai hàm mất mát trong Mục 3.3.3 để học từ mô hình Teacher. Với nhãn giả đã tinh chỉnh, hàm mất mát mục tiêu có thể được biểu diễn theo Công thức 3.16.

$$\begin{aligned} \text{L}_{\text{USL}} = (1 - w_1)\text{L}_{id}(p_i, p'_i, \hat{y}_i) + w_1\text{L}_{kl}(p_i, q_i) \\ + (1 - w_2)\text{L}_{tri}(f_i, y_i) + w_2\text{L}_{stri}(f_i, h_i) \end{aligned} \tag{3.16}$$

Trong hàm mất mát tổng thể, hai trọng số $w_1$, $w_2$, được điều chỉnh là 0.4 và 0.8, được sử dụng để quản lý việc học từ Teacher và học thông tin mới. Sau đó, $p_i$, $p'_i$ và $q_i$ đại diện cho đầu ra lớp phân loại từ các nhánh toàn cục, cục bộ và toàn cục của mô hình Teacher, tương ứng. Cuối cùng, $f_i$ và $h_i$ là các đặc trưng toàn cục được trích xuất từ mô hình Student và Teacher, tương ứng.

## 3.4 Thực nghiệm và đánh giá kết quả

Để chứng minh tính hiệu quả của các phương pháp đề xuất, các thực nghiệm toàn diện được thực hiện trên các trường hợp "Market-to-Duke", "Duke-to-Market", "Market-to-MSMT", "Duke-to-MSMT" với máy tính trang bị một GPU NVIDIA Tesla T4 với 16GB VRAM để trả lời các câu hỏi nghiên cứu.

- RQ1: Việc thích ứng miền mức hình ảnh sử dụng GAN có giúp nâng cao hiệu năng mô hình không?

- RQ2: Việc thích ứng miền mức đặc trưng sử dụng DIM có giúp nâng cao hiệu năng mô hình không?

- RQ3: Việc kết hợp thích ứng miền mức hình ảnh sử dụng GAN và thích ứng miền mức đặc trưng sử dụng DIM có giúp nâng cao hiệu năng mô hình không?

- RQ4: Việc đánh giá chất lượng ảnh để điều chỉnh trọng số huấn luyện có giúp nâng cao hiệu năng mô hình không?

- RQ5: Việc kết hợp thích ứng miền ở mức hình ảnh sử dụng GAN và đánh giá chất lượng ảnh để điều chỉnh trọng số huấn luyện có giúp nâng cao hiệu năng mô hình không?

- RQ6: Việc khai thác các đặc trưng tổng thể có giúp cải thiện hiệu năng của các mô hình không?

- RQ7: Phương pháp tinh chỉnh nhãn giả được đề xuất có khả năng giảm thiểu ảnh hưởng của nhiễu nhãn và cải thiện hiệu năng phân biệt của mô hình không?

- RQ8: Các giá trị tối ưu của các siêu tham số trong phương pháp đề xuất là bao nhiêu và ảnh hưởng của chúng đến hiệu năng tổng thể ra sao?

- RQ9: Hiệu suất tính toán của phương pháp đề xuất so với mô hình cơ sở như thế nào?

- RQ10: Việc kết hợp tất cả các thành phần có giúp cải thiện hiệu năng tổng thể của mô hình tái định danh người thích ứng miền không giám sát so với các phương pháp tiên tiến không?

Các kết quả thực nghiệm tương ứng sẽ cho thấy mức độ đáp ứng của phương pháp đối với từng câu hỏi nghiên cứu.

## 3.4.1   Chi tiết triển khai

### 3.4.1.1   Tham số thích ứng miền ở mức hình ảnh

StarGAN được sử dụng để thực hiện chuyển đổi phong cách ảnh, với các ảnh đầu vào được chuyển đổi kích thước về $256 \times 128$. Việc huấn luyện mô hình GAN được thực hiện bằng thuật toán tối ưu Adam với tốc độ học $3.5 \times 10^{-5}$ và kích thước lô là 16, theo khuyến nghị từ bài báo gốc [21]. Sau khi huấn luyện GAN, mỗi hình ảnh trong miền nguồn sẽ được chuyển đổi thành phong cách của $C$ camera mục tiêu. Cụ thể, trong trường hợp "Market-to-Duke", 8 hình ảnh chuyển đổi phong cách được sinh ra tương ứng với 8 camera khác nhau trong Duke cho mỗi hình ảnh trong Market, và ngược lại, với mỗi hình ảnh trong Duke, 6 hình ảnh mới được sinh ra tương ứng với 6 camera khác nhau trong Market. Trong các trường hợp "Market-to-MSMT" và "Duke-to-MSMT", 15 hình ảnh chuyển đổi phong cách được sinh ra tương ứng với 15 camera khác nhau trong MSMT cho mỗi hình ảnh trong các bộ dữ liệu Market và Duke. Một số kết quả được hiển thị trong Hình 3.8.

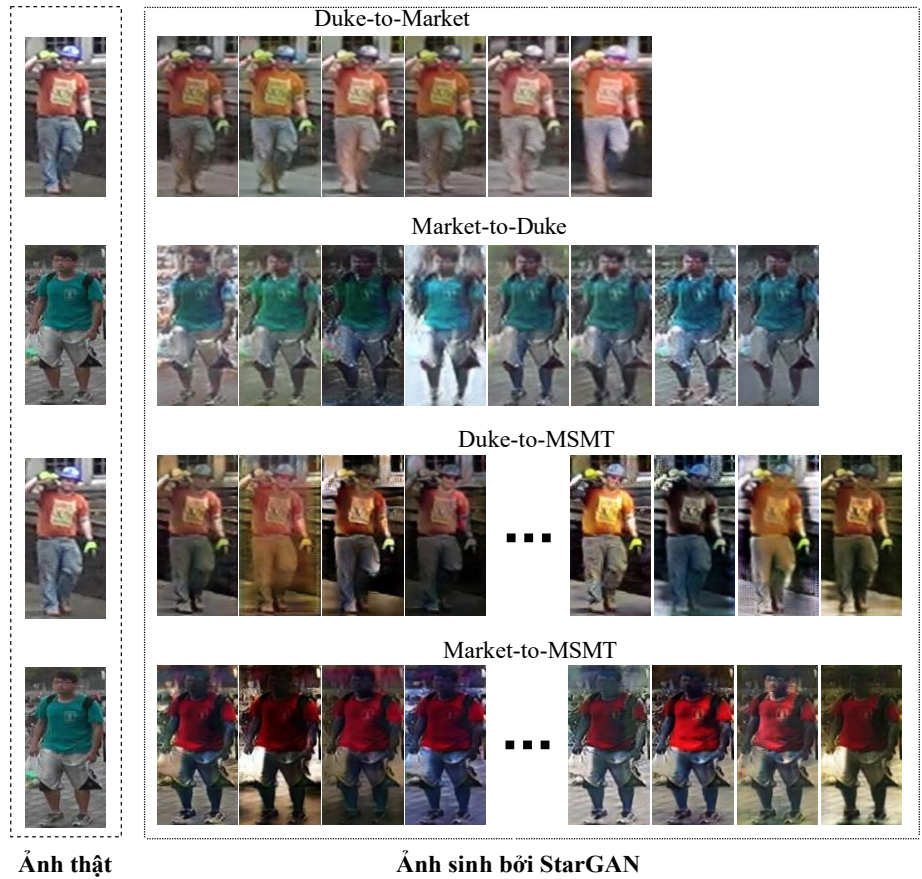

Hình 3.8: Một số ảnh được sinh bởi StarGAN.

#### 3.4.1.2    Tham số huấn luyện có giám sát trên miền nguồn

Quá trình huấn luyện có giám sát trên tập dữ liệu nguồn, mô hình được huấn luyện với kích thước lô là 128, bao gồm 16 danh tính, mỗi danh tính tương ứng với 8 ảnh khác nhau trong mỗi lô. Hơn nữa, mỗi ảnh trước khi huấn luyện được tăng cường theo các kỹ thuật trong 1.2.5. Đối với hàm mất mát bộ ba, tham số margin được cài đặt ở giá trị mặc định là 0.3. Bộ tối ưu Adam được sử dụng với tốc độ học được thiết lập ban đầu là $3.5 \times 10^{-4}$. Quá trình huấn luyện kéo dài trong 80 epoch, trong đó tốc độ học được giảm 10 lần tại epoch thứ 40 và 70. Đối với kỹ thuật đánh giá chất lượng ảnh để điều chỉnh trọng số huấn luyện, giá trị hệ số *h* trong Công thức 3.7 được đặt là 0.33 như trong bài báo gốc [55].

#### 3.4.1.3    Tham số huấn luyện không giám sát trên miền đích

Quá trình huấn luyện không giám sát trên miền đích, mô hình được huấn luyện trong 40 epoch. Tại đầu mỗi epoch, thuật toán DBSCAN được sử dụng để gán nhãn

Bảng 3.2: Hiệu năng cơ sở mô hình "IQAGA"

| Phương pháp | Duke → Market | | | | Market → Duke | | | |
|---|---|---|---|---|---|---|---|---|
| | mAP | R1 | R5 | R10 | mAP | R1 | R5 | R10 |
| *Cơ sở* | 26.2 | 55.3 | 67.8 | 75.2 | 25.8 | 43.7 | 57.9 | 66.2 |
| *+GAN* | 35.1 | 68.6 | 83.1 | 87.8 | 31.5 | 54.2 | 67.6 | 72.5 |
| *+GAN + ASMT* | 36.3 | 70.2 | 84.4 | 88.4 | 32.1 | 55.5 | 68.2 | 73.2 |

Bảng 3.3: Hiệu năng cơ sở của mô hình "DAPRH"

| Phương pháp | Duke → Market | | | | Market → Duke | | | | Market → MSMT | | | | Duke → MSMT | | | |
|---|---|---|---|---|---|---|---|---|---|---|---|---|---|---|---|---|
| | mAP | R1 | R5 | R10 | mAP | R1 | R5 | R10 | mAP | R1 | R5 | R10 | mAP | R1 | R5 | R10 |
| *Cơ sở* | 26.2 | 55.3 | 67.8 | 75.2 | 25.8 | 43.7 | 57.9 | 66.2 | 4.7 | 14.4 | 22.1 | 26.5 | 6.3 | 17.8 | 29.3 | 36.2 |
| *+GAN* | 36.2 | 66.4 | 81.1 | 86.1 | 34.4 | 57.6 | 70.3 | 76.0 | 9.6 | 28.8 | 40.3 | 52.9 | 12.0 | 35.8 | 48.1 | 53.4 |
| *+DIM* | 30.1 | 60.4 | 75.1 | 80.8 | 28.8 | 51.1 | 66.0 | 71.2 | 8.6 | 25.3 | 35.9 | 41.6 | 10.7 | 31.2 | 43.0 | 48.8 |
| *+GAN+DIM* | 36.4 | 68.9 | 83.2 | 87.9 | 35.4 | 58.2 | 72.4 | 77.5 | 12.4 | 36.0 | 49.2 | 54.4 | 16.0 | 43.8 | 56.8 | 62.6 |

giá cho dữ liệu không giám sát. Đối với thuật toán DBSCAN, giá trị $\epsilon$ được thiết lập là 0.6 cho tất cả các trường hợp, số lượng hàng xóm tối thiểu được thiết lập thông qua thực nghiệm lần lượt là 8, 16, và 16 đối với các bộ dữ liệu đích Market, Duke, và MSMT, tương ứng. Cuối cùng, trong cả hai giai đoạn huấn luyện, dữ liệu được lấy mẫu theo chiến lược ngẫu nhiên, với mỗi lô bao gồm 16 danh tính, mỗi danh tính có 8 ảnh, tạo thành kích thước lô là 128.

## 3.4.2 Kết quả học có giám sát trên miền nguồn

### 3.4.2.1 Thích ứng miền mức hình ảnh

Như kết quả ở Bảng 3.2, việc tăng cường dữ liệu giúp cải thiện đáng kể hiệu năng mô hình cơ sở, từ 26.2%, 55.3% lên 35.1%, 68.6% (mAP/Rank-1) trong trường hợp "Duke-to-Market", và từ 25.8%, 43.7% lên 31.5%, 54.2% (mAP/Rank-1) trong thiết lập "Market-to-Duke". Kết quả cho thấy tăng cường dữ liệu bởi StarGAN góp phần hiệu quả trong việc thu hẹp khoảng cách phân phối giữa hai miền.

Mặt khác, thay vì sử dụng tất cả các ảnh sinh ra để làm dữ liệu huấn luyện, chúng tôi đánh giá mô hình với phương án sử dụng tỉ lệ $N : M$ ảnh thực với ảnh tăng cường là 4 : 1. Theo kết quả trong Bảng 3.3, phương án giúp cải thiện hiệu năng mô hình khoảng 10–13% trên cả hai chỉ số mAP và Rank-1. Cụ thể, từ 26.2%, 55.3% lên 36.2%, 66.4% (mAP/Rank-1) trong trường hợp "Duke-to-Market", và từ 25.8%, 43.7% lên 34.4%, 57.6% (mAP/Rank-1) trong thiết lập "Market-to-Duke". Hai kết quả này đã giải thích cho RQ1.

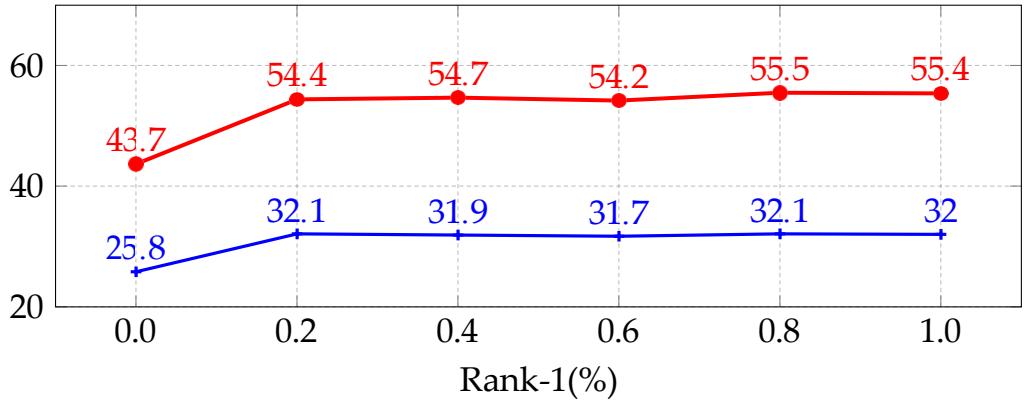

Hình 3.9: Đánh giá trường hợp "Market-to-Duke" với giá trị $\lambda_z$ khác nhau. Đường màu xanh biểu thị giá trị mAP trong khi đường màu đỏ biểu thị giá trị Rank-1.

### 3.4.2.2 Thích ứng miền mức đặc trưng

Tiếp theo, chúng tôi thực nghiệm ảnh hưởng của mô-đun DIM. Kết quả trong Bảng 3.3, cho thấy hiệu năng được cải thiện từ 26.2%, 55.3% lên 30.1% và 60.4% về mAP và Rank-1 trong trường hợp Duke-to-Market. Các trường hợp khác cũng được cải thiện với kết quả tương tự, với kết quả cải thiện khoảng 4% cho mAP và khoảng 8% cho đến 14% đối với Rank-1. Kết quả này đã giải thích cho RQ2.

### 3.4.2.3 Đánh giá chất lượng để điều chỉnh trọng số huấn luyện

Để đánh giá hiệu quả của phương án đánh giá chất lượng ảnh để điều chỉnh trọng số huấn luyện, chúng tôi phân tích tác động của siêu tham số $\lambda_z$ trong Công thức 3.8 bằng cách điều chỉnh giá trị của nó từ 0.0 đến 1.0. Kết quả thực nghiệm được trình bày trong Hình 3.9. Khi $\lambda_z > 0$, hàm mất mát sẽ được điều chỉnh tự động để giảm ảnh hưởng của các hình ảnh chất lượng thấp. Chúng tôi nhận thấy rằng giá trị tốt nhất của $\lambda_z$ là 0.8. So với trường hợp $\lambda_z = 0$, tức là không có cơ chế điều chỉnh trọng số huấn luyện, hiệu năng của mô hình được cải thiện khoảng 0.5-1% trên cả hai chỉ số mAP và Rank-1. Kết quả tốt nhất với giá trị $\lambda_z$ là 0.8 cũng được thể hiện ở hàng thứ 3 Bảng 3.2. Điều này cho thấy tính hiệu quả của hàm mất mát được đề xuất trong việc xử lý các hình ảnh có chất lượng thấp. Điều này giải thích cho RQ3.

### 3.4.2.4 Kết hợp thích ứng miền mức hình ảnh và đánh giá chất lượng để điều chỉnh trọng số huấn luyện

Cuối cùng, chúng tôi tiến hành đánh giá hiệu quả của việc kết hợp thích ứng miền ở mức hình ảnh với cơ chế đánh giá chất lượng ảnh để điều chỉnh trọng số huấn

Bảng 3.4: So sánh với các phương pháp tiên tiến

| Phương pháp | Nơi công bố | Duke → Market | | | | Market → Duke | | | |
|---|---|---|---|---|---|---|---|---|---|
| | | mAP | R1 | R5 | R10 | mAP | R1 | R5 | R10 |
| PTGAN[113] | CVPR18 | - | 38.6 | - | 66.1 | - | 27.4 | - | 50.7 |
| SPGAN+LMP[25] | CVPR18 | 26.9 | 58.1 | 76.0 | 82.7 | 26.4 | 46.9 | 62.6 | 68.5 |
| PGPPM[118] | IEEE.TMM20 | 33.9 | 63.9 | 81.1 | 86.4 | 17.9 | 36.3 | 54.0 | 61.6 |
| CDCSA[102] | ACM.MM22 | <u>34.0</u> | <u>66.8</u> | <u>82.1</u> | <u>87.9</u> | <u>31.4</u> | <u>53.8</u> | <u>67.8</u> | <u>72.2</u> |
| **IQAGA** | | **36.3** | **70.2** | **84.4** | **88.4** | **32.1** | **55.5** | **68.2** | **73.2** |

luyện – đây cũng chính là mục tiêu cốt lõi của phương pháp đề xuất IQAGA. Kết quả được trình bày trong Bảng 3.4.

Các kết quả thực nghiệm cho thấy IQAGA có kết quả tốt hơn các phương pháp tiên tiến, với 36.3% mAP và 70.2% Rank-1 trong trường hợp Duke-to-Market, và 32.1% mAP và 55.5% Rank-1 trong trường hợp Market-to-Duke. So với các phương pháp khác như PTGAN [113], SPGAN+LMP [25], và PGPPM [118], IQAGA thể hiện sự vượt trội rõ rệt trên tất cả các chỉ số. Khi so sánh với CDCSA [102], vốn cũng sử dụng phương pháp dựa trên StarGAN, IQAGA vượt trội hơn từ 0.7% đến 3% trên cả hai chỉ số mAP và Rank-1 trong cả hai kịch bản. Theo đó, IQAGA có tiềm năng ứng dụng trong các tình huống thực tế nhờ hiệu quả và tính đơn giản trên phần cứng hạn chế. Các kết quả này đã giải thích cho RQ4.

### 3.4.2.5 Kết hợp thích ứng miền mức hình ảnh và mức đặc trưng

Bảng 3.3 cho thấy kết quả vượt trội khi sử dụng phương án kết hợp thích ứng miền mức hình ảnh và mức đặc trưng. Cụ thể, trong thiết lập Duke-to-Market, mAP tăng từ 26.2% lên 36.4% và Rank-1 tăng từ 55.3% lên 68.9%. Trong trường hợp Market-to-Duke, mAP và Rank-1 cũng lần lượt tăng từ 25.8% lên 35.4% và từ 43.7% lên 58.2%. Đặc biệt, trong hai thiết lập khó hơn là Market-to-MSMT và Duke-to-MSMT, cách kết hợp cũng cho thấy mức cải thiện ấn tượng: mAP tăng lần lượt từ 4.7% lên 12.4% và từ 6.3% lên 16.0%, trong khi Rank-1 tăng từ 14.4% lên 36.0% và từ 17.8% lên 43.8%. Những kết quả này đã giải thích cho RQ5.

Hơn nữa, khi so sánh với phương pháp kết hợp thích ứng miền ở mức hình ảnh và đánh giá chất lượng ảnh ở Mục 3.4.2.4, kết quả đạt được trong phương án này cho thấy sự cải thiện về hiệu năng. Cụ thể, trong thiết lập Duke-to-Market, hai kết quả mAP và Rank-1 của hai phương án xấp xỉ nhau 36.3% với 36.4%, và 68.6% lên 68.9%. Nhưng đối với thiết lập Market-to-Duke, mAP tăng từ 32.1% lên 35.4% và Rank-1 cải thiện từ 55.2% lên 58.2%. Kết quả này cho thấy rằng việc chuyển từ cơ

Bảng 3.5: Ảnh hưởng của lớp gộp và đặc trưng cục bộ

| Phương pháp | Duke → Market | | | | Market → Duke | | | | Market → MSMT | | | | Duke → MSMT | | | |
|---|---|---|---|---|---|---|---|---|---|---|---|---|---|---|---|---|
| | mAP | R1 | R5 | R10 | mAP | R1 | R5 | R10 | mAP | R1 | R5 | R10 | mAP | R1 | R5 | R10 |
| *Cơ sở* | 81.5 | 92.0 | 96.9 | 98.0 | 58.8 | 74.5 | 82.7 | 85.5 | 29.8 | 55.6 | 66.5 | 73.9 | 32.6 | 60.5 | 73.2 | 79.0 |
| *+GMP* | 84.5 | 93.0 | 97.2 | 98.0 | 69.0 | 81.6 | 90.4 | 92.5 | 32.7 | 60.2 | 72.5 | 77.0 | 34.1 | 62.8 | 74.8 | 79.3 |
| *+GMP+PL* | 85.6 | 93.9 | 98.0 | 98.7 | 70.3 | 82.8 | 91.7 | 94.2 | 34.2 | 63.2 | 75.1 | 79.5 | 35.6 | 65.6 | 77.2 | 81.2 |

chế đánh giá chất lượng ảnh sang chiến lược chọn lọc ảnh kết hợp với DIM đã góp phần nâng cao hiệu quả nhận diện trong giai đoạn học có giám sát trên miền nguồn.

## 3.4.3 Kết quả thích ứng không giám sát trên miền đích

### 3.4.3.1 Đặc trưng tổng thể với GMP

Để trả lời câu hỏi RQ6, chúng tôi đã đánh giá hiệu quả của các kết hợp trong học không giám sát ở giai đoạn II, bao gồm việc sử dụng GMP và đặc trưng cục bộ. Kết quả được trình bày trong Bảng 3.5. Đầu tiên, mô hình được huấn luyện với chỉ có một nhánh toàn cục mà không có tinh chỉnh nhãn và GMP trong kiến trúc như mô hình cơ sở. Các mô hình này đạt được 81.5%, 92.0% về mAP và Rank-1 trong trường hợp Duke-to-Market và 58.8%, 74.5% về mAP và Rank-1 trong trường trường hợp Market-to-Duke.

Thứ hai, khi các bản đồ đặc trưng được kết hợp với lớp GMP để trích xuất các véc-tơ đặc trưng của hình ảnh (*Cơ sở + GMP*), kiến trúc này đạt được độ chính xác mAP là 84.5% và Rank-1 là 93.0% trên trường hợp Duke-to-Market, vượt trội so với hiệu năng của kiến trúc cơ sở. Trong các trường hợp khác, đề xuất cũng đạt được sự cải thiện tương tự. Ưu điểm của GMP là việc chỉ tập trung vào các giá trị đặc trưng có phản hồi cao, giúp tìm ra các đặc trưng phân biệt mạnh mẽ trong ảnh.

Cuối cùng, chúng tôi sử dụng đặc trưng cục bộ để kết hợp (*Cơ sở + GMP + PL*). So với phương án *Cơ sở + GMP*, việc sử dụng đặc trưng cục bộ (PL) mang lại sự cải thiện vừa phải trong hiệu năng của các mô hình chỉ học từ đặc trưng toàn cục là $1 - 2\%$ và $0.5 - 1\%$ về mAP và Rank-1 trên tất cả các trường hợp. Kết quả này chứng tỏ rằng việc huấn luyện với tăng cường nhánh đặc trưng cục bộ có thể bổ sung các thông tin có giá trị mà nhánh toàn cục bỏ qua để nâng cao hiệu quả tổng thể của học không giám sát trong bài toán tái định danh người.

Bảng 3.6: Kết quả của phương pháp tinh chỉnh nhãn giả

| Phương pháp | Duke → Market | | | | Market → Duke | | | | Market → MSMT | | | | Duke → MSMT | | | |
|---|---|---|---|---|---|---|---|---|---|---|---|---|---|---|---|---|
| | mAP | R1 | R5 | R10 | mAP | R1 | R5 | R10 | mAP | R1 | R5 | R10 | mAP | R1 | R5 | R10 |
| Nhãn cứng | 85.6 | 93.9 | 98.0 | 98.7 | 70.3 | 82.8 | 91.7 | 94.2 | 34.2 | 63.2 | 75.1 | 79.5 | 35.6 | 65.6 | 77.2 | 81.2 |
| Tinh chỉnh nhãn | 85.9 | 94.4 | 97.8 | 98.8 | 72.0 | 83.7 | 92.1 | 94.5 | 35.8 | 64.8 | 77.0 | 81.1 | 36.0 | 65.5 | 77.0 | 80.9 |

### 3.4.3.2 Tinh chỉnh nhãn giả

Để đánh giá hiệu quả của phương pháp tinh chỉnh nhãn giả được đề cập trong Mục 3.3.2 đồng thời trả lời câu hỏi nghiên cứu RQ7, chúng tôi tiến hành so sánh giữa các mô hình được huấn luyện với nhãn cứng trong Bảng 3.5 và các mô hình sử dụng phương pháp tinh chỉnh nhãn giả (CRL). Kết quả so sánh hiệu năng được trình bày trong Bảng 3.6. Có thể thấy rằng, CRL mang lại cải thiện về chỉ số mAP và Rank-1 lần lượt là 0.3% và 0.5% trên Duke-to-Market, 1.7% và 1.1% trên Duke-to-Market, cũng như từ 1 – 2% trên Market-to-MSMT và Duke-to-MSMT, so với phương án sử

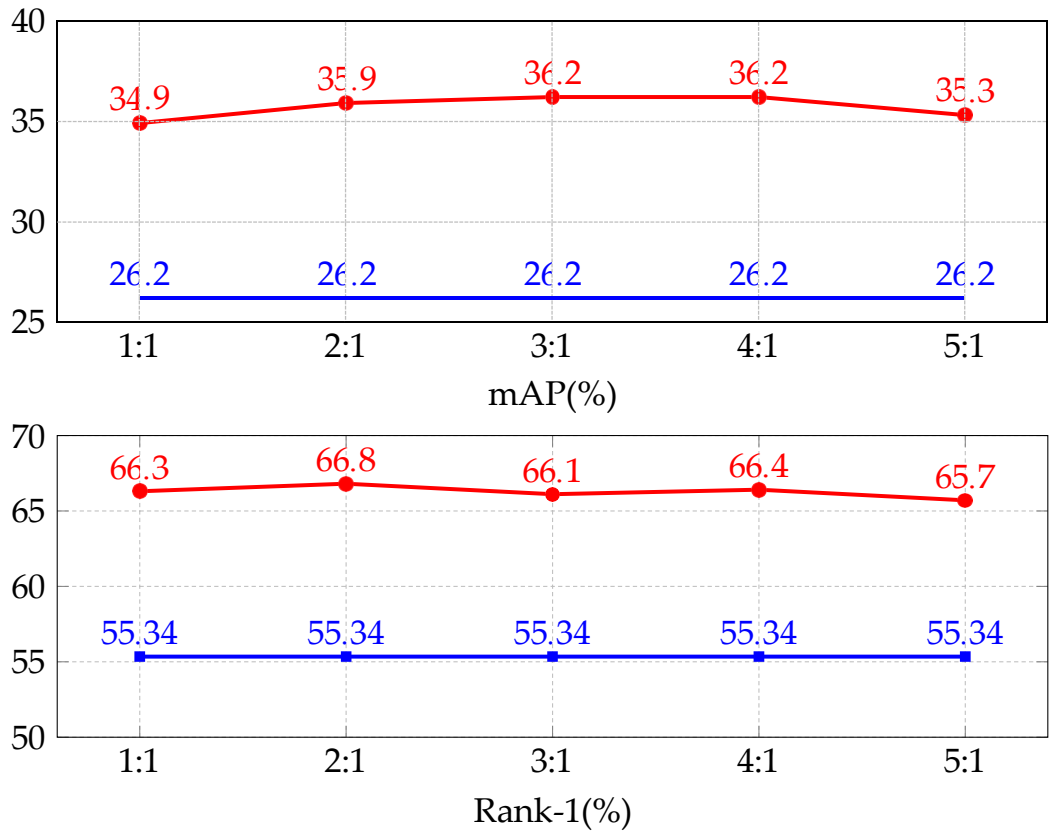

Hình 3.10: Đánh giá tập huấn luyện Market với các tỉ lệ $N : M$ khác nhau trong các lô. Đường màu xanh biểu thị kết quả của mô hình cơ sở, trong khi đường màu đỏ biểu thị kết quả mô hình cơ sở kết hợp với GAN.

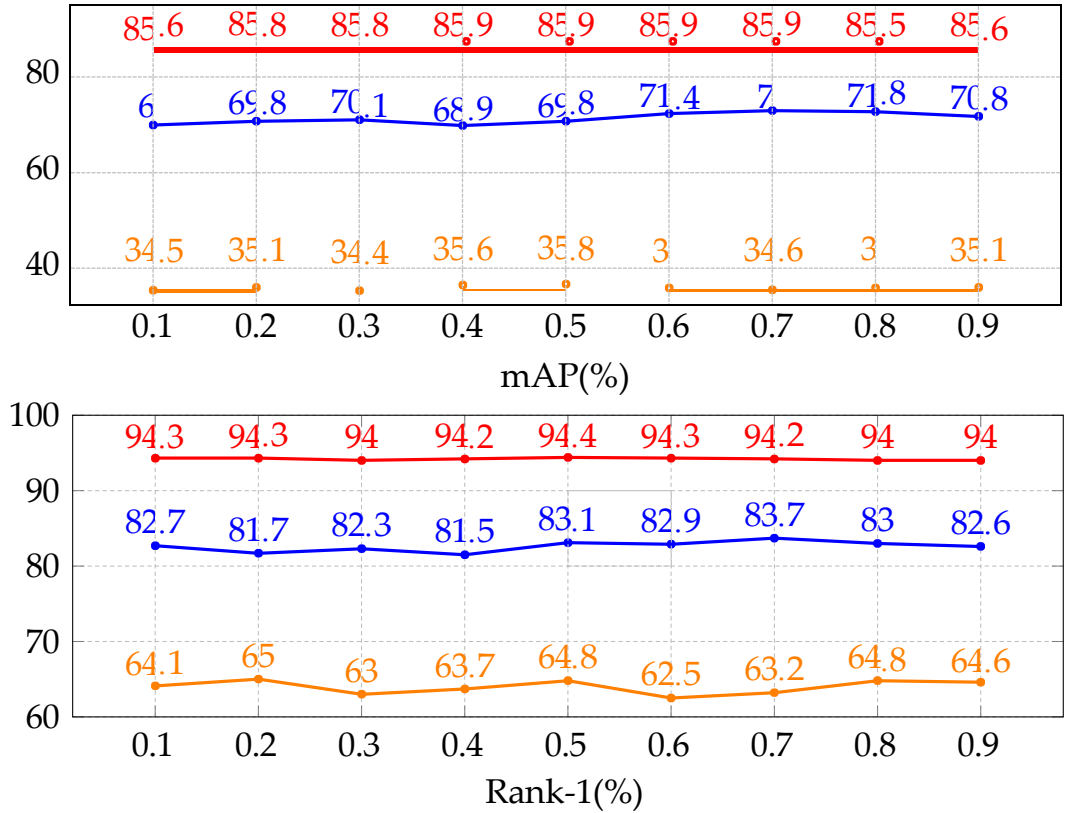

Hình 3.11: Tác động của tham số $\alpha$ trên Market-to-Duke (đường đỏ), Duke-to-Market (đường xanh) và Market-to-MSMT17 (đường cam).

dụng nhãn cứng. Những kết quả này cho thấy CRL có khả năng tạo ra nhãn tinh chỉnh hiệu quả hơn, qua đó góp phần giảm thiểu tác động của nhiễu nhãn trong quá trình huấn luyện mô hình.

### 3.4.3.3 Phân tích giá trị của các siêu tham số

Luận án phân tích giá trị các siêu tham số cũng như ảnh hưởng của chúng đến hiệu năng tổng thể để trả lời cho RQ8.

**Phân tích tỷ lệ $N : M$.** Tỷ lệ $N : M$ là một tham số quan trọng trong việc huấn luyện một mô hình cơ sở mạnh mẽ, trong đó N và M đại diện cho tỷ lệ giữa các mẫu thật và mẫu chuyển đổi phong cách trong mỗi lô đầu vào. Chúng tôi trình bày kết quả thử nghiệm trong Hình 3.10 bằng cách thay đổi tỷ lệ này. Rõ ràng, với các tỷ lệ $N : M$ khác nhau, bộ dữ liệu tổng hợp luôn cải thiện so với phương pháp cơ sở và đạt được kết quả cao nhất khi $N : M = 4 : 1$.

**Phân tích hệ số $\alpha$.** Chúng tôi đánh giá tác động của hệ số $\alpha$ trong Công thức 3.10 bằng cách điều chỉnh giá trị của $\alpha$ từ 0.1 đến 0.9. Các kết quả thực nghiệm trên ba trường hợp trong Hình 3.11 cho thấy rằng $\alpha$ có ảnh hưởng đáng kể đến hiệu năng

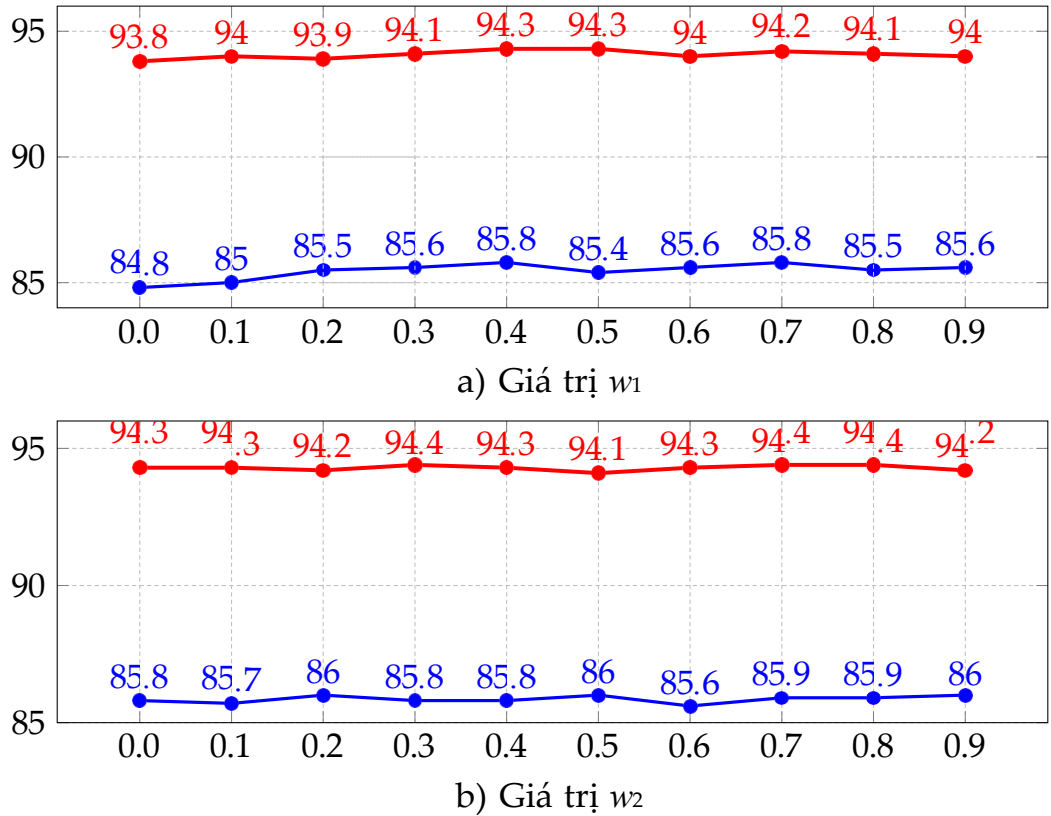

Hình 3.12: Ảnh hưởng của các trọng số điều khiển. Đường màu xanh biểu thị giá trị mAP, trong khi đường màu đỏ biểu thị giá trị Rank-1.

của phương pháp tinh chỉnh nhãn giả, và do đó cần được lựa chọn một cách cẩn thận theo từng miền dữ liệu cụ thể. Đối với trường hợp Market-to-Duke, mô hình đạt hiệu năng cao nhất khi $\alpha = 0.5$, trong khi với Duke-to-Market, giá trị tối ưu là $\alpha = 0.7$. Tương tự, trong trường hợp Market-to-MSMT, chúng tôi cũng thực hiện điều chỉnh siêu tham số $\alpha$ và nhận thấy rằng $\alpha = 0.5$ mang lại hiệu năng tốt nhất.

**Ảnh hưởng của các trọng số điều khiển.** Để giải thích lý do tại sao giá trị của $w_1$ và $w_2$ trong Công thức 3.16 được đặt là 0.4 và 0.8 như đã đề cập trong Mục 3.3.3, chúng tôi tiến hành thực nghiệm để quan sát hiệu năng của DAPRH khi các trọng số điều khiển thay đổi. Kết quả thử nghiệm trên Duke-to-Market được trình bày trong Hình 3.12. Để đơn giản, chúng tôi đầu tiên cố định $w_2 = 0.0$ và điều chỉnh $w_1$ từ 0.0 đến 0.9. Khi $w_1 = 0$, điều này có nghĩa là mô hình Student không được hướng dẫn bởi mô hình Teacher. Như thể hiện trong Hình 3.12(a), tất cả các kết quả chúng tôi thu được khi thay đổi $w_1$ từ 0.1 đến 0.9 đều tốt hơn kết quả khi $w_1 = 0$. Kết quả tốt nhất trên cả hai chỉ số mAP và Rank-1 đạt được khi đặt $w_1 = 0.4$. Sau đó, chúng tôi tiếp tục thay đổi $w_2$ từ 0.0 đến 0.9. hiệu năng cao nhất chúng tôi đạt được là khi đặt $w_2 = 0.8$, như được minh họa trong Hình 3.12(b). Trong các trường hợp khác, việc thiết lập trọng số điều khiển này cũng mang lại kết quả tương tự.

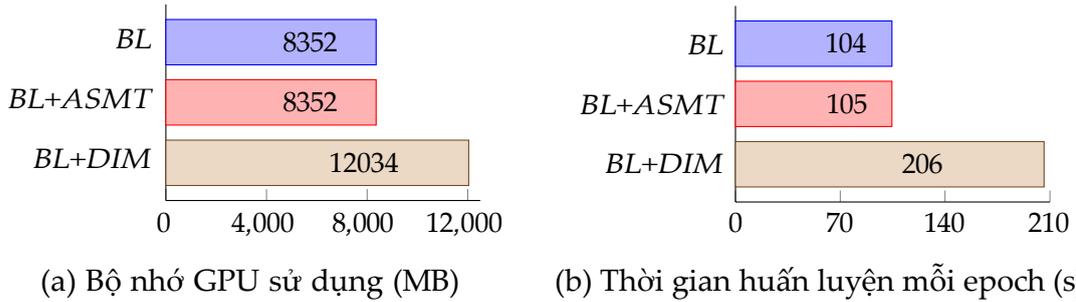

(a) Bộ nhớ GPU sử dụng (MB)    (b) Thời gian huấn luyện mỗi epoch (s)

Hình 3.13: Hiệu suất học không giám sát trên miền nguồn. Trong đó *BL* là mô hình cơ sở; *BL+ASMT* là mô hình cơ sở thêm phương án đánh giá chất lượng ảnh để điều chỉnh trọng số huấn luyện; *BL+DIM* là mô hình cơ sở thêm phương án sử dụng *DIM*.

### 3.4.3.4  Hiệu suất tính toán

Để phân tích thêm tác động thực tiễn của phương pháp đề xuất cũng như giải thích cho RQ9, luận án đánh giá hiệu suất tính toán của mô hình cơ sở với các cải tiến học có giám sát trên miền nguồn và không giám sát trên miền đích bằng cách sử dụng công cụ phân tích tài nguyên của Pytorch để lo lường mức tiêu thụ bộ nhớ GPU tối đa cũng như thời gian huấn luyện mỗi epoch.

**Hiệu suất học có giám sát trên miền nguồn.** Trong trường hợp này, luận án đánh giá hiệu suất tính toán trên mô hình cơ sở (*BL*) tức là chỉ sử dụng hai hàm mất mát phân loại và bộ ba để huấn luyện theo Công thức 3.1. Sau đó, chúng tôi thêm phương án đánh giá chất lượng ảnh để điều chỉnh trọng số huấn luyện (*BL+ASMT*), cũng như phương pháp ánh xạ bất biến miền (*BL+DIM*) nhằm phân tích mức độ ảnh hưởng của từng thành phần đến tài nguyên phần cứng và thời gian huấn luyện. Toàn bộ quá trình đánh giá được thực hiện trên thiết lập Duke-to-Market với kích thước lô là 64, đóng vai trò như là một ví dụ đại diện.

Như minh họa trong Hình 3.13, khi so sánh *BL* với *BL+ASMT*, có thể nhận thấy rằng *BL+ASMT* gần như không tạo thêm chi phí tính toán. Do phép toán đánh giá chất lượng ảnh và điều chỉnh trọng số chỉ bao gồm các phép tính chuẩn hóa đơn giản, mức sử dụng bộ nhớ GPU của hai mô hình là tương đương (8352MB). Bên cạnh đó, thời gian huấn luyện trên mỗi epoch tăng không đáng kể, từ 104 giây lên 105 giây. Trong khi đó, khi tích hợp phương pháp *DIM*, mô hình phải đồng thời trích xuất đặc trưng từ miền đích để thực hiện thích ứng ở mức đặc trưng, dẫn đến mức sử dụng bộ nhớ tăng khoảng 44%, chiếm 12034MB. Thời gian huấn luyện mỗi epoch cũng tăng mạnh, gần như gấp đôi (từ 104 giây lên 206 giây), do cần thêm thời gian cho quá trình trích xuất đặc trưng miền đích cũng như tính toán hàm mất

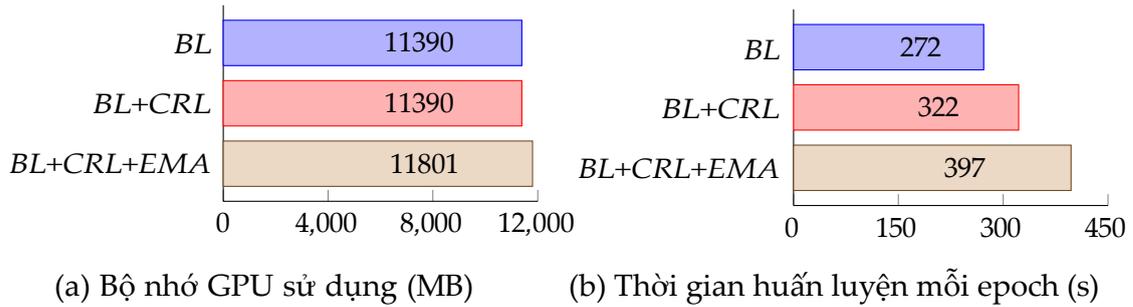

(a) Bộ nhớ GPU sử dụng (MB)  (b) Thời gian huấn luyện mỗi epoch (s)

Hình 3.14: Hiệu suất học không giám sát trên miền đích. Trong đó *BL* là mô hình cơ sở; *BL+CRL* là mô hình cơ sở thêm phương án tinh chỉnh nhãn giản và *BL+CRL+EMA* là tổng thể phương án học không giám sát trên miền đích.

mát *DIM*. Kết quả này cho thấy, mặc dù phương án *BL+DIM* có thể mang lại hiệu quả nhận dạng cao hơn so với *BL+ASMT*, việc áp dụng *DIM* đòi hỏi chi phí phần cứng và thời gian huấn luyện đáng kể hơn.

**Hiệu suất học không giám sát trên miền đích.** Trong trường hợp này, luận án tiến hành đánh giá chi phí tính toán dựa trên mô hình cơ sở được huấn luyện bằng cách áp dụng DBSCAN để phân cụm và gán nhãn giả cho dữ liệu chưa gán nhãn. Sau khi thu được nhãn giả, mô hình cơ sở (*BL*) được tối ưu bằng hai hàm mất mát phân loại và bộ ba. Trên nền tảng này, chúng tôi lần lượt bổ sung kỹ thuật tinh chỉnh nhãn giả (*BL+CRL*) và cơ chế huấn luyện Teacher–Student (*BL+CRL+EMA*) nhằm phân tích mức độ ảnh hưởng của từng thành phần đến chi phí phần cứng và thời gian huấn luyện. Toàn bộ quá trình được thực hiện trên thiết lập Duke-to-Market với kích thước lô bằng 128, đóng vai trò như một ví dụ đại diện.

Như minh họa trong Hình 3.14, khi so sánh *BL* với *BL+CRL*, có thể thấy rằng việc bổ sung *CRL* không làm tăng dung lượng phần cứng, tuy nhiên thời gian huấn luyện mỗi epoch tăng khoảng 18%, từ 272 giây lên 322 giây. Điều này là do quá trình tinh chỉnh nhãn giả làm phát sinh thêm chi phí tính toán. Đối với phương án *BL+CRL+EMA*, mô hình Teacher được bổ sung để điều chỉnh mô hình Student, dẫn đến mức sử dụng bộ nhớ GPU tăng nhẹ, từ 11390MB lên 11801MB. Đồng thời, thời gian huấn luyện mỗi epoch tăng đáng kể khoảng 46%, từ 272 giây lên 397 giây, do phương pháp cần tính toán thêm hàm mất mát điều chỉnh của Teacher và Student cùng với các chi phí phụ trợ liên quan đến mô hình Teacher. Những kết quả này cho thấy mặc dù phương án *BL+CRL+EMA* mang lại hiệu năng cao nhất, nhưng đi kèm với đó là chi phí phần cứng cao hơn và thời gian huấn luyện tăng đáng kể.

**3.4.3.5   So sánh với các phương pháp tiên tiến**

Cuối cùng, luận án so sánh phương án đề xuất DAPRH với các phương pháp UDA tiên tiến khác, như thể hiện trong Bảng 3.7. Điều này giải thích cho RQ10. Cụ thể, kết quả thực nghiệm cho thấy phương pháp đề xuất vượt qua một số phương pháp tiên tiến, đạt được 85.9% mAP và 94.4% độ chính xác Rank-1 trên trường hợp Duke-to-Market, 72.0% mAP và 83.7% Rank-1 trên Market-to-Duke. Đối với hai kịch bản thách thức hơn, phương pháp đề xuất này cũng đạt được kết quả ấn tượng, với 35.8% mAP và 64.8% Rank-1 trên Market-to-MSMT cũng như 36.0% mAP và 65.5% Rank-1 trên Duke-to-MSMT.

So với các phương pháp xuất hiện cách đây vài năm như DIMGLO [65], MMT [30], SpCL [32], DAPRH hoàn toàn vượt trội trên tất cả các bộ dữ liệu. So với các phương pháp UDA nổi tiếng khác như UNRN [125], GLT [126], DAPRH đạt được mAP và độ chính xác Rank-1 cao hơn vài phần trăm trên tất cả các thực nghiệm. Khi so với CMFC [102], DAPRH vượt trội với mức cải thiện 4.9% mAP và 0.8% Rank-1 trong trường hợp Duke-to-Market. DAPRH cũng hoàn toàn vượt qua các giải pháp mới như SECRET [38], MCRN [117], RESL [62], L$F_2$ [26] trong các trường hợp Duke-to-Market, Market-to-MSMT và Duke-to-MSMT về chỉ số mAP.

Tuy nhiên, trong trường hợp Market-to-Duke, DAPRH chỉ đạt kết quả tương đương với phương pháp tiên tiến [62, 117]. Một trong những nguyên nhân chính là quá trình sinh ảnh thực sự hoàn hảo. Cụ thể, các ảnh tăng cường dữ liệu được tạo từ camera của Market sang miền của Duke vẫn còn nhiều hạn chế về chất lượng. Hơn nữa, xét đến các phương pháp đạt kết quả cao nhất và nhì trong thiết lập Market-to-Duke, bao gồm LF$_2$ [26], MCRN [117], AWB [110] và CCTSE [8], có thể thấy rằng các phương pháp này đều sử dụng nhãn giả được tạo thông qua thuật toán phân cụm, đồng thời áp dụng hàm mất mát phân loại và bộ ba làm các hàm mất mát chính để học đặc trưng. Tuy nhiên, các phương pháp này khác nhau ở cách thiết kế trọng số và cách kết hợp các thành phần mất mát đó. Cụ thể, thay vì chỉ đơn giản ghép nối các đặc trưng cục bộ, LF$_2$ áp dụng một mô-đun để tích hợp thông tin đặc trưng hiệu quả hơn. AWB điều chỉnh kiến trúc ResNet-50 bằng cách tích hợp các khối chú ý [110], trong khi MCRN đề xuất mô-đun đa bộ nhớ trọng tâm [117] giúp mô hình thích ứng linh hoạt với các thông tin nhận dạng khác nhau trong từng cụm. Mặt khác, CCTSE giới thiệu cơ chế tự thích ứng, cho phép mô hình điều chỉnh tự động để thích nghi với miền mới hoặc các thay đổi môi trường mà không cần can thiệp thủ công. Trong các thiết lập như Market-to-Duke và Duke-to-MSMT, các phương pháp nói trên thể hiện kết quả cao hơn so với DAPRH. Tuy nhiên, trên các thiết lập khác, DAPRH vẫn giữ được ưu thế nhờ vào sự thiết kế phù hợp.

Bảng 3.7: So sánh các phương pháp tiên tiến trong tái định danh người thích ứng miền không giám sát

| Phương án | Nơi công bố | Duke → Market | | | | Market → Duke | | | | Market → MSMT | | | | Duke → MSMT | | | |
|---|---|---|---|---|---|---|---|---|---|---|---|---|---|---|---|---|---|
| | | mAP | R1 | R5 | R10 | mAP | R1 | R5 | R10 | mAP | R1 | R5 | R10 | mAP | R1 | R5 | R10 |
| DIMGLO[65] | ACM.MM20 | 65.1 | 88.3 | 94.7 | 96.3 | 58.3 | 76.2 | 85.7 | 88.5 | 20.7 | 49.7 | – | 66.1 | 24.4 | 56.5 | – | 70.0 |
| MMT[30] | ICLR20 | 71.2 | 87.7 | 94.9 | 96.9 | 65.1 | 78.0 | 88.8 | 92.5 | 22.9 | 49.2 | 63.1 | 68.8 | 23.3 | 50.1 | 63.9 | 69.8 |
| SpCL[32] | NIPS20 | 76.7 | 90.3 | 96.2 | 97.7 | 68.8 | 82.9 | 90.1 | 92.5 | 26.8 | 53.7 | 65.0 | 69.8 | 26.5 | 53.1 | 65.8 | 70.5 |
| UNRN[125] | AAAI21 | 78.1 | 91.9 | 96.1 | 97.8 | 69.1 | 82.0 | 90.7 | 93.5 | 25.3 | 52.4 | 64.7 | 69.7 | 26.2 | 54.9 | 67.3 | 70.6 |
| CCTSE[8] | IEEE.TIFS21 | 78.4 | 92.9 | 96.9 | 97.8 | <u>72.6</u> | **85.0** | <u>92.1</u> | 93.9 | 33.2 | 62.3 | 74.1 | 78.5 | 34.5 | 63.9 | 75.3 | 79.6 |
| GLT[126] | CVPR21 | 79.5 | 92.2 | 96.5 | 97.8 | 69.2 | 82.0 | 90.2 | 92.8 | 26.5 | 56.6 | 67.5 | 72.0 | 27.7 | 59.5 | 70.1 | 74.2 |
| AWB[110] | IEEE.TIP22 | 81.0 | 93.5 | 97.4 | 98.3 | 70.9 | 83.8 | **92.3** | 94.0 | 29.0 | 57.3 | 70.7 | 75.9 | 29.0 | 57.9 | 71.5 | 76.6 |
| CMFC[102] | ACM.MM22 | 81.0 | <u>94.0</u> | 97.1 | 98.3 | 71.2 | 83.2 | 91.6 | 94.0 | – | – | – | – | – | – | – | – |
| CDCL[100] | KBS23 | 81.5 | 92.8 | <u>97.6</u> | <u>98.7</u> | 70.2 | 82.7 | 91.3 | 93.9 | 29.5 | 59.7 | 71.4 | 75.9 | 30.3 | 61.4 | 74.8 | 77.1 |
| SECRET[38] | AAAI22 | 83.0 | 93.3 | – | – | 69.2 | 82.0 | – | – | 31.7 | 60.0 | – | – | – | – | – | – |
| RESL[62] | AAAI22 | 83.1 | 93.2 | 96.8 | 98.0 | 72.3 | 83.9 | 91.7 | 93.6 | <u>33.6</u> | **64.8** | 74.6 | <u>79.6</u> | 34.2 | 65.2 | 74.6 | 80.1 |
| *LF*₂[26] | ICPR22 | 83.2 | 92.8 | **97.8** | 98.4 | **73.5** | 83.7 | 91.9 | <u>94.3</u> | – | – | – | – | – | – | – | – |
| MCRN[117] | AAAI22 | <u>83.8</u> | 93.8 | 97.5 | 98.5 | 71.5 | <u>84.5</u> | 91.7 | 93.8 | 32.8 | <u>64.4</u> | <u>75.1</u> | 79.2 | <u>35.7</u> | **67.5** | **77.9** | **81.6** |
| **DAPRH** | | **85.9** | **94.4** | **97.8** | **98.8** | 72.0 | 83.7 | <u>92.1</u> | **94.5** | **35.8** | 64.8 | **77.0** | **81.1** | **36.0** | <u>65.5</u> | <u>77.0</u> | <u>80.9</u> |

# 3.5 Kết luận chương

Trong hướng tiếp cận tái định danh người thích ứng miền không giám sát, luận án tập trung vào bốn thách thức chính: (i) sự khác biệt phân phối giữa hai miền dữ liệu, ii) chất lượng ảnh tăng cường dữ liệu (iii) thông tin toàn cục không đủ thông tin và (iiii) nhãn giả không đủ tốt. Do đó, chúng tôi đề xuất hai phương pháp mới, IQAGA và DAPRH, nhằm giải quyết những thách thức này. Phương pháp IQAGA sử dụng StarGAN để tăng cường dữ liệu cũng như giảm sự khác biệt phân phối giữa hai miền dữ liệu và sử dụng kỹ thuật đánh giá chất lượng ảnh để giảm thiểu ảnh hưởng của các ảnh chất lượng xấu lên mô hình. Phương pháp DAPRH ngoài việc sử dụng StarGAN còn tích hợp thêm DIM để thu hẹp khoảng cách phân phối giữa hai miền dữ liệu ở mức đặc trưng. Thứ hai, phương pháp này thêm một nhánh cục bộ trong mô hình và kết hợp GMP và GAP để nắm bắt các đặc trưng quan trọng từ dữ liệu. Cuối cùng, phương án cải thiện nhãn giả được sử dụng dựa trên nhãn mềm thay vì nhãn cứng, cùng với kết hợp huấn luyện trên kiến trúc Teacher-Student giúp nâng cao hiệu năng tái định danh người. Các thực nghiệm toàn diện trên các thiết lập Market-to-Duke, Duke-to-Market, Market-to-MSMT, Duke-to-MSMT đã chứng minh cho hiệu quả của hai phương pháp đề xuất IQAGA và DAPRH trong bài toán định danh người thích ứng miền không giám sát.

Mặc dù các phương pháp đề xuất đã cho kết quả hứa hẹn, nhưng vẫn có một số điểm cần cải thiện. Trước hết, GAN là một công cụ mạnh mẽ trong việc tăng cường dữ liệu, nhưng nó vẫn tạo ra nhiều mẫu chất lượng thấp gây ảnh hưởng tiêu cực đến quá trình huấn luyện của mô hình. Do đó, tối ưu hóa phương pháp GAN để giảm số lượng mẫu chất lượng thấp là một vấn đề cần cải thiện. Bên cạnh đó, hiệu quả của phương pháp tinh chỉnh nhãn giả phụ thuộc phần lớn vào chất lượng phân cụm ban đầu. Khi phân cụm không đủ chính xác, giả định về độ tin cậy của nhãn giả sẽ không còn phù hợp, khiến cho quá trình huấn luyện không giám sát mất ổn định và kém hiệu quả. Điều này đặt ra nhu cầu phát triển các cơ chế tinh chỉnh nhãn giả linh hoạt, thích ứng với mức độ biến thiên của dữ liệu trong quá trình học.

Vì vậy, trong các hướng nghiên cứu tiếp theo, luận án sẽ tập trung vào việc nâng cao chất lượng tăng cường dữ liệu cũng như cải tiến các phương pháp tinh chỉnh nhãn giả, đồng thời tối ưu hóa quy trình huấn luyện tổng thể. Các hướng phát triển tiềm năng có thể bao gồm việc khám phá các kiến trúc mạng mới, đề xuất chiến lược huấn luyện hiệu quả hơn, tích hợp đa dạng kỹ thuật tăng cường dữ liệu, và thiết kế các cơ chế học thích ứng có khả năng ổn định và khái quát hóa tốt hơn trong các môi trường phức tạp.

# Chương 4

# NÂNG CAO HIỆU NĂNG TÁI ĐỊNH DANH NGƯỜI KHÔNG GIÁM SÁT DỰA TRÊN KIẾN TRÚC ViT VÀ THÔNG TIN CAMERA

Trong chương này, luận án trình bày phương pháp nhằm nâng cao hiệu năng tái định danh người trong hướng tiếp cận học không giám sát. Cụ thể, luận án giới thiệu phương pháp "ViTC-UReID", trong đó thay thế mạng cốt lõi CNN bằng Vision Transformer để khai thác đồng thời thông tin đặc trưng toàn cục và cục bộ. Bên cạnh đó, mô hình còn tích hợp thông tin định danh camera nhằm giảm thiểu sự sai lệch miền do khác biệt góc nhìn giữa các thiết bị camera, từ đó nâng cao độ chính xác trong nhận dạng danh tính. Các kết quả nghiên cứu trong chương này được công bố tại công trình [CT3].

## 4.1 Đặt vấn đề

Mặc dù các phương pháp học thích ứng miền không giám sát với dữ liệu từ một miền nguồn có nhãn và một miền đích không có nhãn, đã cho thấy những kết quả tương đối khả quan, tuy nhiên cách tiếp cận này vẫn tồn tại một số hạn chế đáng kể. Thứ nhất, việc phụ thuộc vào dữ liệu từ miền nguồn gây khó khăn trong những trường hợp dữ liệu có nhãn không sẵn có. Thứ hai, dữ liệu miền nguồn không phải lúc nào cũng phù hợp với yêu cầu của thích ứng miền không giám sát, bởi giả định rằng dữ liệu từ hai miền có sự tương đồng về phân phối. Khi sử dụng dữ liệu từ các nguồn khác nhau, sự sai lệch phân phối có thể làm giảm đáng kể hiệu quả mô hình. Thứ ba, quá trình huấn luyện có giám sát trên miền nguồn tiêu tốn thời gian và tài nguyên. Những yếu tố này làm cản trở khả năng chuyển giao và hạn chế hiệu năng của mô hình. Ngược lại, các phương pháp học không giám sát hoàn toàn thể hiện tính linh hoạt và khả năng mở rộng cao hơn, khi chúng có thể được huấn luyện trực tiếp trên tập dữ liệu không nhãn mà không cần phụ thuộc vào miền nguồn bên ngoài, phù hợp hơn với các tình huống triển khai trong thực tế.

Các khung mô hình USL hiện đại thường tích hợp nhiều thành phần mô-đun như thuật toán phân cụm [29], kho bộ nhớ [45, 35], và hàm mất mát tương phản kết [23, 14, 32]. Nhờ hiệu quả vượt trội, cấu trúc này đã trở thành một khuôn mẫu phổ biến trong triển khai huấn luyện USL. Quy trình học của mô hình tái định danh USL thường bao gồm ba bước chính: trích xuất đặc trưng, phân cụm và tạo nhãn giả huấn luyện mô hình với các hàm mất mát phù hợp dựa trên nhãn giả và cập nhật trọng số mô hình phục vụ cho vòng huấn luyện tiếp theo. Qua từng vòng huấn luyện, mạng cốt lõi dần cải thiện khả năng phân biệt đặc trưng một cách hiệu quả và ổn định hơn.

Hiện nay, ResNet-50 đã trở thành kiến trúc nền tảng tiêu chuẩn cho việc trích xuất đặc trưng hình ảnh trong hầu hết các phương pháp tái định danh người nhờ vào hiệu quả tính toán, độ ổn định cao và khả năng thể hiện tốt trên các bộ dữ liệu chuẩn. Tuy nhiên, trong các tình huống thực tế, kiến trúc CNN truyền thống như ResNet-50 bộc lộ một số hạn chế đáng kể. Cụ thể, mô hình này gặp thách thức trong việc mô hình hóa các mối quan hệ không gian dài hạn và sự phụ thuộc giữa các bộ phận cơ thể người, đồng thời thiếu khả năng nắm bắt thông tin ngữ nghĩa và ngữ cảnh cấp cao – những yếu tố đặc biệt quan trọng trong các tình huống nhận diện phức tạp. Ngoài ra, hiệu năng của ResNet-50 có xu hướng suy giảm rõ rệt khi áp dụng trong các điều kiện bất lợi như ánh sáng yếu, đối tượng bị che khuất một phần hoặc sự thay đổi mạnh mẽ về góc nhìn giữa các camera. Những hạn chế này đặt ra nhu cầu nghiên cứu và ứng dụng các kiến trúc hiện đại hơn, tiêu biểu là các mô hình dựa trên Transformer, nhằm nâng cao khả năng biểu diễn ngữ cảnh, tính linh hoạt và hiệu quả tổng quát của hệ thống tái định danh người trong môi trường thực tế. Bên cạnh đó, một thách thức quan trọng khác trong bài toán tái định danh người là hiện tượng khác biệt giữa các camera. Cụ thể, sự khác biệt về góc nhìn, điều kiện ánh sáng, màu sắc và độ phân giải giữa các camera có thể khiến mô hình học sai lệch đặc trưng, ảnh hưởng đến khả năng phân biệt danh tính và làm giảm tính ổn định của hệ thống.

Trước những thách thức đó, chúng tôi đã đề xuất phương pháp "ViTC-UReID" nhằm sử dụng kiến trúc mạng cốt lõi ViT để khai thác hiệu quả thông tin ngữ cảnh toàn cục và chi tiết cục bộ từ hình ảnh. Khác với các kiến trúc CNN truyền thống, ViT tận dụng cơ chế tự chú ý để trích xuất các biểu diễn đặc trưng có ý nghĩa, đảm bảo rằng mọi thông tin quan trọng - bao gồm cả cấu trúc hình học và ngữ nghĩa - đều được khai thác phục vụ cho nhiệm vụ tái định danh người một cách hiệu quả và chính xác hơn. Bên cạnh đó, chúng tôi còn đề xuất tích hợp thông tin định danh camera vào quá trình huấn luyện mô hình. Việc đưa thông tin định danh camera vào mô hình giúp giảm thiểu sự sai lệch miền gây ra bởi sự khác biệt về góc nhìn, điều kiện ánh sáng và độ phân giải giữa các camera. Cụ thể, bằng cách cập nhật và sử dụng các tâm cụm đặc trưng cho từng camera, phương pháp của chúng tôi tăng cường tính nhất quán của đặc trưng trong cùng một nguồn và giữa các nguồn camera khác nhau, từ đó cải thiện khả năng tổng quát hóa và thích ứng trong các môi trường thực tế đa dạng.

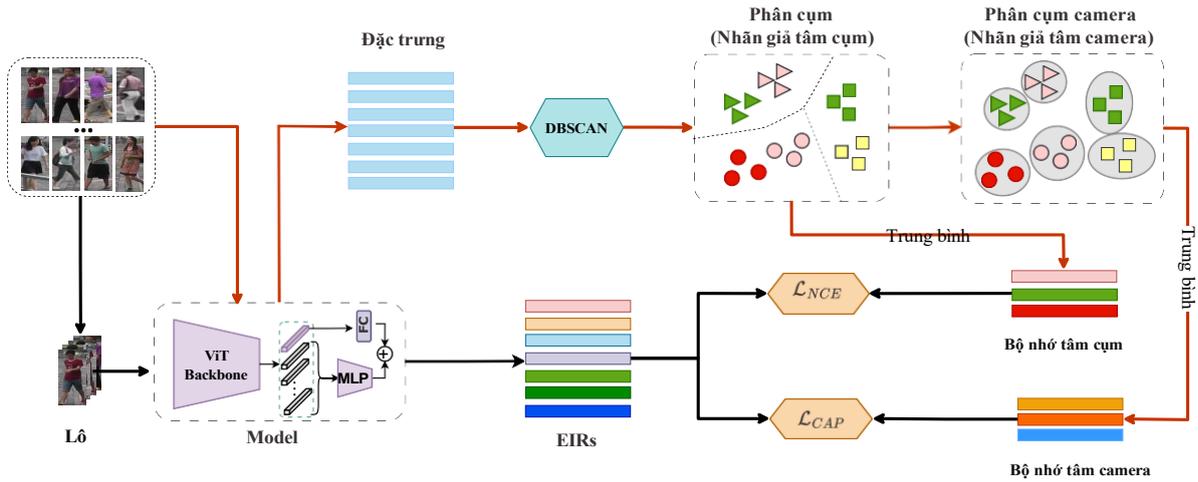

Hình 4.1: Kiến trúc phương pháp đề xuất ViTC-UReID.

Kết quả thực nghiệm trên ba bộ dữ liệu tái định danh phổ biến gồm Market, MSMT và CUHK cho thấy phương pháp đề xuất đạt hiệu năng cạnh tranh so với nhiều kỹ thuật tiên tiến hiện nay, qua đó khẳng định tính hiệu quả và tiềm năng ứng dụng của hướng tiếp cận này trong các hệ thống thực tế.

## 4.2  Phương pháp đề xuất

Phương pháp ViTC-UReID được phát triển dựa trên hai ý tưởng chính: i) tăng cường biểu diễn ảnh trong không gian đặc trưng dựa trên mạng cốt lõi ViT; ii) tích hợp thông tin định danh camera. Phương pháp đề xuất được minh họa trong Hình 4.1. Cụ thể, trong giai đoạn khởi tạo (mũi tên màu cam), tất cả ảnh huấn luyện được đưa qua mô hình tái định danh để trích xuất đặc trưng. Các đặc trưng này sau đó được phân cụm sử dụng DBSCAN để sinh ra các nhãn giả theo cụm. Tiếp theo, thuật toán gom cụm theo thông tin định danh camera được áp dụng để sinh ra nhãn giả theo định danh camera. Các nhãn giả này được sử dụng để khởi tạo hai kho bộ nhớ: kho bộ nhớ tâm cụm và kho bộ nhớ tâm camera thông qua việc trung bình hóa các véc-tơ đặc trưng theo từng cụm. Trong giai đoạn huấn luyện chính (**mũi tên màu đen**), các dữ liệu đã gán nhãn giả được chia thành từng lô và đưa vào mô hình tái định danh để thu được các đặc trưng tăng cường EIR. Các đặc trưng này được so sánh với các tâm cụm trong hai kho bộ nhớ để tính toán hai hàm mất mát tương phản là $L_{NCE}$ và $L_{CAP}$. Hai hàm mất mát này sẽ hướng dẫn mô hình học được các đặc trưng vừa có tính phân biệt cao, vừa bất biến với sự thay đổi giữa các camera, từ đó nâng cao hiệu năng tái định danh người không giám sát. Các thành phần trong phương pháp đề xuất sẽ được trình bày chi tiết trong các mục tiếp theo.

### 4.2.1 Mô hình cơ sở

Với một tập dữ liệu không gán nhãn D, bước đầu tiên là áp dụng thuật toán DBSCAN [29] để gán nhãn giả cho từng mẫu $x_i \in$ D theo Công thức 4.1.

$$C = \text{DBSCAN}(\{\Omega(x_i) \mid x_i \in D\}) \tag{4.1}$$

trong đó, $\Omega(x_i)$ biểu thị đặc trưng hình ảnh được trích xuất tương ứng với mẫu $x_i$, trong khi $C = \{y_i\}_{i=0}^{N}$ là tập các nhãn giả được gán cho $N$ mẫu trong tập dữ liệu D.

Thuật toán DBSCAN phân cụm các đặc trưng hình ảnh đã trích xuất thành hai loại: thuộc cụm (các mẫu được phân cụm) và ngoài cụm (các mẫu không được phân cụm), dựa trên mật độ điểm trong không gian đặc trưng. Các điểm không được phân cụm này sẽ bị loại bỏ trước khi huấn luyện để duy trì độ tin cậy của nhãn. Trong khi đó, các điểm thuộc cụm sẽ được gán nhãn giả phục vụ cho việc tính toán các hàm mất mát và tối ưu hóa mô hình. Cách tiếp cận này cho phép mô hình học được các biểu diễn có ý nghĩa mà không cần đến nhãn thủ công, qua đó đặc biệt phù hợp với các bối cảnh ứng dụng thực tế nơi dữ liệu có nhãn khan hiếm.

Để tối ưu hóa biểu diễn đặc trưng, hầu hết các phương pháp tái định danh người không giám sát gần đây sử dụng học tương phản trong phạm vi từng lô. Tuy nhiên, việc chọn mẫu cùng định danh và khác định danh chỉ trong phạm vi lô hiện tại gây ra nhiều hạn chế. SpCL [32] khắc phục điều này bằng cách sử dụng một bộ nhớ toàn cục lưu trữ các đặc trưng và cập nhật dần trong quá trình huấn luyện. Dẫu vậy, vì chỉ một phần mẫu được cập nhật trong mỗi vòng lặp, nên tiến độ cập nhật không đồng đều, dẫn đến sự lệch trong phân phối đặc trưng. Do đó, Dai và cộng sự [23] giải quyết vấn đề này bằng cách tính các tâm cụm thông qua trung bình đặc trưng của từng cụm theo Công thức 4.2.

$$c_k = \frac{1}{|C_k|} \sum_{f_i \in C_k} f_i \tag{4.2}$$

trong đó, $f_i$ là véc-tơ đặc trưng của mẫu $i$, $c_k$ biểu thị cụm thứ $k$, và $|\cdot|$ biểu thị số lượng véc-tơ đặc trưng $i$ trong mỗi cụm.

Các tâm cụm này sau đó được lưu trữ trong một cấu trúc bộ nhớ gọi là kho bộ nhớ tâm cụm, được sử dụng để hỗ trợ cơ chế cập nhật trong quá trình huấn luyện. Cách tiếp cận này đảm bảo sự cập nhật đặc trưng tâm cụm một cách ổn định và giảm thiểu sự dịch chuyển phân phối, từ đó tạo ra các biểu diễn nhất quán và mạnh mẽ hơn cho bài toán tái định danh người. Trong quá trình huấn luyện, hàm mất mát ClusterNCE [23] được sử dụng để tính toán độ tương đồng giữa đặc trưng truy vấn $q$ và tất cả các tâm cụm. Hàm mất mát này được biểu diễn theo Công thức 4.3.

$$L_{NCE} = -\log \frac{\exp(q \cdot c_+/\tau)}{\sum_{k=1}^{K} \exp(q \cdot c_k/\tau)} \tag{4.3}$$

trong đó, $c_+$ biểu thị tâm cụm cùng định danh với truy vấn $q$, tức là cụm mà $q$ thuộc về, $c_k$ là véc-tơ biểu diễn của cụm thứ $k$, và $\tau$ là siêu tham số điều chỉnh.

Hàm mất mát ClusterNCE điều phối quá trình huấn luyện, nhằm tối ưu sự căn chỉnh đặc trưng và nâng cao hiệu năng của mô hình. Ngoài ra, trong mỗi vòng lặp huấn luyện, các đặc trưng tâm cụm sẽ được cập nhật bằng đặc trưng truy vấn với cơ chế theo Công thức 4.4.

$$c_k \leftarrow m \cdot c_k + (1 - m) \cdot q \tag{4.4}$$

trong đó, $c_k$ là tâm cụm của cụm $k$, $q$ là đặc trưng truy vấn và $m$ là hệ số cập nhật.

Tham số $m$ điều chỉnh mức độ nhất quán giữa đặc trưng tâm cụm và đặc trưng của mẫu truy vấn gần nhất. Khi $m$ tiến dần về 0, tâm cụm được cập nhật sẽ càng tiệm cận với đặc trưng truy vấn mới nhất. Cơ chế này cho phép mô hình cập nhật tâm cụm một cách hiệu quả theo từng lô dữ liệu, giúp các tâm cụm phản ánh chính xác hơn trạng thái phân bố đặc trưng hiện tại. Nhờ đó, mô hình có thể học được các biểu diễn đặc trưng ổn định và thích ứng tốt hơn trong quá trình huấn luyện không giám sát.

## 4.2.2 Tăng cường biểu diễn đặc trưng hình ảnh

Phương pháp đề xuất sử dụng mạng cốt lõi ViT để trích xuất đặc trưng hình ảnh. Cụ thể, với một ảnh đầu vào $I$ có kích thước $R^{H \times W \times C}$, ảnh sẽ được chia thành các mảnh không chồng lắp nhau với số lượng $M = H \times W/S^2$, trong đó $S$ là kích thước của mỗi mảnh. Các mảnh này sau đó được thêm thông tin nhúng tuyến tính và trở thành các tham số có thể học được. Một token đặc biệt [CLS] được thêm vào đầu chuỗi để biểu diễn thông tin toàn ảnh. Tiếp theo, tập các mảnh được ký hiệu là $P = \{p_{cls}, p_1, p_2, ..., p_{M-1}, p_M\}$ được đưa vào các khối Transformer của bộ mã hóa ảnh. Quá trình này tạo ra một chuỗi biểu diễn ảnh có $D$ chiều theo Công thức 4.5.

$$F' = \{f'_{cls}, f'_1, f'_2, ..., f'_{M-1}, f'_M\} = \text{Transformer Block (P)} \tag{4.5}$$

Các biểu diễn này sau đó được đưa vào một lớp kết nối đầy đủ để thu được các đặc trưng chi tiết theo Công thức 4.6.

$$F = \{f_{cls}, f_1, f_2, ..., f_{M-1}, f_M\}$$
$$\text{với } f_i = FC(f'_i) \text{ và } f'_i \in F' \tag{4.6}$$

Đặc trưng toàn cục của toàn bộ ảnh được ký hiệu là $f_{cls}$, trong khi các đặc trưng cục bộ của từng mảnh ảnh được ký hiệu là $\{f_i \mid (i = 1, 2, ..., M)\}$. Trong toàn quá trình huấn luyện và đánh giá, chỉ các đặc trưng toàn cục được sử dụng để tính toán giá trị hàm mất mát và khoảng cách giữa các ảnh nhằm phục vụ truy xuất. Mặc dù các đặc trưng toàn cục cung cấp một cái nhìn tổng quát về nội dung và ngữ cảnh của ảnh, chúng có thể bỏ qua các chi tiết đặc thù từ đặc trưng cục bộ như kết cấu và màu sắc, vốn đóng vai trò quan trọng trong việc nhận dạng chính xác đối tượng [87].

Do đó, chúng tôi đề xuất sử dụng thêm đặc trưng cục bộ quan trọng chứa thông tin hữu ích và bổ sung chúng vào đặc trưng toàn cục, tạo thành đặc trưng toàn thể sử dụng trong quá trình huấn luyện. Ví dụ, từ Công thức 4.5 và 4.6, với một ảnh $I$, ta có thể thu được hai tập đặc trưng $F$ và $F'$, cùng với bản đồ tự chú ý $\mathbf{A} \in \mathbb{R}^{(1+M) \times (1+M)}$ từ tầng cuối cùng của bộ mã hóa.

Điểm tương quan giữa đặc trưng toàn cục và các đặc trưng cục bộ được ký hiệu là $m = \mathbf{A}[0, 1:] \in \mathbb{R}^M$. Sau đó, $\mathbf{K}$ phần trăm các đặc trưng có điểm số cao nhất được chọn từ $M$ biểu diễn cục bộ trong tập đặc trưng $F'$ theo Công thức 4.7.

$$\mathbf{F}^* = \{f'_j \text{ khi } m_j \text{ thuộc về top-K}\} \qquad (4.7)$$

Cuối cùng, đặc trưng hình ảnh được tăng cường thông qua một lớp perceptron nhiều tầng (MLP), bao gồm hai lớp kết nối đầy đủ theo Công thức 4.8.

$$V = \text{MaxPool}(MLP(\|F^*\|)) \oplus f_{cls} \qquad (4.8)$$

trong đó, $\| * \|$ biểu thị chuẩn hóa toán tử và $\oplus$ là phép nối. Chúng tôi sử dụng phép gộp cực đại để làm nổi bật thông tin cục bộ và chuyển chúng thành các véc-tơ có cùng chiều với tập đặc trưng $F$. Dựa trên đó, chúng tôi sẽ thay thế đặc trưng toàn cục ban đầu $f_{cls}$ trong quá trình tính hàm mất mát và truy xuất bằng đặc trưng đã được tăng cường $V$.

## 4.2.3 Kết hợp thông tin định danh camera

Thách thức cốt lõi trong bài toán tái định danh người là làm sao để nhận dạng chính xác cùng một người qua các camera khác nhau. Dựa trên nghiên cứu của Wang và các cộng sự [108], trong phần này, việc học các thông tin định danh camera liên miền được tập trung khai thác, nhằm mục tiêu học được các đặc trưng nhận dạng nhất quán giữa các camera khác nhau.

Cụ thể, cơ chế học theo thông tin định danh camera đã được áp dụng, trong đó mỗi cụm danh tính được xác định thông qua thuật toán DBSCAN sẽ được chia tiếp thành các cụm con theo từng camera, dựa trên định danh camera tương ứng. Mỗi cụm con đại diện cho cùng một danh tính nhưng chỉ đến từ một camera cụ thể. Điều này cho phép xây dựng các nhãn giả đáng tin cậy hơn, vì các mẫu từ cùng một camera thường có độ biến thiên ngoại hình thấp hơn.

Cụ thể, với mỗi tập dữ liệu, tập hợp các camera được ký hiệu là $C = \{c_1, c_2, \ldots, c_n\}$. Một mẫu $x$ thuộc cả cụm $a$ và camera $b$ được kí hiệu là $x_a^b$ với véc-tơ đặc trưng biểu diễn của $x_a^b$ được ký hiệu là $f_a^b$. Khi đó, tâm cụm camera $c_a^b$ được xác định là trung bình của các véc-tơ đặc trưng tương ứng với tất cả các mẫu trong cụm $a$ được chụp từ camera $b$ và được tính bằng Công thức 4.9.

$$c_a^b = \frac{1}{|X_a^b|} \sum_{x \in X_a^b} f(x) \qquad (4.9)$$

trong đó, $X_a^b$ là tập hợp tất cả các mẫu thuộc cụm $a$ được chụp bởi camera $b$, và $f(x)$ là véc-tơ đặc trưng biểu diễn của mẫu $x$. Tương tự như các tâm cụm, các tâm cụm camera này cũng được lưu trữ trong một cấu trúc bộ nhớ, gọi là kho bộ nhớ tâm camera.

Từ đó, hàm mất mát tương phản giữa các camera được định nghĩa trong Công thức 4.10.

$$L_{CAP} = -\frac{1}{|P(i)|} \sum_{j \in P(i)} \log \frac{\exp(q \cdot c_d^j / \tau_c)}{\exp(q \cdot c^j / \tau_c) + \sum_{k \in N(i)} \exp(q \cdot c_k / \tau_c)} \qquad (4.10)$$

trong đó, $P$ và $N$ lần lượt là tập cùng định danh và khác định danh thông tin định danh camera, và $\tau_c$ là siêu tham số điều chỉnh.

## 4.2.4  Hàm mục tiêu huấn luyện

Để tối ưu hóa mô hình, hai hàm mất mát $L_{NCE}$ và $L_{CAP}$ được sử dụng trong quá trình huấn luyện (tương ứng trong Công thức 4.3 và Công thức 4.10). Hàm mục tiêu cuối cùng trong quá trình huấn luyện được biểu diễn trong Công thức 4.11.

$$L_{OBJ} = L_{NCE} + \lambda * L_{CAP} \qquad (4.11)$$

trong đó, $\lambda$ là hệ số dùng để điều chỉnh mức độ ảnh hưởng của thành phần CAP trong quá trình huấn luyện.

Toàn bộ quá trình huấn luyện được minh họa trong Thuật toán 4.1. Cụ thể, đầu vào thuật toán bao gồm tập dữ liệu không nhãn $D$; mô hình $\Theta$ được khởi tạo từ trọng số tiền huấn luyện LUPerson; các siêu tham số $\lambda$, *top-K* và số vòng lặp tối đa max_epoch. Trước hết, mô hình $\Theta$ được sử dụng để trích xuất đặc trưng cho toàn

---

**Thuật toán 4.1** ViTC–UReID

---

**Đầu vào:** Dữ liệu huấn luyện $D = \{I_i\}_{i=1}^M$; mô hình $(\Theta)$ với trọng số của mô hình tiền huấn luyện LUPerson; siêu tham số: $\lambda$, *top-K*; max_epoch.

1: **for each** epoch $\in$ [1, max_epoch] **do**
2:     Trích xuất đặc trưng: $V \leftarrow \Theta(D)$
3:     Sinh nhãn giả theo cụm: $L \leftarrow DBSCAN(V)$
4:     Sinh nhãn giả theo thông tin camera: $P \leftarrow CamCluster(L)$
5:     Khởi tạo kho bộ nhớ tâm cụm: $M \leftarrow Init(V, L)$
6:     Khởi tạo kho bộ nhớ tâm camera: $C \leftarrow Init(V, P)$
7:     **for each** $x = \{(I_i, L_i, P_i)\}_{i=1}^B \in zip(D, L, P)$ **do**
8:         $F_i \leftarrow \Theta(I_i)$
9:         $V_i \leftarrow$ Tăng cường biểu diễn đặc trưng theo Công thức 4.8 từ $F_i$;
10:        L$_1 \leftarrow$ L$_{NCE}(V_i, L_i, M)$;
11:        L$_2 \leftarrow$ L$_{CAP}(V_i, P_i, C)$;
12:        L$_{OBJ}$ = L$_1 + \lambda *$ L$_2$ ;
13:        $\Theta$ = Tối ưu hóa$(\Theta,$ L$_{OBJ})$;
14:     **end for**
15: **end for**

**Đầu ra:** Mô hình tái định danh với tham số tối ưu.

---

bộ tập dữ liệu, từ đó sinh nhãn giả theo cụm $L$ bằng thuật toán DBSCAN và nhãn giả dựa trên thông tin định danh camera $P$. Hai loại nhãn này tiếp tục được dùng để khởi tạo bộ nhớ tâm cụm $M$ và bộ nhớ tâm camera $C$, đóng vai trò như các đại diện hỗ trợ quá trình học đặc trưng. Tiếp theo, thuật toán xây dựng các lô dữ liệu dựa trên ảnh trong $D$, nhãn cụm $L$ và nhãn camera $P$. Đối với mỗi lô, mô hình trích xuất đặc trưng và thực hiện tăng cường biểu diễn; các đặc trưng tăng cường này được sử dụng để tính hai hàm mất mát L$_{NCE}$ và L$_{CAP}$. Hai hàm mất mát này được kết hợp tuyến tính để hình thành hàm mục tiêu tổng thể, dùng để cập nhật tham số mô hình cho đến khi đạt số vòng lặp max_epoch. Kết quả đầu ra của thuật toán là một mô hình tái định danh được tối ưu hoá, trong đó không gian đặc trưng trở nên phân tách và ổn định hơn nhờ việc khai thác đồng thời khả năng tăng cường biểu diễn đặc trưng hình ảnh và thông tin định danh theo camera.

# 4.3 Thực nghiệm và đánh giá

Để chứng minh hiệu quả của phương pháp đề xuất ViTC-UReID, chúng tôi tiến hành các thực nghiệm toàn diện trên ba bộ dữ liệu tiêu biểu là Market, MSMT và CUHK (cụ thể là CUHK(L)) với máy tính trang bị một GPU NVIDIA RTX 4060 Ti với 16GB VRAM để trả lời các câu hỏi nghiên cứu sau.

- RQ1: Việc sử dụng tăng cường biểu diễn đặc trưng hình ảnh có giúp cải thiện hiệu năng tái định danh người không?

- RQ2: Việc sử dụng thông tin định danh camera có giúp cải thiện hiệu năng tái định danh người không?

- RQ3: Việc kết hợp tăng cường biểu diễn đặc trưng hình ảnh và sử dụng thông tin camera có giúp cải thiện hiệu năng tái định danh người không? Khi đó, giá trị tối ưu của tham số $\lambda$ trong Công thức 4.11 là bao nhiêu?

- RQ4: Hiệu suất tính toán của mô hình đề xuất so với mô hình cơ sở như thế nào?

- RQ5: Phương pháp đề xuất có hiệu năng vượt trội hơn các phương pháp tiên tiến trong bài toán tái định danh người không giám sát không?

Các phần tiếp theo sẽ trình bày chi tiết về quá trình huấn luyện mô hình, kết quả thực nghiệm, cũng như phân tích để trả lời các câu hỏi nghiên cứu đã được đặt ra.

## 4.3.1 Chi tiết cài đặt

Ảnh huấn luyện được chuẩn hóa về kích thước 256x128 pixel và được tăng cường dữ liệu theo các kỹ thuật trong Mục 1.2.5. Kiến trúc mô hình dựa trên ViT-B/16 với khoảng 86 triệu tham số làm bộ mã hóa ảnh và được khởi tạo bằng trọng số từ mô hình LUPerson đã huấn luyện trước [44]. Trong suốt quá trình huấn luyện, bộ tối ưu SGD được sử dụng trong 60 epoch với kích thước lô là 128 và tốc độ học ban đầu là $10^{-3}$. Tốc độ học sẽ giảm 10 lần tại các epoch 30 và 50.

## 4.3.2 Thực nghiệm lược bỏ

**Phân tích mô hình cơ sở:** Việc khởi tạo mạng cốt lõi ViT từ các tham số đã được huấn luyện từ tập dữ liệu không gán nhãn quy mô lớn LUPerson [44] đã được chứng minh là một hướng tiếp cận hiệu quả trong các bài toán tái định danh người. Mạng

Bảng 4.1: Nghiên cứu lược bỏ trên bộ dữ liệu Market and MSMT

| Phương pháp | Market | | | | MSMT | | | |
|---|---|---|---|---|---|---|---|---|
| | mAP | R1 | R5 | R10 | mAP | R1 | R5 | R10 |
| *Pretrained* | 11.3 | 34.2 | 49.7 | 57.3 | 4.7 | 20.5 | 29.5 | 34.1 |
| *Cơ sở* | 90.7 | 96.4 | 98.8 | 99.4 | 53.3 | 78.9 | 88.0 | 90.8 |
| *Cơ sở + EIR* | 92.5 | 96.5 | 98.8 | 99.5 | 55.7 | 80.6 | 88.7 | 91.3 |
| *Cơ sở + CAP* | 92.0 | 96.6 | 98.9 | 99.5 | 62.9 | 85.4 | 92.1 | 94.0 |
| *Cơ sở + EIR + CAP* | 92.8 | 97.1 | 99.1 | 99.5 | 63.6 | 85.8 | 92.3 | 94.1 |

cốt lõi ViT được huấn luyện trước đóng vai trò là bộ trích xuất đặc trưng mạnh mẽ, có khả năng nắm bắt các chi tiết thị giác đặc trưng từ ảnh người, từ đó tăng đáng kể độ bền vững và khả năng khái quát của mô hình.

Chiến lược tương tự cũng đã được sử dụng trong các công trình như TransReID-SSL [82] và TMGF [60], cho thấy việc tận dụng các mạng cốt lõi ViT đã được huấn luyện trước là một kỹ thuật được chấp nhận rộng rãi. Tuy nhiên, để đánh giá tác động thực sự của quá trình khởi tạo này, chúng tôi thiết lập một mô hình cơ sở bằng cách đánh giá hiệu năng của mô hình khi chỉ sử dụng trọng số huấn luyện trước từ LUPerson mà chưa tinh chỉnh.

Kết quả ban đầu, như thể hiện trong Bảng 4.1, cho thấy rằng mặc dù mô hình tiền huấn luyện *Pretrained* mang lại một điểm khởi đầu mạnh, nhưng hiệu năng lại kém khi được áp dụng trực tiếp lên các tập dữ liệu theo miền cụ thể, chỉ đạt 34.2% Rank-1 và 11.3% mAP trên Market, và 20.5% Rank-1 và 4.7% mAP trên MSMT. Điều này nhấn mạnh tầm quan trọng của việc tinh chỉnh để mô hình có thể thích nghi với các miền chuyên biệt.

Tiếp theo, chúng tôi tiến hành tinh chỉnh mô hình đã được khởi tạo ban đầu bằng phương pháp cơ bản sử dụng DBSCAN và huấn luyện chỉ với hàm mất mát $L_{NCE}$ như mô tả trong Mục 4.2. Kết quả của mô hình cơ sở được trình bày ở hàng thứ hai của Bảng 4.1, cho thấy vai trò then chốt của việc sử dụng khởi tạo bằng trọng số mô hình đã được huấn luyện.

Sau khi tinh chỉnh, mô hình đạt được những cải thiện hiệu năng đáng kể, với 96.4% Rank-1 và 90.7% mAP trên tập Market, và 78.9% Rank-1 cùng 53.3% mAP trên tập MSMT. Những kết quả này xác nhận rằng trong khi bước tiền huấn luyện cung cấp khả năng trích xuất đặc trưng tổng quát cho mô hình, thì quá trình tinh

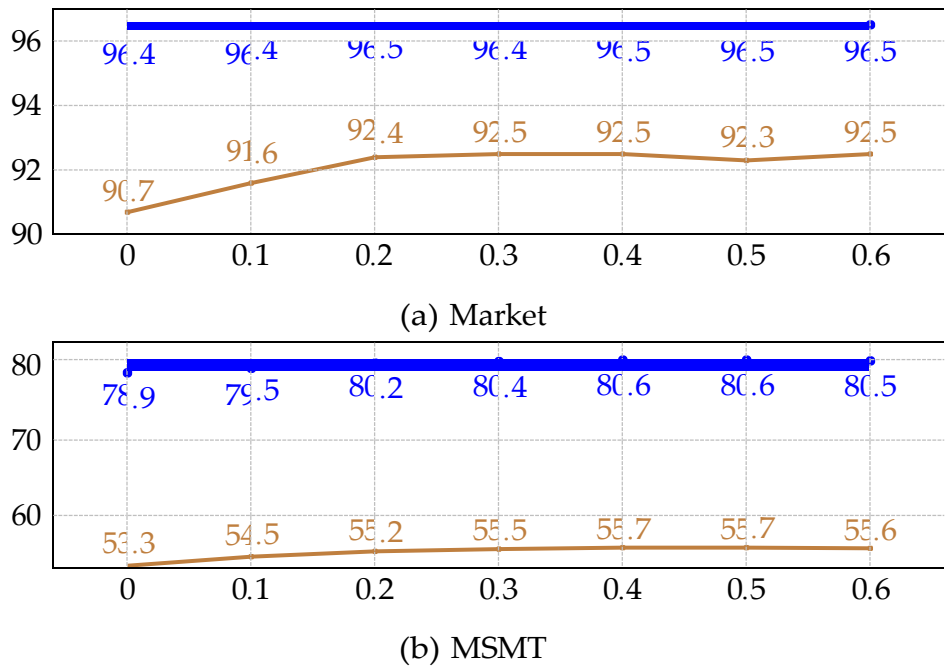

Hình 4.2: Hiệu năng R1 và mAP đối với giá trị top-K.

chính là yếu tố thiết yếu để tối ưu hóa hiệu năng trên từng miền dữ liệu cụ thể. Qua đó, mô hình điều chỉnh lại cách biểu diễn đặc trưng sao cho phù hợp hơn với các đặc điểm riêng biệt của từng bộ dữ liệu, từ đó nâng cao độ chính xác trong truy xuất danh tính.

**Kết quả kết hợp các thành phần:** Trong phần này, các thành phần được đề xuất được đánh giá một cách hệ thống, bắt đầu với mô-đun tăng cường biểu diễn đặc trưng hình ảnh (EIR), tiếp theo là mô-đun tích hợp thông tin định danh camera (CAP), và cuối cùng là đánh giá hiệu quả khi kết hợp đồng thời cả hai mô-đun trong cùng một hệ thống huấn luyện.

Đầu tiên, EIR được thêm vào nhằm làm tăng thêm độ mạnh mẽ của biểu diễn đặc trưng bằng cách nhấn mạnh các chi tiết quan trọng, từ đó nâng cao độ chính xác trong truy vấn. Như thể hiện trong Bảng 4.1, việc tích hợp EIR giúp cải thiện hiệu năng mô hình lên khoảng 2% ở cả hai chỉ số Rank-1 và mAP. Điều này đã giải thích cho RQ1. Đáng chú ý, mức cải thiện lớn nhất được ghi nhận trên tập dữ liệu MSMT. Khác với Market và CUHK, MSMT bao gồm các hình ảnh được chụp trong nhiều điều kiện môi trường khác nhau, khiến các mô hình truyền thống dễ bỏ qua các đặc trưng quan trọng. Sự đa dạng này có thể cản trở việc truy xuất danh tính, do mô hình không thể tập trung vào các thuộc tính có liên quan nhất đến ngoại hình của một cá nhân. Khi tích hợp EIR, những đặc trưng quan trọng nhất của ảnh người sẽ được làm nổi bật trong biểu diễn đặc trưng, bảo đảm sự tích hợp cân bằng giữa thông tin cục bộ và toàn cục. Do đó, sự cải tiến này góp phần đáng kể vào việc

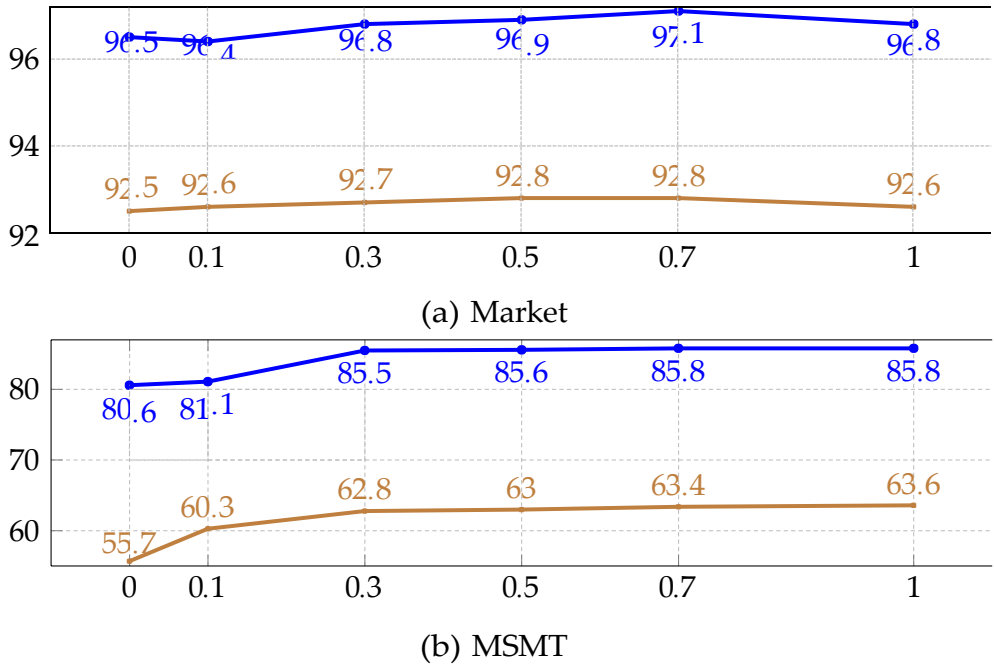

(a) Market

(b) MSMT

Hình 4.3: Hiệu năng R1 và mAP đối với các giá trị $\lambda$.

nâng cao độ chính xác truy vấn trong các tình huống phức tạp.

Ngoài ra, chúng tôi cũng thực nghiệm với top-K phần trăm đặc trưng cục bộ tốt nhất được sử dụng để tạo EIR trong Hình 4.2. Hình minh họa cho thấy EIR đạt độ ổn định khi top-K vượt quá 0.2. Nếu đặt top-K quá nhỏ (top-K < 0.2), mô hình có thể thiếu thông tin cần thiết để tạo ra cải thiện đáng kể. Ngược lại, nếu chọn top-K quá cao (≥ 0.5) sẽ làm tăng độ phức tạp tính toán mà hiệu năng lại không được cải thiện tương xứng. Dựa trên quan sát này, chúng tôi chọn top-K là 0.4 để đạt hiệu năng tối ưu với thời gian huấn luyện hợp lý.

Để giải thích cho RQ2, mô hình được tinh chỉnh bằng cách loại bỏ mô-đun EIR và bổ sung hàm mất mát CAP. Kết quả thu được được trình bày trong hàng thứ tư của Bảng 4.1. Giống như EIR, CAP mang lại một mức cải thiện nhẹ trên tập dữ liệu Market. Tuy nhiên, nó giúp cải thiện đáng kể hiệu năng trên tập MSMT. Quan sát này cho thấy tầm quan trọng của thông tin định danh camera như một yếu tố then chốt giúp mô hình học được các biểu diễn có tính phân biệt cao hơn. Thông qua việc tích hợp CAP, mô hình có khả năng thích nghi tốt hơn với sự biến đổi về góc nhìn giữa các camera, từ đó nâng cao khả năng phân biệt hiệu quả giữa các định danh.

Cuối cùng, cả EIR và CAP được tích hợp vào quá trình huấn luyện. So với mô hình cơ sở, ViTC-UReID thể hiện mức cải thiện hiệu năng rõ rệt. Cụ thể, ViTC-UReID đạt được 97.1% Rank-1 và 92.8% mAP trên tập dữ liệu Market, trong khi trên tập MSMT, mô hình đạt 85.8% Rank-1 và 63.6% mAP. Để đạt được các kết quả này, giá trị $\lambda$ đã được điều chỉnh nhằm đánh giá mức ảnh hưởng của thành phần CAP. Như được minh họa trong Hình 4.3, các giá trị tốt nhất lần lượt là 0.7 và 1.0, mang lại hiệu năng tối ưu và ổn định nhất trên các tập dữ liệu Market và MSMT. Mặt khác, các kết quả này cũng xác nhận hiệu quả của việc kết hợp EIR và CAP, làm nổi bật vai trò bổ trợ lẫn nhau của chúng trong việc cải thiện quá trình học biểu diễn và độ chính xác truy xuất trên các tập dữ liệu đa dạng. Điều này giải thích cho RQ3.

### 4.3.3 Hiệu suất tính toán

Để đánh giá thêm tác động thực tiễn của phương pháp đề xuất cũng như giải thích cho RQ4, luận án tiến hành so sánh hiệu suất tính toán của ViTC-UReID với phương pháp cơ sở trên tập dữ liệu Market, sử dụng kích thước lô là 128 như một trường hợp đại diện. Mặc dù việc tích hợp hai thành phần EIR và CAP giúp cải thiện hiệu năng tổng thể trong bài toán tái định danh người, nhưng đồng thời cũng làm gia tăng chi phí tính toán trong quá trình huấn luyện.

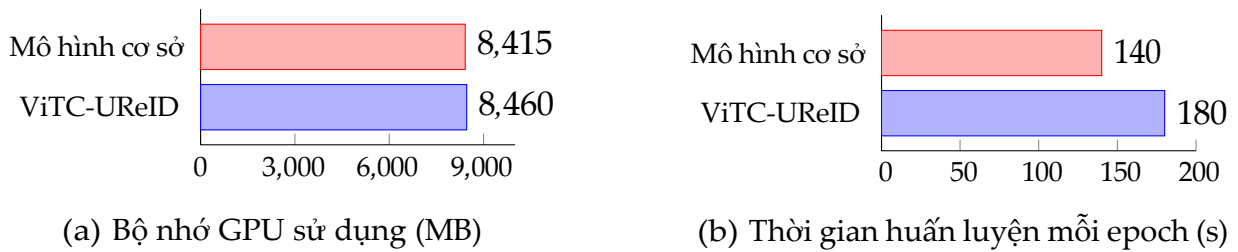

(a) Bộ nhớ GPU sử dụng (MB)      (b) Thời gian huấn luyện mỗi epoch (s)

Hình 4.4: So sánh hiệu suất tính toán trên bộ dữ liệu Market.

Sử dụng công cụ theo dõi tài nguyên của PyTorch, chúng tôi đo lường dung lượng bộ nhớ GPU được sử dụng và thời gian huấn luyện cho mỗi epoch. Như thể hiện trong Hình 4.4, mức sử dụng bộ nhớ GPU giữa hai phương pháp gần như tương đương, với ViTC-UReID tiêu thụ 8460 MB – chỉ cao hơn một chút so với 8415 MB của phương pháp cơ sở. Về thời gian huấn luyện, ViTC-UReID cần 180 giây cho mỗi epoch, so với 140 giây của mô hình cơ sở, tương ứng với mức tăng 28,5%. Chi phí tăng thêm này chủ yếu đến từ việc tích hợp cơ chế tăng cường đặc trưng cục bộ trong bộ mã hóa ViT và thông tin camera.

Mặc dù có phát sinh chi phí huấn luyện, sự cải thiện rõ rệt về mAP và độ chính xác Rank-n đã cho thấy sự đánh đổi này là hoàn toàn hợp lý. Hơn nữa, độ ổn định cao hơn trong huấn luyện cùng với khả năng học biểu diễn tốt hơn của ViTC-UReID đóng vai trò then chốt trong việc đảm bảo độ chính xác và khả năng khái quát hóa trong các tình huống triển khai thực tế và môi trường có nhiều camera khác nhau.

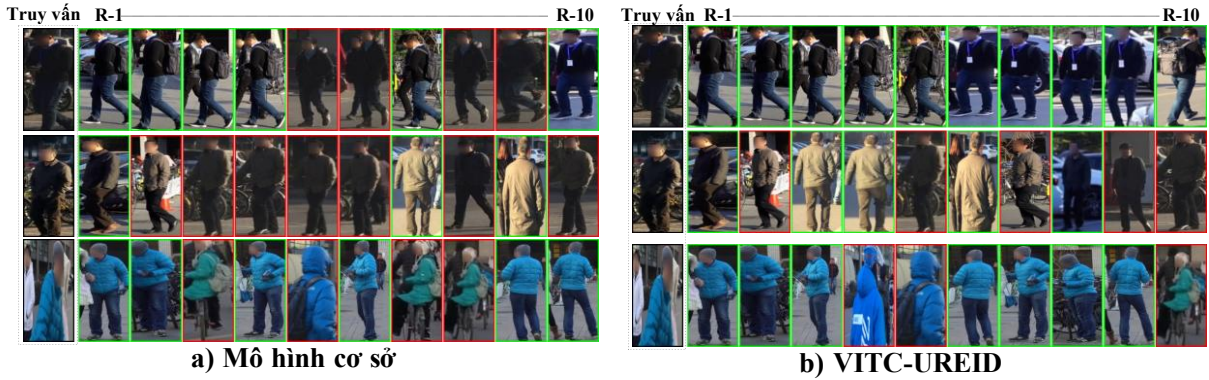

**a) Mô hình cơ sở**   **b) VITC-UREID**

Hình 4.5: So sánh 10 kết quả đầu tiên của truy xuất trên tập dữ liệu MSMT.

### 4.3.4    Phân tích trực quan

Để minh họa trực quan hơn cho hiệu quả của ViTC-UReID, chúng tôi thực hiện phân tích trực quan trên bộ dữ liệu đại diện MSMT. Các kết quả truy xuất tiêu biểu được trình bày trong Hình 4.5 nhằm đánh giá định tính hiệu quả của phương pháp được đề xuất. Cụ thể, trong Hình 4.5, 10 ảnh đầu tiên của quá trình truy xuất đã được thể hiện để so sánh giữa mô hình cơ sở và ViTC-UReID. Ở phía bên trái, các kết quả truy xuất từ mô hình cơ sở được hiển thị, trong khi phía bên phải là các ảnh được truy xuất bởi ViTC-UReID. Các ảnh so khớp chính xác với danh tính truy vấn được đánh dấu bằng khung màu xanh lá cây. Trong phần lớn các trường hợp, các ảnh của đối tượng truy vấn đã được truy xuất chính xác hơn bởi ViTC-UReID, cho thấy hiệu quả cải thiện rõ rệt về khả năng xếp hạng so với mô hình cơ sở. Đặc biệt, các ảnh so khớp đúng thường được xếp ở vị trí cao hơn trong danh sách truy xuất, từ đó khẳng định được tính ưu việt của phương pháp được đề xuất trong việc nâng cao hiệu năng trong bài toán tái định danh người không giám sát.

### 4.3.5    So sánh với các phương pháp tiên tiến

Kết quả thực nghiệm trong Bảng 4.2 cho thấy hiệu năng vượt trội của phương pháp đề xuất khi so sánh với các phương pháp tiên tiến và đã giải thích cho RQ5.

**So sánh giữa các phương pháp không giám sát dựa trên CNN và ViT.**

Trên tập dữ liệu Market, các phương pháp không giám sát dựa trên kiến trúc CNN (bao gồm ICE [14], CC [23], PPLR [20] và ISE [123]) thường đạt độ chính xác Rank-1 trong khoảng 92% đến 95%, với giá trị mAP dao động từ 83.0% đến 85.3%. Ngược lại, các phương pháp không giám sát dựa trên ViT như TransReID [37], PASS [136], TMGF [60], ACFL-VIT [50], TCMM [137], và PCL-CLIP [59] thường đạt độ chính xác Rank-1 cao hơn một chút (khoảng 94.8% đến 96.0%) và cải thiện đáng kể chỉ số mAP (từ 88.4% đến 90.5%).

Trên tập dữ liệu MSMT, xu hướng này còn rõ rệt hơn: trong khi các phương pháp dựa trên CNN chỉ đạt giá trị mAP ở mức khá khiêm tốn (ví dụ, PPLR chỉ đạt 42.2%), thì nhiều phương pháp dựa trên ViT đã nâng hiệu năng lên mức cao đáng kể. Những kết quả này cho thấy rằng biểu diễn đặc trưng dựa trên ViT có khả năng nắm bắt tốt hơn thông tin toàn cục và ngữ cảnh, từ đó mang lại lợi thế rõ rệt về hiệu năng trong các bài toán tái định danh người không giám sát.

Bảng 4.2: So sánh với các phương pháp tiên tiến. † sử dụng thông tin camera

| Mạng cốt lõi | Phương pháp | Nơi công bố | Market | | | | MSMT | | | | CUHK | | | |
|---|---|---|---|---|---|---|---|---|---|---|---|---|---|---|
| | | | mAP | R1 | R5 | R10 | mAP | R1 | R5 | R10 | mAP | R1 | R5 | R10 |
| *Phương pháp có giám sát* | | | | | | | | | | | | | | |
| CNN | ABNET+ NFormer[107] | CVPR22 | 93.0 | 95.7 | – | – | 62.2 | 80.8 | – | – | 79.1 | 80.6 | – | – |
| | ProNet++[109] | CoRR23 | 90.2 | 96.0 | – | – | 65.5 | 85.4 | – | – | 82.7 | 85.2 | – | – |
| VIT | TransReID[37] | ICCV21 | 89.5 | 95.2 | – | – | 69.4 | 86.2 | – | – | – | – | – | – |
| | CLIP-ReID | AAAI23 | 89.6 | 95.5 | – | – | 73.4 | 88.7 | – | – | – | – | – | – |
| | TMGF†[60] | WACV23 | 91.9 | 96.3 | 98.9 | 99.3 | 70.3 | 88.2 | 94.1 | 95.4 | – | – | – | – |
| *Phương pháp không giám sát* | | | | | | | | | | | | | | |
| CNN | CC[23] | ACCV22 | 83.0 | 92.9 | 97.2 | 98.0 | 33.0 | 62.0 | 71.8 | 76.7 | – | – | – | – |
| | PPLR†[20] | CVPR22 | 84.4 | 94.3 | 97.8 | 98.6 | 42.2 | 73.3 | 83.5 | 86.5 | – | – | – | – |
| | ISE[123] | CVPR22 | 85.3 | 94.3 | 98.0 | 98.8 | 37.0 | 67.6 | 77.5 | 81.0 | – | – | – | – |
| VIT | PASS[136] | ECCV22 | 88.5 | 94.9 | – | – | 41.0 | 67.0 | – | – | – | – | – | – |
| | TMGF†[60] | WACV23 | 89.5 | 95.5 | <u>98.0</u> | <u>98.7</u> | 58.2 | 83.3 | 90.2 | 92.1 | – | – | – | – |
| | PCL-CLIP†[59] | Arxiv23 | 88.4 | 94.8 | <u>98.0</u> | <u>98.7</u> | **65.5** | <u>84.9</u> | **92.0** | <u>94.0</u> | – | – | – | – |
| | ACFL-VIT[50] | PR24 | 89.1 | 95.1 | – | – | 45.7 | 70.1 | – | – | – | – | – | – |
| | TCMM[137] | Arxiv25 | <u>90.5</u> | <u>96.0</u> | – | – | 52.0 | 78.4 | – | – | – | – | – | – |
| VIT | ViTC-UReID | Our | **92.8** | **97.1** | **99.1** | **99.3** | <u>63.6</u> | **85.8** | **92.3** | **94.1** | **89.8** | **91.1** | **95.1** | **97.1** |

**So sánh các phương pháp không giám sát tích hợp thông tin camera:** Một số phương pháp không giám sát đã tích hợp thông tin camera nhằm xử lý sự không nhất quán giữa các góc nhìn camera, được ký hiệu bằng dấu thánh giá (†). Trong nhóm các phương pháp dựa trên CNN, PPLR thể hiện hiệu năng cạnh tranh với độ chính xác Rank-1 đạt 92.8% và mAP đạt 84.4% trên tập dữ liệu Market.

Với các phương pháp dựa trên ViT, có hai phương pháp - TMGF và PCL-CLIP - khai thác thông tin định danh camera. Trong đó, TMGF đạt Rank-1 là 95.5% và mAP là 89.5%, còn PCL-CLIP đạt kết quả cao hơn với Rank-1 là 94.8% và mAP là 88.4% trên bộ dữ liệu Market.

Những thông tin này xác nhận rằng việc tích hợp thông tin định danh camera trong các khung học không giám sát giúp cải thiện khả năng phân biệt đặc trưng giữa các góc nhìn camera khác nhau. Điều này đặc biệt có ích khi xử lý các tập dữ liệu có số lượng camera lớn như MSMT.

**So sánh với các phương pháp huấn luyện hoàn toàn có giám sát:** Khi so sánh với các phương pháp không giám sát, các phương pháp có giám sát hoàn toàn vẫn giữ được lợi thế trên nhiều bộ đánh giá. Trong nhóm các phương pháp dựa trên CNN, các mô hình như ABNET+NFormer đạt độ chính xác Rank-1 là 95.7% với mAP ấn tượng 93.0% trên tập Market.

Tương tự, trong nhóm các phương pháp có giám sát dựa trên ViT, TMGF (với dữ liệu có nhãn) và TransReID duy trì hiệu năng cao, với TMGF đạt Rank-1 là 96.3% và mAP là 91.9% trên Market. Mặc dù các phương pháp có giám sát thường cho kết quả vượt trội nhờ vào dữ liệu được gán nhãn, nhưng khoảng cách hiệu năng đang dần thu hẹp, đặc biệt là khi các kỹ thuật không giám sát tiếp tục phát triển.

Đối với tập dữ liệu CUHK, với mô hình sử dụng mạng cốt lõi ViT, phương pháp đề xuất vượt trội hơn nhiều so với các phương pháp dựa trên CNN, dù chỉ được huấn luyện theo cách không giám sát.

## 4.4 Kết luận chương

Trong chương này, luận án đề xuất phương pháp ViTC-UReID sử dụng ViT nhằm khai thác cơ chế tự chú ý trong việc đồng thời thu nhận thông tin ngữ cảnh toàn cục và chi tiết cục bộ. Bên cạnh đó, thông tin định danh camera được tích hợp vào quá trình huấn luyện, qua đó giảm thiểu sự sai lệch gây ra bởi sự khác biệt về góc nhìn, độ phân giải, và điều kiện chiếu sáng giữa các camera. Phương pháp đề xuất được kiểm chứng qua các bộ dữ liệu đánh giá phổ biến như Market, MSMT, và CUHK cho thấy khả năng vượt trội so với nhiều phương pháp USL tiên tiến - đặc biệt là các phương pháp có tích hợp thông tin định danh camera - và dần tiệm cận hiệu năng của các phương pháp có giám sát. Mặc dù đạt được hiệu quả ấn tượng, phương pháp đề xuất vẫn tồn tại một số hạn chế nhất định. Cụ thể, quá trình sinh nhãn giả chưa thực sự đáng tin cậy trong các môi trường nhiễu hoặc có chất lượng hình ảnh thấp. Bên cạnh đó, kiến trúc Vision Transformer đòi hỏi tài nguyên tính toán lớn, gây khó khăn trong việc triển khai trên các hệ thống bị giới hạn phần cứng.

Trong tương lai, luận án định hướng phát triển theo ba hướng chính: cải thiện độ chính xác của nhãn giả thông qua các chiến lược gán nhãn ổn định hơn; nghiên cứu các kiến trúc Transformer khác phù hợp với môi trường triển khai thực tế; và mở rộng chiến lược học đặc trưng dựa trên thông tin camera để thích ứng tốt hơn với các biến động môi trường. Những định hướng này kỳ vọng sẽ tiếp tục thu hẹp khoảng cách giữa các phương pháp tái định danh người không giám sát và có giám sát, từ đó mở rộng khả năng ứng dụng của tái định danh trong các hệ thống giám sát thông minh.

# KẾT LUẬN VÀ HƯỚNG PHÁT TRIỂN

## Đóng góp chính

Trong bối cảnh đô thị thông minh và sự phổ biến của hệ thống camera giám sát, nhu cầu giám sát và nhận diện danh tính trong không gian công cộng ngày càng cấp thiết. Một trong những hướng tiếp cận quan trọng phục vụ mục tiêu này là bài toán tái định danh người, nhằm xác định và nhận diện lại danh tính cụ thể của một người qua tập hình ảnh thu được từ nhiều camera. Khác với nhận dạng khuôn mặt, tái định danh người khai thác các đặc trưng toàn thân như màu trang phục, hình dáng hoặc phụ kiện. Tuy nhiên, bài toán vẫn đối mặt với nhiều thách thức do sự biến đổi ngoại hình dưới ảnh hưởng của góc nhìn, vật cản, ánh sáng và môi trường. Nhằm giải quyết những vấn đề này, luận án đề xuất ba đóng góp chính, hướng tới mục tiêu tổng thể là nâng cao hiệu năng tái định danh người trong các điều kiện thực tế khác nhau.

- Đóng góp 1: Trong hướng tiếp cận học có giám sát, luận án đề xuất phương pháp "SCM-ReID" nhằm nâng cao hiệu năng tái định danh người thông qua việc kết hợp hàm mất mát tương phản có giám sát với bốn hàm mất mát khác: phân loại, trung tâm, bộ ba và bộ ba trọng tâm. Phương pháp này tận dụng ưu điểm của hàm mất mát tương phản có giám sát trong việc so sánh mỗi mẫu với đồng thời nhiều mẫu cùng và khác danh tính, giúp mô hình học được biểu diễn đặc trưng phân biệt cao và khả năng khái quát tốt hơn so với các hàm mất mát truyền thống. Hơn nữa, việc đồng thời sử dụng năm hàm mất mát cho phép mô hình khai thác các ưu điểm bổ sung của từng hàm, từ đó xây dựng được không gian đặc trưng có khả năng phân biệt mạnh giữa các danh tính.

- Đóng góp 2: Trong hướng tiếp cận học thích ứng miền không giám sát, luận án đề xuất phương pháp "IQAGA" nhằm thu hẹp khoảng cách phân phối giữa miền nguồn và miền đích bằng cách sử dụng GAN, đồng thời tích hợp thuật toán đánh giá chất lượng ảnh với điều chỉnh trọng số nhằm giảm thiểu tác động từ các ảnh chất lượng thấp trong quá trình huấn luyện có giám sát trên miền nguồn. Tiếp theo, luận án đề xuất phương pháp "DAPRH", không chỉ sử dụng GAN mà còn tích hợp DIM để thu hẹp sự khác biệt giữa hai miền ở mức đặc trưng. Ngoài ra, phương pháp còn tập trung vào việc học không giám sát trên miền đích bằng cách khai thác kết hợp các đặc trưng toàn thể, cùng với kỹ thuật tinh chỉnh nhãn giả dựa trên khoảng cách đến tâm cụm, từ đó nâng cao chất lượng biểu diễn và tăng cường khả năng phân biệt của mô hình.

- Đóng góp 3: Trong hướng tiếp cận học không giám sát, luận án đề xuất phương pháp "ViTC-UReID" sử dụng mạng cốt lõi Vision Transformer kết hợp với đặc trưng cục bộ để tạo ra đặc trưng toàn thể, đồng thời tích hợp thông tin định danh camera nhằm học các đặc trưng phân biệt theo từng camera. Việc sử dụng đặc trưng toàn thể giúp mô hình khai thác hiệu quả các mối quan hệ dài hạn trong không gian hình ảnh, từ đó học được các biểu diễn có khả năng khái quát tốt hơn trong điều kiện không đồng nhất giữa các ảnh quan sát. Bên cạnh đó, việc tích hợp thông tin định danh camera cho phép mô hình nhận biết và điều chỉnh theo sự khác biệt giữa các camera. Nhờ đó, ViTC-UReID không chỉ cải thiện độ chính xác trong việc truy xuất danh tính mà còn tăng tính ổn định khi triển khai trong các hệ thống đa camera thực tế.

Từ những kết quả đạt được qua ba đóng góp, có thể khẳng định rằng luận án đã hoàn thành mục tiêu đề ra ban đầu là đề xuất phương pháp mới để nâng cao hiệu năng tái định danh người theo cả ba hướng tiếp cận học có giám sát, thích ứng miền không giám sát và không giám sát.

# Bàn luận

Kết quả thực nghiệm trong các chương phản ánh hiệu quả của ba hướng tiếp cận chính được nghiên cứu và đề xuất trong luận án, bao gồm: học có giám sát (SCM-ReID), học thích ứng miền không giám sát (IQAGA, DAPRH), và học không giám sát (ViTC-UReID). Mỗi hướng tiếp cận được thiết kế nhằm giải quyết các bài toán thực tiễn khác nhau.

Đầu tiên, trong hướng học có giám sát, phương pháp SCM-ReID đạt hiệu năng cao nhờ tận dụng được ưu điểm của hàm mất mát tương phản có giám sát kết hợp với các hàm mất mát phân loại, trung tâm, bộ ba, bộ ba trọng tâm. Cụ thể, kết quả đánh giá trên bộ dữ liệu Market, mô hình đạt mAP 98.84% và Rank-1 98.66%; trên bộ dữ liệu CUHK03(L) và CUHK03(D), mAP đạt lần lượt là 96.92% và 96.53%. Tuy nhiên, do phụ thuộc hoàn toàn vào dữ liệu gán nhãn, phương pháp này khó triển khai trong các môi trường mới hoặc dữ liệu không có nhãn.

Tiếp theo, trong hướng học thích ứng miền không giám sát, mô hình được huấn luyện trên dữ liệu miền nguồn có nhãn và đánh giá trên miền đích không nhãn. Để thu hẹp khoảng cách giữa hai miền, các kỹ thuật thích ứng miền ở mức hình ảnh và mức đặc trưng thường được sử dụng. Tuy nhiên, khi mô hình bị tối ưu hóa quá mức cho miền nguồn, hiện tượng quá khớp có thể làm suy giảm khả năng tổng quát và dẫn đến hiệu năng không ổn định. Điều này nhấn mạnh tầm quan trọng của việc thiết kế chiến lược huấn luyện hợp lý, thay vì tối ưu hóa cực đoan cho miền nguồn mà bỏ qua tính đáp ứng ở miền đích.

Với nhận định trên, luận án đề xuất phương pháp IQAGA chỉ sử dụng thêm hai kỹ thuật là thu hẹp khoảng cách giữa hai miền ở mức hình ảnh bằng GAN và đánh giá chất lượng hình ảnh trong quá trình huấn luyện. Tuy nhiên, kết quả đạt được vẫn còn tương đối hạn chế với mAP 36.3% và Rank-1 70.2% đối với thiết lập Duke-to-Market; mAP 31.2% và Rank-1 55.5% đối với thiết lập Market-to-Duke. Điều này cho thấy việc chỉ tối ưu mô hình trên miền nguồn là không đủ để đảm bảo hiệu năng cao khi chênh lệch miền lớn.

Để cải thiện điều này, luận án đề xuất phương pháp DAPRH theo hướng kết hợp giữa học có giám sát trên miền nguồn và học không giám sát trên miền đích thông qua các kỹ thuật: kết hợp đặc trưng, phân cụm và tinh chỉnh nhãn giả. Kết quả cho thấy, DAPRH đạt mAP 85.9% và Rank-1 94.4% trong thiết lập Duke-to-Market, đồng thời duy trì hiệu năng cao trên các thiết lập khó như Market-to-MSMT và Duke-to-MSMT. Đây là minh chứng cho thấy việc huấn luyện với dữ liệu không nhãn tại miền đích mang lại hiệu quả. Tuy nhiên, khoảng cách miền vẫn là yếu tố giới hạn hiệu năng khi so với học có giám sát.

Theo xu hướng hiện nay, hướng học không giám sát hoàn toàn được xem là tiềm năng cho các hệ thống triển khai thực tế, nhờ loại bỏ hoàn toàn yêu cầu về nhãn dữ liệu. Do đó, luận án đề xuất phương pháp ViTC-UReID theo hướng tiếp cận này. Trên tập dữ liệu Market, mô hình đạt mAP 92.8% và Rank-1 97.1%; trên MSMT, đạt mAP 63.6% và Rank-1 85.8%; và trên CUHK, đạt mAP 89.8% và Rank-1 91.1%. ViTC-UReID tận dụng sức mạnh biểu diễn của kiến trúc Vision Transformer, thông tin đặc trưng cục bộ, và kết hợp với thông tin camera. Nhờ đó, mô hình không chỉ vượt trội hơn các phương pháp không giám sát sử dụng CNN mà còn tiệm cận hiệu năng của nhiều mô hình có giám sát. Điều này cho thấy tiềm năng của hướng học không giám sát khi được kết hợp với các kiến trúc mạng cốt lõi mạnh và thông tin phụ trợ phù hợp.

Trong bối cảnh triển khai thực tế, ba hướng tiếp cận chính trong tái định danh người đều có những ưu điểm và hạn chế nhất định. Học có giám sát thường đạt độ chính xác cao và ổn định nhờ được huấn luyện trên dữ liệu gán nhãn, nên đặc biệt phù hợp với các hệ thống giám sát trong môi trường được kiểm soát tốt như sân bay hoặc tòa nhà, nơi có thể thu thập và gán nhãn trước dữ liệu. Tuy nhiên, chi phí gán nhãn lớn và sự suy giảm hiệu năng khi có khác biệt miền dữ liệu lại là rào cản cho việc mở rộng. Thích ứng miền không giám sát giải quyết phần nào hạn chế này khi tận dụng dữ liệu gán nhãn từ miền nguồn và dữ liệu không gán nhãn từ miền đích, giúp giảm chi phí gán nhãn và tăng tính khái quát trong môi trường triển khai mới.

Dẫu vậy, hiệu quả của UDA phụ thuộc mạnh vào dữ liệu miền nguồn và quá trình thích ứng có thể làm tăng độ phức tạp tính toán, gây khó khăn cho ứng dụng thời gian thực. Trong khi đó, học không giám sát hứa hẹn khả năng tận dụng trực tiếp dữ liệu thu thập mà không cần gán nhãn, nhờ đó thích hợp với các hệ thống quy mô lớn, liên tục sinh dữ liệu mới. Tuy nhiên, hạn chế về độ chính xác, sự nhạy cảm với nhiễu và yêu cầu kỹ thuật xử lý phức tạp như phân cụm và gán nhãn giả không ổn định khiến hướng tiếp cận này khó đáp ứng ngay các ứng dụng đòi hỏi độ tin cậy cao. Như vậy, mỗi hướng tiếp cận đều mang lại giá trị riêng cho triển khai thực tế và việc lựa chọn phù hợp phụ thuộc vào yêu cầu cụ của từng vấn đề.

# Hạn chế

Mặc dù các phương pháp đề xuất đã cải thiện hiệu năng tái định danh người theo từng hướng tiếp cận, nhưng vẫn tồn tại những hạn chế nhất định.

- Hạn chế 1: Theo hướng tiếp cận có giám sát, việc tích hợp hàm mất mát tương phản có giám sát làm tăng độ nhạy kết quả với kích thước lô, đồng thời yêu cầu cao hơn về bộ nhớ và chi phí tính toán so với mô hình cơ sở. Bên cạnh đó, phương pháp đề xuất vẫn chưa xử lý hiệu quả các tình huống khó như che khuất nghiêm trọng, thay đổi góc nhìn lớn, hoặc các trường hợp có ngoại hình tương đồng cao giữa các danh tính. Ngoài ra, mô hình còn phụ thuộc vào mạng cốt lõi ResNet-50 và cách kết hợp hàm mất mát được thiết lập cố định, hạn chế khả năng thích ứng và mở rộng.

- Hạn chế 2: Theo hướng tiếp cận thích ứng miền không giám sát, phương pháp đề xuất sử dụng GAN để tăng cường dữ liệu bổ sung cho bộ dữ liệu thực. Tuy nhiên, GAN có xu hướng sinh ra nhiều mẫu có chất lượng thấp, ảnh hưởng đến quá trình huấn luyện và làm giảm hiệu năng mô hình. Bên cạnh đó, các phương pháp tinh chỉnh nhãn giả thường giả định tỷ lệ nhiễu trong nhãn là thấp, dẫn đến hiệu quả kém trong các trường hợp nhãn giả sai lệch phổ biến, từ đó làm suy giảm khả năng học và độ chính xác của mô hình.

- Hạn chế 3: Theo hướng tiếp cận không giám sát, khi sử dụng ViT làm mạng cốt lõi sẽ sinh ra yêu cầu tính toán cao, từ đó gây thách thức trong việc triển khai ở các hệ thống giới hạn tài nguyên. Một thách thức nữa là việc phụ thuộc vào chất lượng sinh nhãn giả có thể làm phát sinh nhiễu, đặc biệt trong các môi trường hình ảnh có tính tương đồng cao hoặc chất lượng thấp, điều này yêu cầu các phương pháp tinh chỉnh nhãn giả phù hợp.

# Hướng phát triển

Dựa trên những hạn chế đã phân tích, một số hướng nghiên cứu tiềm năng có thể được phát triển nhằm tiếp tục nâng cao hiệu năng của mô hình tái định danh người. Thứ nhất, việc áp dụng các hàm mất mát mới hoặc kết hợp linh hoạt nhiều hàm mất mát khác nhau có thể giúp tối ưu hóa không gian đặc trưng, từ đó tăng cường khả năng phân biệt giữa các danh tính. Thứ hai, cần cải thiện chất lượng ảnh sinh bởi GAN thông qua việc bổ sung các ràng buộc hợp lý trong quá trình sinh ảnh hoặc thay thế GAN bằng các kỹ thuật tăng cường dữ liệu tiên tiến hơn, chẳng hạn như mô hình khuếch tán, nhằm nâng cao tính đa dạng và biểu diễn của dữ liệu huấn luyện. Thứ ba, có thể áp dụng một chiến lược tinh chỉnh nhãn giả hiệu quả hơn để tăng hiệu năng mô hình trong quá trình huấn luyện không giám sát. Cuối cùng, một hướng nghiên cứu quan trọng là mở rộng đánh giá mô hình trên các bộ dữ liệu hiện đại như LUPerson, LaST, ENTIRe-ID, cũng như các tập dữ liệu thu thập từ hệ thống camera giám sát thực tế – nơi tồn tại các điều kiện khắc nghiệt hơn về ánh sáng, thời tiết, góc nhìn và sự biến thiên ngoại hình. Các đánh giá này không chỉ giúp xác định rõ giới hạn của mô hình hiện tại mà còn hỗ trợ định hướng hợp lý cho những cải tiến về kiến trúc và cơ chế huấn luyện, từ đó nâng cao khả năng ứng dụng trong môi trường triển khai thực tế.

# DANH MỤC
# CÔNG TRÌNH KHOA HỌC

## Tạp chí

[CT1] **Dang H. Pham**, Anh D. Nguyen, and Hoa N. Nguyen. "GAN-based data augmentation and pseudo-label refinement with holistic features for unsupervised domain adaptation person re-identification". In: Knowledge- Based Systems 288 (2024), p. 111471. issn: 0950-7051. doi: 10.1016/j.knosys.2024.111471. (SCI-E, Q1-Scopus).

[CT2] **Dang Hai Pham** and Hoa Ngoc Nguyen. "SCM-ReID: enhancing person re-identification by supervised contrastive–metric learning and hybrid loss optimization". In: Journal of Electronic Imaging 34.4 (2025), p. 043001. doi: 10.1117/1.JEI.34.4.043001. (SCI-E, Q3-Scopus).

[CT3] **Hai Dang Pham**, Ngoc Tu Nguyen, and Ngoc Hoa Nguyen. "ViTC-UReID: Enhancing unsupervised person ReID with vision transformer image encoder and camera-aware proxy learning". In: Journal of Computer Science and Cybernetics 41.3 (Sept. 2025), pp. 265–284. doi: 10.15625/1813-9663/23018.

## Hội nghị

[CT4] **Dang H. Pham**, Anh D. Nguyen, Long V. Vu, and Hoa N. Nguyen. "IQAGA: Image Quality Assessment-Driven Learning with GAN-Based Dataset Augmentation for Cross-Domain Person Re-Identification". In: Proceedings of the 12th International Symposium on Information and Communication Technology. 2023, pp. 63–70. isbn: 9798400708916. doi: 10.1145/3628797.3628961.

[CT5] Anh D. Nguyen, **Dang H. Pham**, and Hoa N. Nguyen. "GAN-Based Data Augmentation and Pseudo-label Refinement for Unsupervised Domain Adaptation Person Re-identification". In: Computational Collective Intelligence. 2023, pp. 591–605. isbn: 978-3-031-41456-5. doi: 10.1007/978-3-031-41456-5_45.

# TÀI LIỆU THAM KHẢO